\documentclass[letterpaper, 10pt, conference]{ieeeconf}

\IEEEoverridecommandlockouts    %
\makeatletter
\let\NAT@parse\undefined
\makeatother

\usepackage[utf8]{inputenc}
\usepackage[T1]{fontenc}      %

\usepackage{amsmath,amssymb,amsfonts}  %
\usepackage{bm}               %
\usepackage{booktabs}         %
\usepackage{cite}
\usepackage{float}            %
\usepackage{gensymb}          %
\usepackage{graphicx}
\usepackage{microtype}        %
\usepackage{multirow}
\usepackage{physics}          %
\usepackage{placeins}         %
\usepackage{subcaption}       %
\usepackage[table]{xcolor}    %
\usepackage{xcolor}
\usepackage{xspace}           %
\usepackage{xstring}          %
\usepackage{hhline}

\definecolor{citecolor}{RGB}{34,139,34}  %
\usepackage[breaklinks=false,bookmarks=true,hypertexnames=true,pagebackref=true,colorlinks,citecolor=citecolor]{hyperref}

\newcommand{\datasetName}{AMV-Bench}
\newcommand{\totalKm}{482}
\newcommand{\totalSnippets}{116}
\newcommand{\totalHours}{21}

\newcommand{\poselow}{\xi}
\newcommand{\pose}{\bm{\poselow}}
\newcommand{\linpose}{\pose^\ell}
\newcommand{\cbpose}{\pose^c}

\newcommand{\rotlow}{\omega}
\newcommand{\rot}{\bm{\rotlow}}
\newcommand{\translow}{v}
\newcommand{\trans}{\bm{\translow}}

\newcommand{\mappt}{\mathbf{X}_{j}}
\newcommand{\timet}[1]{\bar{t}_{#1}}
\newcommand{\mf}[1]{\text{MF}_{#1}}
\newcommand{\kmf}[1]{\text{KMF}_{#1}}
\newcommand{\Exp}{\text{Exp}}
\newcommand{\Log}{\text{Log}}

\newcommand{\bR}{\mathbb{R}}

\renewcommand{\deg}{{$^{\circ}$}}

\newcommand{\se}[1]{\mathfrak{se}(#1)}
\newcommand{\SE}[1]{\mathbb{SE}(#1)}

\newcommand{\SO}[1]{\mathbb{SO}(#1)}

\newcommand{\be}{\mathbf{e}}

\newcommand{\bT}{\mathbf{T}}

\newcommand{\cM}{\mathcal{M}}
\newcommand{\cC}{\mathcal{C}}

\newcommand{\rt}{RPE-T}
\newcommand{\rr}{RPE-R}
\newcommand{\ate}{ATE}
\newcommand{\sr}{SR}
\newcommand{\best}[1]{\textbf{#1}}
\newcommand{\second}[1]{\underline{#1}}
\newcommand{\yes}{$\checkmark$}

\makeatletter
\DeclareRobustCommand\onedot{\futurelet\@let@token\@onedot}
\def\@onedot{\ifx\@let@token.\else.\null\fi\xspace}

\def\eg{\emph{e.g}\onedot} 
\def\ie{\emph{i.e}\onedot} 
 
\def\etc{\emph{etc}\onedot}

\makeatother

\def\BibTeX{{\rm B\kern-.05em{\sc i\kern-.025em b}\kern-.08em
    T\kern-.1667em\lower.7ex\hbox{E}\kern-.125emX}}

\newcommand{\lidar}{{LiDAR}}

\definecolor{can_color}{RGB}{117,112,179} %
\definecolor{shenlong_color}{RGB}{27,158,119} %
\definecolor{raquel_color}{RGB}{217,95,2}     %
\definecolor{andrei_color}{RGB}{231,41,138}   %
\definecolor{weichiu_color}{RGB}{102,166,30}  %
\definecolor{joyce_color}{RGB}{30, 144, 255} %

\title{\LARGE \bf
Asynchronous Multi-View SLAM
}

\author{
    Anqi Joyce Yang$^{\ast 1,2}$, Can
    Cui$^{\ast 1,3}$, Ioan Andrei Bârsan$^{\ast 1,2}$, Raquel Urtasun$^{1,2}$, Shenlong Wang$^{1,2}$%
    \thanks{$^\ast$Denotes equal contribution. Work done during
    Can's internship at Uber.}
    \thanks{$^{1}$Uber Advanced Technologies Group}
    \thanks{$^{2}$University of Toronto, \newline\texttt {\{ajyang, iab, urtasun, slwang\}@cs.toronto.edu}}
    \thanks{$^{3}$University of Waterloo, \texttt {c23cui@uwaterloo.ca}}
}%

\begin{document}
\maketitle

\newif\ifarxiv
\arxivtrue

\ifarxiv
  \newcommand{\suppName}{appendix}
  \thispagestyle{plain}
  \pagestyle{plain}
\else
  \newcommand{\suppName}{supplementary material}
  \thispagestyle{empty}
  \pagestyle{empty}
\fi

\renewcommand{\thefootnote}{\alph{footnote}}
\begin{abstract}
  Existing multi-camera SLAM systems assume synchronized shutters for all
  cameras, which is often not the case in practice.
  In this work, we propose a
  generalized multi-camera SLAM formulation which  accounts for asynchronous
  sensor observations. Our framework integrates a %
  continuous-time motion model to
  relate information across asynchronous multi-frames during tracking, local
  mapping, and loop closing.
  For evaluation, we collected \textit{\datasetName{}}, a challenging new SLAM dataset covering
  \totalKm{} km of driving recorded using our asynchronous multi-camera robotic platform.
  \datasetName{} is over an order of magnitude larger than previous
  multi-view HD
  outdoor SLAM datasets, and covers diverse and challenging motions and environments.
  Our experiments emphasize the necessity of asynchronous sensor modeling,
  and show that the use of multiple cameras is critical
  towards robust and accurate SLAM in challenging outdoor scenes.
  \ifarxiv%
    For additional information, please see the project website at:\\
    \centerline{\url{https://www.cs.toronto.edu/~ajyang/amv-slam}}\unskip
  \else
    The supplementary material is located at:
    \url{https://www.cs.toronto.edu/~ajyang/amv-slam}
  \fi
\end{abstract}

\section{Introduction}

\begin{figure*}
  \centering
  \includegraphics[width=0.95\linewidth]{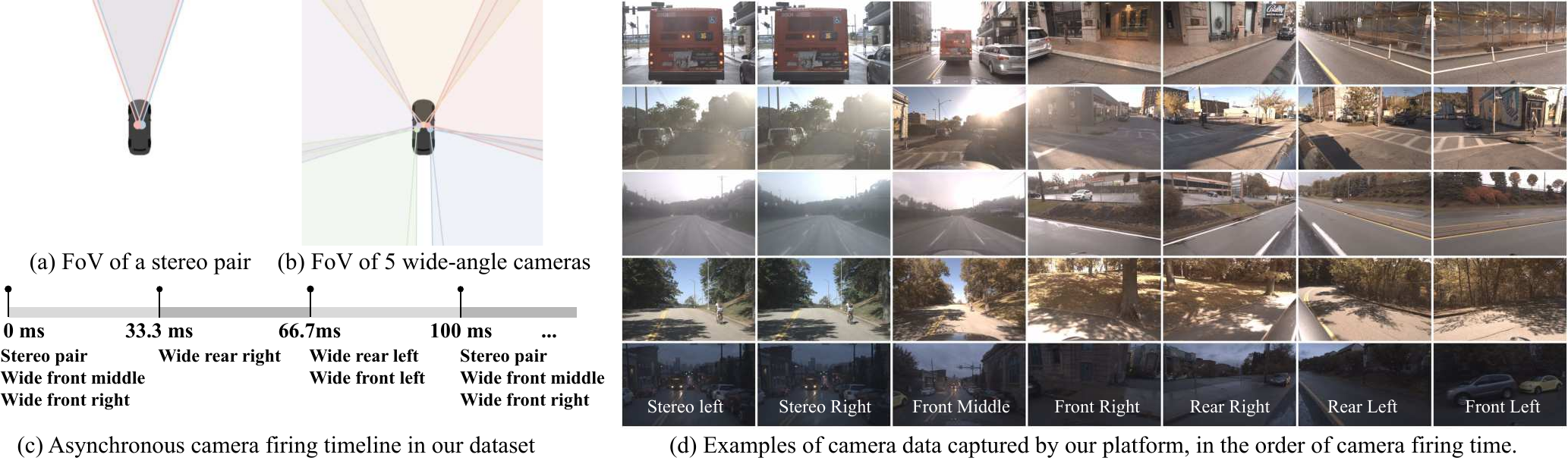}
  \caption{
    The asynchronous multi-camera rig in~\datasetName{}, containing a stereo pair and
    five wide-angle cameras. The cameras are synced to a LiDAR, with the
    asynchronous firing schedule shown in (c). %
    The sample images highlight challenging scenarios like occlusions, sunlight
    glare, low-textured highways, 
    shadows on the road, 
    and low-light rainy environments.}%
  \label{fig:camera_layout_and_data_samples_v1_0_0}
  \vspace{-5mm}
\end{figure*}

Simultaneous Localization and Mapping (SLAM) is the task of localizing an autonomous agent in unseen
environments by building a map at the same time.
SLAM is a fundamental part of %
many technologies ranging from augmented reality to photogrammetry and robotics.
Due to the
availability of camera sensors and the rich information they provide, camera-based SLAM, or visual SLAM,
has been widely studied and applied in robot navigation.

Existing visual SLAM methods~\cite{engel2014lsd, engel2015stereolsd,engel2017direct, mur2015orb, mur2017orb}
and benchmarks~\cite{kitti, euroc, wang2020tartanair}
mainly focus on either monocular or stereo camera settings.
Although lightweight,
such configurations %
are prone to tracking failures caused by occlusion, dynamic objects, lighting
changes and textureless scenes, all of which are common in the real world.
Many of these challenges can be attributed to the narrow field of view typically used
(Fig.~\ref{fig:camera_layout_and_data_samples_v1_0_0}a).
Due to their larger field of view (Fig.~\ref{fig:camera_layout_and_data_samples_v1_0_0}b),
wide-angle or fisheye lenses~\cite{intel-realsense,kitti360}
or multi-camera rigs~\cite{Heng-RSS-14, tribou2015multi, harmat2015multi, urban2016multicol, Liu2018IROS, skydiox2}
can significantly increase the robustness of visual SLAM systems~\cite{Liu2018IROS}. %

Nevertheless, using multiple cameras comes with its own set of challenges.
Existing stereo~\cite{mur2017orb} or
multi-camera~\cite{Heng-RSS-14, tribou2015multi, harmat2015multi, urban2016multicol, Liu2018IROS}
SLAM literature assumes synchronized shutters for all cameras and adopts
discrete-time trajectory modeling based on this assumption.
However, in practice different cameras are not always triggered at the
same time,
either due to technical limitations, or by design.
For instance, the camera shutters could be synchronized to another sensor, such as a
spinning LiDAR (\eg, Fig.~\ref{fig:camera_layout_and_data_samples_v1_0_0}c),
which is a common set-up in self-driving~\cite{lyft2019,waymo_open_dataset,nuscenes2020,zhou2020da4ad}.
Moreover, failure to account for the robot motion in between the firing of the
 cameras could lead to %
 localization failures. %
Consider a car driving along a highway at 30m/s (108km/h).  %
  Then in a short 33ms camera firing interval, the vehicle would travel one meter, which is significant
when centimeter-accurate pose estimation is required.
As a result, a need arises for a generalization of  multi-view visual SLAM to be
agnostic to camera timing, while being scalable and robust to real-world
conditions.

  In this paper we formalize the {\it asynchronous multi-view SLAM (AMV-SLAM)} problem.
Our first contribution is a general framework for AMV-SLAM, which, to the best of our knowledge,
is the first full asynchronous continuous-time multi-camera visual SLAM system for large-scale outdoor
environments.
Key to
this formulation is (1) the concept of asynchronous multi-frames, which group
input images %
from multiple asynchronous cameras, and (2) the integration of a continuous-time motion model,
which relates spatio-temporal information from
asynchronous multi-frames
for joint continuous-time trajectory estimation.

  Since there is no public asynchronous multi-camera SLAM dataset,
our second contribution is \textit{\datasetName}, %
a novel large-scale
dataset with high-quality ground-truth.
\datasetName{} was collected during a full
year in Pittsburgh, PA, %
and includes challenging conditions
such as low-light scenes, occlusions, fast driving
(Fig.~\ref{fig:camera_layout_and_data_samples_v1_0_0}d), and complex
maneuvers like three-point turns and reverse parking.
Our experiments show that multi-camera configurations are critical in overcoming
adverse conditions in large-scale outdoor scenes. In addition, we show that
asynchronous sensor modeling is crucial,
as treating the cameras as synchronous leads to 30\% higher failure rate and $4\!\times{}$ the local pose errors
compared to asynchronous modeling.

\section{Related Work}%
\label{sec:related}

\subsubsection{Visual SLAM / Visual Odometry}
SLAM has been a core area of research in
  robotics since the %
  1980s~\cite{Bailey2006a,
  leonard1990phd, davison1998, burgard2000, montemerlo2002fastslam}.
  The comprehensive survey by Cadena et al.~\cite{Cadena2016past} provides
  a detailed overview of SLAM. %
Modern visual SLAM approaches can be divided into direct and indirect methods.
Direct methods like DTAM~\cite{newcombe2011dtam},
LSD-SLAM~\cite{engel2014lsd}, and DSO~\cite{engel2017direct} estimate motion
and map parameters by directly optimizing over pixel intensities (photometric error)~\cite{Lucas81aniterative, irani99allabout}.
Alternatively, indirect methods, which are the focus of this work,
minimize the re-projection energy (geometric error)~\cite{Triggs00bundleadjustment}
{over an intermediate representation
obtained from raw images.}
A common subset of these are feature-based methods like
PTAM~\cite{klein2007parallel} and ORB-SLAM~\cite{mur2015orb} which represent
raw observations as sets of keypoints.

\subsubsection{Multi-View SLAM}
Monocular~\cite{klein2007parallel,engel2014lsd,mur2015orb,engel2017direct} and
stereo~\cite{engel2015stereolsd,mur2017orb,wenzel20204seasons} are the most common visual SLAM configurations.  However, many applications
could benefit from a much wider field of view for better perception and situation awareness
during navigation.
Several multi-view SLAM approaches have been
proposed~\cite{sola2008fusing,hee2013motion,tribou2015multi,harmat2015multi,urban2016multicol,Liu2018IROS,
meng2018dense, zhang2018vins,ye2019robust,seok2020rovins}. Early
filtering-based approaches~\cite{sola2008fusing} treat multiple cameras
as independent sensors, %
and fuse their observations
using a central Extended Kalman Filter. 
Recent optimization-based multi-camera SLAM
systems \cite{tribou2015multi,harmat2015multi,urban2016multicol,Liu2018IROS} extend  monocular PTAM~\cite{klein2007parallel},
ORB-SLAM~\cite{mur2015orb} and DSO~\cite{engel2017direct} respectively to
synchronized multi-camera rigs to jointly estimate ego-poses at discrete
timestamps. Multi-view visual-inertial systems such as VINS-MKF~\cite{zhang2018vins}
and ROVINS~\cite{seok2020rovins} also assume synchronous camera timings.

\subsubsection{Continuous-time Motion Models}

Continuous-time motion models help relate sensors %
triggered at arbitrary times
while moving, with applications such as calibration~\cite{furgale2013iros,furgale2015batch},
target tracking~\cite{li2003survey, crouse2015basic} and motion planning~\cite{zefran1996continuous, mukadam2016gaussian}.
Continuous-time SLAM typically focuses on %
visual-inertial fusion~\cite{lovegrove2013spline,
ovren2019trajectory},
rolling-shutter cameras~\cite{hedborg2012rolling,kerl2015dense,kim2016direct,schubert2018direct,schops2019bad,ovren2019trajectory}
or LiDARs~\cite{zhang2014loam,alismail2014,behnke2018lidar,wong2020data}.
Klingner et al.~\cite{google2013sfm} propose a continuous-time Structure-from-Motion framework for multiple synchronous
rolling-shutter cameras.
The key component for continuous-time trajectory modeling is choosing a family
of functions that is both flexible and reflective of the kinematics.
A common approach is fitting parametric functions over the states, %
\eg, piecewise linear functions~\cite{google2013sfm,schubert2018direct},
spirals~\cite{zeng2019end},
wavelets~\cite{anderson2014hierarchical},
or
B-splines~\cite{lovegrove2013spline,sommer2019efficient,hug2020conceptualizing}.
Other approaches represent trajectories through non-parametric methods such as Gaussian
Processes%
~\cite{furgale2013iros,furgale2015batch,anderson2015batch,dong2018sparse,tang2019white,wong2020data}.

\section{Notation}%
\label{sec:preliminaries}

\subsubsection{Coordinate Frame}
We denote a coordinate frame $x$ with $\mathcal{F}_x$. $\mathbf{T}_{yx}$ is the
rigid transformation
that maps homogeneous points from $\mathcal{F}_x$ to $\mathcal{F}_y$. In this work we use three coordinate frames:
the world frame $\mathcal{F}_w$, the
moving robot's body
frame $\mathcal{F}_b$,
and the camera frame $\mathcal{F}_k$ associated with
each camera $\mathcal{C}_k$. 

\subsubsection{Pose and Motion}
The pose of a 3D rigid body can be represented as a rigid transform from %
$\mathcal{F}_b$ to %
$\mathcal{F}_w$ as follows:
\[
\mathbf{T}_{wb}= \begin{bmatrix} \mathbf{R}_{wb} & \mathbf{t}_w \\ \mathbf{0}^T & 1 \\ \end{bmatrix} \in \SE{3}
\text{\ \ with \ \ }  \mathbf{R}_{wb}
\in \SO{3},
\mathbf{t}_w \in \mathbb{R}^3
\]
where $\mathbf{R}_{wb}$ is the $3\times3$ rotation matrix,
$\mathbf{t}_w$ is the translation vector, and 
$\SE{3}$ and $\SO{3}$ are the Special Euclidean
and Special Orthogonal Matrix Lie Groups respectively.
We define the trajectory of a 3D rigid body  as a
function $\bT_{wb}(t): \mathbb{R}\rightarrow\SE{3}$ over the time
domain $t\in\mathbb{R}$.

\subsubsection{Lie Algebra Representation} For optimization purposes,
a 6-DoF minimal pose representation associated with the Lie Algebra $\se{3}$ of
the matrix group $\SE{3}$ is widely adopted.
It is a vector $\pose
= [\,\trans^T \,\,\, \rot^T\,]^T  \in \mathbb{R}^6$, where $\trans \in \mathbb{R}^3$
and $\rot \in \mathbb{R}^3$ encode the translation and rotation components respectively.
We use the uppercase $\Exp$ (and, conversely,
$\Log$) to
convert $\pose\in\mathbb{R}^6$ to $\mathbf{T}\in\SE{3}$:
$\Exp(\pose) = \exp(\pose^\wedge) = \mathbf{T}$,
where $\exp$ is the matrix exponential, and $\pose^\wedge = \begin{bmatrix} \rot_\times & \trans \\ \bm{0}^T & 0 \\
\end{bmatrix} \in \se{3}$ with
$\rot_\times$ being the $3\times3$ skew-symmetric matrix of $\rot$.
\subsubsection{Motion Model}
We use superscripts to denote the type of motion models.
$c$ and $\ell$ represent the cubic B-spline motion model
$\bT^c(t)$ and the linear motion model $\bT^\ell(t)$ respectively.

\section{Asynchronous Multi-View SLAM}%
\label{sec:asynchronous_multi_view_slam}

\begin{figure*}
  \centering
  \includegraphics[width=0.95\linewidth]{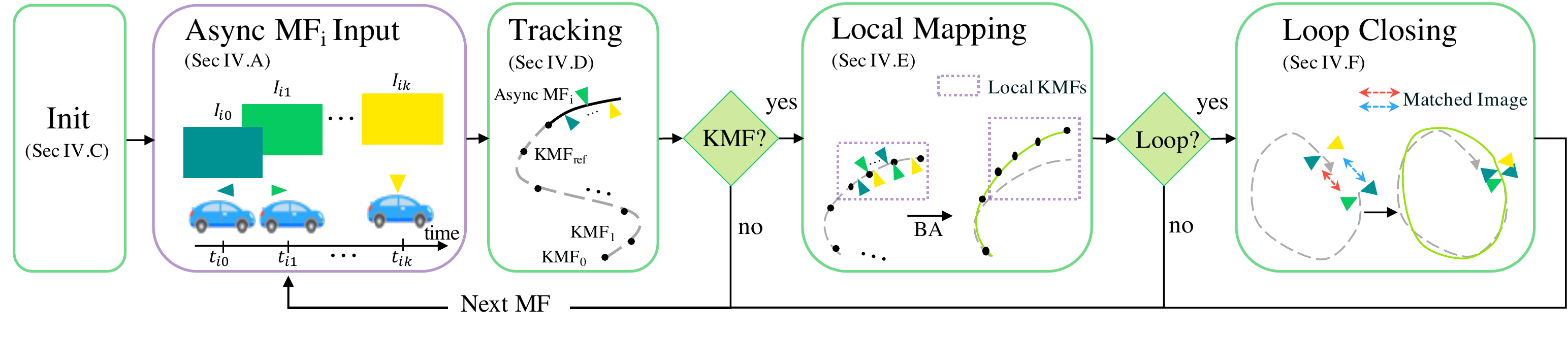}
  \vspace{-2mm}
  \caption{Overview. Initialization is followed by
  tracking, local mapping and loop closing. %
  MF=Multi-Frame; KMF=Key MF.}%
  \label{fig:pipeline}
  \vspace{-6mm}
\end{figure*}

\begin{figure}
  \centering
  \vspace{-0mm}
  \includegraphics[width=\linewidth]{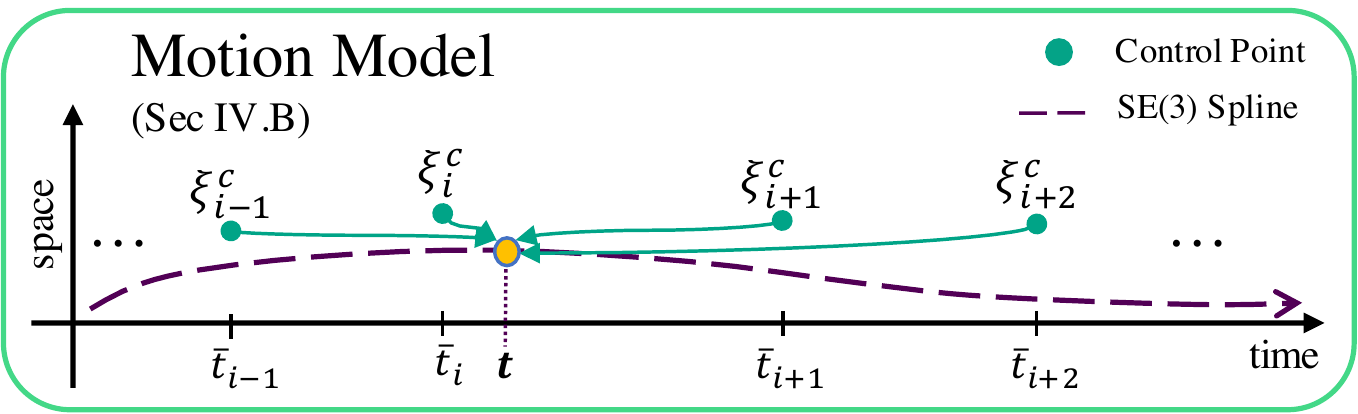}
  \vspace{-5mm}
  \caption{Illustration of the cubic B-spline model.
  The pose $\mathbf{T}^{c}_{wb}(t)$ at time $t \in [\timet{i}, \timet{i+1}]$
  is defined by four control poses associated with key multi-frames
  indexed at $i-1, i, i+1, i+2$.}%
  \label{fig:motion_model}
  \vspace{-6mm}
\end{figure}

We consider the asynchronous multi-view SLAM problem where the observations are
captured by multiple cameras %
triggered at arbitrary times with respect to each other. %
Each camera $\cC_k$ is
assumed to be a calibrated pinhole camera
with intrinsic matrix $\mathbf{K}_k$, and
extrinsics encoded by the mapping
$\mathbf{T}_{kb}$
from the body frame $\mathcal{F}_b$ to camera frame $\mathcal{F}_k$. %
The input to the problem is
a sequence of image and capture timestamp pairs $\{(I_{ik}, t_{ik})\}_{\forall i}$,
associated with each camera $\cC_k$. The goal is then to estimate
the robot trajectory $\bT_{wb}(t)$ in the world frame. As a byproduct we also estimate a
map $\cM$ of the 3D structure of the environment as a set of points.

Our system follows the standard visual SLAM structure of
initialization coupled with the three-threaded tracking, local mapping, and loop closing,
with the key difference that we generalize to multiple cameras with asynchronous timing
via \textit{asynchronous multi-frames} (Sec.~\ref{sec:asyncmf}) and a continuous-time motion model (Sec.~\ref{sec:motionmodel}).
In particular,
after initialization (Sec.~\ref{sec:init}), %
tracking (Sec.~\ref{sec:tracking})
takes each incoming multi-frame as input, infers its motion  parameters,
and decides whether to promote it as a \textit{key} multi-frame (KMF). For efficiency,
only KMFs are used during local mapping
(\mbox{Sec.~\ref{sec:mapping}}) and loop closing
(Sec.~\ref{sec:loopclosure}).
When a new KMF is selected,
the local mapping module refines
poses and map points over a recent window of KMFs
to ensure local consistency, while the loop closing module detects when
a previously-mapped area is being revisited and corrects the drift to enhance global consistency.
See Fig.~\ref{fig:pipeline} for an overview.

\subsection{Asynchronous Multi-Frames}
\label{sec:asyncmf}
Existing synchronous multi-view systems~\cite{urban2016multicol} group
multi-view images captured at the same time into a \textit{multi-frame}
as input. %
However, this cannot be directly applied
when the firing time of each sensor varies.
To generalize to asynchronous camera timings, we introduce the concept of
\textit{asynchronous multi-frame}, which groups images that
are captured closely (\eg, within 100ms) in time.
In Fig.~\ref{fig:camera_layout_and_data_samples_v1_0_0}
each asynchronous multi-frame contains the
images taken during a single spinning LiDAR sweep at 10 Hz.
Contrasting to synchronous multi-frames~\cite{urban2016multicol} that store images
and a discrete pose estimated at a single timestamp, each \textit{asynchronous} multi-frame
$\mf{i}$ stores: (1) a set of image and capture time pairs $\{(I_{ik}, t_{ik})\}$
indexed by associated camera $\mathcal{C}_k$, and (2) continuous-time
motion model parameters %
to recover the estimated trajectory. %

\subsection{Continuous-Time Trajectory Representation}
\label{sec:motionmodel}

To associate the robot pose with observations that could be made at arbitrary times,
we formulate the overall robot
trajectory as a \textit{continuous-time}
function $ \mathbf{T}_{wb}(t) : \bR \to \SE{3} $, rather than %
discrete poses.
We
exploit a \emph{cumulative} cubic B-spline function \cite{lovegrove2013spline} %
as
  the first and second derivatives of this parameterization are smooth and computationally efficient
  to evaluate~\cite{sommer2019efficient}.
The cumulative structure is necessary for accurate
on-manifold interpolation in $\SE{3}$~\cite{kim1995general,lovegrove2013spline}.
Following~\cite{lovegrove2013spline}, given a knot vector $\mathbf{b} = [t_0, t_1,\ldots, t_7] \in \mathbb{R}^8$,
a cumulative cubic B-spline trajectory $\mathbf{T}_{wb}^c(t)$ over $t\in[t_3, t_4)$
is defined by four control poses $\cbpose_0\!:\!\cbpose_3 \in \mathbb{R}^6$~\cite{lovegrove2013spline}.
In our framework, we associate each key multi-frame $\kmf{i}$ with a control
pose.
In addition, since the KMFs do not necessarily distribute evenly in
time, we use a non-uniform knot vector. 
For each $\kmf{i}$, we define the \textit{representative time}
$\timet{i}$ %
as the median
of all image capture times $t_{ik}$, and define the knot vector as
$\mathbf{b}_i = [\timet{i-3}, \timet{i-2}, \ldots, \timet{i+4}] \in \mathbb{R}^8$.
Then, the spline trajectory over the interval
$t \in [\timet{i}, \timet{i+1})$ can be expressed as a function of four control poses
$\cbpose_{i-1}$, $\cbpose_{i}$, $\cbpose_{i+1}$, $\cbpose_{i+2}$:
\begin{equation}
  \mathbf{T}_{wb}^c(t) =
  \Exp \left(\cbpose_{i-1} \right)
  \prod_{j=1}^{3} \Exp \left( \widetilde{B}_{j,4}(t) \pmb{\Omega}_{i-1+j} \right),
\end{equation}
where $\pmb{\Omega}_{i-1+j}\!=\!\Log(\Exp(\cbpose_{i-2+j})^{-1}\Exp(\cbpose_{i-1+j}))$ %
is the relative pose between control poses, and
$\widetilde{B}_{j,4}(t)\!=\!\sum_{l=j}^{3} B_{l,4}(t) \in \mathbb{R}$
is the cumulative basis function, where
the basis function $B_{l,4}(t)$ is computed with the knot vector $\mathbf{b}_i$
using the de Boor-Cox %
formula~\cite{de1972calculating, cox1972numerical}. See
Fig.~\ref{fig:motion_model} for an illustration, and the \suppName{} for more details.

\subsection{Initialization}
\label{sec:init}
The system initialization assumes
that there exists a pair of cameras that share a reasonable overlapping field
of view and fire very closely in time (\eg, a synchronous stereo pair, present
in most autonomous vehicle setups~\cite{nuscenes2020,agarwal2020ford,geyer2020a2d2}).
At the system startup, we create the first MF %
with the associated camera images and capture times, select it
as the first KMF, set the representative time $\timet0$ to
the camera pair firing time, the control pose $\cbpose_0$ to
the origin of the world frame, and initialize the map with %
points triangulated from the %
camera pair.
Map points from other camera images are created during mapping
after the second KMF %
is inserted.

\subsection{Tracking}
\label{sec:tracking}
During tracking, we estimate the continuous pose of an incoming multi-frame $\mf{i}$
by matching it with the most recent KMF. We then decide
whether $\mf{i}$ should be selected as a KMF for
map refinement and future tracking. Following~\cite{mur2015orb}, we formulate
pose estimation and map refinement as an indirect geometric energy minimization problem
based on sparse image features. %
\subsubsection{Feature Matching}

For each image in the new MF, we identify its reference images in the reference
KMF as images captured by the same camera or any camera %
sharing a reasonable overlapping
field of view. We extract sparse 2D keypoints and associated descriptors
from the new images and match them against the reference image keypoints
to establish associations with existing 3D map points.
We denote the set of matches as
$\{(\mathbf{u}_{i,k,j}, \mappt)\}_{\forall(k,j)}$,
where $\mathbf{u}_{i,k,j} \in \mathbb{R}^2$
is the 2D keypoint extracted from image $I_{ik}$ in $\mf{i}$, %
and $\mathbf{X}_j \in \mathbb{R}^3$ is the matched 3D map point %
in the world frame.

\subsubsection{Pose Estimation}

Cubic B-splines are effective for modeling the overall trajectory,
but directly using them in tracking entails estimating four 6-DoF control poses
that define motion not only in the new MF, but also in
the existing trajectory. Therefore, more information is needed for a stable estimation.
For computational efficiency, we instead use a simpler and less expressive
continuous-time linear motion model $\bT^{\ell}_{wb}(t)$ during tracking, whose parameters
are later %
used to initialize the cubic B-spline model in Sec.~\ref{sec:mapping}.
Specifically, we estimate $\mf{i}$ pose $\linpose_i\!\in\!\mathbb{R}^6$ at the representative timestamp $\timet{i}$,
and evaluate the continuous pose with linear interpolation and extrapolation:
$\bT^{\ell}_{wb}(t) =\Exp(\linpose_i)\left({\Exp(\linpose_i)}^{-1}\bT^c_{wb}(\timet{\text{ref}})\right)^{\alpha}$,
where $\alpha=\frac{\timet{i}- t}{\timet{i} - \timet{\text{ref}}}$ and $\timet{\text{ref}}$ is the representative timestamp
of the reference KMF.

Coupled with the obtained multi-view correspondences, we formulate pose
estimation for $\linpose_i$ as a constrained, asynchronous,
multi-view case of the perspective-n-points (PnP) problem, in which a geometric
energy is minimized:
\begin{equation}
\begin{aligned}
	E_{geo}(\linpose_i) =
		\sum_{(k, j)}
    \rho\left(
      \left\|\mathbf{e}_{i,k,j}(\linpose_i)\right\|^2_{\mathbf{\Sigma}_{i,k,j}^{-1}}
    \right),
\label{eq:async_tracking_energy}
\end{aligned}
\end{equation}
where $\be_{i, k, j}(\linpose_i)\!\in\!\mathbb{R}^2$ is the reprojection error
of the matched correspondence pair $(\mathbf{u}_{i, k, j}, \mappt)$, and
$\mathbf{\Sigma}_{i, k, j} \in \mathbb{R}^{2 \times 2}$
is a diagonal covariance matrix denoting the uncertainty of the match.
$\rho$ denotes a robust norm, with Huber loss used in practice. %
The reprojection error for the pair is defined as:
\begin{equation} \label{eq:async_tracking_reproj_error}
	\mathbf{e}_{i,k, j}(\linpose_i)
		= \mathbf{u}_{i,k, j} - \pi_k \big( \mappt, \mathbf{T}^\ell_{wb}(t_{ik}) \mathbf{T}^{-1}_{kb} \big),
\end{equation}
where $\pi_k( \cdot , \cdot) : \mathbb{R}^3 \times \mathbb{SE}(3) \to \mathbb{R}^2$
is the perspective projection function of camera $\mathcal{C}_k$,
$\mathbf{T}_{kb}$ is the respective camera extrinsics matrix,
and $\bT^\ell_{wb}(t)$ is the linear model used only during tracking
to initialize the cubic B-spline model later.

We initialize $\linpose_i$
by linearly extrapolating poses from the previous two
multi-frames $\mf{i-2}$ and $\mf{i-1}$ based on a constant-velocity model.
To achieve robustness against outlier map point associations, we wrap
the above optimization in a RANSAC loop, where only a minimal number of
$(\mathbf{u}_{i, k, j}, \mappt)$
are sampled to obtain each hypothesis.
We solve the optimization with the Levenberg-Marquardt (LM) algorithm within each RANSAC iteration.
Given our initialization, %
the problem converges in a few steps in each RANSAC iteration.

\subsubsection{Key Multi-Frame Selection}
We use a hybrid key frame selection scheme based on estimated motion~\cite{klein2007parallel} and map
point reobservability~\cite{mur2015orb}. Namely, the current MF is registered as a KMF
if the tracked pose has a local translational or rotational change above a
certain threshold, or if the ratio of map points reobserved in a number of
cameras is below a certain threshold. 
In addition, to better condition the shape of cubic B-splines, a new KMF is
regularly inserted during long periods of little change in motion and scenery
(\eg, when the robot stops).
Empirically we find reobservability-only
heuristics insufficient during very fast motions
in low-textured areas (\eg, fast highway driving),
but show their combination with
the motion-based heuristic
to be robust to such scenarios.

\subsection{Local Mapping}
\label{sec:mapping}
When a new KMF is selected, we run local bundle adjustment to refine the 3D map
structure and minimize drift accumulated from tracking errors in recent frames.
Map points %
are then created and culled to reflect the latest changes.

\subsubsection{Bundle Adjustment}
We use windowed bundle adjustment to refine poses and map points in
a set of recent KMFs. %
Its formulation is similar to the pose estimation problem, except extended to a window
of $N$ key frames $\{ \cbpose_i \}_{1 \le i \le N} $ to jointly estimate a set of
control poses %
and map points:
\begin{equation}
\label{eq:async_ba_energy}
  E_{geo}(\{\cbpose_i \}, \{\mappt\})\kern-0.1em = \kern-0.6em
    \sum_{(i, k, j)}\kern-0.3em
    \rho\left(
      \left\|
        \mathbf{e}_{i, k, j}(\{\cbpose_i\}, \mappt)
      \right\|_{\mathbf{\Sigma}_{i, k, j}^{-1}}^2
    \right).
\end{equation}
Note that unlike tracking, we now refine the estimated local trajectory with the
cubic B-spline model $\mathbf{T}_{wb}^c(t)$ parameterized by control poses
$\{\cbpose_i\}$. We initialize the control pose $\cbpose_N$ of the
newly-inserted key multi-frame with $\linpose_N$ estimated in tracking.
For observations made after $\timet{N-1}$, their
pose evaluation would involve control poses $\cbpose_{N+1}$ and $\cbpose_{N+2}$
and knot vector values $\timet{N+1\le p \le N+4}$ which do not yet exist.
To handle such boundary cases, 
we represent these future control poses and timestamps
as a linear extrapolation function of existing
control poses and timestamps.
We again minimize Eq.~\eqref{eq:async_ba_energy}
with the LM algorithm. %

\subsubsection{Map Point Creation and Culling}
With a newly-inserted KMF, we triangulate new map points with the refined
poses and keypoint matches from
overlapping image pairs both within the same KMF and across neighboring KMFs.
To increase robustness against dynamic objects
and outliers, we cull map points that are behind the cameras or have a reprojection
errors above a certain threshold.

\subsection{Loop Closing}
\label{sec:loopclosure}
The loop closing module detects when the robot revisits an area, and
corrects the accumulated drift to achieve global consistency in mapping
and trajectory estimation. With a wider field of view, multi-view SLAM
systems can detect loops that are
encountered at arbitrary angles.

We extend the previous DBoW3~\cite{GalvezTRO12DBoW3}-based loop detection
algorithm~\cite{mur2015orb} with a multi-view similarity check and a multi-view
geometric verification. To perform loop closure, we integrate the
cubic B-spline motion model to formulate an asynchronous, multi-view case of the pose
graph optimization problem. 
Please see the \suppName{} for details.

\newcommand{\yy}{\checkmark}

\begin{table}[]
  \centering
  \caption{An overview of major SLAM datasets. Legend:
    W = diverse weather, G = large geographic diversity, MV = multi-view, HD
    = vertical resolution $\ge1080$. $^\ast$total travel distance not explicitly %
    released at the time of writing.}
  \vspace{-1mm}
  \label{tab:datasets}
  \begin{tabular}{lrrrrrr}
    \toprule
    Name                                  & Total km      & Async   & W   & G   & MV  & HD  \\
    \midrule
    KITTI-360~\cite{kitti360}          &  74         &     &     &     & \yy & \yy \\
    RobotCar~\cite{maddern2017robotcar}   & 1000          &     & \yy &     & \yy &     \\
    Ford Multi-AV~\cite{agarwal2020ford}  & n/A*          &     & \yy &     & \yy & \yy \\
    A2D2~\cite{geyer2020a2d2}             & $\approx$100* &     & \yy & \yy & \yy & \yy \\
    4Seasons~\cite{wenzel20204seasons}    & 350           &     & \yy & \yy &     &     \\
    EU Long-Term~\cite{yan2020eu}         & 1000          &     & \yy &     & \yy & \yy \\
    \midrule
    Ours                                  & \totalKm{}    & \yy & \yy & \yy & \yy & \yy  \\
    \bottomrule
  \end{tabular}
  \vspace{-5mm}
\end{table}

\section{Dataset}%
\label{sec:dataset}

{%

Much of the recent progress in computer vision and robotics has been driven by the
existence of large-scale high-quality
datasets~\cite{russakovsky2015imagenet,kitti,lin2014microsoft,maddern2017robotcar}.
However, in the field of SLAM, in spite of their large number,
previous datasets have been insufficient for evaluating asynchronous multi-view
SLAM systems due to
either scale, diversity, or sensor configuration limitations.
Such datasets either emphasize a specific canonical route over a long
period of time, foregoing geographic
diversity~\cite{maddern2017robotcar,agarwal2020ford,yan2020eu}, do not have a surround camera
configuration critical for robustness%
~\cite{kitti,maddern2017robotcar,wenzel20204seasons},
or lack the large scale necessary for evaluating
safety-critical SLAM systems~\cite{kitti,nclt,kitti360,geyer2020a2d2}.
Furthermore, none of the existing SLAM datasets feature multiple
asynchronous cameras to directly evaluate an AMV-SLAM system.

To address these limitations, we propose \textit{\datasetName},
a novel large-scale asynchronous multi-view SLAM dataset recorded using
a fleet of SDVs in Pittsburgh, PA over the span of one year.
Table~\ref{tab:datasets} shows a high-level comparison between the proposed
dataset and other similar SLAM-focused datasets.
Please see the \suppName{} for details.
}

\subsubsection{Sensor Configuration}

Each vehicle is equipped with seven cameras, wheel
encoders, an IMU, a consumer-grade GPS receiver, and an HDL-64E \lidar{}.
The \lidar{} data is only used to compute the ground-truth poses. 
There are
five wide-angle cameras spanning most of the vehicle's surroundings,
and an additional forward-facing stereo pair.
All intrinsic and extrinsic calibration parameters are computed in
advance using a set of checkerboard calibration patterns.
All images are rectified to a pinhole camera model.

Each camera has an RGB resolution of 1920$\times$1200 pixels,
and uses a global shutter.
The five wide-angle cameras are hardware
triggered in sync with the rotating \lidar{} at an average frequency of 10Hz,
firing asynchronously with respect to each other. %
Fig.~\ref{fig:camera_layout_and_data_samples_v1_0_0} illustrates the camera
configuration.%

\subsubsection{Dataset Organization}
The dataset contains \totalSnippets{} sequences spanning \totalKm{}km and
$\totalHours{}$h. %
All sequences are recorded during daytime or twilight.
Each sequence ranges
from 4 to 18 minutes, with a wide variety of scenarios including %
(1) diverse environments (busy streets, %
highways, residential and rural areas)  %
(2) diverse weather ranging from sunny days to heavy precipitation;
(3) diverse motions with varying speed (highway, urban traffic,
parking lots),
trajectory loops, and maneuvers
including U-turns and reversing.
Please refer to Fig.~\ref{fig:camera_layout_and_data_samples_v1_0_0}d for examples.
\begin{table}
  \centering
  \caption{Baseline methods. M=monocular, S=stereo, A=all cameras.
  \rt (cm/m), \rr (rad/m), \ate (m), AUC(\%).}
  \vspace*{-2mm}
  \label{tab:quantitative_baseline}
  \addtolength{\tabcolsep}{-2pt}
  \begin{tabular}{l|rr|rr|rr|r}
    \toprule
    \multirow{2}{*}{Method} &
    \multicolumn{2}{c|}{\rt} &
    \multicolumn{2}{c|}{\rr} &
    \multicolumn{2}{c|}{\ate} &
    \multirow{2}{*}{\sr(\%)} \\
    & med & AUC & med & AUC & med & AUC & \\
    \midrule
    DSO-M~\cite{gao2018ldso} & 42.72 & 28.08 & 8.02E-05 & 54.23 & 594.39 & 44.67 & 62.67 \\
    ORB-M~\cite{mur2015orb} & 34.00 & 25.66 & 5.49E-05 & 63.77 & 694.37 & 42.65 & 64.00 \\
    ORB-S~\cite{mur2017orb} & 1.85 & 65.70 & 3.29E-05 & 70.47 & 30.74 & 74.31 & 77.33 \\
    Sync-S & \second{1.30} & \second{77.54} & \second{2.91E-05} & \second{78.37} & \second{24.53} & \second{77.44} & \second{84.00} \\
    Sync-A~\cite{urban2016multicol} & 2.15 & 68.46 & 3.47E-05 & 70.47 & 58.18 & 75.01 & 74.67 \\
    Ours-A & \best{0.35} & \best{88.63} & \best{1.13E-05} & \best{88.17} & \best{6.13} & \best{88.82} & \best{92.00} \\
    \bottomrule
  \end{tabular}
  \vspace{-5mm}
  \addtolength{\tabcolsep}{2pt}
\end{table}

The dataset is split geographically into train, validation, and test
subsets (65/25/26 sequences), as %
shown in the \suppName{}. %
The ground-truth poses are obtained using an offline HD map-based global
optimization leveraging IMU, odometer, GPS, and \lidar{}.
The maps are built from multiple runs through the same area,
ensuring ground-truth accuracy. %

\newcommand{\rpeUnit}{E-05}

\begin{table*}[]
  \centering
  \caption{(Left) Ablation on motion models.
      (Right) Ablation on cameras, all with the cubic B-spline model;
      s=stereo, wf=wide-front, wb=wide-back.
      All ablations performed with loop closing disabled.%
  }%
  \addtolength{\tabcolsep}{-1.0pt}
  \begin{tabular}[t]{l|cc|cc|cc|c}
    \toprule
    \multirow{2}{*}{} &
    \multicolumn{2}{c|}{\rt} &
    \multicolumn{2}{c|}{\rr} &
    \multicolumn{2}{c|}{\ate} &
    \multirow{2}{*}{\sr} \\
    & med & AUC & med & AUC & med & AUC & \\
    \midrule
    Synch. & 1.97 & 69.46 & 2.96\rpeUnit & 73.39 & 55.24& 75.11 & 70.7 \\
    Linear & \second{0.41} & \second{87.76} & \best{1.11\rpeUnit}
           & \second{88.39} & \best{6.09} & \second{88.31} & \second{89.3} \\
    B-spline & \best{0.35} & \best{88.79} & \best{1.11\rpeUnit} & \best{88.47}
             & \second{6.53} & \best{89.04} & \best{92.0} \\
    \bottomrule
  \end{tabular}\hfill%
  \begin{tabular}[t]{ccc|cc|cc|cc|c}
    \toprule
    \multicolumn{3}{c|}{Cameras} &
    \multicolumn{2}{c|}{\rt} &
    \multicolumn{2}{c|}{\rr} &
    \multicolumn{2}{c|}{\ate} &
    \multirow{2}{*}{\sr} \\
    s & wf & wb & med & AUC & med & AUC & med & AUC & \\
    \midrule
    \yes&  & & 0.70 & 79.86 & 1.93\rpeUnit & 80.48 & 11.44 & 75.75 & 88.0 \\
    \yes& \yes & & \second{0.41} & \second{84.88} & \second{1.21\rpeUnit}
        & \second{85.86} & \second{9.00} & \second{84.92} & \second{90.7} \\
    \yes& \yes & \yes & \best{0.35} & \best{88.79} & \best{1.11\rpeUnit} & \best{88.47} & \best{6.53} & \best{89.04} & \best{92.0} \\
    \bottomrule
  \end{tabular}
  \label{tab:unified-ablation}
  \vspace{-3mm}
\end{table*}

\section{Experiments}
\label{sec:experiments}
We evaluate our method on the proposed \datasetName{} dataset.
We first show that it outperforms
several popular SLAM methods~\cite{mur2015orb,mur2017orb, urban2016multicol, gao2018ldso}.
Next, we perform ablation studies highlighting the
importance of asynchronous modeling, the use of multiple cameras, the
impact of loop closure
and, finally, the differences between feature extractors.

\newcommand{\qualtraj}[2]{\includegraphics[width=0.163\linewidth,trim={#1 0.0cm 0 0.0cm},clip]{#2}}

\begin{figure*}[]
  \centering
  \hspace{-1.8cm}
  \includegraphics[width=0.32\linewidth]{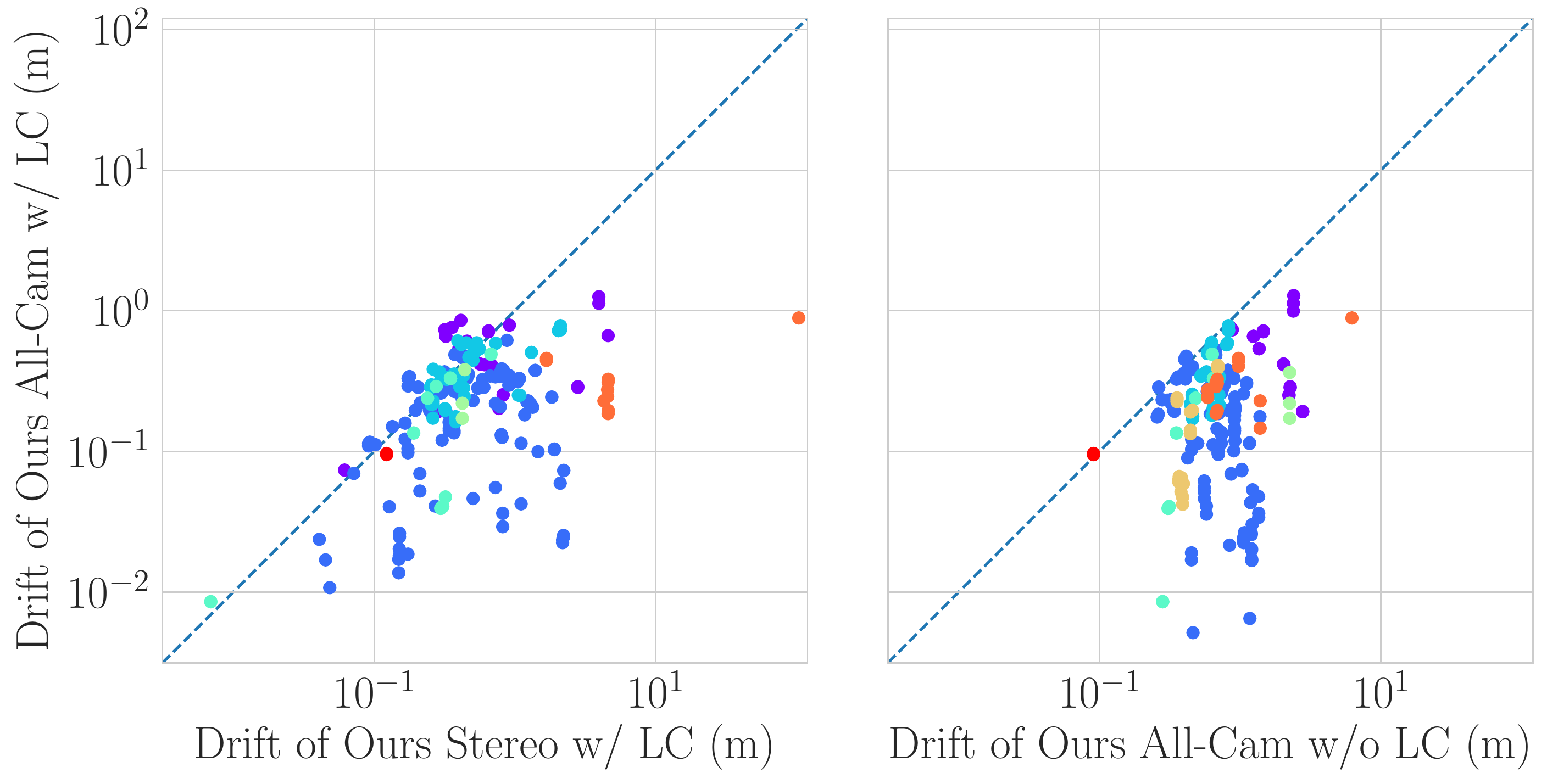}
  \qualtraj{0cm}{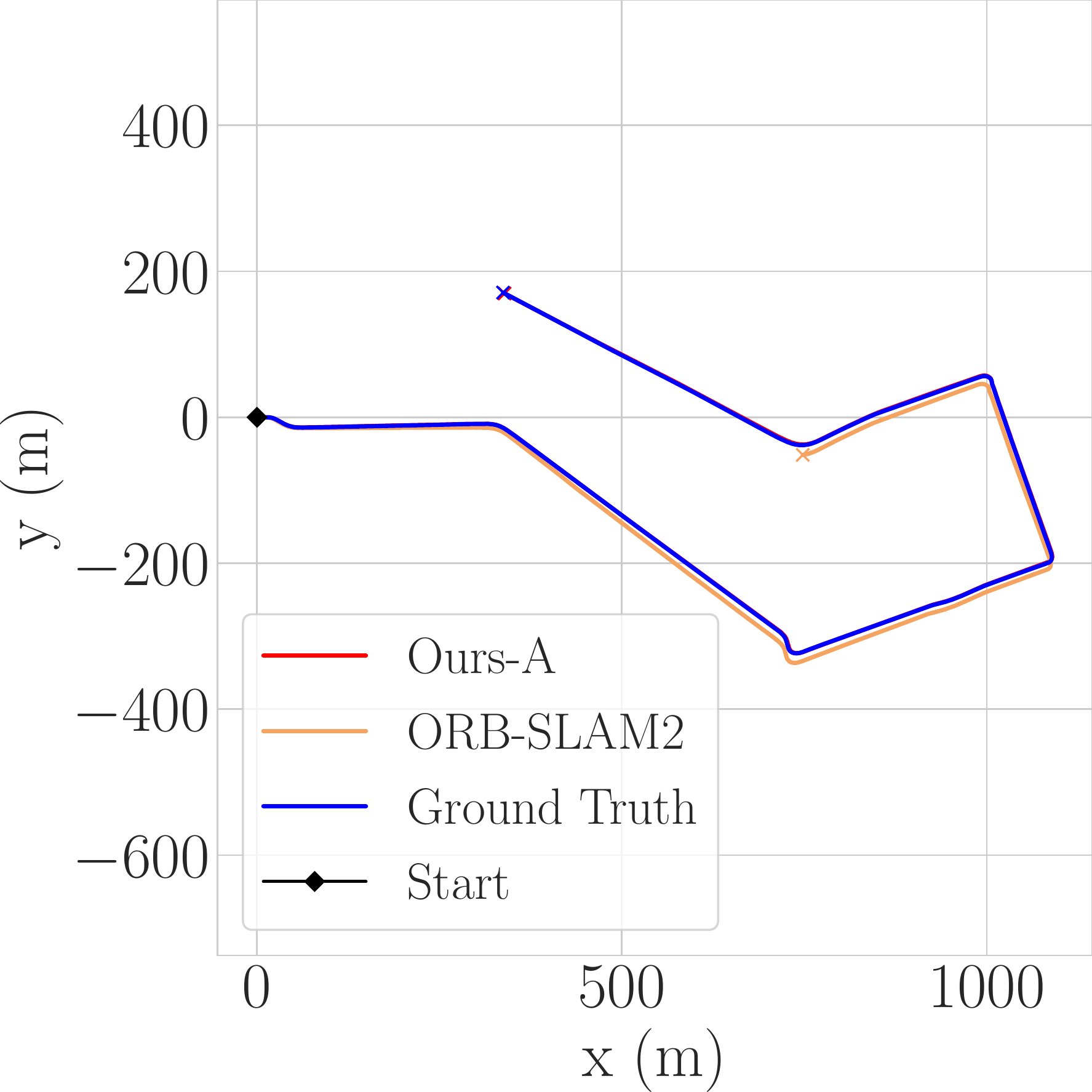}~%
  \qualtraj{0.0cm}{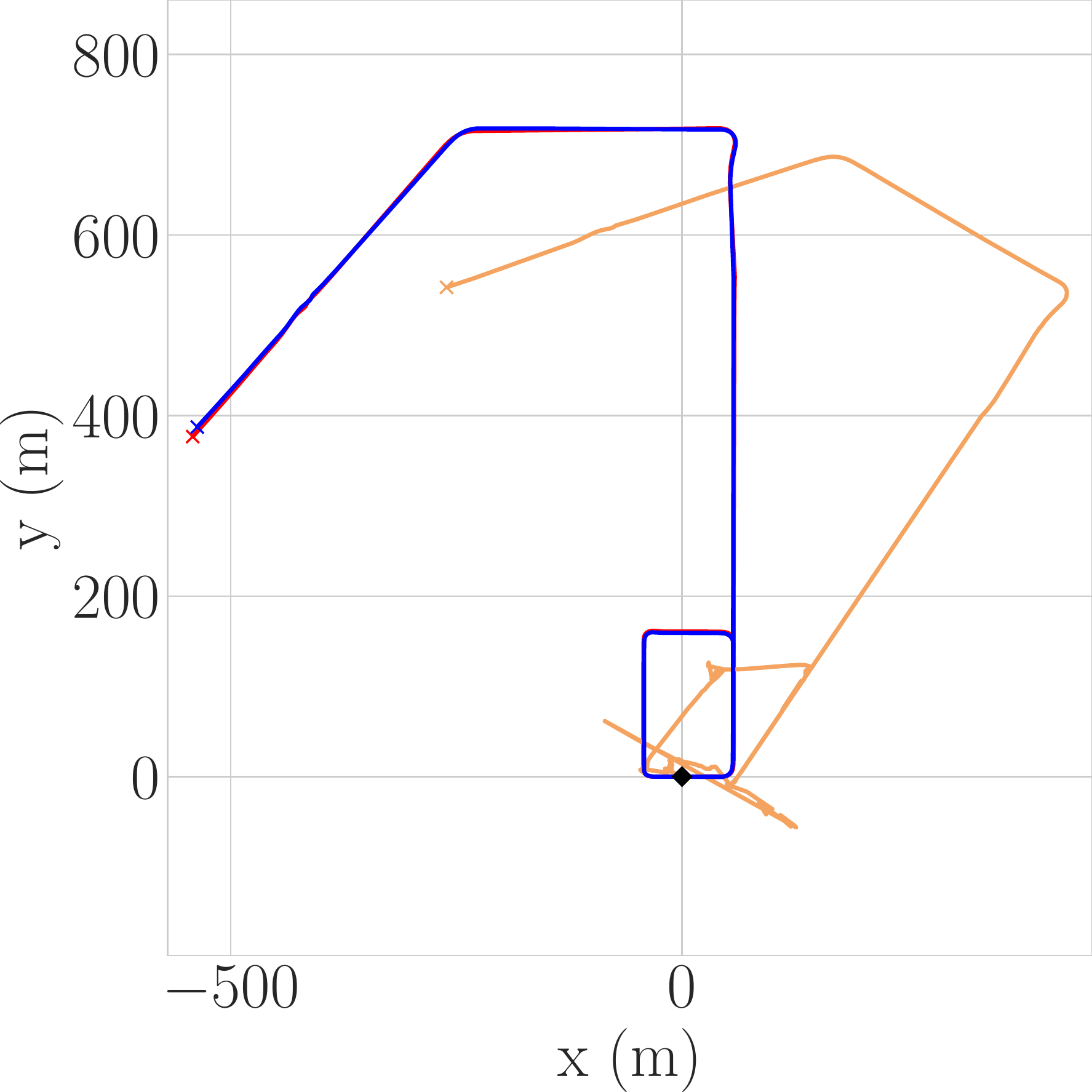}~%
  \qualtraj{0.0cm}{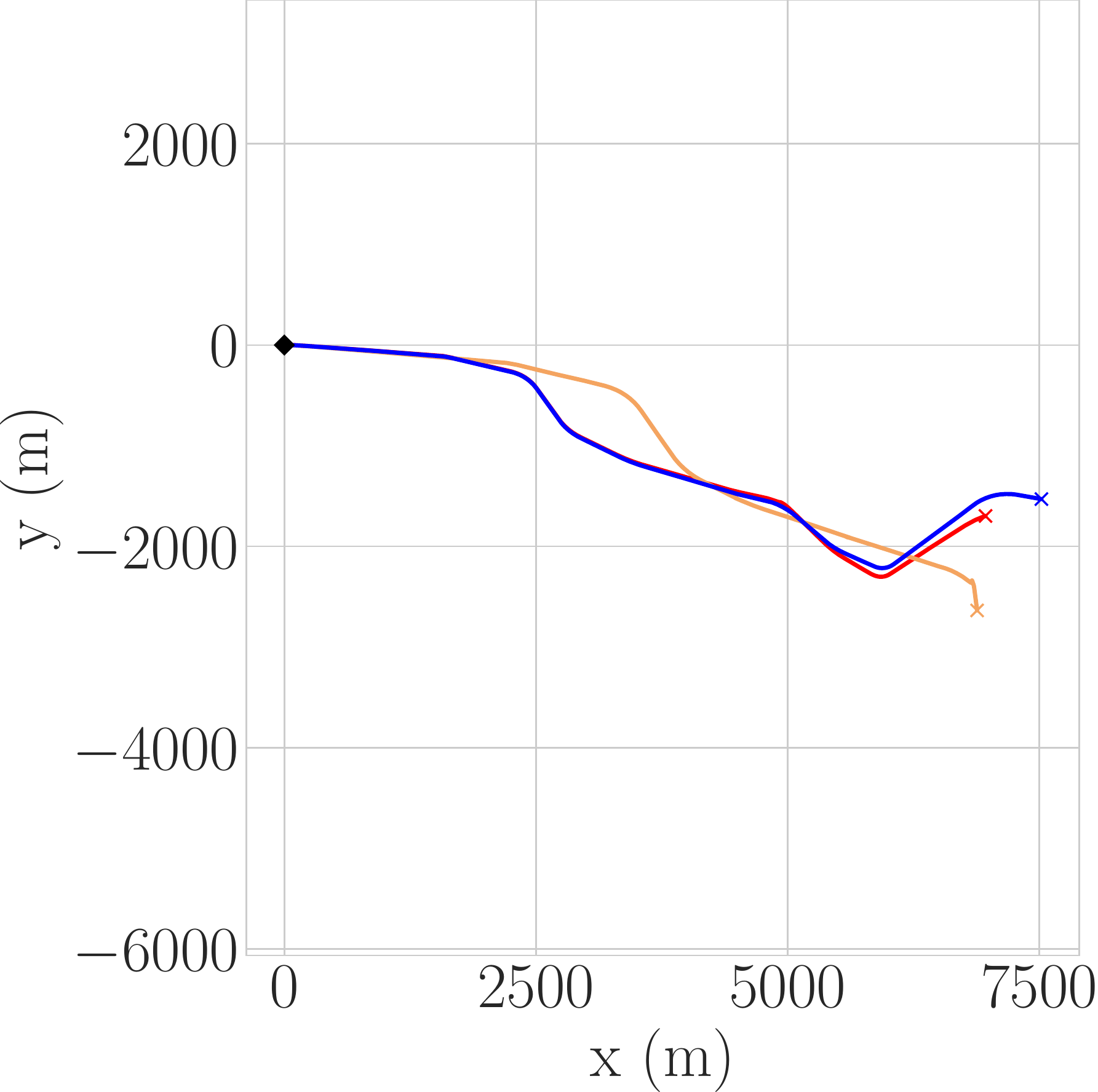}~%
  \qualtraj{0.0cm}{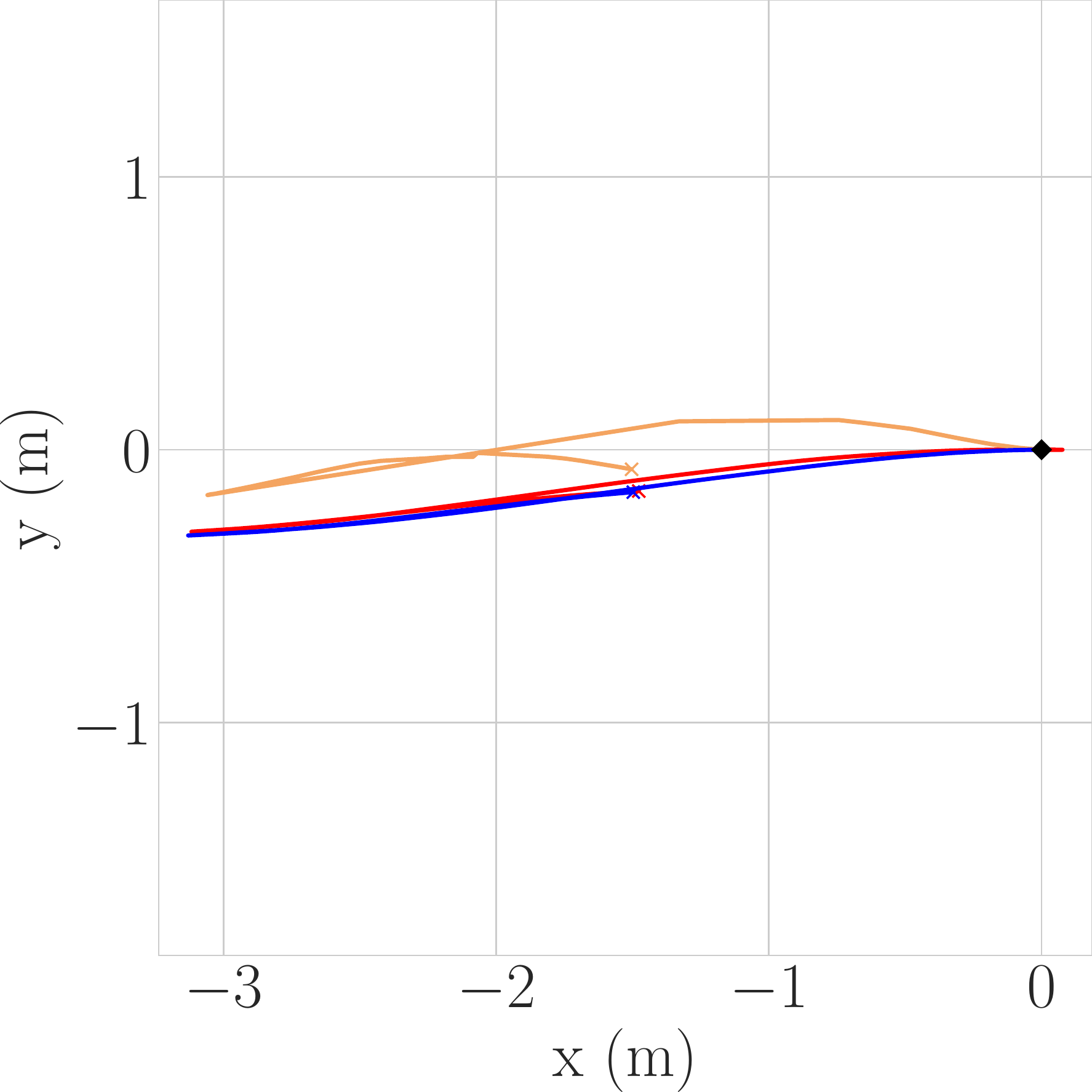}
  \hspace{-1.5cm}
  \caption{(Left) Pose drift (1) after multi-view/stereo loop closure;
  (2) with/without multi-view loop closure.
  Colors correspond to different sequences.
  (Right) Qualitative results. Rightmost is a maneuver reversing into
  a parallel parking spot.%
}%
\label{fig:lc_drift_with_qual_traj}
\vspace{-5mm}
\end{figure*}

\subsection{Experimental Details}

\subsubsection{Implementation Details}
All images are downsampled to 960$\times$600 for both our method as well as all
baselines.
In our system, we extract 1000 ORB~\cite{orb} keypoints from each image,
using grid-based sampling~\cite{mur2015orb} to encourage homogeneous distribution.
Matching is
performed with nearest neighbor + Lowe's ratio test~\cite{sift} with threshold 0.7.
A new KMF is inserted either (1) when the estimated local translation against reference KMF
is over 1m, or local rotation is over 1$^\circ$, or (2) when under 35\% of the map points are re-observed in
at least two camera frames, or (3) when a KMF hasn't been inserted for 20 consecutive MFs.
(3) is necessary to model the spline trajectory when the robot stays
stationary. %
We perform bundle adjustment over a recent window of size $N=11$.

\subsubsection{Experiment Set-Up \& Metrics}
We use the training set (65 sequences) for hyperparameter tuning, and
the validation set (25 sequences) for testing. To account for stochasticity, we run each
 experiments three times.
We use three classic metrics: absolute trajectory error (ATE)~\cite{sturmiros12},
relative pose error (RPE)~\cite{kitti}, and success rate (SR).
SR is the fraction of sequences that were successfully completed without
SLAM failures such as tracking
loss or repeated mapping failures.
We evaluate ATE at 10Hz and RPE at 1Hz. For each method, we report the mean SR
over the three trials.

For a large-scale dataset like \datasetName, it is impractical to list the
ATE and RPE errors for each sequence.
To aggregate results, for each method
we report the median and the area under a cumulative error curve (AUC)
of the errors in all trials at all evaluated timestamps.
Missing entries
due to SLAM failures
are padded with infinity.
AUC is computed between 0 and a given threshold, which we set to be 20cm/m, 5E-04rad/m and 1km for RPE-T, RPE-R and ATE respectively.

\subsection{Results}
\subsubsection{Quantitative Comparison}
We compare our method with multiple popular SLAM methods, including monocular DSO~\cite{engel2017direct,gao2018ldso},
and monocular~\cite{mur2015orb} and stereo~\cite{mur2017orb} ORB-SLAM.
We use the front middle camera in the monocular setting.
We also compare with our discrete-time motion model implementation,
using (1) only the stereo cameras, and (2) all 7 cameras, which corresponds
to a multi-view sync baseline~\cite{urban2016multicol}.
All methods are run with loop closure. %
Table~\ref{tab:quantitative_baseline} shows that our dataset is indeed challenging,
with third-party baselines finishing under 80\% of the validation sequences.
Our method significantly outperforms the rest in terms of accuracy and
robustness. 
Our stereo-sync baseline performs better than~\cite{mur2017orb}
mostly due to the more robust key frame selection strategy.

\subsubsection{Motion Model Ablation}
We run the following experiments using all cameras but with different motion models:
(1) a synchronous discrete-time model;
(2) an asynchronous
linear motion model; %
(3) an asynchronous cubic B-spline model, which is our proposed solution.
Loop closing is disabled for simplicity. Table~\ref{tab:unified-ablation} shows that
the wrong synchronous timing assumption
finishes about 30\% fewer sequences and has
local errors that are $4\times$ as high compared to the main system.
Furthermore, trajectory modeling with cubic B-splines
consistently
outperforms the less expressive linear model.

\subsubsection{Camera Ablation}
We experiment with different camera combinations %
with the
same underlying cubic B-spline motion model. We disable loop closing
for simplicity.
Table~\ref{tab:unified-ablation} indicates a performance boost in all metrics
with more cameras covering a wider field of view.

\subsubsection{Loop Closure}
We first study the effect of multiple cameras on loop detection. %
Out of 9 validation sequences containing loops, our full method using all cameras could
detect 8 loops with 100\% precision, while our stereo loop detection implementation was only able to detect 6 loops closed in the same direction.
The leftmost subfigure of Fig.~\ref{fig:lc_drift_with_qual_traj}
compares the drift after multi-view vs.\ stereo loop closure.
To further highlight the reduction in global trajectory drift,
the second subfigure in Fig.~\ref{fig:lc_drift_with_qual_traj} depicts the drift relative to the ground
truth with and without loop closure at every multi-frame where loop closure was performed.

\subsubsection{Features}

Indirect
SLAM typically uses classic keypoint
extractors~\cite{orb,sift}, yet
recently, learning-based extractors
~\cite{superpoint2018, lfnet2018, d2net19, rfnet2019, r2d22019}
have shown promising results. %
We benchmark
a set of classic~\cite{orb,rootsift} and learned~\cite{superpoint2018,r2d22019} keypoint extractors.
For learned methods, we directly run the provided pre-trained models without re-training. Loop closure is disabled for simplicity.
Our results show that SIFT and SuperPoint %
finish more sequences than ORB ($98.7\%/98.7\%$ SR vs.\ $92.0\%$ for ORB) and have higher %
ATE AUC ($91.5\%/96.3\%$ vs.\ $89.0\%$), with the caveat that feature extraction is
much slower. %

\subsubsection{Runtime}
\label{exp:run_time}
Unlike monocular or stereo SLAM methods,
asynchronous multi-view SLAM requires processing multiple camera images
(seven in our setting). This introduces a linear multiplier
to the complexity of the full SLAM pipeline, including feature extraction,
tracking, bundle adjustment, and loop closure. Thus, despite the significant
improvement on robustness and accuracy, our current implementation 
is not able to achieve real-time operation. Improving 
the
efficiency of AMV-SLAM %
is thus an important area for future research.

\subsubsection{Qualitative Results}
Fig.~\ref{fig:lc_drift_with_qual_traj} plots trajectories
of our method, ORB-SLAM2~\cite{mur2017orb} and the ground truth in selected validation
sequences. Our approach outperforms ORB-SLAM2 and visually aligns well with the GT trajectories in most cases.
We also showcase a failure case from a rainy highway sequence.
For additional quantitative and qualitative results, please see the \suppName{}.

\section{Conclusion}

In this paper, we formalized the problem of multi-camera SLAM with
asynchronous shutters.
Our framework groups input images into asynchronous multi-frames,
and extends feature-based SLAM to the asynchronous multi-view
setting using a cubic B-spline continuous-time motion model.

To evaluate AMV-SLAM systems, we proposed a new large-scale asynchronous multi-camera
outdoor SLAM dataset, \datasetName{}.
Experiments on this dataset
highlight the necessity of the asynchronous sensor modeling, and the
importance of using multiple cameras to achieve robustness and accuracy in
challenging real-world conditions.

\ifarxiv
  \newpage  %
  \section*{APPENDIX}
  \addcontentsline{toc}{section}{Appendix}
  \setcounter{section}{0}
  \renewcommand{\thesection}{\Alph{section}}
  \section{Overview}

The \suppName{} covers additional information on:

\begin{enumerate}
  \item 
    Our method, specifically the details of the linear motion model used during
    tracking, more mathematical details of cubic B-splines,
    as well as further details on the loop closing module.

  \item
    Our dataset and its geographic splits, providing a more in-depth comparison
    to other related SLAM benchmarks.

  \item
    Our experiments, showcasing additional 
    ablation studies, quantitative tables,
    qualitative results and discussions. 

\end{enumerate}

\ifarxiv\else
\noindent\textbf{Acknowledgments}\hspace{0.25cm}
  The authors would like to thank Julieta Martinez, Davi Frossard, and Wei-Chiu
Ma for their valuable input on the writing of the paper, Jack Fan for his
contributions to the metadata code used to select the segments in the
benchmark, and Rui Hu for his help on improving the learned feature
inference speed. We would also like to thank Prof. Tim Barfoot for the detailed
and thoughtful feedback on the project and paper, as well as our anonymous 
ICRA 2021 reviewers for their thorough comments and suggestions.

\fi

  \section{Method}%
\label{sec:method}

\subsection{Asynchronous Multi-Frames}
\begin{figure}[t]
  \centering
  \includegraphics[width=\linewidth]{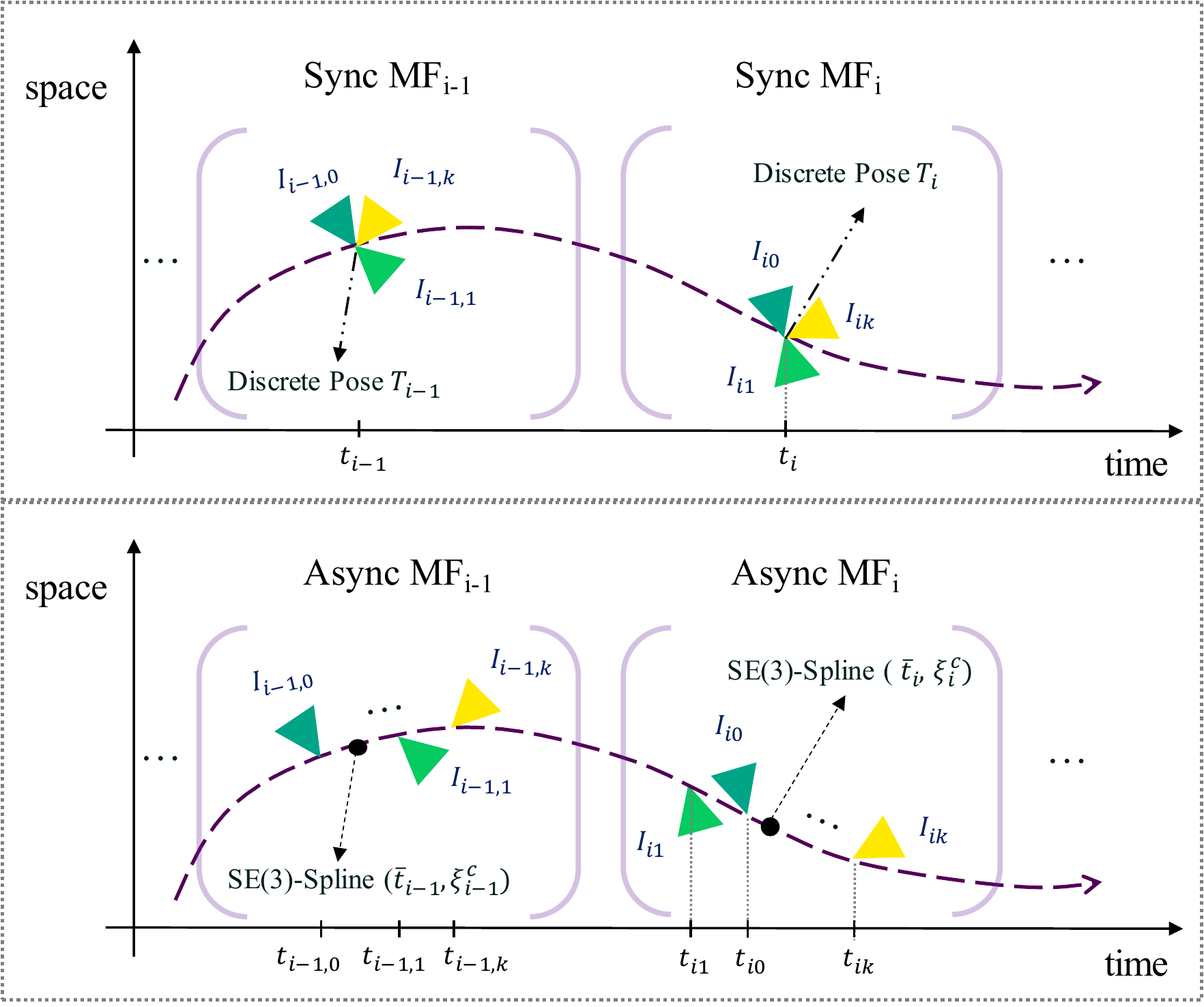}
  \caption{Illustration of the synchronous multi-frame vs. the
  asynchronous multi-frame. For simplicity, in this illustration
  each async multi-frame is assumed to be a key multi-frame.
  Each key multi-frame is associated with cubic B-spline motion
  parameters to define the overall trajectory.}
  \label{fig:kmf}
\end{figure}
We provide an illustration for the concept of 
an asynchronous multi-frame compared to a synchronous multi-frame in
Fig.~\ref{fig:kmf}.

\subsection{Linear Motion Model}
During tracking, we estimate poses in the current asynchronous multi-frame
with a linear motion model, denoted by the superscript $\ell$.
In general, given timestamps $t_1 \le t_2$, and respective associated
poses $\mathbf{T}_1$, $\mathbf{T}_2$, poses at any timestamp $t$ could be linearly interpolated or
extrapolated as
\begin{equation}
\begin{split}
  \mathbf{T}^{\ell}(t) &= \mathbf{T}_2(\mathbf{T}_2^{-1}\mathbf{T}_1)^{\alpha} \\
      & = \mathbf{T}_2\Exp(\alpha\Log(\mathbf{T}_2^{-1}\mathbf{T}_1)), \quad
      \text{where }\alpha = \frac{t_2 - t}{t_2 - t_1}.
\end{split}
\end{equation}
In the context of multi-frames, for each multi-frame $\mf{i}$ with the representative
timestamp $\timet{i}$, we define the linear pose parameter $\pose^{\ell}_i \in \mathbb{R}^6$
to represent the minimal 6-DoF robot pose in the world frame at $\timet{i}$.
It follows that poses at any timestamp $t$ within $\mf{i}$ could be evaluated with

\begin{equation}
\begin{split}
    \mathbf{T}^{\ell}_{wb}(t) &= \bT^{\ell}_{wb}(\timet{i})\Exp\left(\frac{\timet{i}
    - t}{\timet{i}
- \timet{\text{ref}}}\Log\left({\bT^{\ell}_{wb}(\timet{i})}^{-1}\bT^{c}_{wb}(\timet{\text{ref}})\right)\right) \\
&= \Exp(\linpose_i)\Exp\left(\frac{\timet{i}
- t}{\timet{i}
- \timet{\text{ref}}}\Log\left(\Exp(-\linpose_{i})\mathbf{T}_{wb}^c(\timet{\text{ref}})\right)\right),
\end{split}
\end{equation}
where $\timet{\text{ref}}$ and $\mathbf{T}_{wb}^c(\timet{\text{ref}})$
are the respective representative timestamp and evaluated cubic B-spline pose at $\timet{\text{ref}}$
of the reference multi-frame $\mf{\text{ref}}$.
In practice, new MFs are tracked against a reference
\textit{key} multi-frame, so \emph{ref} refers to the MF id of the most recent KMF.

\subsection{Cubic B-Spline Model}
We use a cumulative cubic B-spline motion model over key multi-frames to
represent the overall trajectory. We use the linear motion model parameters
estimated during tracking to initialize cubic B-spline control points, 
and refine the spline trajectory during mapping and loop closing.
In general,
given $n+1$ control points $\cbpose_0, \ldots, \cbpose_n \in \mathbb{R}^6$,
and a knot vector $\mathbf{b} \in \mathbb{R}^{n+k+1}$,
the cumulative B-spline of order $k$ is defined as:
\begin{equation}
  \label{eq:bspline}
 \mathbf{T}_{wb}^c(t) =
 \Exp \left(\widetilde{B}_{0,k}(t) \cbpose_{0} \right)
 \prod_{i=1}^{n} \Exp \left( \widetilde{B}_{i,k}(t) \pmb{\Omega}_i \right),
\end{equation}
where $\pmb{\Omega}_i = \Log\left(\Exp(\cbpose_{i-1})^{-1} \Exp(\cbpose_i)\right) \in \mathbb{R}^6$
is the relative pose in Lie Algebra twist coordinate form between control poses
$\cbpose_{i-1}$ and $\cbpose_i$. The superscript $c$ is used to denote the
cubic B-spline motion model.
The cumulative basis function $\widetilde{B}_{i,k}(t) = \sum_{j=i}^{n} B_{j,k}(t) \in \mathbb{R}$
is the sum of basis function $B_{j,k}(t)$.
Based on the knot vector $\mathbf{b}=\begin{bmatrix}b_0 &\ldots &b_{n+k}\end{bmatrix}$,
the basis function $B_{j,k}(t)$ is computed using the de Boor-Cox recursive
formula~\cite{de1972calculating, cox1972numerical},
with the base case
\[B_{p,1}(t) = \begin{cases}
   1 \qquad \text{if $t \in [b_p, b_{p+1}]$}\\
   0 \qquad \text{otherwise}
 \end{cases}.\]
 For $q \ge 2$,
 \[B_{p,q}(t) = \frac{t-b_p}{b_{p+q-1}-b_p}B_{p,q-1}(t) + \frac{b_{p+q}-t}{b_{p+q}-b_{p+1}}B_{p+1,q-1}(t).\]
 More intuitively, each $\mathbf{T}_{wb}^c(t)$ can be interpreted as an $n$-way
 interpolation between the control points $\cbpose_i$ with respective interpolation weight
 $B_{i,k}(t)$.
 However, instead of directly interpolating
 $\mathbf{T}_{wb}^c(t) = \prod_{i=0}^n \Exp(B_{i,k}(t)\cbpose_i)$,
 we use the cumulative formulation in Eq.~\ref{eq:bspline}
 for accurate on-manifold interpolation in $\SE{3}$~\cite{kim1995general,lovegrove2013spline}.

Since we use cubic B-splines, $n=3$ and $k=4$. In the context of multi-frames,
we associate each \textit{key} multi-frame $\text{KMF}_i$ with a control point
$\cbpose_i$.
In addition, since the key multi-frames are not necessarily distributed evenly in
time, we cannot utilize a uniform knot vector (as typically employed
for modeling rolling-shutter
cameras~\cite{lovegrove2013spline} and LiDARs~\cite{behnke2018lidar}). Instead, we
associate each $\text{KMF}_i$ with a \textit{representative timestamp} $\timet{i}$
as the median of all image capture times $t_{ik}$ (with the exception of
initialization,
where $\timet{0}$ is defined at the firing time of the overlapping camera pair).
We define a non-uniform knot vector according to the representative timestamps. %
Specifically, we define
$\mathbf{b} = \begin{bmatrix} \timet{i-3} & \timet{i-2} &\ldots & \timet{i+4} \end{bmatrix} \in \mathbb{R}^8$.

\subsection{Loop Detection}
When a new KMF is selected, we run loop detection to check if a previously-seen area is being revisited.
For computational efficiency, we only perform loop detection if the most recent
successful loop correction took place at least a minimal number of KMFs ago.
In our implementation we set this threshold to 30.

For a newly-inserted query $\text{KMF}_q$, the loop candidate $\text{KMF}_l$ must
pass an odometry check, a multi-camera DBoW3-based~\cite{GalvezTRO12DBoW3}
similarity check, and a multi-camera geometric verification. We detail this process
in the following subsections.
\subsubsection{Odometry Check}
To avoid false loop detection when the robot is staying in the same area,
the odometry check ensures that the robot must have traveled a minimal distance
since the loop candidate frame. In addition, a minimal time and a minimal number
of key frames must have passed since the candidate frame as well.
The traveling distance is computed based on
the estimated trajectory.
The time and key frame count conditions serve as complements in the case when the
estimated traveling distance is unreliable.
In our experiment,
we set the traveling distance threshold to 30m, time to 5s, and the number of
KMFs to 30. Note that the KMF threshold in odometry check is different from
the KMF threshold described at the beginning of the loop detection section.
The former specifies that a candidate KMF must be older than 30 KMFs ago,
while the latter dictates that we will only perform loop detection for
the current query KMF if the latest loop closing happened at least 30 KMFs ago.

\subsubsection{Similarity check}
For candidates passing the odometry check, we perform a multi-view version of the
single-view DBoW3-based similarity check described in ORB-SLAM~\cite{mur2015orb}. 
The key idea is that images in the loop candidate KMF and the query KMF
should have similar appearance. We perform similarity detection
with the bag-of-words techniques DBoW3 for place recognition~\cite{GalvezTRO12DBoW3}.
For each key multi-frame, we concatenate features extracted from
all images in the multi-frame to build the DBoW3 vocabulary and compute the DBoW3 vector.
Note that the simple concatenation does not take into account of the fact that cameras
can be asynchronous, but we argue that the same area should have
similar appearance within the short camera firing time interval, and
false positives will be filtered by the stricter geometric verification
that factors in asynchronous sensors in the next step.

During the similarity check, we first compute the DBoW3 similarity score between
the query KMF and the neighboring KMFs that are included in the associated local
bundle adjustment window,
and we denote the minimum similarity score as $m$. Next, we compute the
similarity score between the query KMF and all available candidate KMFs
and denote the top score as $t$. Then all the candidates that pass the
similarity check must have a similarity score that is greater than $\max(0.01, m, 0.9 * t)$.

\subsubsection{Geometric check}
For each remaining candidate KMF, we perform a geometric check by directly solving
for a relative pose between cameras in the candidate KMF and cameras in the query KMF.
To identify the camera pairs to be matched, let us consider a setting where the camera
rig contains a set of $M$ cameras covering a
combined 360$^\circ$ FoV, with the cameras denoted as $\cC_1,\ldots,\cC_M$ in the clockwise order.
We also assume that the robot is on the ground plane in this setting, \ie, the roll
and pitch angles of the robot poses remain the same when the robot revisits
the same area.
Since a loop can be encountered at an arbitrary yaw angle, there are a total of $M$ possible
scenarios of how the multiple cameras between the candidate and the query frame can be matched,
where in scenario $i$, each camera $\mathcal{C}_j$ in the candidate frame
is matched to camera $\mathcal{C}_{(j+i)\%M}$ in the query frame.

For each possible matching scenario involving $M$ pairs of cameras,
we solve for a discrete relative pose between each camera pair.
Specifically, for each pair of cameras, we first perform keypoint-based sparse image matching between
the associated image pair by fitting an essential matrix in
a RANSAC~\cite{fischler1981random} loop.
If the number of inlier matches passes a certain threshold,
we associate the inlier matches with the existing 3D map points.
Note that different from tracking, here we draw associations in both directions:
2D keypoints in the loop candidate image are associated to 3D map points observed
in the query image, and vice versa.

If the number of such keypoint-to-map-point correspondences passes a threshold, 
we estimate a single relative pose in $\SE{3}$ between
the two cameras.
Following~\cite{mur2015orb}, we perform pose estimation with Horn's method in a RANSAC loop, where within each
RANSAC iteration we sample a minimal number of matches, and solve for the discrete pose by
minimizing a reprojection error.
The hypothesis with the most number of inliers is the final estimate.

The geometric check passes if at least a certain number of camera pairs have a minimum
number of inliers for the relative pose estimation. In our full system, we perform
geometric check with the $M=5$ wide cameras covering the surroundings of the vehicle.
We consider a geometric check to be
successful if there exists a matching scenario where
we can successfully estimate the discrete relative pose for
at least 2 pairs of cameras, where for each camera pair there are
at least 20 inlier correspondence pairs during
sparse image matching, 20 pairs of 2D-3D associations,
and 20 pairs of inlier correspondences
after the relative pose estimation.
If there are multiple matching scenarios that pass the check, we
select the configuration with the most successfully matched
camera pairs and the most inlier correspondences.

The multi-camera geometric verification outputs
$\{(\mathcal{C}_{k_l}, \mathcal{C}_{k_q}, \mathbf{T}_{b_{k_q},b_{k_l}})\}$, which
is a set of triplets denoting the camera indices of the matched camera pairs
between the loop and
the query frames, along with $\mathbf{T}_{b_{k_q},b_{k_l}}$, which is
an estimated rigid-body transformation from the body frame $\mathcal{F}_b$ at the camera
capture time $t_{lk_l}$ to $\mathcal{F}_b$ at time $t_{qk_q}$.

\subsection{Loop Correction}
\newcommand{\Erel}{E_\text{rel}}
\newcommand{\Ereg}{E_\text{reg}}

\begin{figure*}
  \centering
  \includegraphics[width=.8\linewidth]{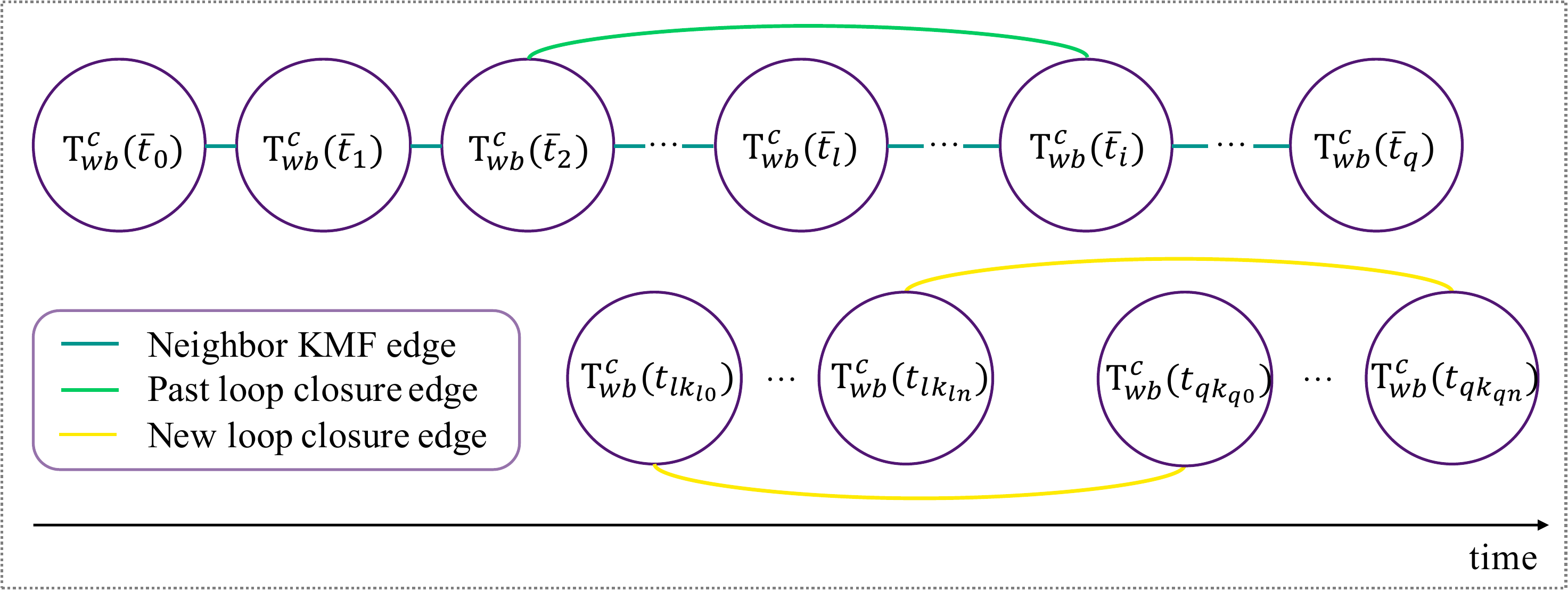}
  \caption[Illustration of the loop correction pose graph]
  {Illustration of the loop correction pose graph.
  The nodes correspond to robot poses at the representative timestamps
  of all KMFs + capture times of the matched cameras in the new loop closure KMFs. The
  edges consist of neighboring KMF edges, past loop closing edges,
  and the new multi-view loop closing edges. If $n$ camera pairs are successfully
  matched during the loop detection stage, then there should be $n$ new loop closing edges.}
  \label{fig:factor_praph}
\end{figure*}

If a loop candidate $\kmf{l}$ passes all loop detection checks,
we perform loop correction with the geometric verification output.
We first build an asynchronous multi-view version of the pose
graph in ORB-SLAM~\cite{mur2015orb}. Each node $\alpha$
of the pose graph is encoded by a timestamp $t_\alpha$
representing the underlying robot pose $\mathbf{T}_{wb}(t_\alpha)$.
Each edge $(\alpha, \beta)$
encodes the relative pose constraint $\mathbf{T}_{\beta\alpha}$
representing the rigid transformation from
$\mathbf{T}_{wb}(t_\alpha)$ to $\mathbf{T}_{wb}(t_\beta)$.

Specifically, in our pose graph, the nodes are associated with times at
$\{\timet{i}\}_{\forall\kmf{i}} \cup \{t_{lk_l}, t_{qk_q}\}_{\forall(k_l, k_q)}$,
\ie, the representative timestamps of all existing KMFs, as well as the
camera times from matched camera pairs in the geometric verification output.
The edges are comprised of: (1) neighboring edges connecting adjacent KMF nodes
at times $\timet{i-1}$ and $\timet{i}$,
(2) past loop closure edges connecting nodes associated with past query and loop closure KMFs,
and (3) the new loop closure edges between nodes at time $t_{lk_l}$
and $t_{qk_q}$.
For edges in case (1) and (2), we compute the relative pose $T_{\beta\alpha}$
by evaluating $(\mathbf{T}^c_{wb}(\timet{\beta}))^{-1}\mathbf{T}^c_{wb}(\timet{\alpha})$
based on the current trajectory. For (3), we use the discrete poses $\mathbf{T}_{b_{k_q},b_{k_l}}$
estimated in the geometric verification step in loop detection.
Please refer to Fig.~\ref{fig:factor_praph} for an illustration of the pose
graph.

We denote the local KMF windows spanning the query and
loop frames as the \emph{welding windows}. In our implementation they are the
same size as the local bundle adjustment window.
To correct the global drift, we perform a pose graph optimization (PGO) over the
continuous-time cubic B-spline trajectory.
To better anchor the trajectory, the control poses in the
welding window associated to $\text{KMF}_l$ are fixed during the pose graph
optimization, where the following objective is minimized:
\begin{equation}
  E_{\text{PGO}}(\{\cbpose_i\}) = \Erel(\{\cbpose_i\}) + \Ereg(\{\cbpose_i\}),
\end{equation}
where
\begin{equation}
\begin{split}
  &\Erel(\{\cbpose_i\}) =
  \sum_{(\alpha, \beta)}{
    \rho\left(\left\|\be_{\alpha,\beta}^T(\{\cbpose_i\})\right\|^2_{\mathbf{\Sigma}_{\alpha,\beta}^{-1}}\right)
  },\\
  &\text{with $\be_{\alpha,\beta}(\{\cbpose_i\}) = \Log(\mathbf{T}_{\beta\alpha}(\mathbf{T}_{wb}^c(t_\alpha))^{-1}\mathbf{T}^c_{wb}(t_\beta))\in\mathbb{R}^6$}
\end{split}
\end{equation}
sums over the relative pose errors of each edge weighted by an uncertainty term $\mathbf{\Sigma}^{-1}_{\alpha,\beta}$, and
\begin{equation}
\begin{split}
  &\Ereg(\{\cbpose_i\}) = \sum_{i}{\rho\left(\left\|\mathbf{r}_i^T(\{\cbpose_i\})\right\|^2_{\mathbf{\Lambda}^{-1}_i}\right)},\\
  &\text{with ${\mathbf{r}_i^T(\{\cbpose_i\}) = \Log\left(\bT_i^{-1}\bT_{wb}^c(\timet{i})\right)}\in\mathbb{R}^6$}
\end{split}
\end{equation}
is a unary regularization term weighted by uncertainty $\mathbf{\Lambda}^{-1}_i$ to anchor
each KMF's representative pose at
$\bT_i=\bT_{wb}^c(\timet{i})$ evaluated before the optimization.
Empirically, we set the diagonal entries of
both $\mathbf{\Sigma}^{-1}_{\alpha,\beta}$ and $\mathbf{\Lambda}^{-1}_i$ to 1.0.
The regularization term helps to better constrain the optimization especially when
there is a large loop (\ie, $q$ is much bigger than $l$) and a large
amount of drift to correct. $\rho$ is the robust norm and we again use
the Huber loss in practice.
The energy term is minimized with
the LM algorithm.

After PGO, we next update the map points with the adjusted trajectory.
If a map point is observed in multiple images,
we update the map point position with the median of all pose corrections
related to the map point.
Note that ideally, we would want to solve a global bundle adjustment problem
that jointly refines the entire trajectory and all map points at the same time.
However, with a long trajectory and many observations from multi-view cameras,
global bundle adjustment becomes computational expensive or even infeasible.
The two-stage process described above, where we first optimize the poses and
then update the map points, is a light-weight approximation that is sufficient under most
circumstances.

Furthermore, note some new map points in the query KMF window have been created during
recent local mappings, but they may correspond to points already triangulated in the previous pass through
the revisited area.
As a result, we next deduplicate the re-triangulated map points.
We first match image pairs in the candidate KMF welding
window and the query KMF welding window to identify and fuse these map points.
We then perform a local bundle adjustment over the motion and map points corresponding
to the two
welding windows. The purpose of the local bundle adjustment is to refine both map points and
control poses in the query KMF window. To anchor candidate poses,
we freeze the control poses corresponding to the candidate welding
window in the optimization.
  \section{Dataset}

\begin{figure}[]
    \centering
    \includegraphics[width=0.9\linewidth]{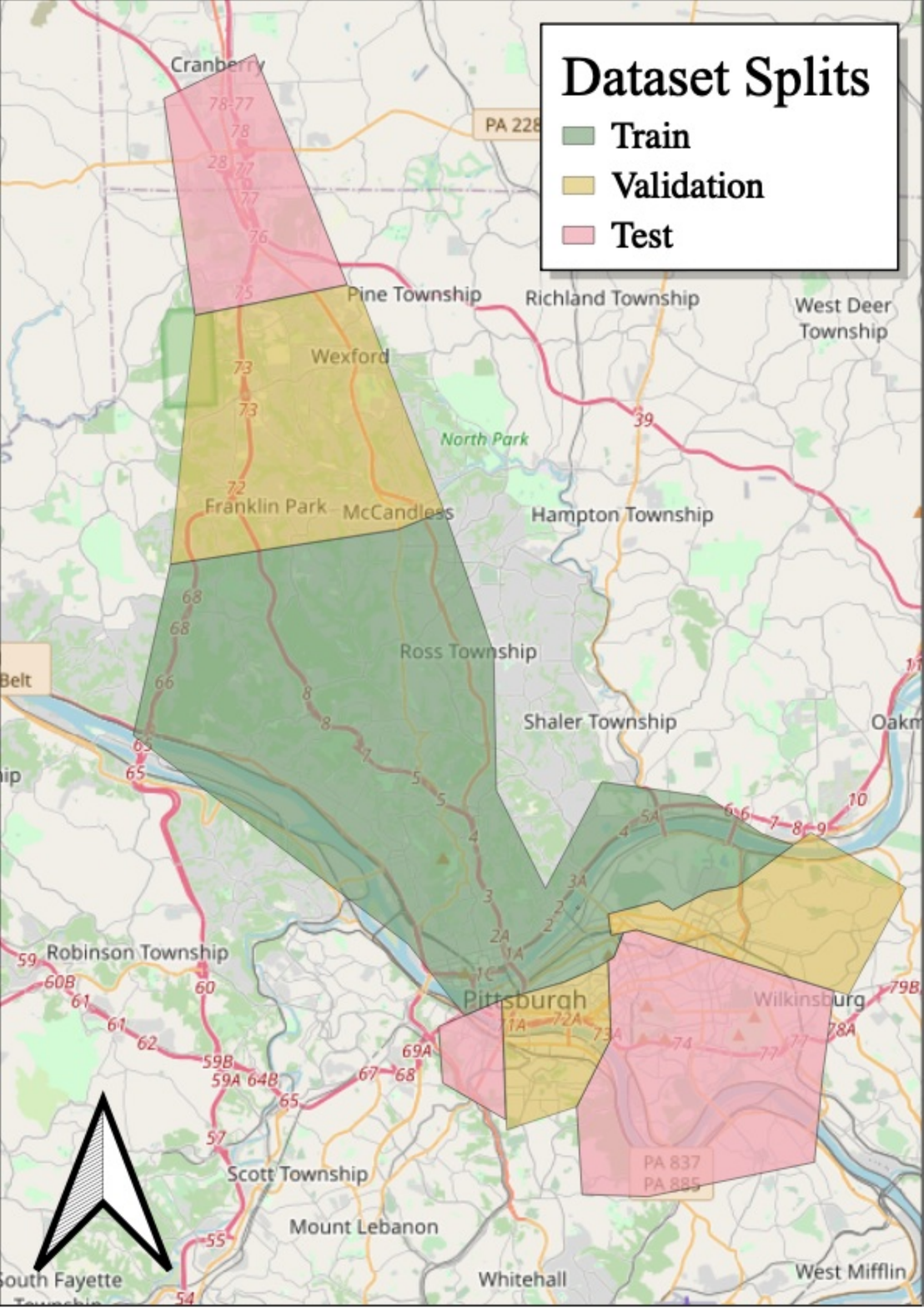}
    \caption{The geographic splits of the proposed datasets. They are designed
    such that train, val, and test each covers a balanced blend of environment
    types (highway, urban, industrial, residential, etc.). The splits are also
    selected to have similar distributions of weather, loop closures, etc.}%
    \label{fig:geosplits}
  \end{figure}

\subsection{Existing SLAM Benchmarks}
As described in the main paper, existing SLAM datasets fall short in terms of
geographic diversity, modern sensor layouts, or scale. In this section, we
describe the most relevant modern SLAM datasets together with their primary
limitations.

The KITTI Odometry Benchmark~\cite{kitti} covers 40km of driving through
Karlsruhe, Germany using a vehicle equipped with a stereo camera pair,
a 64-beam LiDAR, IMU, and RTK-based ground truth. However, most sequences in
the dataset have relatively small numbers of dynamic objects, and all the data
is captured in sunny or overcast weather, which is not representative of the
variety of conditions which can be encountered by commercial AVs. The NCLT
dataset~\cite{nclt} covers a larger distance in the University of Michigan
campus using a custom-built robotic Segway equipped with three LiDARs an IMU and an
omnidirectional camera. While the scale of the dataset is large, its geographic
diversity is lacking, being constrained to a university campus, while at the
same time not capturing the same challenging motion patterns which would be
encountered by an SDV. 

The Oxford RobotCar~\cite{maddern2017robotcar} dataset covers over 1000km
of driving in challenging conditions containing a large number of dynamic objects
as well as strong weather and lighting variation. However, since it is focused on a single
primary trajectory it lacks geographic diversity. Moreover, it does not provide
360$^\circ$ camera data in HD, which is critical for SDVs. Similarly, the Ford
Multi-AV Dataset~\cite{agarwal2020ford} contains a large volume of data but is
focused on a relatively limited 60km route which is traversed repeatedly by
multiple SDVs. The A2D2 Dataset~\cite{geyer2020a2d2} contains a high-resolution
multi-camera multi-LiDAR setup optimized for 3D reconstruction. However, the
approximate total scale of A2D2 is still on the order of a few hours of
driving, which is insufficient for evaluating robust SLAM system in a wide
variety of challenging conditions.

Finally, the recent 4Seasons dataset~\cite{wenzel20204seasons}
covers diverse areas in a wide range of environments (highway,
industrial, residential, etc.) over a long time period, but lacks the HD
multi-view sensor array common in commercial SDVs.

Note that in the dataset overview table in the main paper we label A2D2 and
Ford Multi-AV as non-asynchronous dataset on the basis that their cameras are
not described as following the LiDAR or any custom firing pattern causing more
than 2--3ms of delay. While Ford Multi-AV does have cameras firing at different
frame rates, with the higher-resolution front stereo operating at 5Hz and the
other cameras at 15Hz, we do not consider this as a true asynchronous setting.

Additionally, even though datasets such as Waymo Open~\cite{waymo_open_dataset} and
nuScenes~\cite{nuscenes2020} include asynchronous cameras, they are focused on
perception tasks and contain short sequences (e.g., less than a minute each).
Therefore, we
do not consider them in our evaluation as they are too short to robustly
evaluate SLAM algorithms. %

\subsection{Dataset Details}
The dataset has been selected using a semi-automatic curation process
to ensure all splits cover similar categories of environments, while at the
same time having substantial appearance differences between the splits.

Table~\ref{tab:dataset-stats} shows the high-level statistics of the train,
validation, and test partitions of the dataset. Figure~\ref{fig:geosplits}
shows the geographic regions of the splits.

The cameras are all RGB and have a resolution of 1920$\times$1200 pixels,
a global shutter; the (asynchronous) shutter timestamps are recorded with
the rest of the data.
The wide-angle and narrow-angle cameras have rectified FoVs of
76.6\deg$\times$ 52.6\deg and 37.8\deg$\times$ 24.2\deg, respectively.

Furthermore, the 6-DoF ground-truth
was generated using an offline HD-map based localization system, which enables
SLAM systems to be evaluated at the centimeter level.

\begin{table}
  \centering
  \caption{The proposed dataset, \datasetName{}, in numbers.}
  \label{tab:dataset-stats}
  \begin{tabular}{lrrr}
    \toprule
    \textbf{Split} & \textbf{Sequences} & \textbf{Distance (km)} & \textbf{Time (h)} \\
    \midrule
    Train       &  65   &  281  & 12  \\
    Validation  &  25   &  103  & 4   \\
    Test        &  26   &   98  & 5   \\
    \midrule
    Total & \totalSnippets{} & \totalKm{} & \totalHours{} \\
    \bottomrule
  \end{tabular}
\end{table}

  \section{Additional Experiments}%
\label{sec:additional_experiments}

\subsection{Full Implementation Details}
All images are downsampled to 960$\times$600 for both our method as well as all
baselines.
In our system, we extract 1000 ORB~\cite{orb} keypoints from each image,
using grid-based sampling~\cite{mur2015orb} to encourage homogeneous distribution.
Image matching is performed with nearest neighbor + Lowe's ratio test~\cite{sift}
with a ratio threshold of 0.7.
The initial 2D matches between each image pair are additionally filtered by
fitting an essential matrix with RANSAC. The inlier correspondences are
used (1) as input to the multi-view PnP during tracking, (2) for new map point
triangulation during mapping, (3) for geometric verification during loop closing.

In our system, we use the synchronous stereo camera pair during initialization.
During tracking, we match image pairs captured by the same camera.
During new map point creation, we match images captured by the same camera between the new KMF
and four previous KMFs, and triangulate new map points based on the 2D matches.
We additionally
triangulate new map points from the stereo cameras within the new key multi-frame.
Note that we do not match between different wide cameras in the same multi-frame
due to little overlap between them and ORB's reduced performance in wide baseline image
matching settings. Please see Sec.~\ref{sec:additional_quant_experiments} for details.

During tracking, we randomly sample 7 pairs of correspondences within each PnP RANSAC loop
to solve for a hypothesis.
The pose estimate hypothesis with the most number of inliers becomes the final
estimate.

During key frame selection, a new KMF is inserted either (1) when the estimated local translation against reference KMF
is over 1m, or local rotation is over 1$^\circ$, or (2) when under 35\% of the map points are re-observed in
at least two camera frames, or (3) when a KMF hasn't been inserted for 20 consecutive MFs.
Note that (3) is necessary to model the spline trajectory when the robot stays
stationary.

We perform bundle adjustment over a recent window of size $N=11$.
We cull map points with reprojection error over 1.5 pixels.

Following~\cite{mur2015orb}, the uncertainty weighting $\Sigma$ in both tracking
and bundle adjustment is based on the scale level where ORB features are extracted.
Keypoints extracted from larger/coarser scale levels are less precise and therefore
correspond to higher uncertainty and lower weighting during pose estimation.

During the pose graph optimization in loop closure, the uncertainty weighting for
the relative pose error terms and the regularization terms are all set to 1.0.

In our system and all our ablation implementations,
we declare a tracking failure if the estimated pose parameters
yield under 12 inlier PnP correspondences in total. We declare bundle adjustment
failure and not apply the bundle adjustment update
if after bundle adjustment any of the pose parameters is changed by
more than 6 meters or 20 degrees. We selected these values empirically based on
training set performance.
We stop the system in the middle of processing a sequence
if there are at least 5 successive tracking failures, or if there are at least 5
successive bundle adjustment failures.

We use the Ceres Solver~\cite{ceres-solver} for modeling and solving
the non-linear optimization problems arising in tracking, bundle adjustment, and loop closure.

\subsection{Third-Party Baseline Experiment Details}
For all ORB-SLAM~\cite{mur2015orb,mur2017orb} experiments, we lowered the default tracking
failure threshold from 30 matching inliers to 10 matching inliers. This is to
increase the tolerance for tracking failures, as the system with 10--30 matching
inliers was
able to complete larger portions of the training sequences without much compromise of
local tracking errors.
Apart from the inlier threshold,, we use all other default hyperparameters
provided for the KITTI experiments by the authors\footnote{\url{https://github.com/raulmur/ORB_SLAM2}},
which extract 2000 ORB
keypoints per image, while we only extract 1000 ORB keypoints for our system and all our
baseline and ablation study implementations.

We use the default KITTI hyperparameters for LDSO~\cite{gao2018ldso}
provided by the authors\footnote{\url{https://github.com/tum-vision/LDSO}}.

For the keypoint extractor ablation study, we extract 1000 keypoints and
associated features from each image for all methods. We match features with Lowe's ratio test.
The ratio is tuned on the training set. We use 0.7 for ORB~\cite{orb} and
RootSIFT~\cite{sift, rootsift}, and 0.8 for SuperPoint~\cite{superpoint2018} and
R2D2~\cite{r2d22019}. For SuperPoint, we run the provided pre-trained
model~\footnote{\url{https://github.com/rpautrat/SuperPoint}}. For R2D2,
we run the provided pre-trained {\tt r2d2\_WASF\_N16} model~\footnote{\url{https://github.com/naver/r2d2}}.

\subsection{Metrics}
In the additional experiments, aside from reporting the median and AUC for
the aggregated ATE and RPE results, we also report the ATE and RPE errors
at the 90th percentile, \ie $x$ with $f(x) = 0.9$ where $f$ is the cumulative error curve.
Compared to median (the 50th percentile), the 90th percentile metric better
characterizes outlier behavior, and the AUC metric gives a better characterization
of the overall performance, while being able to model system failures.

\subsection{Quantitative Results}
In this subsection we show additional details on the main paper results,
as well as additional ablation studies.
\label{sub:quantitative_results}

\subsubsection{Detailed main paper results}
\label{sec:additional_quant_experiments}
\begin{table*}
  \centering
  \caption{Baseline methods. M=monocular, S=stereo, A=all cameras.}
  \label{tab:quantitative_baseline}
  \begin{tabular}{l|rrr|rrr|rrr|r}
    \toprule
    \multirow{2}{*}{\textbf{Method}} &
    \multicolumn{3}{c|}{\rt\space(cm/m)} &
    \multicolumn{3}{c|}{\rr\space(rad/m)} &
    \multicolumn{3}{c|}{\ate\space(m)} &
    \multirow{2}{*}{\sr\space(\%)} \\
    & @0.5 & @0.9 & AUC(\%) & @0.5 & @0.9 & AUC(\%) & @0.5 & @0.9 & AUC(\%) & \\
    \midrule
    LDSO-M~\cite{gao2018ldso} & 42.72 & - & 28.08 & 8.02E-05 & - & 54.23 & 594.39 & - & 44.67 & 62.67 \\
    ORB-M~\cite{mur2015orb} & 34.00 & - & 25.66 & 5.49E-05 & - & 63.77 & 694.37 & - & 42.65 & 64.00 \\
    ORB-S~\cite{mur2017orb} & 1.85 & - & 65.70 & 3.29E-05 & - & 70.47 & 30.74 & - & 74.31 & 77.33 \\
    Sync-S & \second{1.30} & - & \second{77.54} & \second{2.91E-05} & - & \second{78.37} & \second{24.53} & - & \second{77.44} & \second{84.00} \\
    Sync-A & 2.15 & - & 68.46 & 3.47E-05 & - & 70.47 & 58.18 & - & 75.01 & 74.67 \\
    Ours-A & \best{0.35} & \best{1.99} & \best{88.63} & \best{1.13E-05} & \best{6.50E-05} & \best{88.17} & \best{6.13} & \best{322.95} & \best{88.82} & \best{92.00} \\
    \bottomrule
  \end{tabular}
\end{table*}

\begin{figure*}
    \centering
    \includegraphics[width=0.32\linewidth]{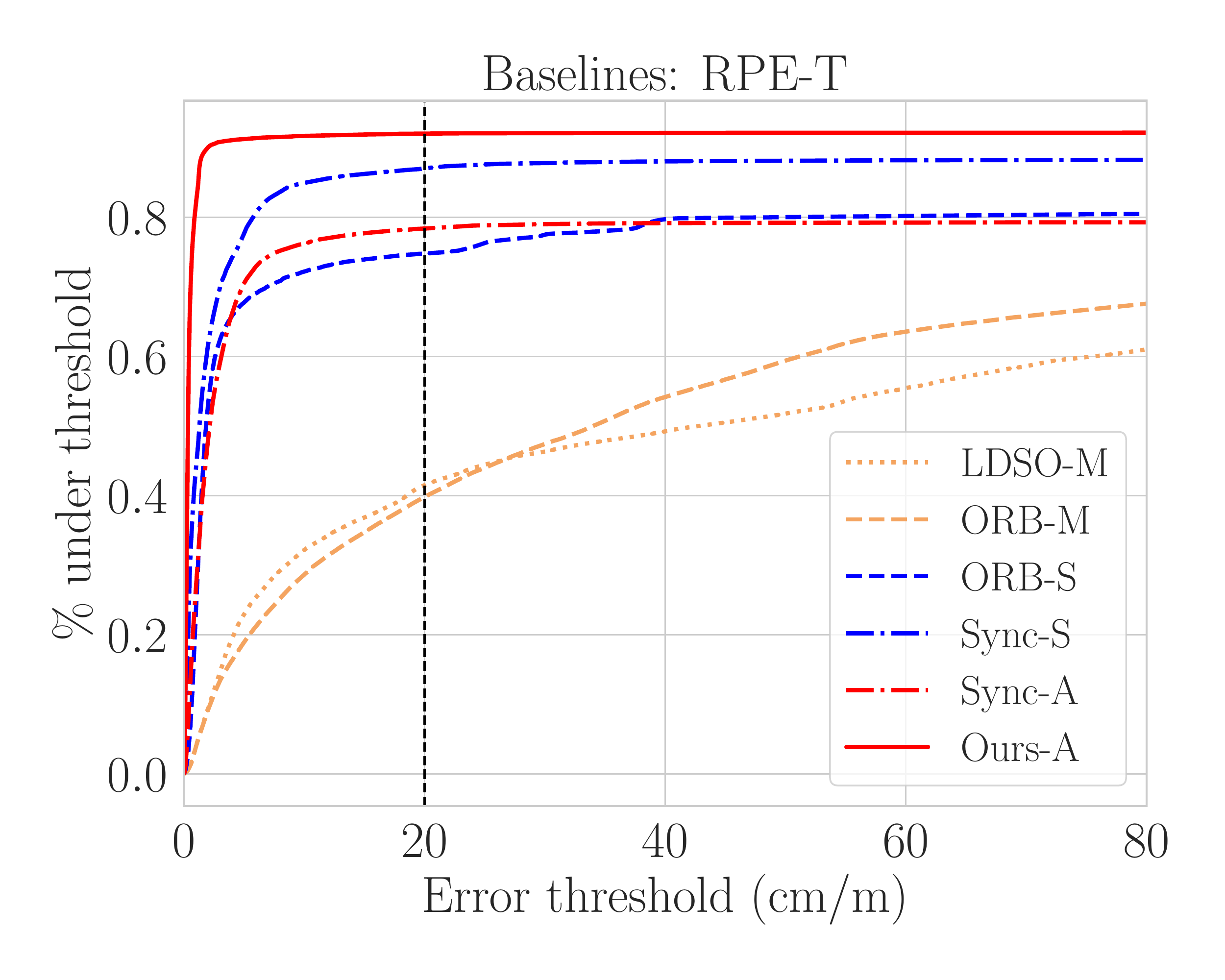}
    \includegraphics[width=0.32\linewidth]{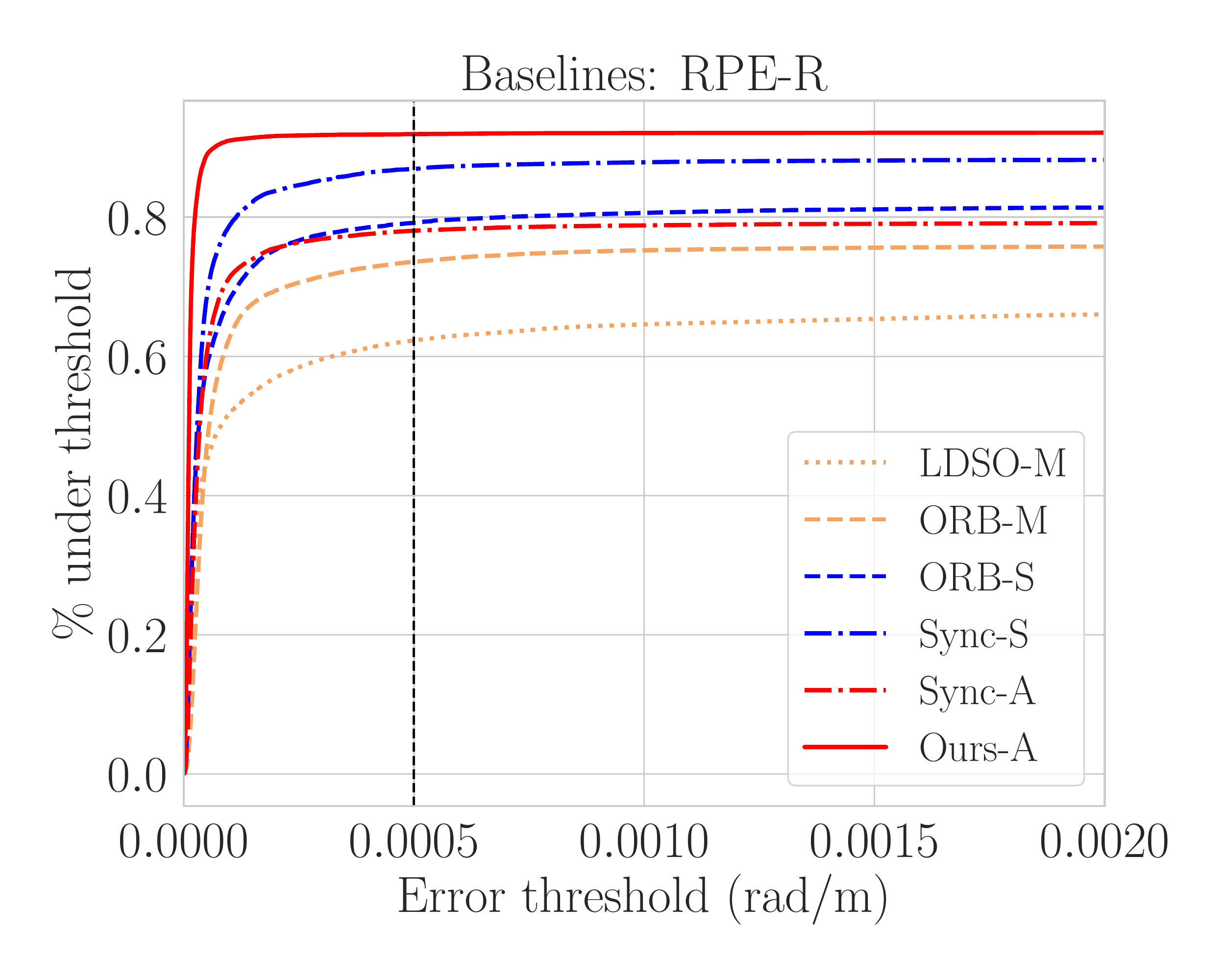}
    \includegraphics[width=0.32\linewidth]{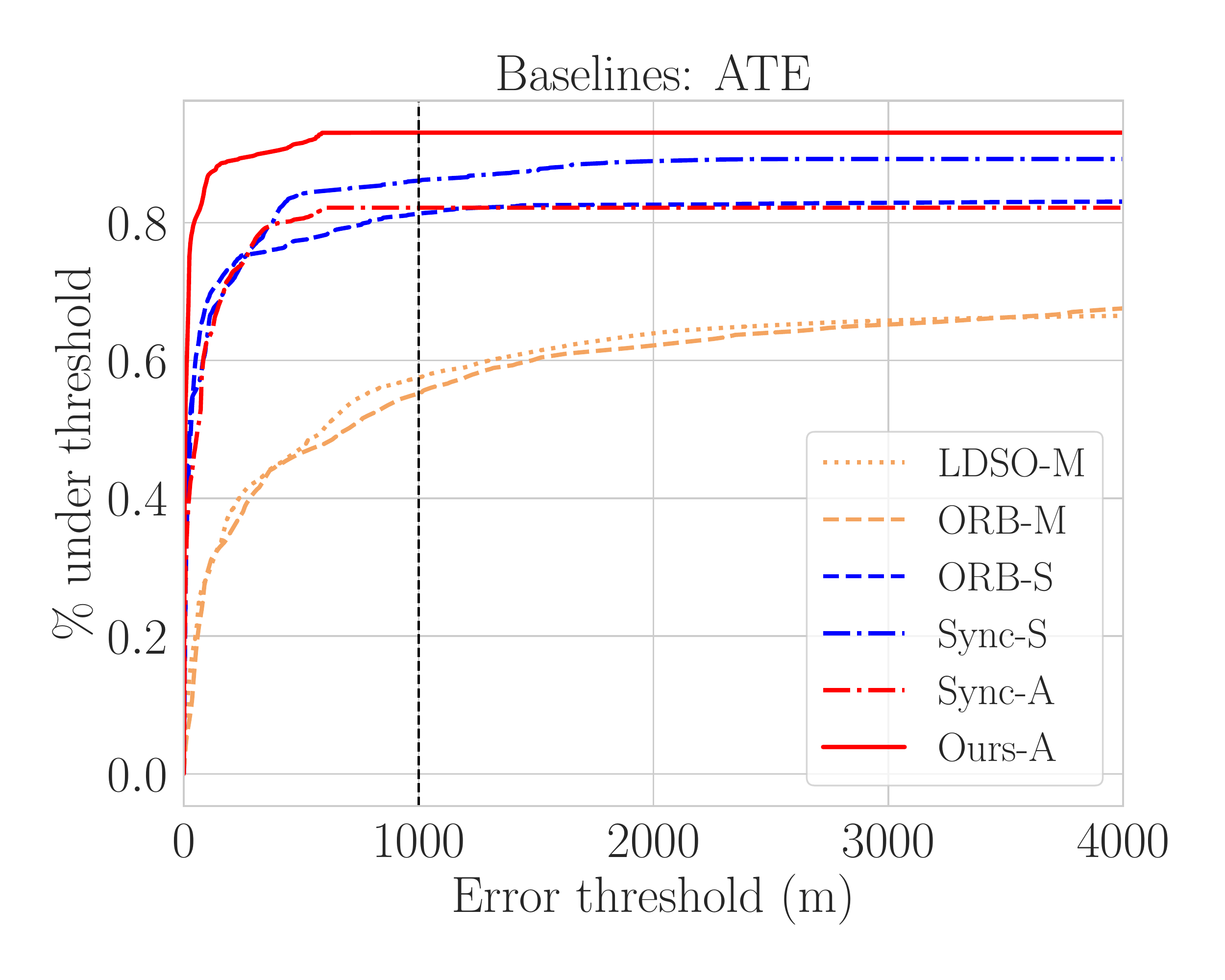}
    \caption{Cumulative error curves comparing all baseline methods
    and our full system, with loop closure.
    \label{fig:cum_curve_baselines_lc}}
\end{figure*}

\begin{table*}
  \centering
  \caption{Baseline methods, all in the visual odometry (VO) mode
  with loop closing (and relocalization in ORB-SLAM) disabled.}
  \label{tab:quantitative_baseline_vo}
  \begin{tabular}{l|ccc|ccc|ccc|c}
    \toprule
    \multirow{2}{*}{\textbf{Method}} &
    \multicolumn{3}{c|}{\rt\space(cm/m)} &
    \multicolumn{3}{c|}{\rr\space(rad/m)} &
    \multicolumn{3}{c|}{\ate\space(m)} &
    \multirow{2}{*}{\sr\space(\%)} \\
    & @0.5 & @0.9 & AUC(\%) & @0.5 & @0.9 & AUC(\%) & @0.5 & @0.9 & AUC(\%) & \\
    \midrule
    DSO-M~\cite{engel2017direct} & 30.99 & - & 32.93 & 3.88E-05 & - & 58.98 & 801.99 & - & 41.87 & 64.00 \\
    ORB-M (VO)~\cite{mur2015orb} & 45.22 & - & 20.33 & 4.93E-05 & - & 64.91 & 849.64 & - & 40.62 & 58.67 \\
    ORB-S (VO)~\cite{mur2017orb} & 2.24 & - & 64.91 & 2.99E-05 & - & 72.30 & 45.28 & - & 74.61 & 66.67 \\
    Sync-S (VO) & \second{1.27} & - & \second{77.77} & \second{2.80E-05} & - & \second{78.72} & \second{24.07} & - & \second{77.64} & \second{85.33} \\
    Sync-A (VO) & 1.97 & - & 69.46 & 2.96E-05 & - & 73.39 & 55.24 & - & 75.11 & 70.67 \\
    Ours-A (VO) & \best{0.35} & \best{2.14} & \best{88.79} & \best{1.11E-05} & \best{6.30E-05} & \best{88.47} & \best{6.53} & \best{299.30} & \best{89.04} & \best{92.00} \\    
    \bottomrule
  \end{tabular}
\end{table*}

\begin{figure*}
    \centering
    \includegraphics[width=0.32\linewidth]{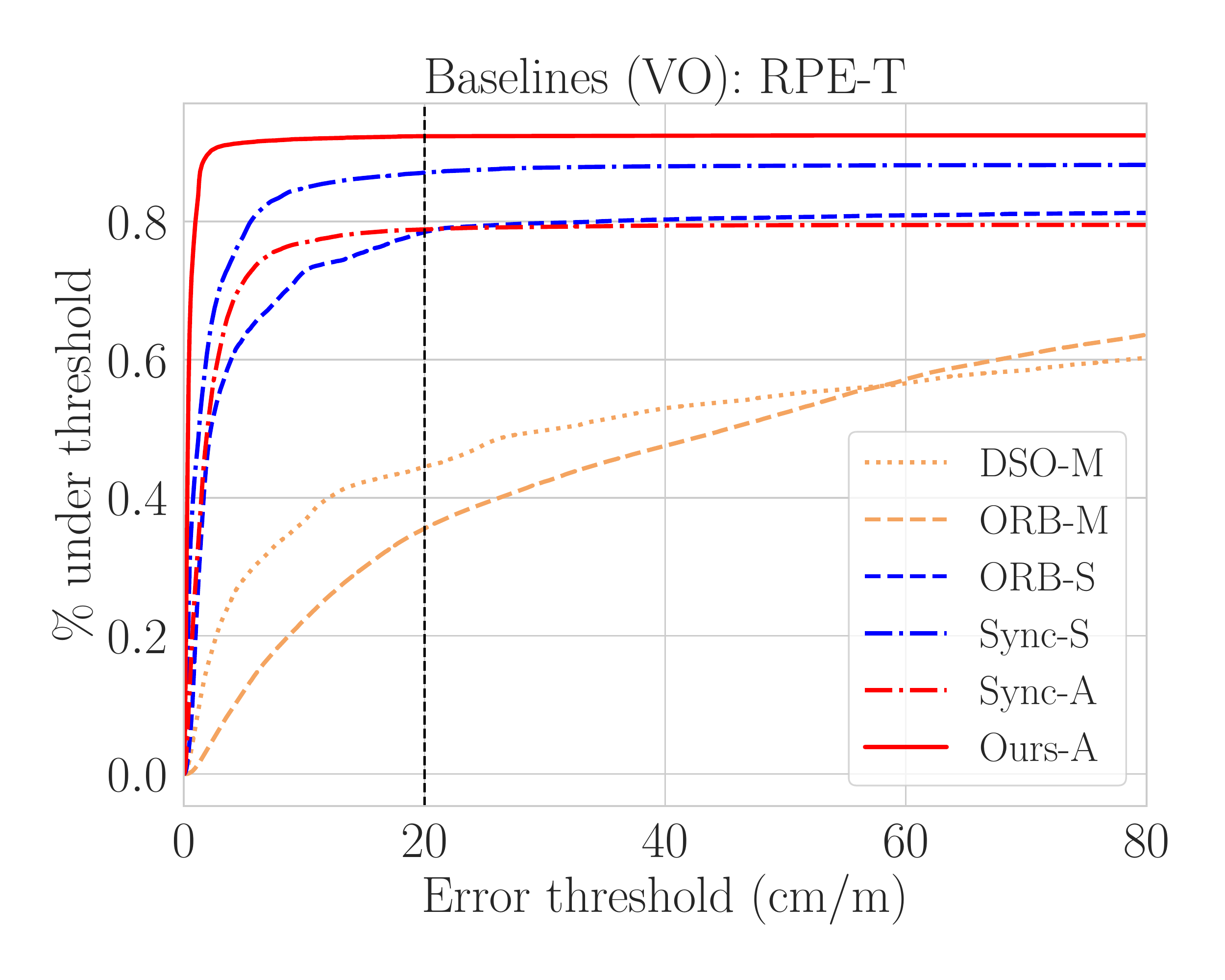}
    \includegraphics[width=0.32\linewidth]{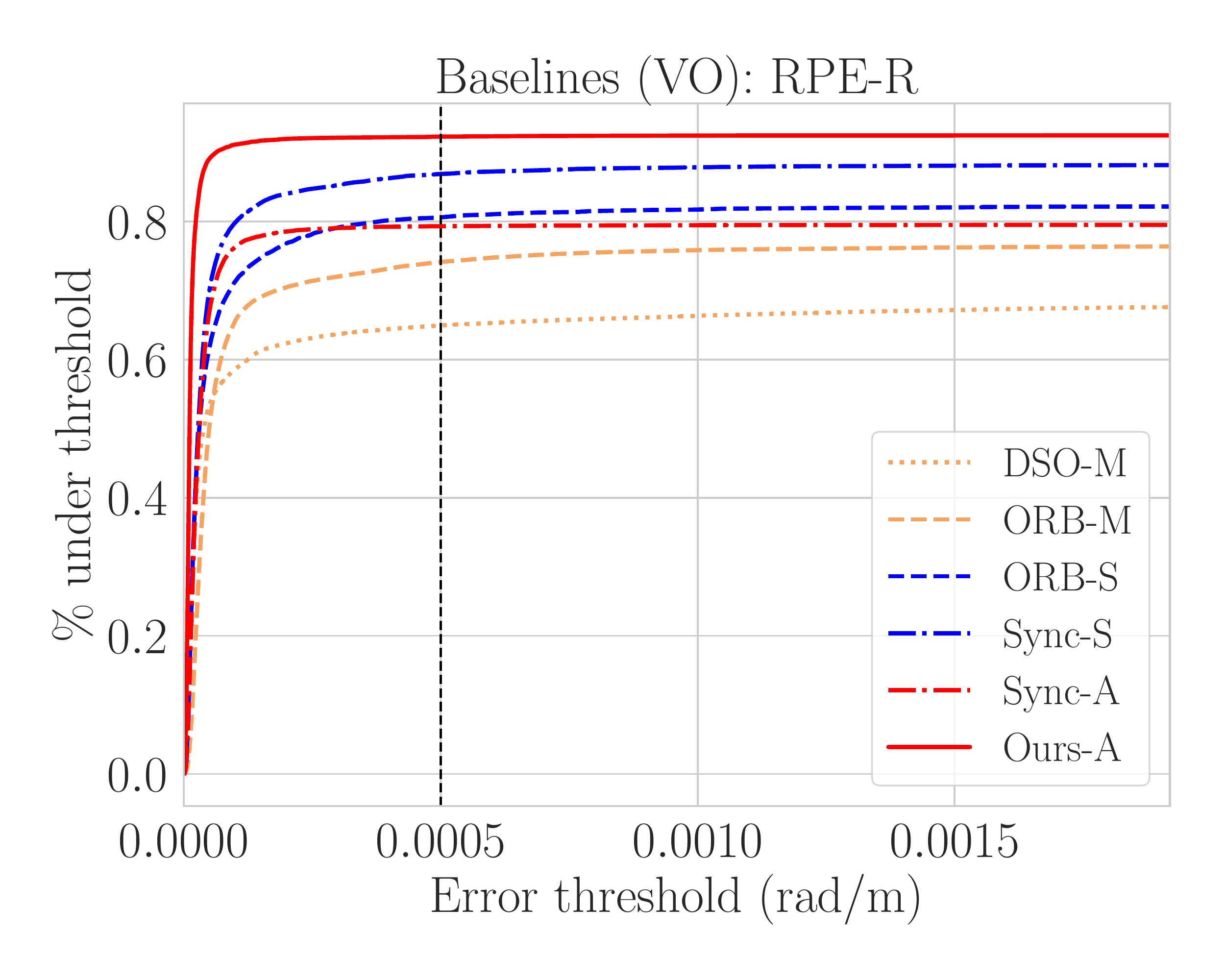}
    \includegraphics[width=0.32\linewidth]{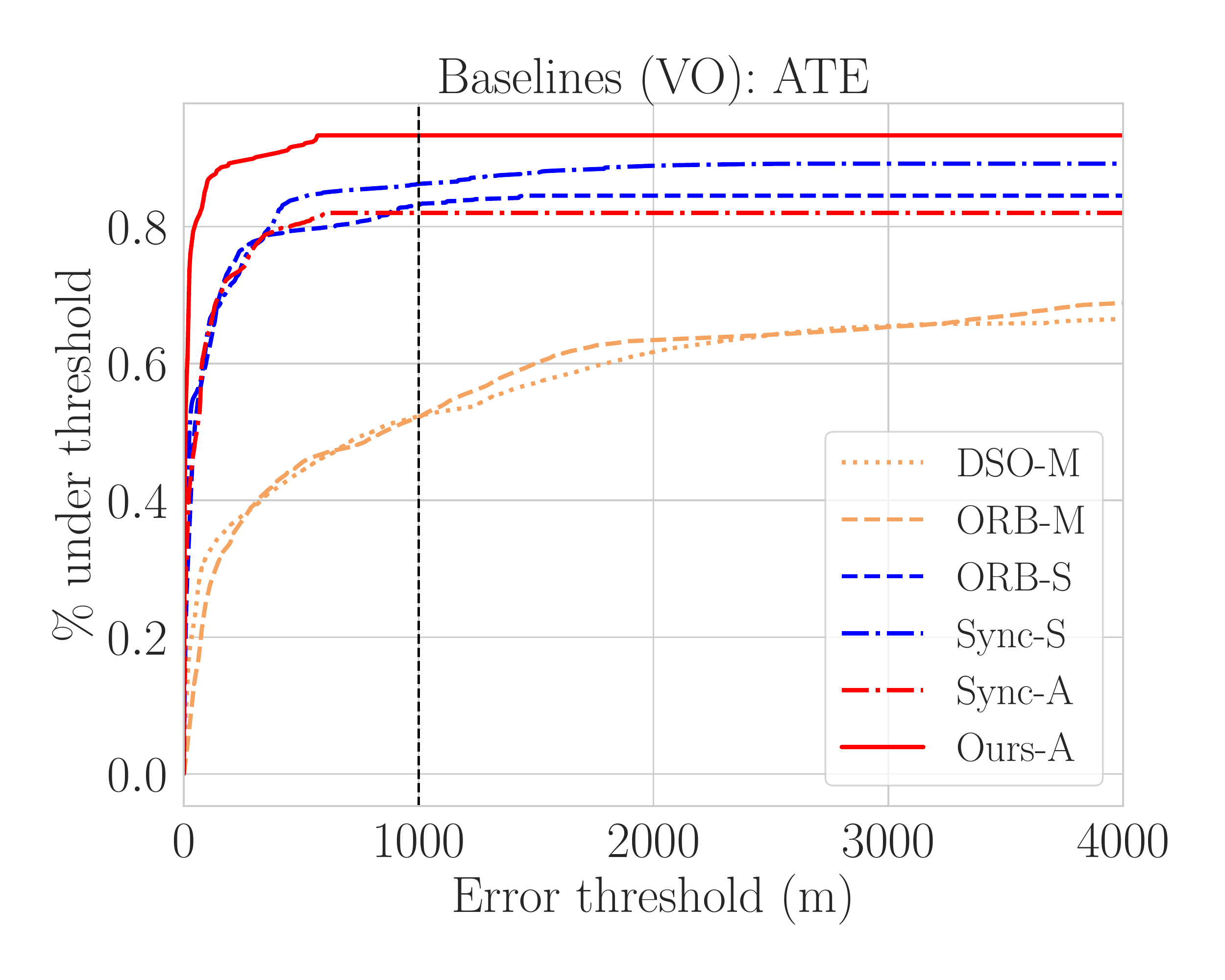}
    \caption{Cumulative error curves comparing all baseline methods
    and our full system, without loop closure.
    \label{fig:cum_curve_baselines_vo}}
\end{figure*}

\begin{table*}
  \centering
  \caption{Ablation study on loop closure on 8 validation sequences where
  loop closing was performed.}
  \label{tab:ablation-loop-closure}
  \begin{tabular}{l|ccc|ccc|ccc|c}
    \toprule
    \multirow{2}{*}{\textbf{Method}} &
    \multicolumn{3}{c|}{\rt\space(cm/m)} &
    \multicolumn{3}{c|}{\rr\space(rad/m)} &
    \multicolumn{3}{c|}{\ate\space(m)} &
    \multirow{2}{*}{\sr\space(\%)} \\
    & @0.5 & @0.9 & AUC(\%) & @0.5 & @0.9 & AUC(\%) & @0.5 & @0.9 & AUC(\%) & \\
    \midrule
    Ours-A (VO) & \best{0.27} & 0.99 & 96.67 & \best{9.51E-06} & \best{3.07E-05} & 95.73 & 3.87 & 25.08 & 97.67 & \best{100.00} \\
    Ours-A & 0.28 & \best{0.92} & \best{96.78} & 9.83E-06 & 3.19E-05 & \best{95.88} & \best{2.97} & \best{20.58} & \best{97.73} & \best{100.00} \\
    \bottomrule
  \end{tabular}
\end{table*}
\begin{table*}
  \centering
  \caption{Ablation study on motion models. All experiments in visual odometry (VO) mode with loop closing disabled.
  Split indicates interpolation in $\SO{3}$ and $\bR^3$ separately~\cite{sommer2019efficient} instead of
  jointly in $\SE{3}$.}
  \label{tab:ablation-motion}
  \begin{tabular}{l|ccc|ccc|ccc|c}
    \toprule
    \multirow{2}{*}{\textbf{Method}} &
    \multicolumn{3}{c|}{\rt\space(cm/m)} &
    \multicolumn{3}{c|}{\rr\space(rad/m)} &
    \multicolumn{3}{c|}{\ate\space(m)} &
    \multirow{2}{*}{\sr\space(\%)} \\
    & @0.5 & @0.9 & AUC(\%) & @0.5 & @0.9 & AUC(\%) & @0.5 & @0.9 & AUC(\%) & \\
    \midrule
    Sync Assumption & 1.97 & - & 69.46 & 2.96E-05 & - & 73.39 & 55.24 & - & 75.11 & 70.67 \\
    Linear (Split) & \second{0.36} & \second{2.51} & \second{88.08} & \best{1.10E-05} & 7.96E-05 & 87.79 & 6.25 & 580.23 & 87.43 & \best{92.00} \\
    Linear & 0.41 & 2.71 & 87.76 & \second{1.11E-05} & \best{5.99E-05} & \second{88.39} & \second{6.09} & \second{429.86} & \second{88.31} & 89.33 \\
    Cubic B-Spline (Split) & 0.38 & 3.34 & 87.86 & 1.12E-05 & 9.01E-05 & 87.69 & \best{5.15} & 588.96 & 87.34 & \best{92.00} \\
    Cubic B-Spline & \best{0.35} & \best{2.14} & \best{88.79} & \second{1.11E-05} & \second{6.30E-05} & \best{88.47} & 6.53 & \best{299.30} & \best{89.04} & \best{92.00} \\
    \bottomrule
  \end{tabular}
\end{table*}

\begin{figure*}
    \centering
    \includegraphics[width=0.32\linewidth]{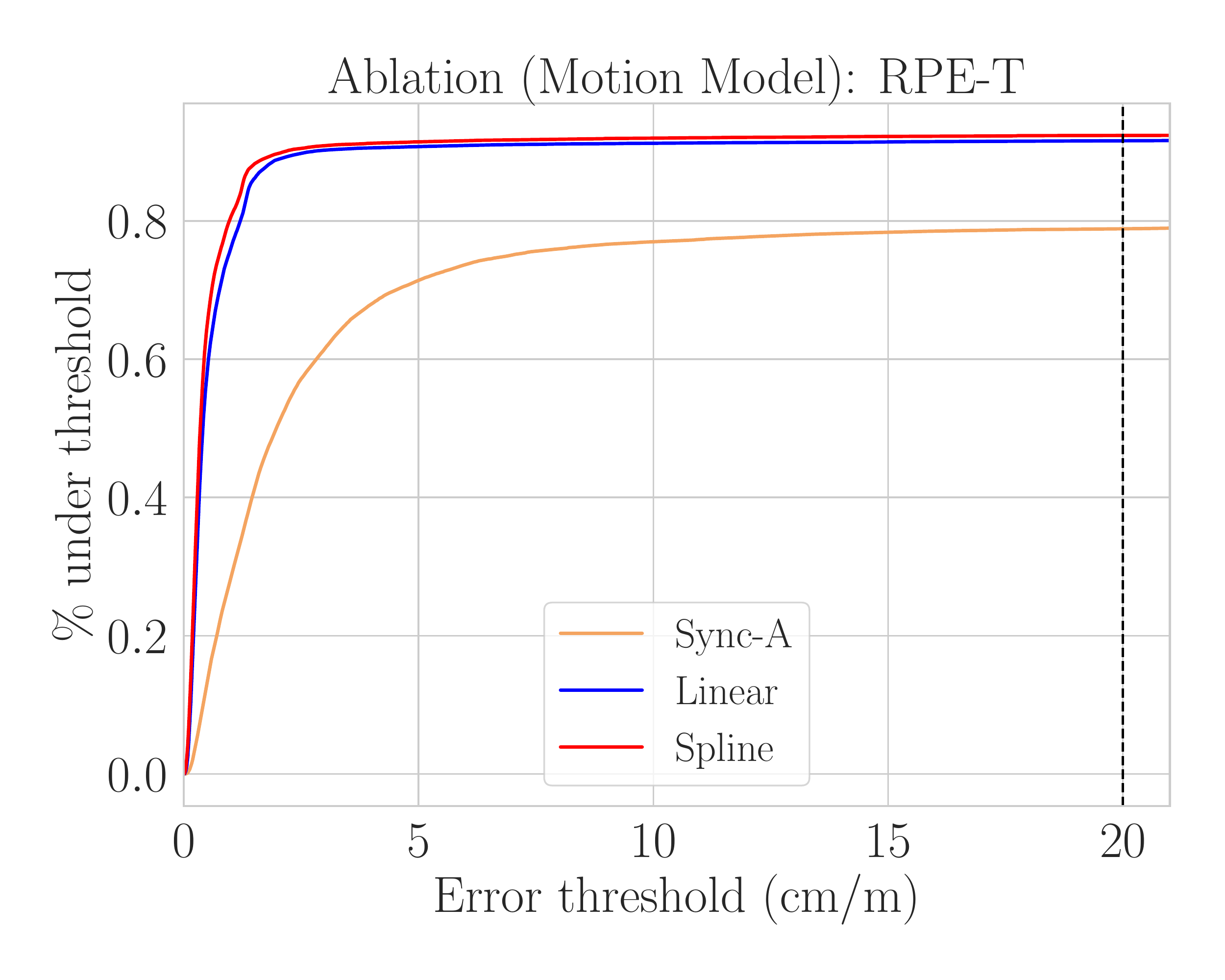}
    \includegraphics[width=0.32\linewidth]{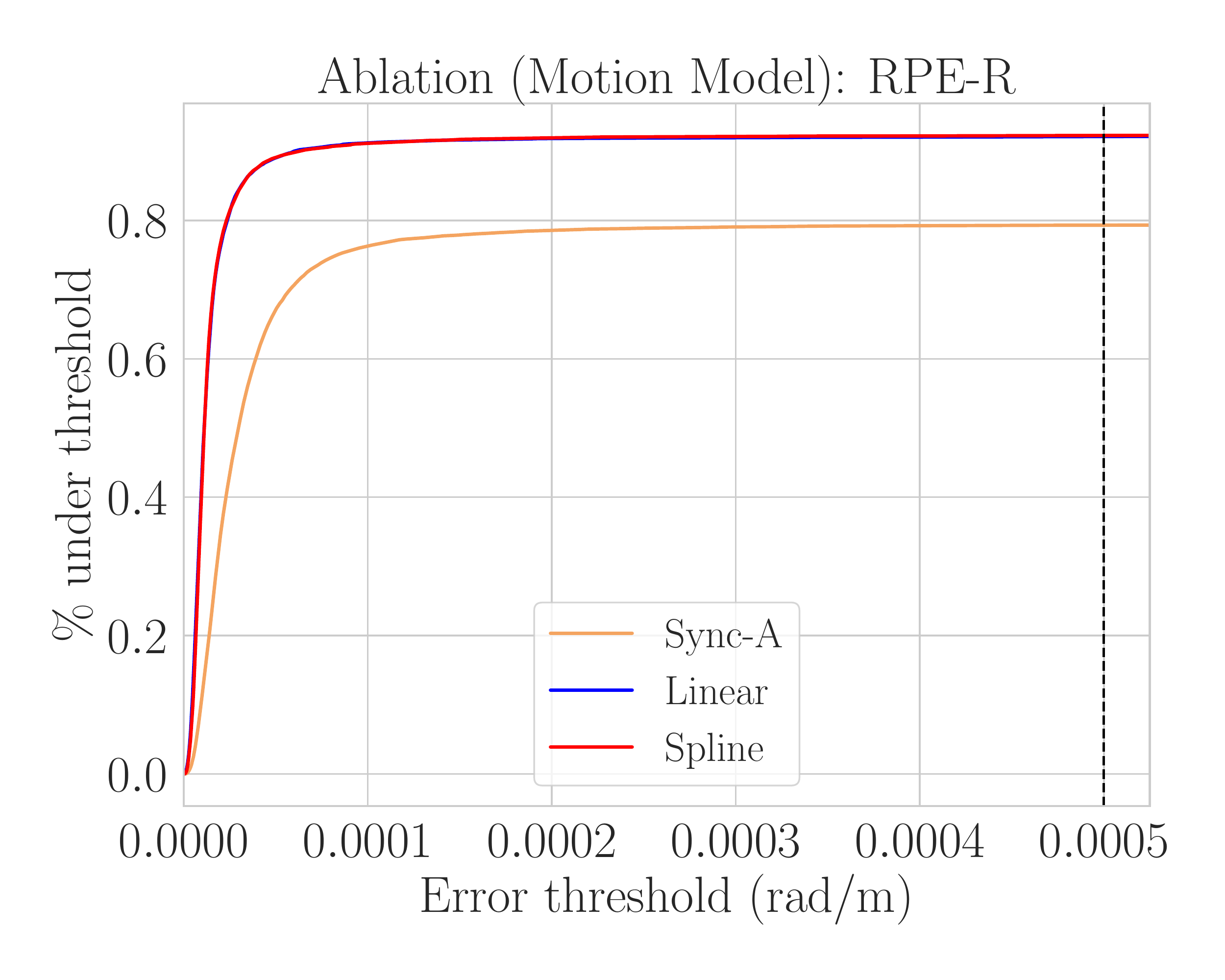}
    \includegraphics[width=0.32\linewidth]{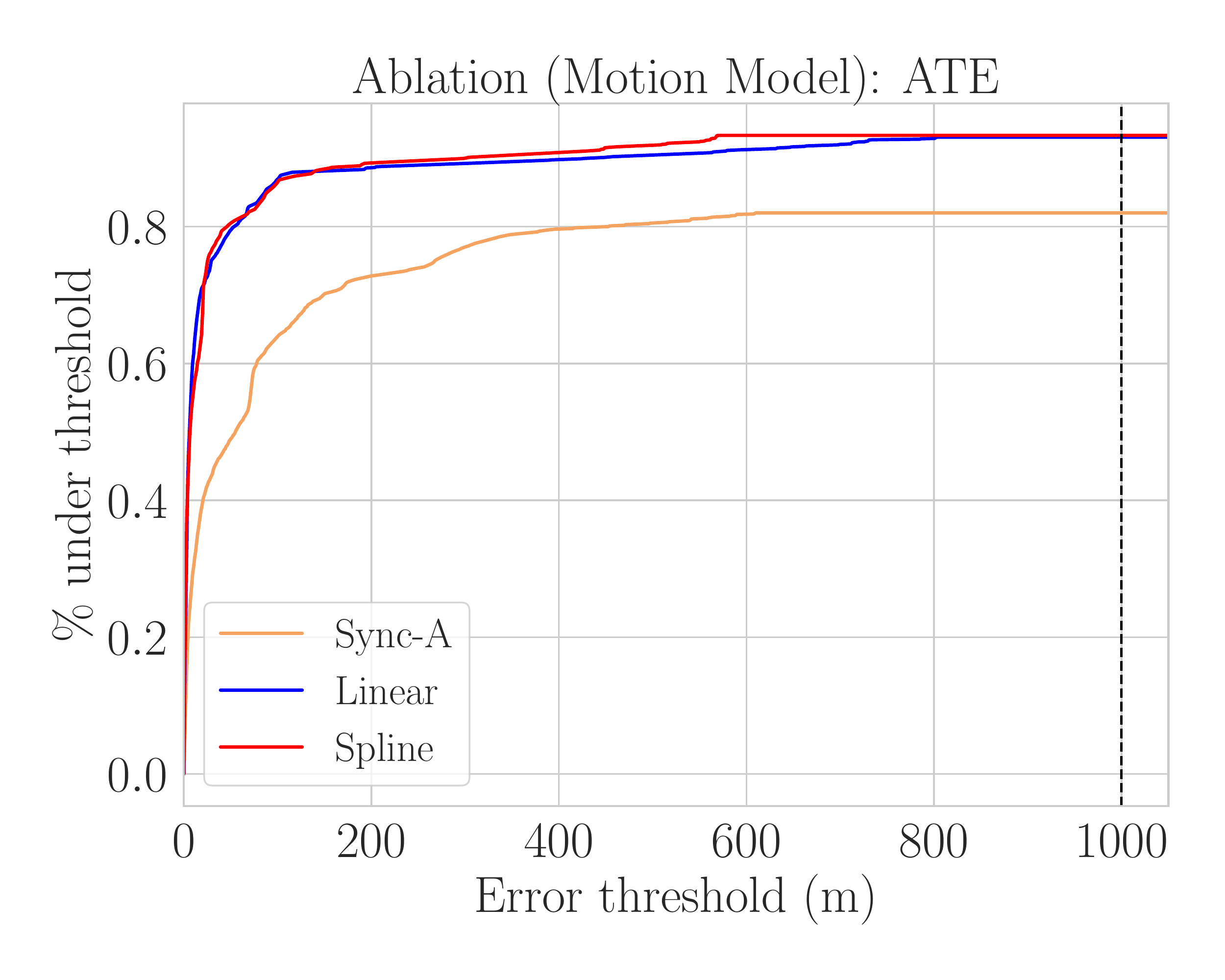}
    \includegraphics[width=0.32\linewidth]{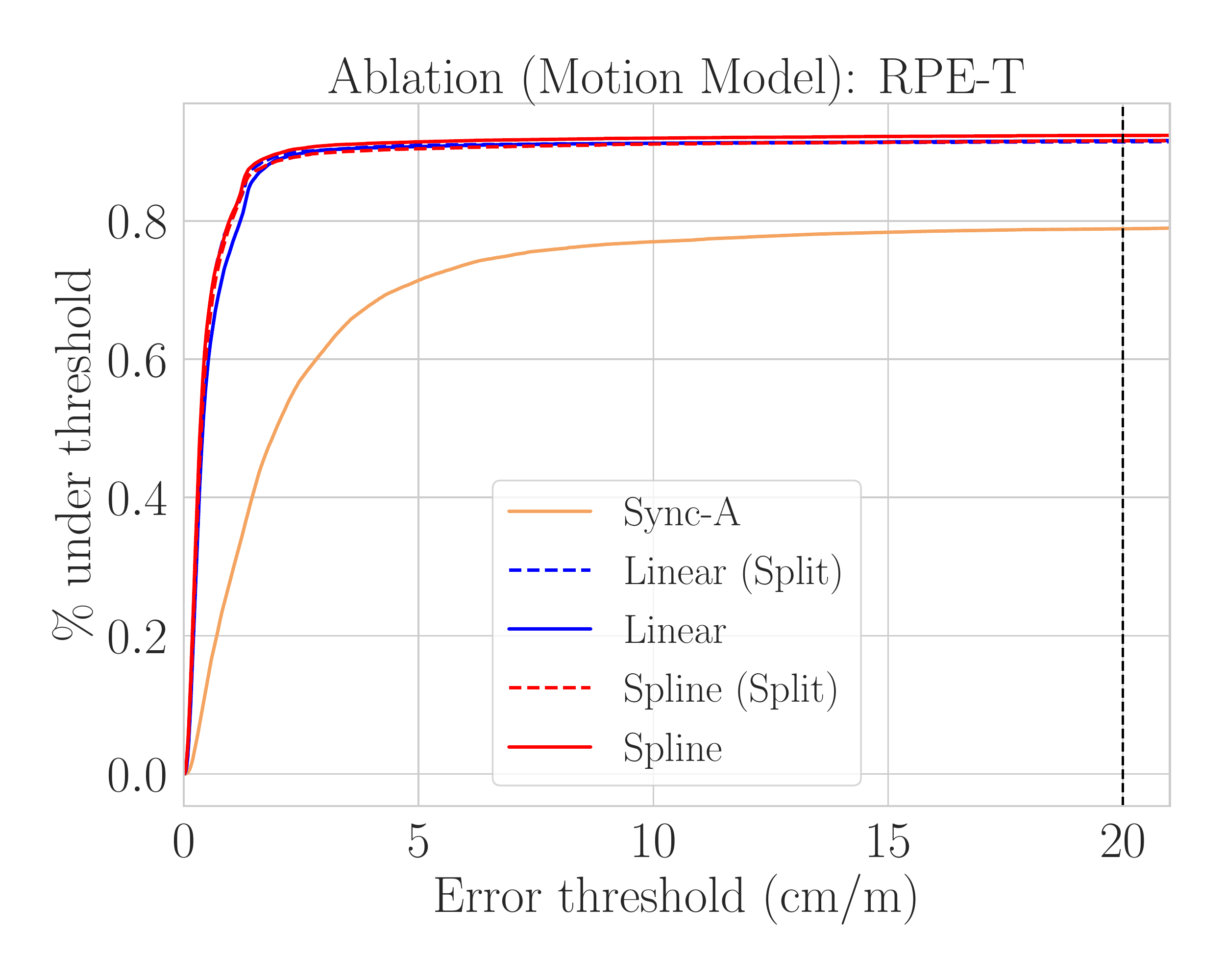}
    \includegraphics[width=0.32\linewidth]{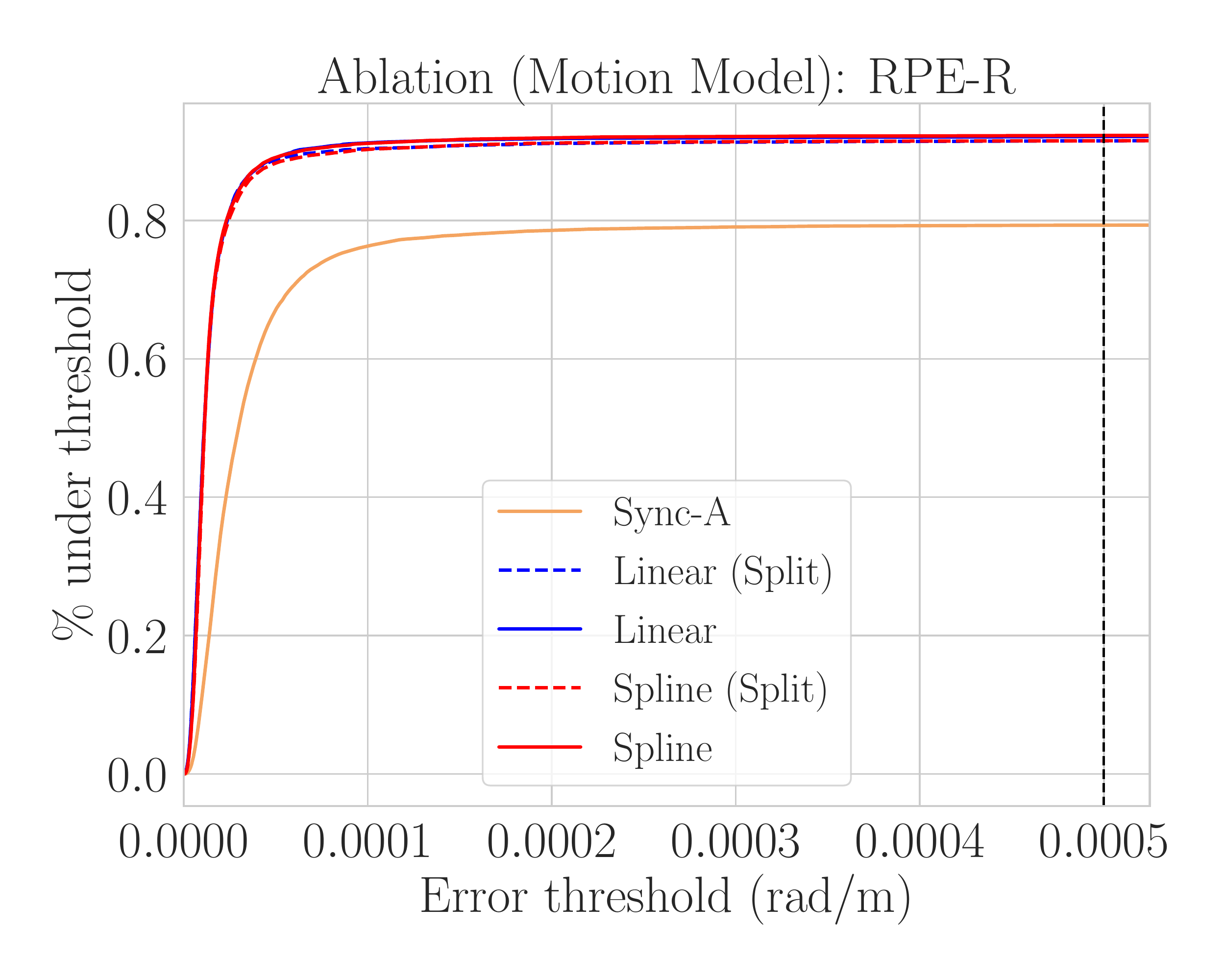}
    \includegraphics[width=0.32\linewidth]{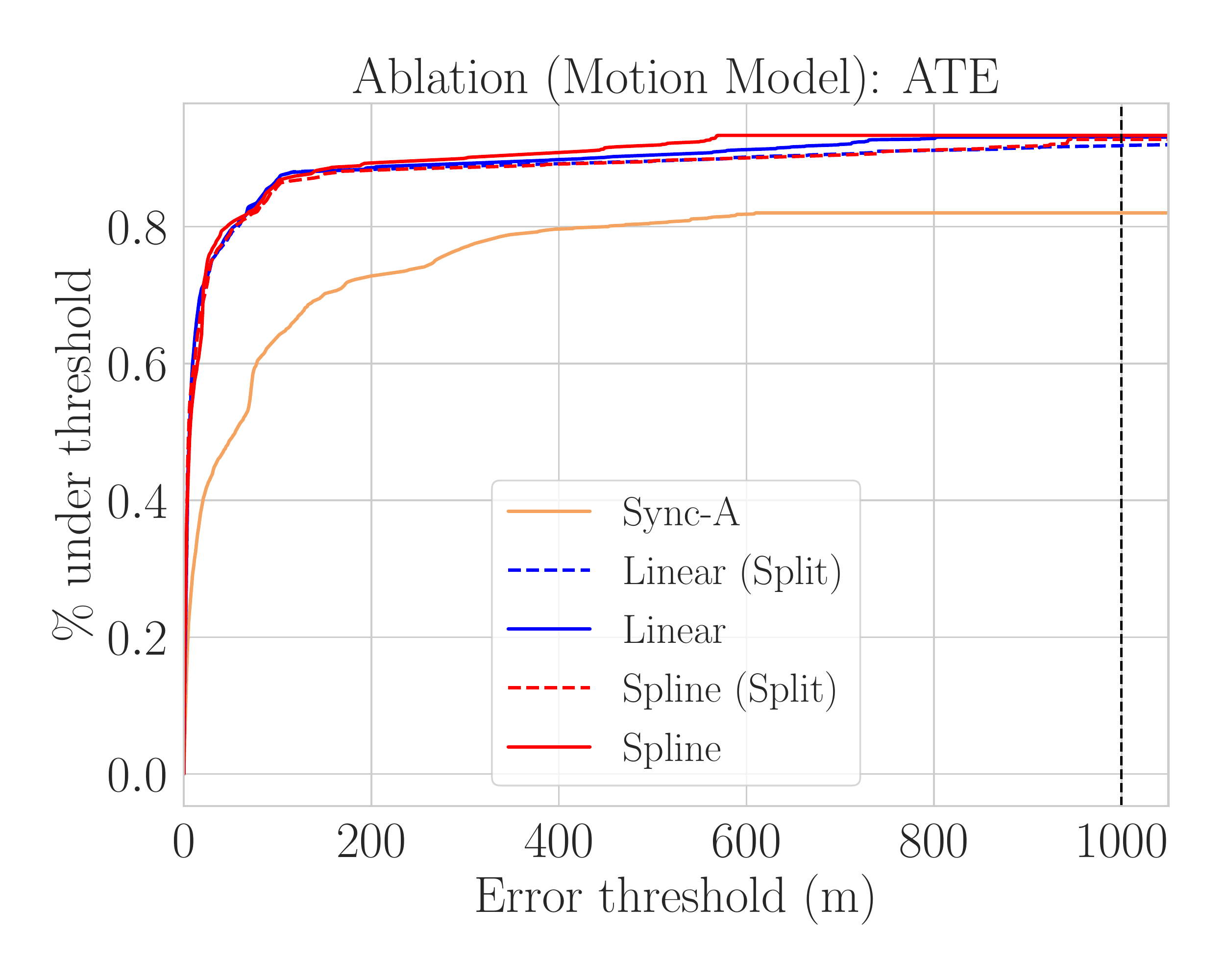}
    \caption{Cumulative error curves of the motion model ablation study. (Top) The three
    comparisons in the main paper. (Bottom) All comparisons in the additional
    experiments.
    \label{fig:cum_curve_motion}}
\end{figure*}

\begin{table*}
  \centering
  \caption{Ablation study on camera rigs in the VO mode, all initialized 
  with the stereo cameras.
  s = stereo, wf = wide-front, wb = wide-back, \underline{\yes} is used for
  intra-frame new map point creation during mapping.
  The last row in the ORB table represents the main system.
\label{tab:ablation-num-cameras}}
  \begin{tabular}{cccc|ccc|ccc|ccc|c}
    \toprule
    &
    \multicolumn{3}{c|}{Camera Config} &
    \multicolumn{3}{c|}{\rt\space(cm/m)} &
    \multicolumn{3}{c|}{\rr\space(rad/m)} &
    \multicolumn{3}{c|}{\ate\space(m)} &
    \multirow{2}{*}{\sr\space(\%)} \\
    & s & wf & wb & @0.5 & @0.9 & AUC(\%) & @0.5 & @0.9 & AUC(\%) & @0.5 & @0.9 & AUC(\%) & \\
    \midrule
    \multirow{6}{*}{\rotatebox[origin=c]{90}{ORB~\cite{orb}}}
    &\underline{\yes}&  & & 0.70 & - & 79.86 & 1.93E-05 & - & 80.48 & 11.44 & - & 75.75 & 88.00 \\                                                                                                                                 
    & \underline{\yes}& \yes & & 0.41 & 8.52 & 84.88 & 1.21E-05 & 2.48E-04 & 85.86 & 9.00 & 802.88 & 84.92 & 90.67 \\                                                                                                                
    &  &\underline{\yes}& & 6.07 & - & 49.00 & 4.58E-05 & - & 52.96 & 53.76 & - & 55.05 & 57.33 \\                                                                                                                             
    & & \underline{\yes}& \yes & 1.16 & - & 63.94 & 1.65E-05 & - & 74.34 & 18.63 & - & 72.10 & 74.67 \\                                                                                                                              
    & \underline{\yes}& \underline{\yes}& \yes & \second{0.36} & \second{3.43} & \second{88.43} & \second{1.12E-05} & \best{5.35E-05} & \best{88.60} & \best{5.95} & \best{298.08} & \best{89.05} & \best{92.00} \\                           
    & \underline{\yes}& \yes & \yes & \best{0.35} & \best{2.14} & \best{88.79} & \best{1.11E-05} & \second{6.30E-05} & \second{88.47} & \second{6.53} & \second{299.30} & \second{89.04} & \best{92.00} \\ 
    \midrule
    \multirow{6}{*}{\rotatebox[origin=c]{90}{SuperPoint~\cite{superpoint2018}}}
    &\underline{\yes}&  & & 0.64 & 19.44 & 82.97 & 1.59E-05 & 5.77E-04 & 83.09 & 16.40 & 463.41 & 85.09 & 82.67 \\
    & \underline{\yes}& \yes & & 0.44 & 1.62 & 92.83 & \best{1.01E-05} & 3.48E-05 & 93.43 & 8.33 & 106.31 & 93.92 & 97.33 \\
    & & \underline{\yes}&  & 1.06 & - & 78.32 & 2.88E-05 & - & 76.73 & 21.08 & - & 73.48 & 88.00 \\
    & & \underline{\yes}& \yes & 0.54 & 1.85 & 89.42 & 1.84E-05 & 6.51E-05 & 88.31 & 10.62 & 413.52 & 87.67 & 96.00 \\
    & \underline{\yes}& \underline{\yes}& \yes & \best{0.38} & \best{1.22} & \best{95.66} & 1.04E-05 & \second{2.63E-05} & \second{95.86} & \best{5.38} & \second{86.67} & \best{96.38} & \best{100.00} \\
    & \underline{\yes}& \yes & \yes & \second{0.41} & \second{1.28} & \second{95.28} & \second{1.03E-05} & \best{2.54E-05} & \best{95.90} & \second{6.83} & \best{78.06} & \second{96.34} & \second{98.67} \\
    \bottomrule
  \end{tabular}
\end{table*}

\begin{figure*}
    \centering
    \includegraphics[width=0.32\linewidth]{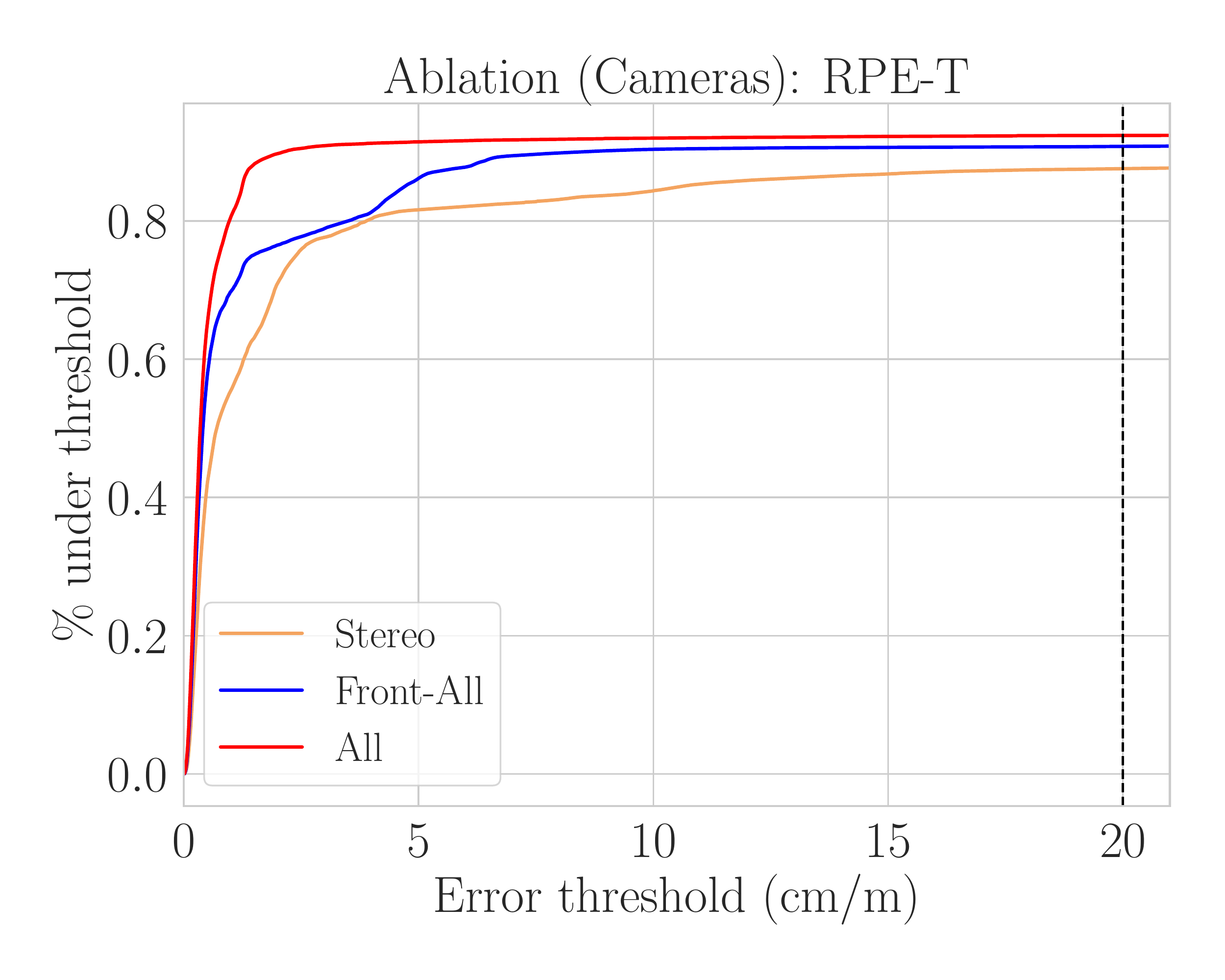}
    \includegraphics[width=0.32\linewidth]{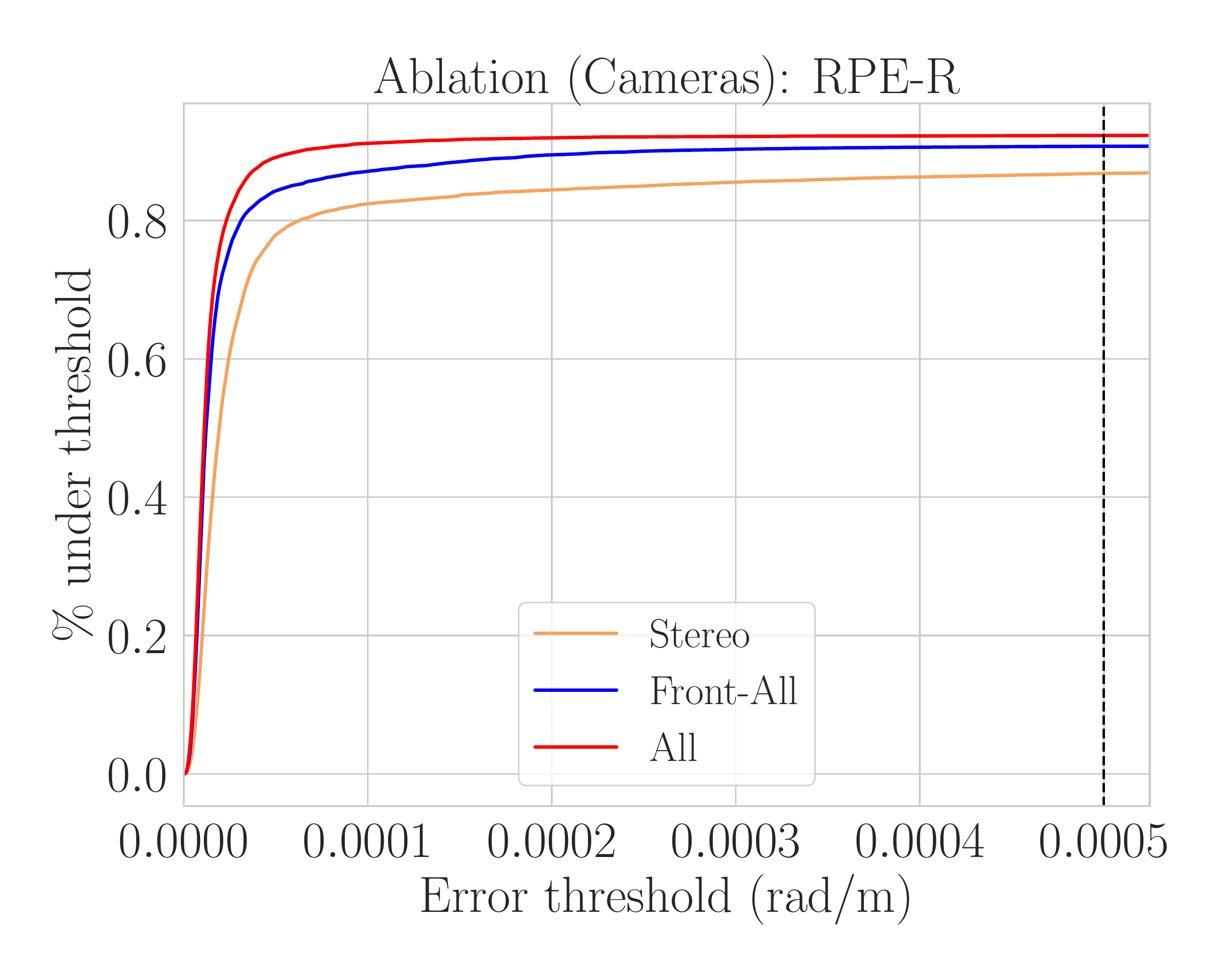}
    \includegraphics[width=0.32\linewidth]{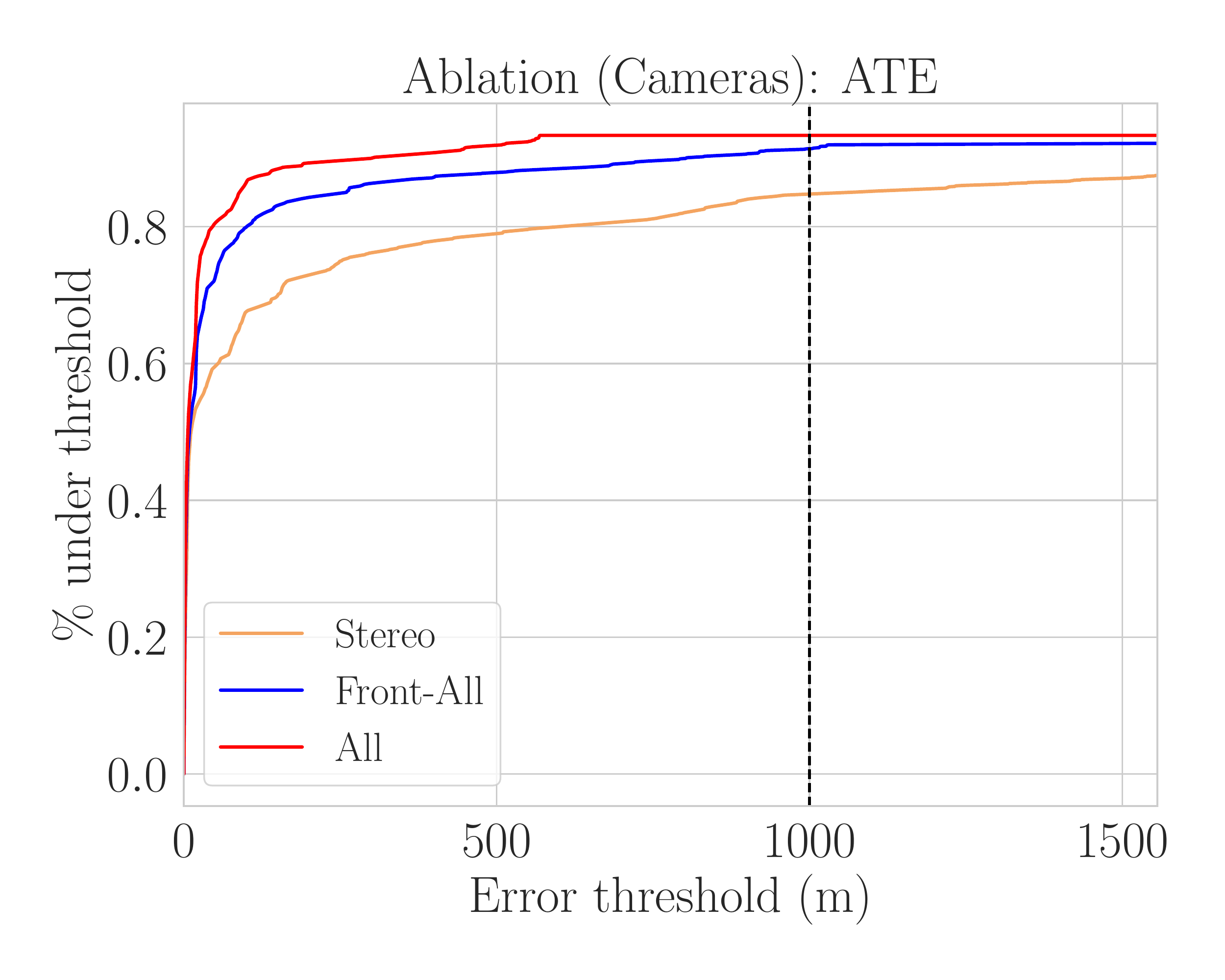}
    \includegraphics[width=0.32\linewidth]{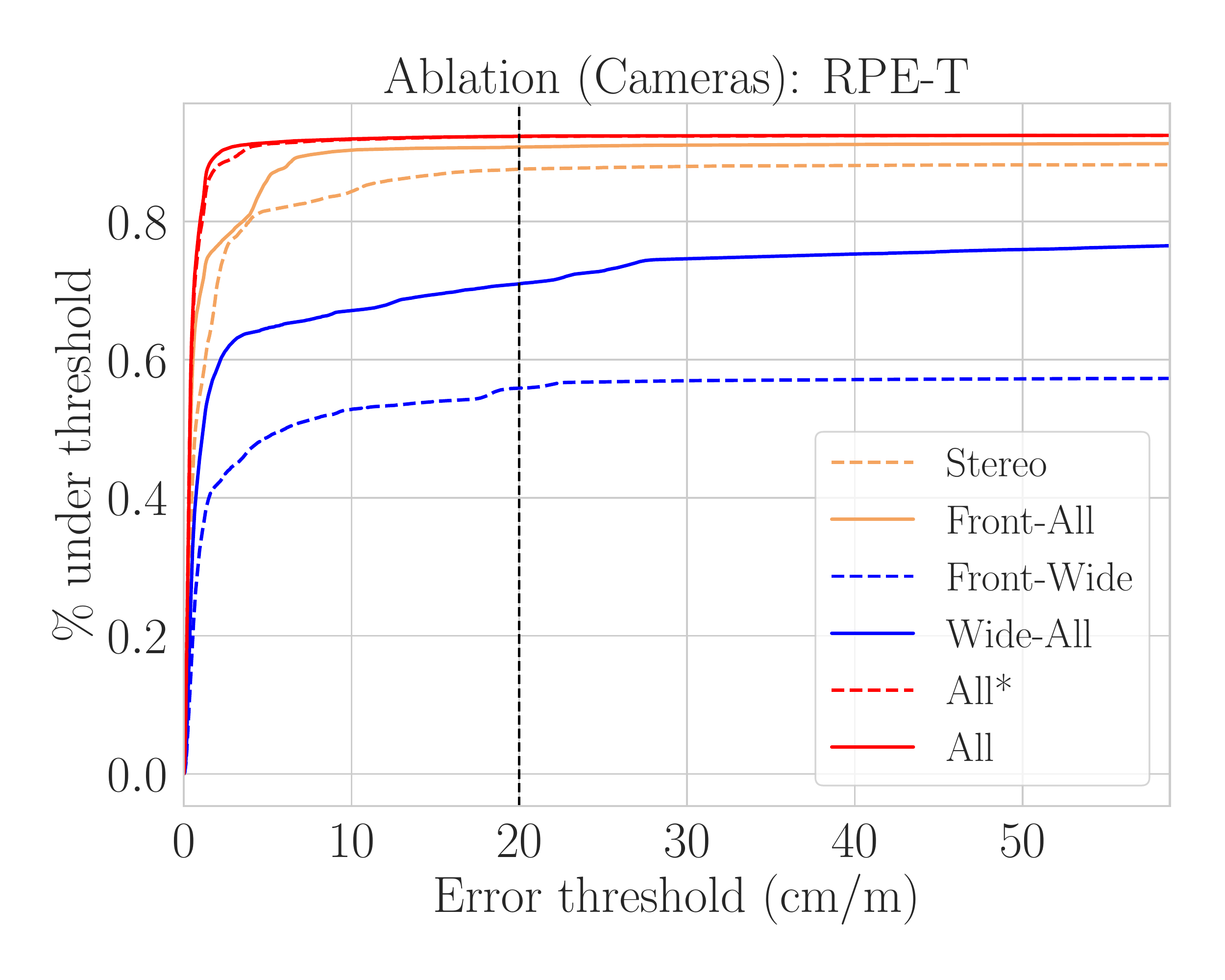}
    \includegraphics[width=0.32\linewidth]{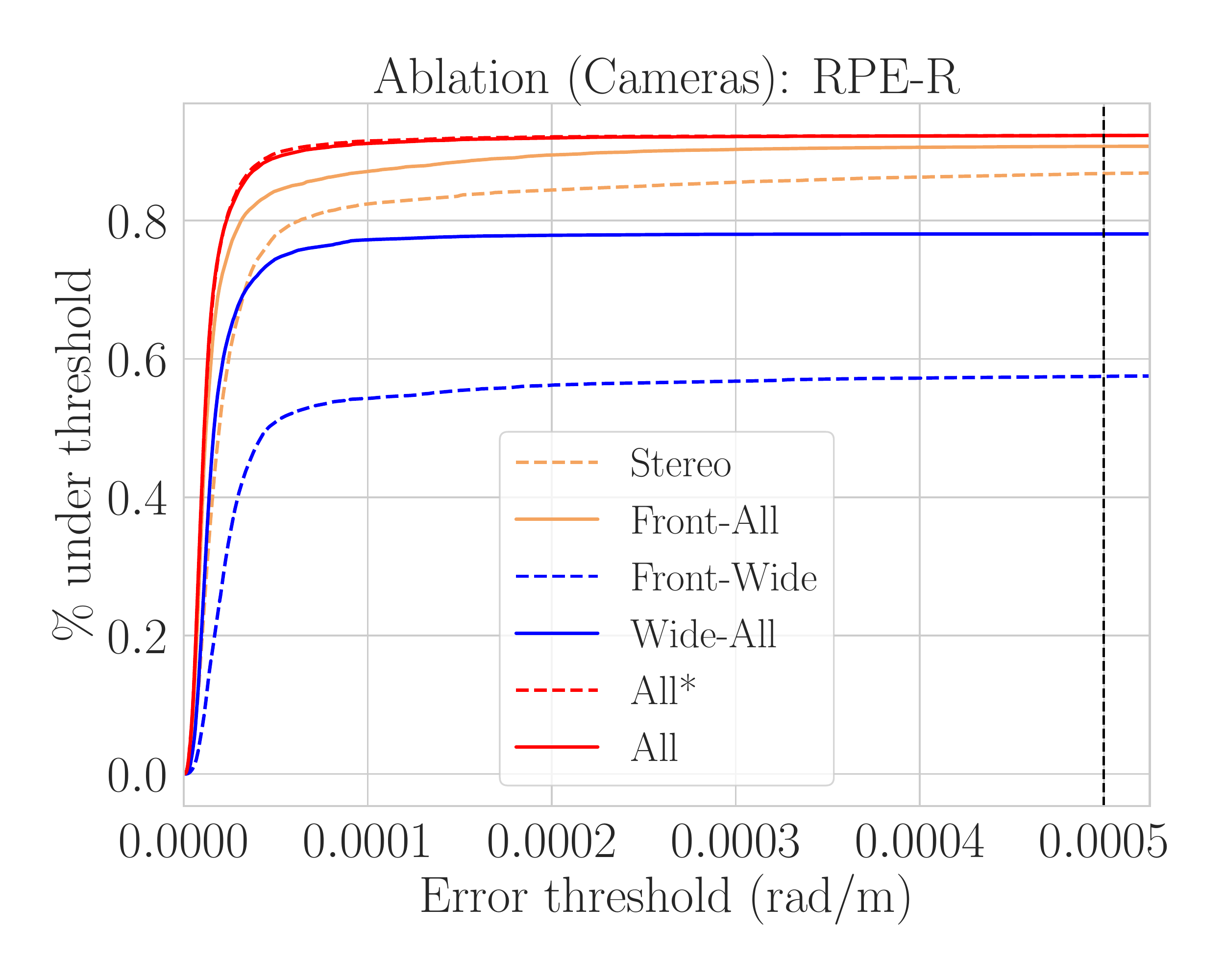}
    \includegraphics[width=0.32\linewidth]{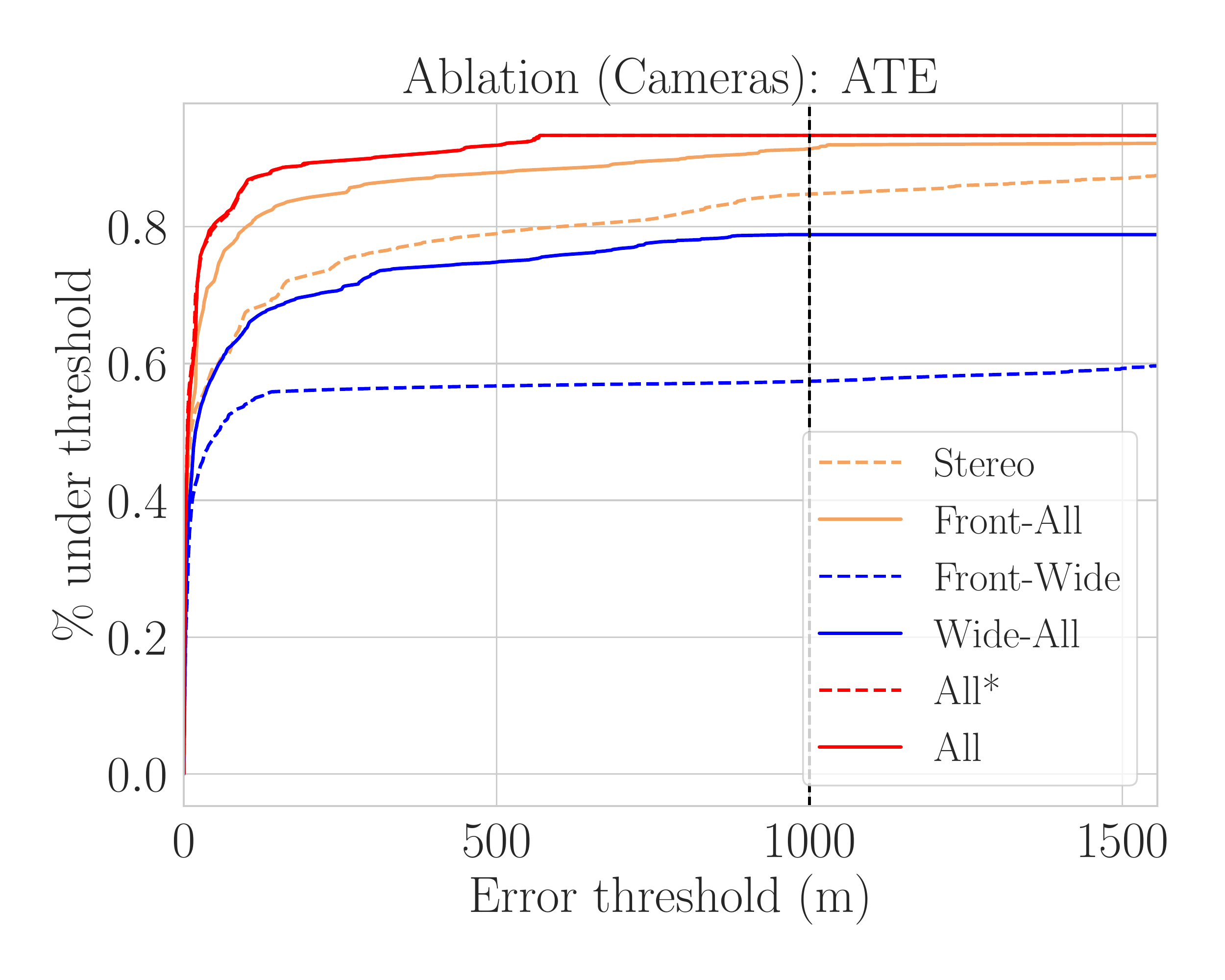}
    \includegraphics[width=0.32\linewidth]{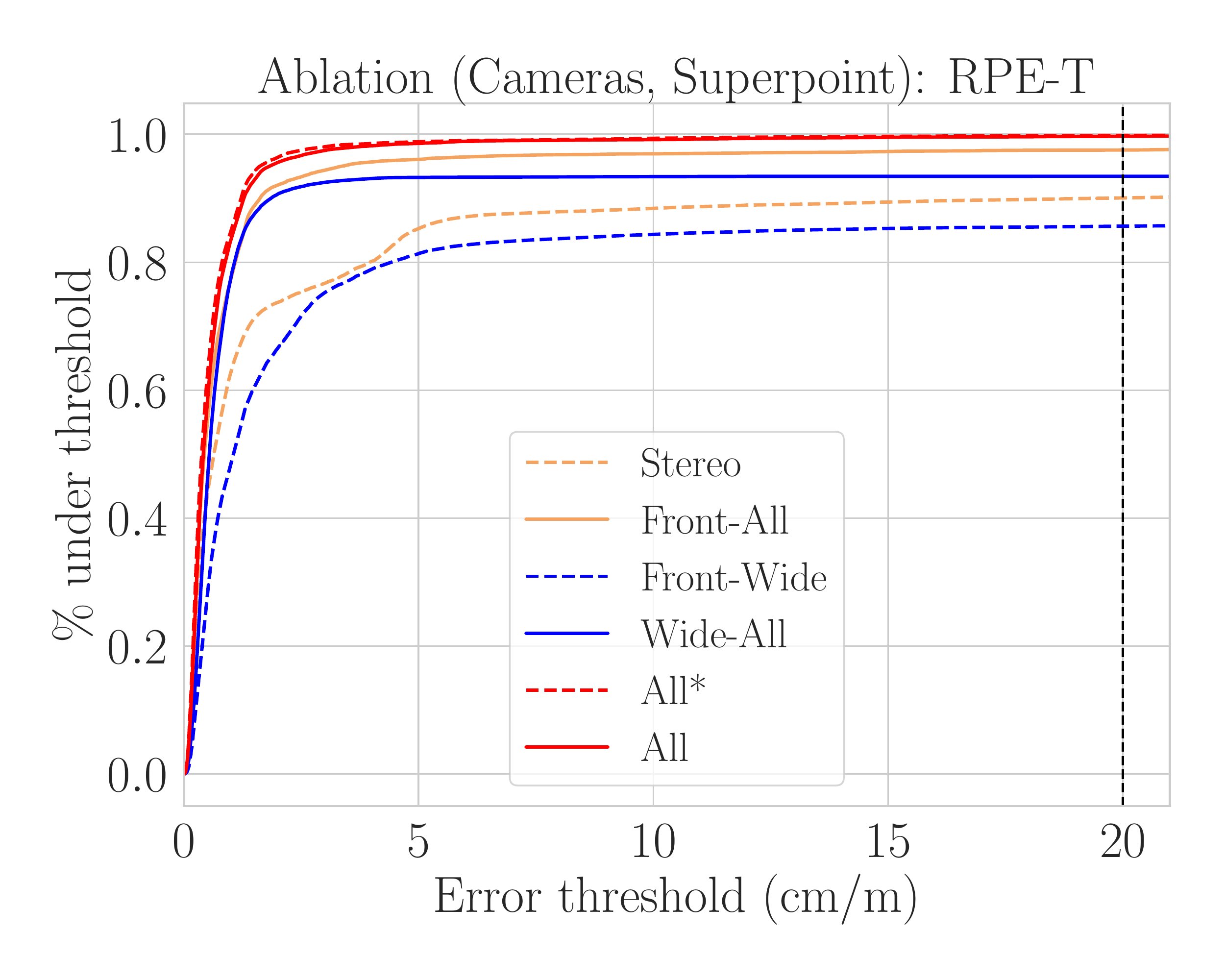}
    \includegraphics[width=0.32\linewidth]{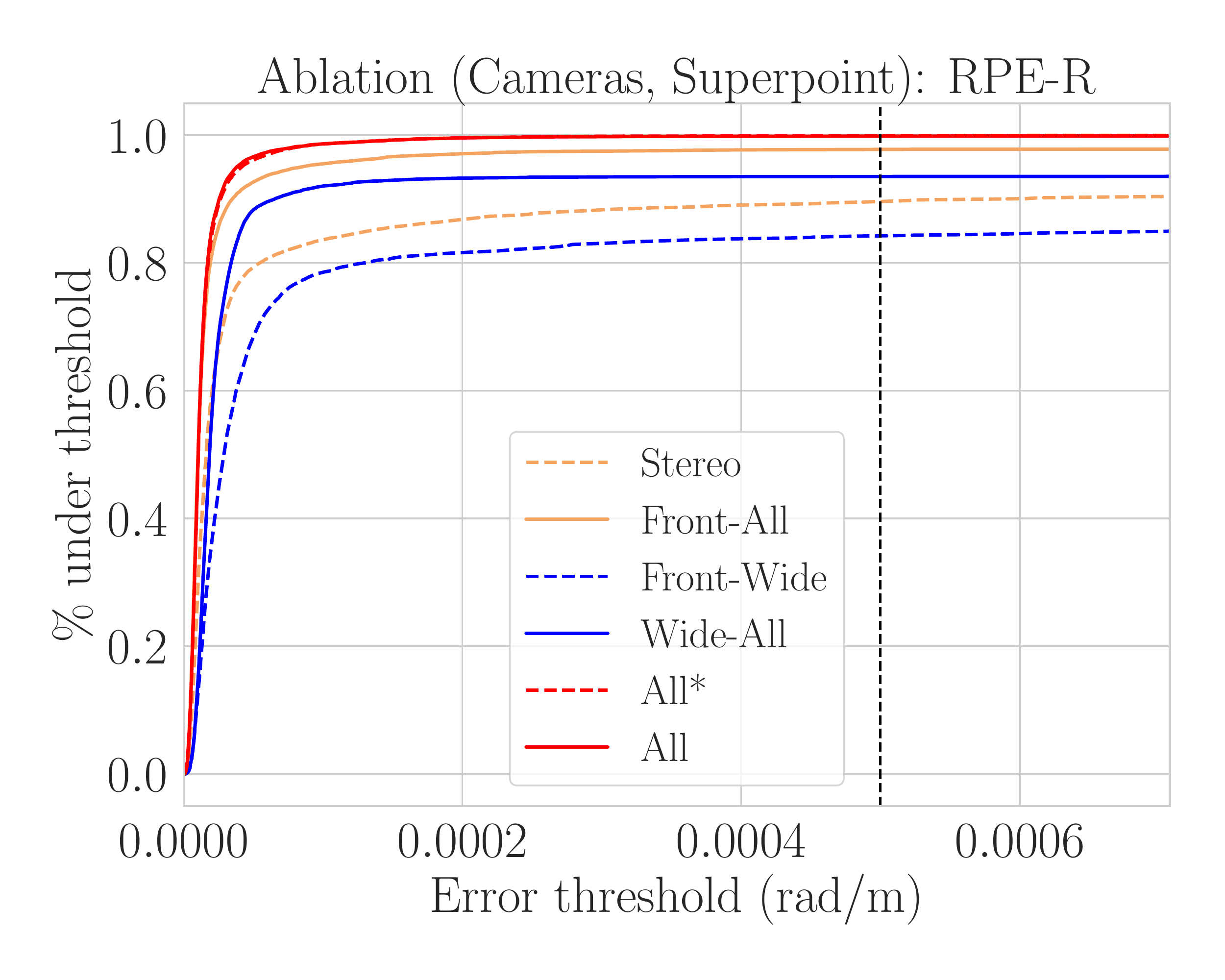}
    \includegraphics[width=0.32\linewidth]{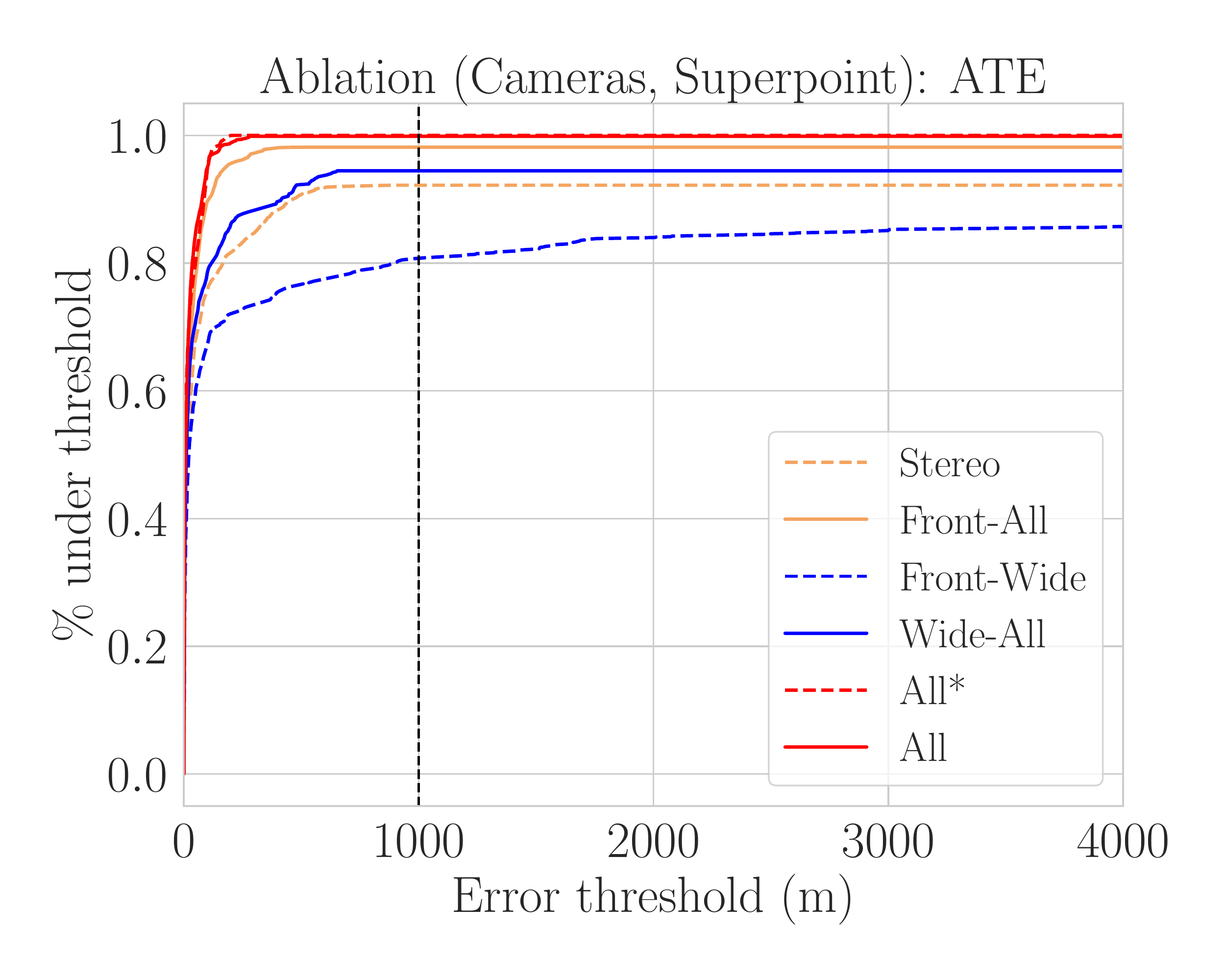}
    \caption{Cumulative error curves of the camera ablation study. (Top) The three
    configurations in the main paper. (Middle) All camera configurations
    in the additional experiments. (Bottom) All camera configurations
    with SuperPoint in place of ORB as the keypoint extractor in the additional
    experiments. Camera configuration legend order corresponds to the
    order in the table.
    \label{fig:cum_curve_camera}}
\end{figure*}

\begin{table*}
  \centering
  \caption{Ablation study for keypoint extractors in the VO mode.
  Time is the average feature extraction time per image
    using 24 CPU cores.}
  \label{tab:ablation-feature-extractor}
  \begin{tabular}{l|r|rrr|rrr|rrr|r}
    \toprule
    \multirow{2}{*}{\textbf{Method}} &
    \multirow{2}{*}{Time\space(s)} &
    \multicolumn{3}{c|}{\rt\space(cm/m)} &
    \multicolumn{3}{c|}{\rr\space(rad/m)} &
    \multicolumn{3}{c|}{\ate\space(m)} &
    \multirow{2}{*}{\sr\space(\%)} \\
    & & @0.5 & @0.9 & AUC(\%) & @0.5 & @0.9 & AUC(\%) & @0.5 & @0.9 & AUC(\%) & \\
    \midrule
    RootSIFT~\cite{rootsift} & \second{0.10} & \second{0.41} & 2.26 & \second{94.06} & \second{1.11E-05} & \second{2.69E-05} & \second{95.84} & \second{6.66} & \second{123.72} & \second{91.50} & \best{98.67} \\
    SuperPoint~\cite{superpoint2018} & 0.35 & \second{0.41} & \best{1.28} & \best{95.28} & \best{1.03E-05} & \best{2.54E-05} & \best{95.90} & 6.83 & \best{78.06} & \best{96.34} & \best{98.67} \\
    R2D2~\cite{r2d22019} & 20.50& \second{0.41} & - & 84.09 & 1.40E-05 & - & 83.72 & 7.42 & - & 86.62 & 88.00 \\
    ORB~\cite{orb} & \best{0.01}  & \best{0.35}& \second{2.14} & 88.79 & \second{1.11E-05} & 6.30E-05 & 88.47 & \best{6.53} & 299.30 & 89.04 & 92.00 \\
    \bottomrule
  \end{tabular}
\end{table*}

\begin{figure*}
    \centering
    \includegraphics[width=0.30\linewidth]{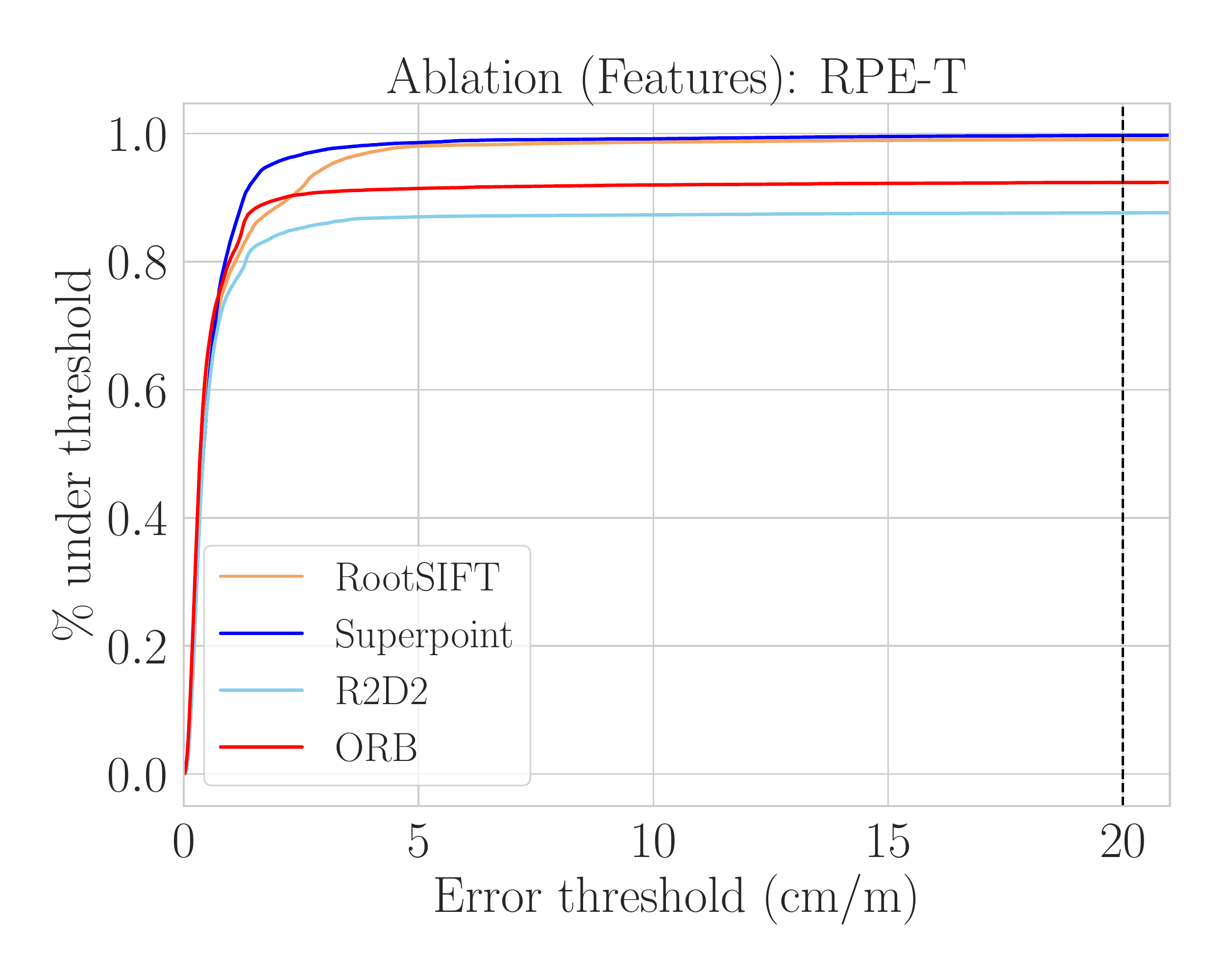}%
    \hfill
    \includegraphics[width=0.30\linewidth]{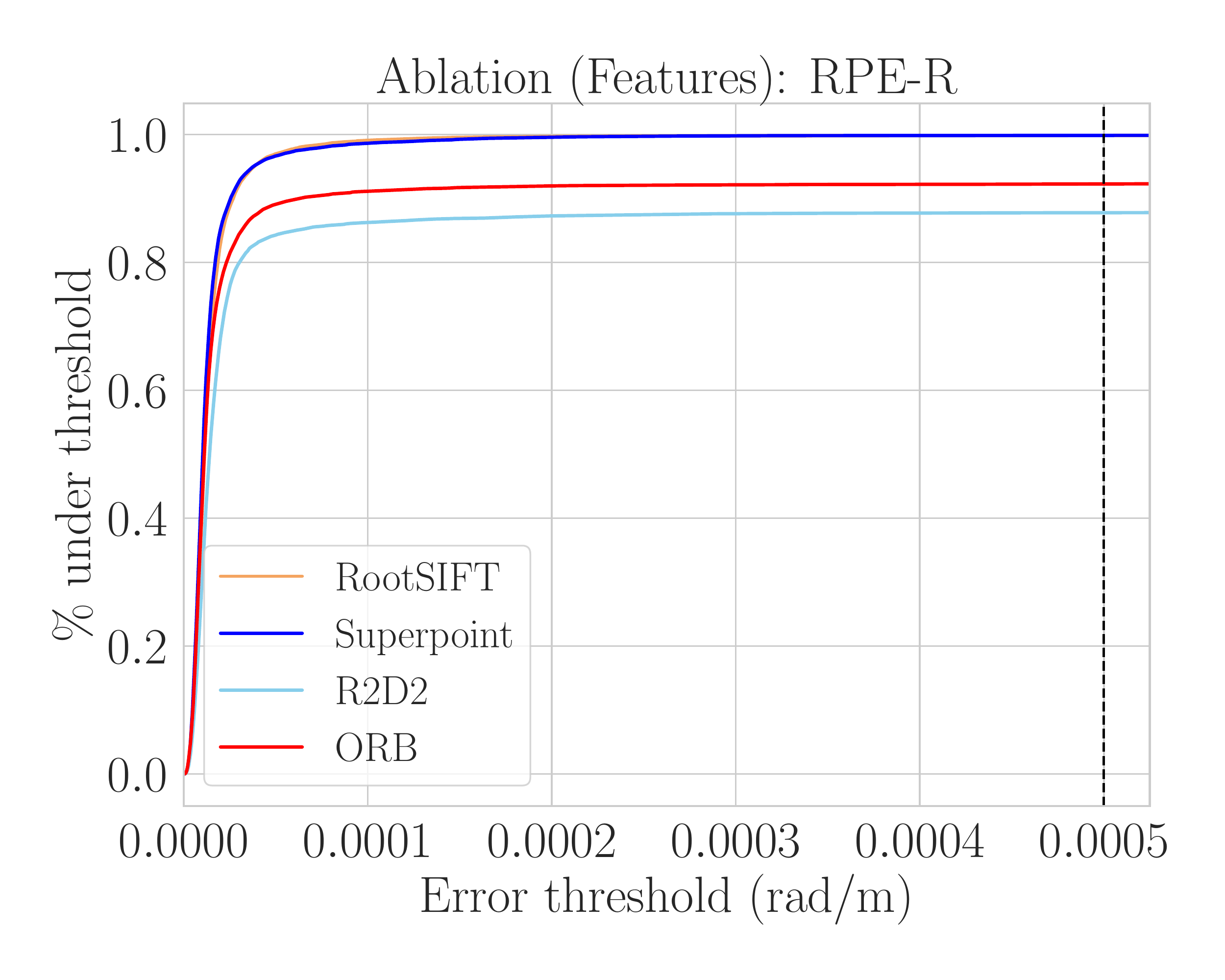}%
    \hfill
    \includegraphics[width=0.30\linewidth]{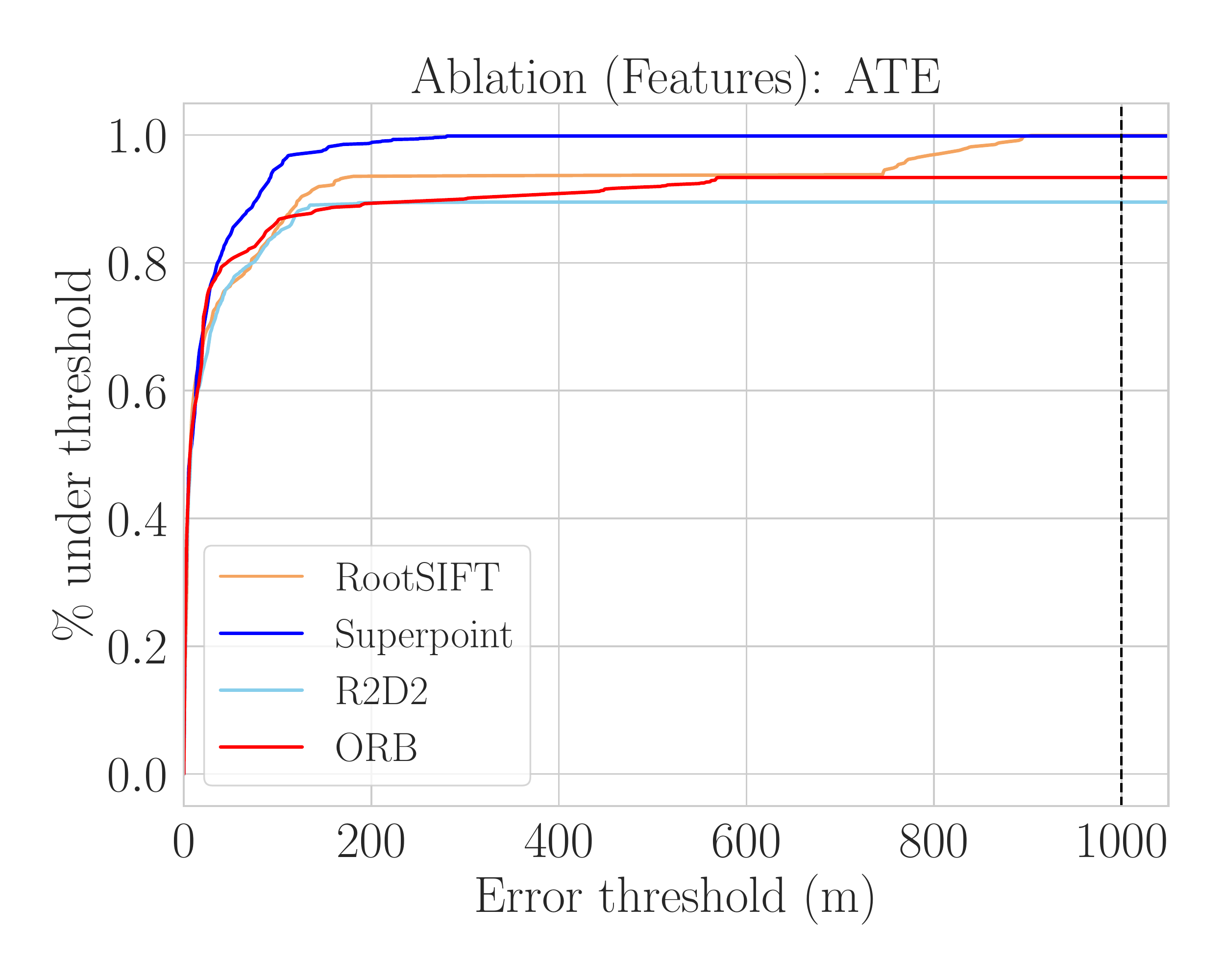}%
    \caption{Cumulative error curves of the keypoint extractor ablation study.
      Our experiments show that while ORB features still remain competitive,
      SuperPoint features lead to the best overall performance, especially in
      terms of translational error. However, this comes at a much higher
      computational cost.
    \label{fig:cum_curve_features}}
    \vspace{-3mm}
\end{figure*}

\begin{table*}
    \centering
    \caption{Per sequence errors of all baselines and our method. Errors averaged over
    all three trials at all evaluated timestamps. - denotes at least one
    trial did not successfully complete the sequence. RPE-T(cm/m),
    RPE-R(rad/m), ATE(m).}
    \label{tab:full-seq-others}
    \rowcolors{2}{blue!10}{white}
    \scalebox{0.81}{
    \addtolength{\tabcolsep}{-3pt}
    \setlength{\arrayrulewidth}{.1em}
    \begin{tabular}{|l|rrr|rrr|rrr|rrr|rrr|rrr|}
      \hline
      \cellcolor{white}&
      \multicolumn{6}{c|}{Monocular} &
      \multicolumn{6}{c|}{Stereo} &
      \multicolumn{6}{c|}{All-Camera} \\
      \hhline{~-|-|-|-|-|-|-|-|-|-|-|-|-|-|-|-|-|-}
      \cellcolor{white}&
      \multicolumn{3}{c|}{LDSO-M} &
      \multicolumn{3}{c|}{ORB-M} &
      \multicolumn{3}{c|}{ORB-S} &
      \multicolumn{3}{c|}{Sync-S} &
      \multicolumn{3}{c|}{Sync-A} &
      \multicolumn{3}{c|}{Ours-A} \\
      \multirow{-3}{*}{\cellcolor{white}Sequence} & RPE-T & RPE-R & ATE & RPE-T & RPE-R & ATE & RPE-T & RPE-R & ATE
      & RPE-T & RPE-R & ATE & RPE-T & RPE-R & ATE & RPE-T & RPE-R & ATE \\
      \hline
      day\_no\_rain\_0 & - & - & - & - & - & - & 2.95 & 8.44E-05 & 40.96 & \second{0.51} & \second{1.70E-05} & \best{6.17} & - & - & - & \best{0.46} & \best{1.25E-05} & \second{7.38} \\
      day\_no\_rain\_1 & 46.45 & 1.61E-03 & 225.75 & 1.25 & \second{2.78E-05} & 5.43 & 1.29 & 3.07E-05 & 5.09 & \second{0.90} & 6.57E-05 & \second{2.74} & 1.66 & 6.55E-05 & 8.43 & \best{0.22} & \best{1.00E-05} & \best{0.55} \\
      day\_no\_rain\_2 & - & - & - & 5.46 & 4.89E-05 & 52.75 & 1.72 & 7.37E-05 & 12.04 & \second{0.41} & \second{2.45E-05} & \second{5.54} & - & - & - & \best{0.29} & \best{1.20E-05} & \best{3.53} \\
      day\_no\_rain\_3 & 21.63 & 2.63E-05 & 382.04 & - & - & - & 1.53 & 3.33E-05 & 13.03 & \second{0.33} & 2.00E-05 & \second{2.95} & 0.85 & \second{1.71E-05} & 5.18 & \best{0.17} & \best{1.16E-05} & \best{2.71} \\
      day\_no\_rain\_4 & - & - & - & 31.93 & 4.33E-05 & 426.90 & \best{0.64} & \best{2.17E-05} & \second{8.85} & \second{0.96} & 3.61E-05 & 10.17 & - & - & - & 0.96 & \second{2.57E-05} & \best{3.74} \\
      day\_no\_rain\_5 & - & - & - & - & - & - & \second{1.06} & \second{2.77E-05} & 6.64 & 1.25 & 5.11E-05 & \second{6.49} & 2.61 & 5.28E-05 & 7.83 & \best{0.44} & \best{1.55E-05} & \best{1.18} \\
      day\_no\_rain\_6 & 4.11 & 1.13E-04 & \best{17.04} & 1.74 & 4.91E-05 & 47.99 & - & - & - & \second{0.79} & \second{4.01E-05} & 24.63 & 3.27 & 2.00E-04 & 73.84 & \best{0.24} & \best{1.24E-05} & \second{20.11} \\
      day\_no\_rain\_7 & 0.72 & 3.35E-05 & 1.75 & 2.74 & 1.52E-04 & 50.96 & 0.64 & \best{1.58E-05} & 1.41 & \second{0.28} & 1.84E-05 & \best{0.38} & 0.50 & \second{1.61E-05} & 1.80 & \best{0.23} & 1.64E-05 & \second{0.49} \\
      day\_no\_rain\_8 & - & - & - & 44.84 & 7.88E-05 & 411.75 & - & - & - & 1.82 & 4.80E-05 & 26.81 & \second{1.63} & \second{3.13E-05} & \second{23.36} & \best{0.40} & \best{1.28E-05} & \best{5.30} \\
      day\_no\_rain\_9 & 17.47 & 4.31E-04 & 74.46 & 26.15 & 1.47E-04 & 571.95 & - & - & - & \second{0.37} & \second{1.82E-05} & \second{2.70} & - & - & - & \best{0.33} & \best{1.70E-05} & \best{2.07} \\
      day\_no\_rain\_10 & 31.34 & 4.76E-05 & 110.64 & 14.70 & 3.54E-05 & 73.36 & 1.12 & 3.73E-05 & 7.43 & \second{0.33} & 1.66E-05 & 2.91 & 0.41 & \second{1.55E-05} & \second{2.66} & \best{0.27} & \best{9.05E-06} & \best{1.96} \\
      day\_no\_rain\_11 & 1.18 & 2.50E-05 & 11.79 & 2.47 & 3.16E-05 & 82.23 & 0.73 & 2.50E-05 & 3.12 & \second{0.26} & 1.53E-05 & \second{1.70} & 0.28 & \second{1.49E-05} & 2.86 & \best{0.24} & \best{1.12E-05} & \best{1.38} \\
      \hline
      day\_rain\_0 & 23.37 & \second{2.41E-05} & 652.45 & - & - & - & - & - & - & \second{1.59} & 5.73E-05 & \second{65.56} & 4.06 & 2.88E-05 & 73.63 & \best{0.31} & \best{1.33E-05} & \best{12.56} \\
      day\_rain\_1 & - & - & - & 19.60 & 3.43E-05 & 228.22 & - & - & - & \second{0.49} & \second{3.33E-05} & \second{7.85} & - & - & - & \best{0.31} & \best{1.52E-05} & \best{3.01} \\
      day\_rain\_2 & \second{8.45} & \second{2.75E-05} & \second{34.99} & 19.52 & 8.92E-05 & 82.56 & 11.85 & 6.23E-04 & 130.85 & - & - & - & - & - & - & \best{0.77} & \best{2.05E-05} & \best{4.91} \\
      day\_rain\_3 & - & - & - & \second{31.31} & \second{1.11E-04} & \second{219.47} & - & - & - & - & - & - & - & - & - & \best{0.37} & \best{1.15E-05} & \best{14.46} \\
      day\_rain\_4 & 23.91 & 3.19E-04 & 138.17 & 29.49 & 1.94E-04 & 173.28 & \second{5.19} & 1.79E-04 & \second{27.15} & 6.59 & \second{4.39E-05} & 83.15 & - & - & - & \best{0.72} & \best{2.14E-05} & \best{7.21} \\
      day\_rain\_5 & 11.88 & \second{2.40E-05} & 25.51 & 4.65 & 2.71E-05 & 8.62 & 10.32 & 1.92E-04 & 18.70 & \second{0.44} & 3.60E-05 & \second{0.86} & 1.86 & 3.28E-05 & 2.90 & \best{0.16} & \best{1.23E-05} & \best{0.51} \\
      \hline
      hwy\_no\_rain\_0 & - & - & - & - & - & - & - & - & - & \second{2.21} & 4.02E-05 & 218.65 & 2.51 & \second{3.25E-05} & \second{190.31} & \best{0.41} & \best{1.52E-05} & \best{52.94} \\
      hwy\_no\_rain\_1 & - & - & - & - & - & - & 2.92 & \second{3.26E-05} & \second{49.79} & 3.21 & 9.19E-05 & 888.29 & \second{2.91} & 4.01E-05 & 92.70 & \best{0.59} & \best{1.49E-05} & \best{22.54} \\
      hwy\_no\_rain\_2 & - & - & - & - & - & - & - & - & - & \second{3.08} & \second{3.30E-05} & \best{324.44} & 3.34 & 3.56E-05 & 395.12 & \best{0.45} & \best{1.40E-05} & \second{377.00} \\
      hwy\_no\_rain\_3 & - & - & - & - & - & - & 9.04 & 7.69E-05 & 852.20 & 5.53 & 7.38E-05 & 185.56 & \second{3.38} & \second{5.60E-05} & \second{84.41} & \best{0.48} & \best{1.50E-05} & \best{36.58} \\
      hwy\_rain\_0 & - & - & - & - & - & - & - & - & - & \second{0.74} & 2.57E-05 & 171.41 & 1.38 & \second{1.86E-05} & \second{168.77} & \best{0.31} & \best{9.69E-06} & \best{63.98} \\
      hwy\_rain\_1 & - & - & - & - & - & - & - & - & - & - & - & - & - & - & - & - & - & - \\
      hwy\_rain\_2 & - & - & - & - & - & - & - & - & - & - & - & - & - & - & - & - & - & - \\
      \hline
    \end{tabular}
    \addtolength{\tabcolsep}{3pt}
  }
\end{table*}

\begin{table*}
  \centering
  \caption{Ablation study for the impact of the KMF-selection heuristics on
  the system performance.}
  \label{tab:ablation-keyframe}
  \begin{tabular}{lcc|rrr|rrr|rrr|r}
    \toprule
    \multirow{2}{*}{\textbf{Method}} &
    \multicolumn{2}{c}{\textbf{KMF Heuristics}} &
    \multicolumn{3}{c|}{\rt\space(cm/m)} &
    \multicolumn{3}{c|}{\rr\space(rad/m)} &
    \multicolumn{3}{c|}{\ate\space(m)} &
    \multirow{2}{*}{\sr\space(\%)} \\
    & reobservability & motion & @0.5 & @0.9 & AUC(\%) & @0.5 & @0.9 & AUC(\%) & @0.5 & @0.9 & AUC(\%) & \\
    \midrule
    Sync-S & \yes & & 2.90 & - & 65.24 & 3.28E-05 & - & 71.39 & 68.56 & - & 68.02 & 82.67 \\
    Sync-S & \yes & \yes & \best{1.30} & - & \best{77.54} & \best{2.91E-05} & - & \best{78.37} & \best{24.53} & - & \best{77.44} & \best{84.00} \\
    \bottomrule
  \end{tabular}
\end{table*}

\begin{table*}
  \centering
  \caption{Number of KMFs selected per validation sequence, comparing ORB-SLAM2, ours stereo
  with reobservability-only heuristics, and ours stereo with a combined heuristics.
  Empty cells correspond to unfinished sequences.
  \label{tab:ablation-keyframe-numbers}}
  \begin{tabular}{lccc}
    \toprule
    sequence & ORB-S~\cite{mur2017orb} & Ours-S (r-only) & Ours-S (combined) \\
    \midrule
    day\_no\_rain\_0 & 1718 & 1322 & 1759 \\
    day\_no\_rain\_1 & 2249 & 1645 & 2054 \\
    day\_no\_rain\_2 & 3795 & 2370 & 2690 \\
    day\_no\_rain\_3 & 1245 & 969 & 1399 \\
    day\_no\_rain\_4 & 2364 & 1734 & 2143 \\
    day\_no\_rain\_5 & - & 1455 & 2034 \\
    day\_no\_rain\_6 & 3373 & - & 2983 \\
    day\_no\_rain\_7 & 501 & 464 & 633 \\
    day\_no\_rain\_8 & 1724 & 1121 & 2104 \\
    day\_no\_rain\_9 & - & 1598 & 1744 \\
    day\_no\_rain\_10 & 1263 & 982 & 1505 \\
    day\_no\_rain\_11 & 1593 & 1107 & 1559 \\
    \midrule
    day\_rain\_0 & - & 1522 & 2838 \\
    day\_rain\_1 & - & 1902 & 3043 \\
    day\_rain\_2 & 1097 & - & - \\
    day\_rain\_3 & 1952 & 2181 & - \\
    day\_rain\_4 & 2520 & 2339 & 3119 \\
    day\_rain\_5 & 383 & 408 & 531 \\
    \midrule
    hwy\_no\_rain\_0 & 3646 & 3507 & 6532 \\
    hwy\_no\_rain\_1 & 2400 & 2094 & 3994 \\
    hwy\_no\_rain\_2 & 3358 & 2495 & 4632 \\
    hwy\_no\_rain\_3 & 2124 & 2242 & 4484 \\
    hwy\_rain\_0 & 1835 & 2316 & 5132 \\
    hwy\_rain\_1 & - & - & - \\
    hwy\_rain\_2 & - & - & - \\
    \bottomrule
  \end{tabular}
\end{table*}

Table~\ref{tab:quantitative_baseline} and Table~\ref{tab:quantitative_baseline_vo}
compare (1) third-party baselines, (2) our implementation of the synchronous baselines,
and (3) our main system, in the SLAM mode and visual odometry (VO) mode respectively.
In the VO mode we disable loop closing, and relocalization in ORB-SLAM. %
Figs.~\ref{fig:cum_curve_baselines_lc} and~\ref{fig:cum_curve_baselines_vo} plot
the respective cumulative error curves.

Note that some metrics of our full system with loop closure in Table~\ref{tab:quantitative_baseline}
are slightly worse than those without loop closure in Table~\ref{tab:quantitative_baseline_vo}.
The difference can also be observed in the DSO experiments.
The difference in our system is due to the stochasticity of the trials,
instead of loop closing failures.
To support our claim, in the main paper
we plot the drift relative to the ground truth with and without loop closure to
show that loop closing successfully reduced global drifts at all key multi-frames where
loop closing was performed.
Furthermore, Table~\ref{tab:ablation-loop-closure} compares all metrics on the 8/25
validation sequences where loop closure was performed, and shows that the 8 loop closing
sequences did not contribute to the metrics differences over all 25 validation sequences.

Table~\ref{tab:ablation-motion} showcases the motion model ablation study results.
For simplicity, loop closure is disabled.
The asynchronous linear
motion model setting represents the trajectory with a linear motion model parameterized
by a 6-DoF pose $\linpose_i$ at each representative timestamp $\timet{i}$. The motion model
is explained in detail in Sec.~\ref{sec:method}. Similar to the main system, we estimate
linear motion model parameter during tracking, and we jointly refine the linear motion model
parameters along with the map points during bundle adjustment,
with the reprojection energy similar to that of tracking. Our experiments show
that modeling the vehicle motion using cubic B-splines leads to improved
performance due to the splines' ability to impose a realistic motion prior on
the estimated trajectories.

In the additional motion model ablation results, we also compare
with linear and cubic B-spline models that perform
interpolation in $\SO{3}$ and $\bR^3$ separately, instead of jointly in
$\SE{3}$. Previous works~\cite{haarbach2018spline,ovren2018trajectory,
ovren2019trajectory,sommer2019efficient} have shown that the split
interpolation formulation is generally better in terms of both computation
time and trajectory representation.

The results show our main system with
cubic B-spline model and full interpolation in $\SE{3}$ has the best performance overall.
The split-interpolation motion models have slightly worse performance
in our experiments, most likely because full interpolation
in $\SE{3}$ is more apt at modeling curvy trajectories (\eg, during turns).

Table~\ref{tab:ablation-num-cameras} and Fig.~\ref{fig:cum_curve_camera}
compare with additional camera configurations during tracking and mapping.
In the additional camera configurations, during local mapping
we additionally match and triangulate new map points
between the wide front left and wide front middle cameras, and between
the wide front middle and wide front right cameras within the same key multi-frame.
Note that in the settings without stereo cameras, we still use stereo cameras (only)
for initialization.

Configurations without stereo
cameras have worse performance, and we argue this is due to ORB's poor performance
in the much harder wide-baseline image matching problem posed by little overlap
between the wide front cameras during new map point creation.
The table shows that with keypoint extractors such as SuperPoint~\cite{superpoint2018}
in place of ORB, this performance gap is significantly narrowed.

Table~\ref{tab:ablation-feature-extractor} and Fig.~\ref{fig:cum_curve_features}
show the full results of the keypoint extractor
ablation study. Compared to ORB, SIFT and SuperPoint finish more sequences and have
better ATE, with the caveat that feature extraction takes more time.

\subsubsection{Per sequence results}
Table~\ref{tab:full-seq-others} shows per-sequence errors comparing
baselines and our method, all with loop closure. We report the mean over all three trials. If
at least one trial did not complete the sequence successfully, we do not report
results for that sequence.

\subsubsection{KMF heuristics ablation}
To study the effect of the combined KMF selection heuristics that factors in
both map point reobservability and motion, we perform an ablation study
where we run our stereo + synchronous discrete-time
motion model implementation with a reobservability-only KMF
heuristic.
Table~\ref{tab:ablation-keyframe} shows that our stereo implementation
with a reobservability-only heuristic performs worse than the
stereo system with the more robust combined heuristic. 
Table~\ref{tab:ablation-keyframe-numbers}
compares the number of key frames inserted by ORB-SLAM2, our stereo sync with a
reobservability-only heuristic,
and our stereo sync with a reobservability+motion heuristic.
The key frame numbers are taken from a randomly-selected
trial. The table shows
that our reobservability-only heuristic in general inserts fewer key frames
than ORB-SLAM2, and that the combined heuristic selects almost twice as many
key frames during highway sequences, when the vehicle is driving very fast
in a highly repetitive scene.

\subsection{Qualitative Results}%
\label{sub:qualitative_results}

\subsubsection{Qualitative Trajectories}%
\begin{figure*}
    \centering
    \includegraphics[width=0.24\linewidth]{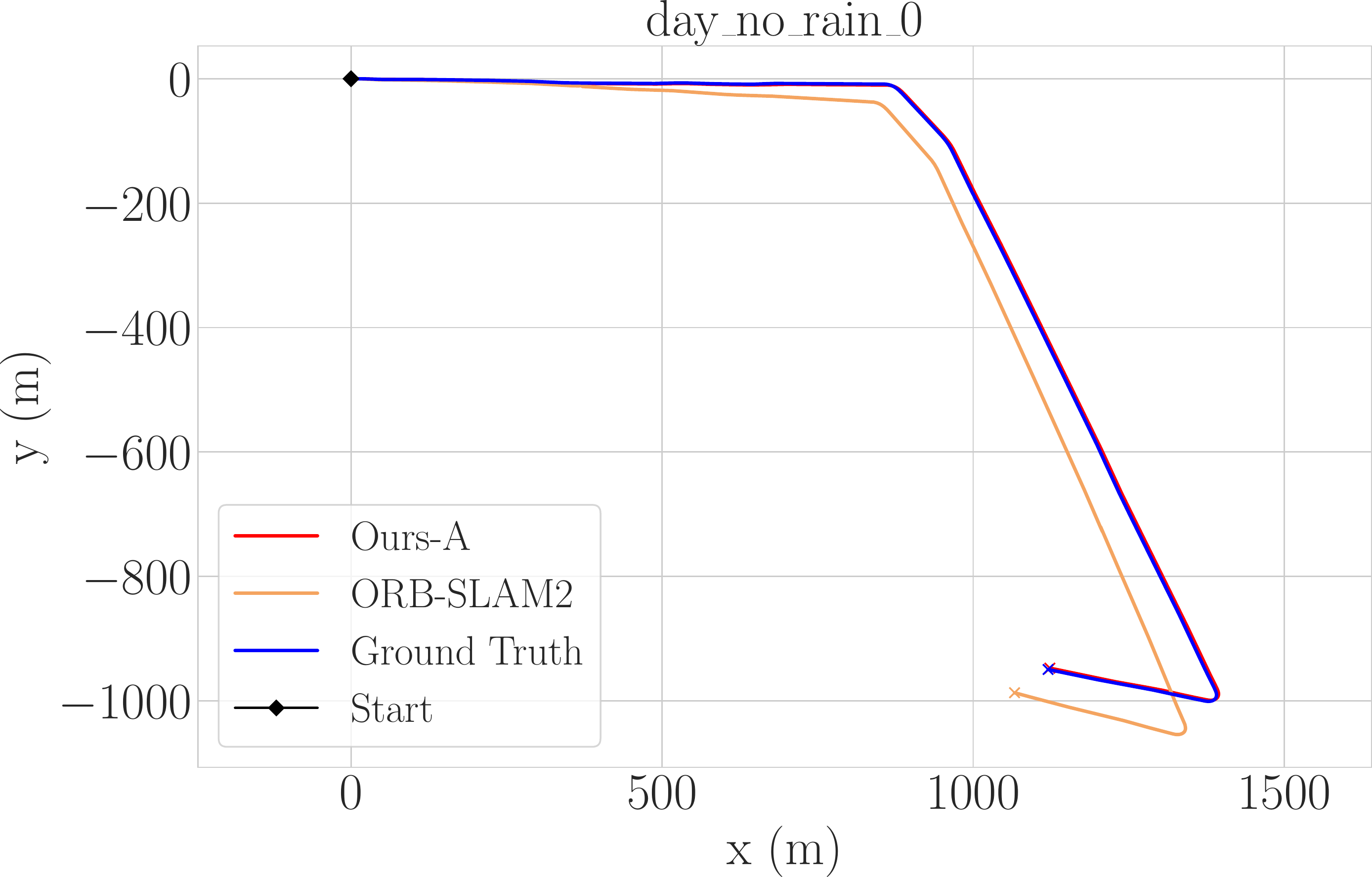}
    \includegraphics[width=0.24\linewidth]{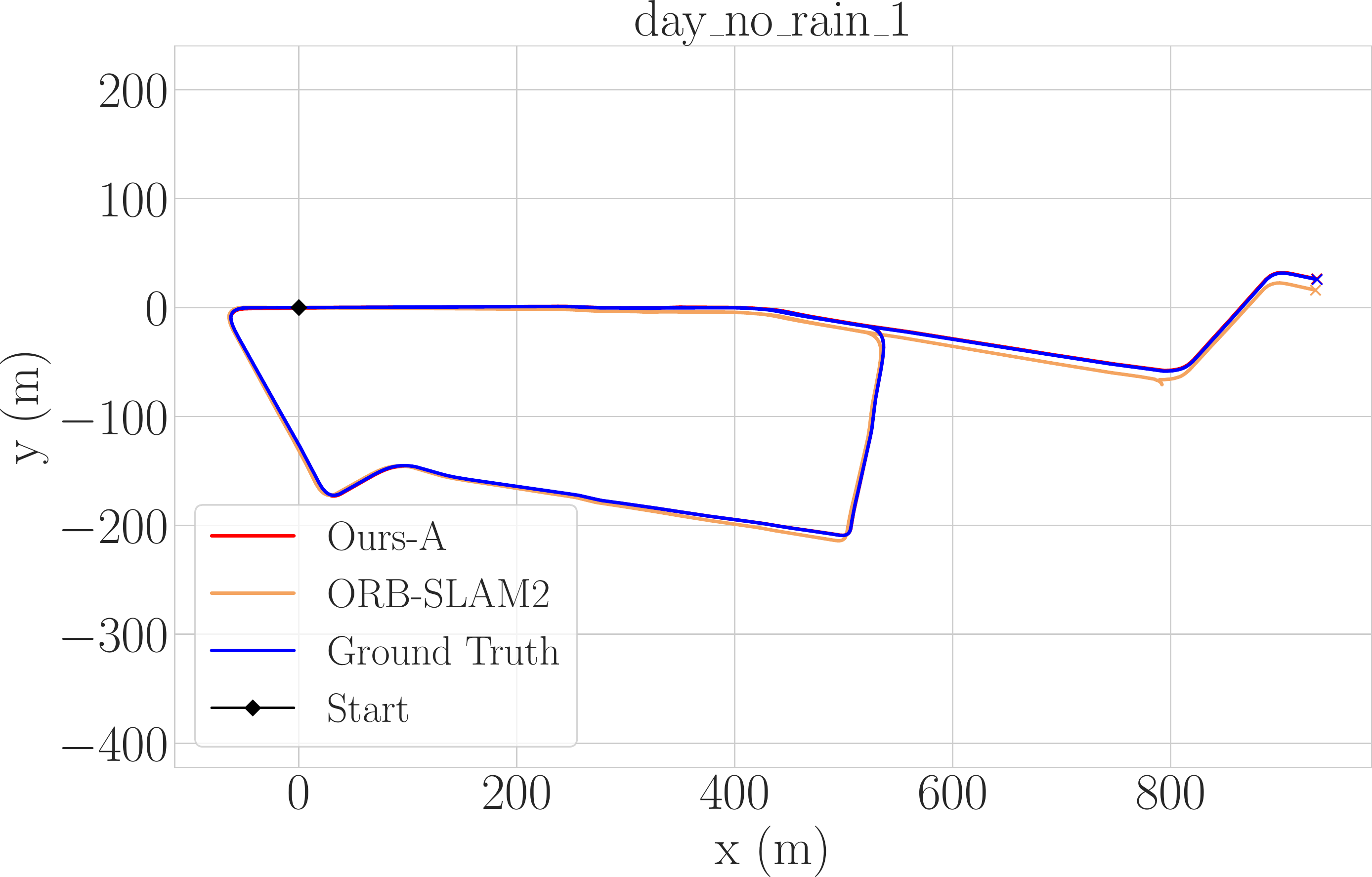}
    \includegraphics[width=0.24\linewidth]{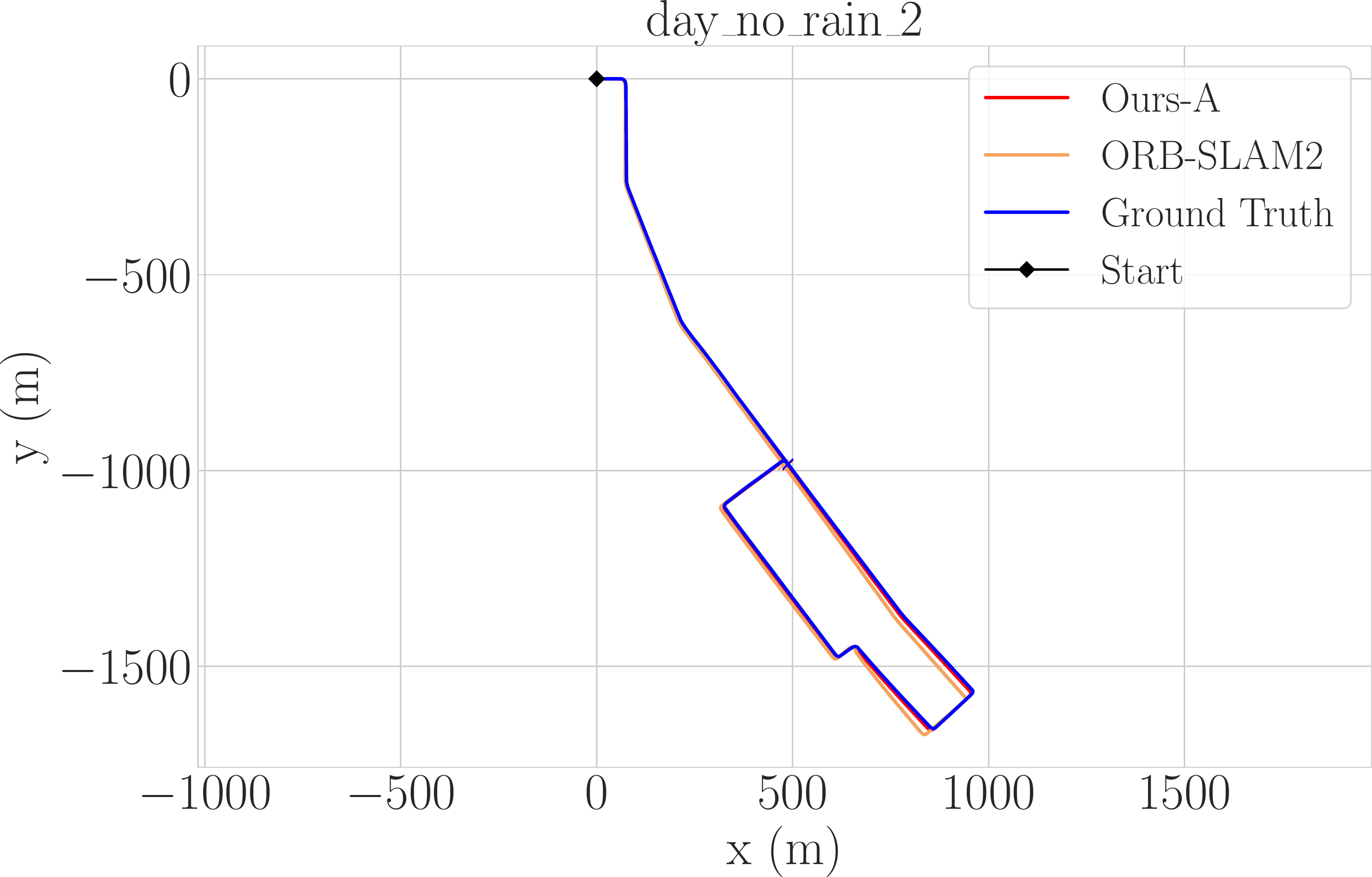}
    \includegraphics[width=0.24\linewidth]{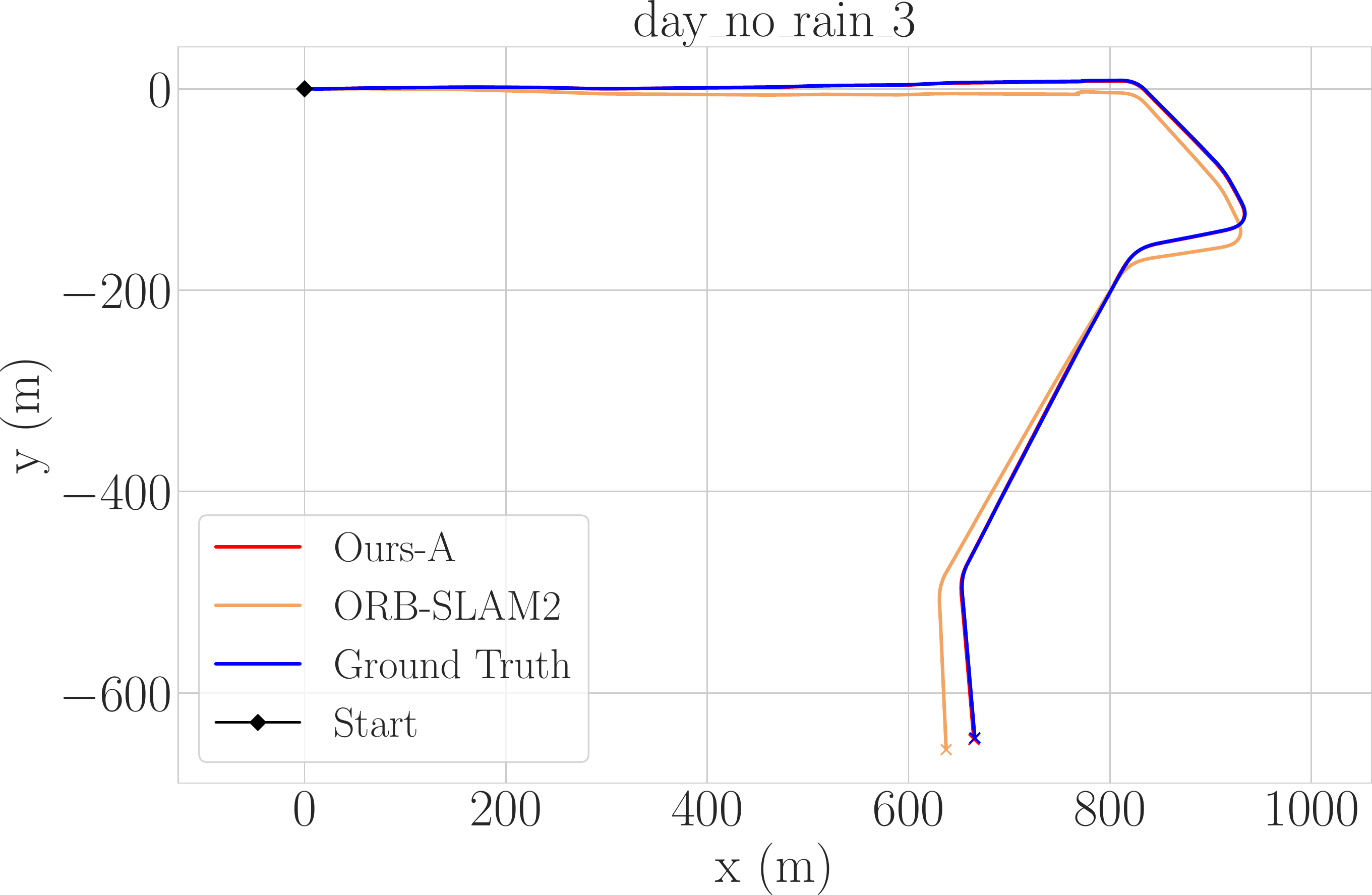}
    \includegraphics[width=0.24\linewidth]{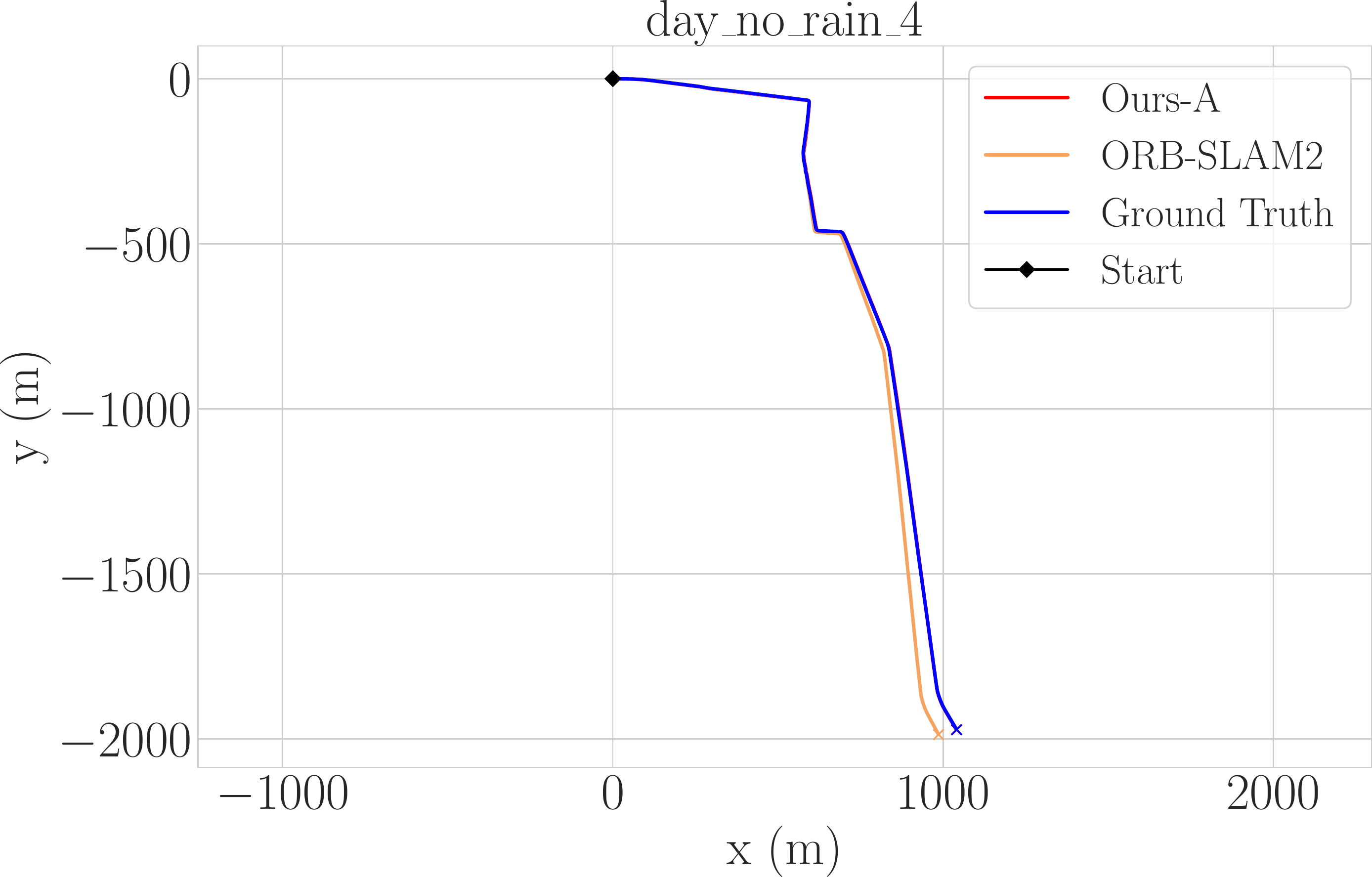}
    \includegraphics[width=0.24\linewidth]{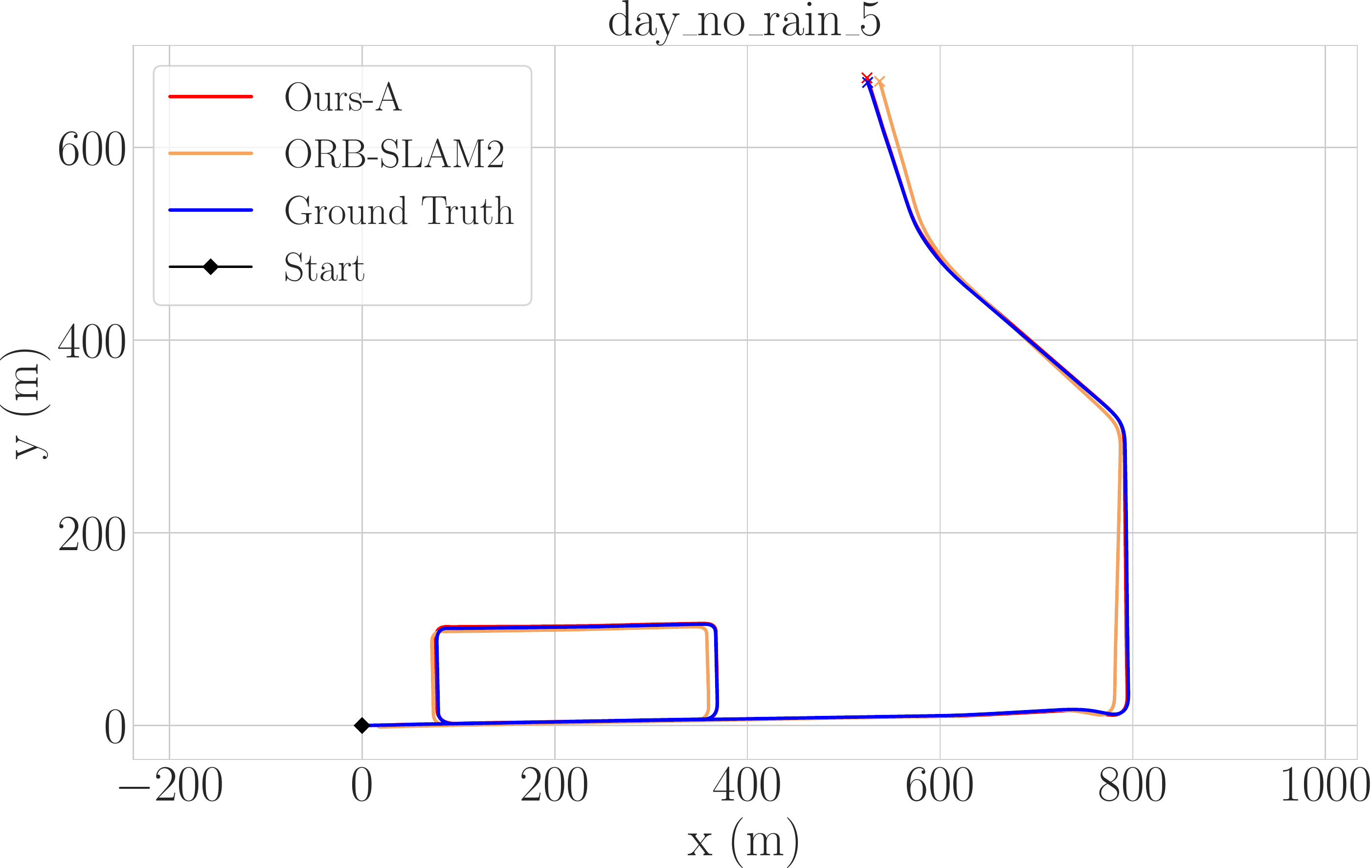}
    \includegraphics[width=0.24\linewidth]{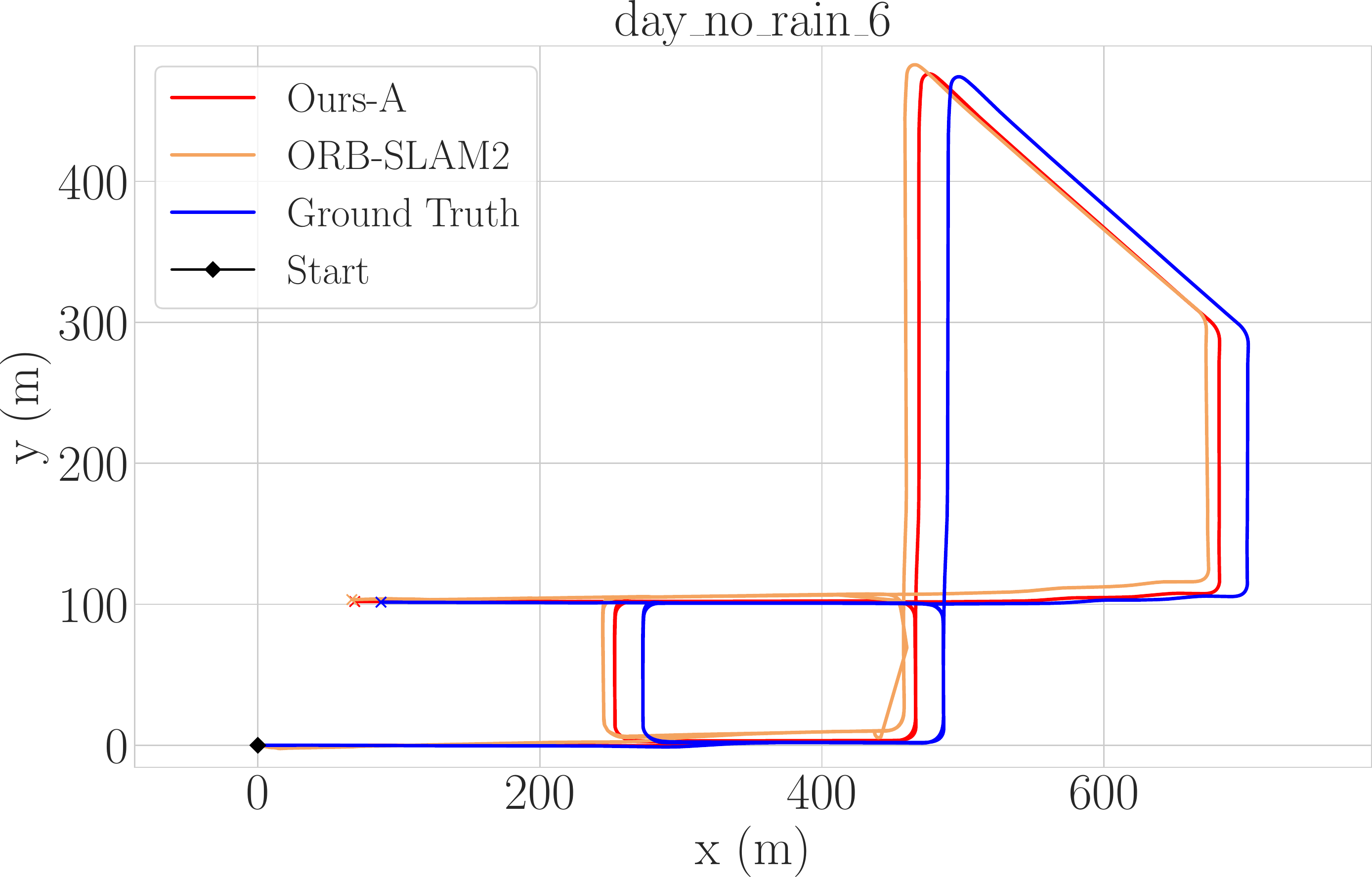}
    \includegraphics[width=0.24\linewidth]{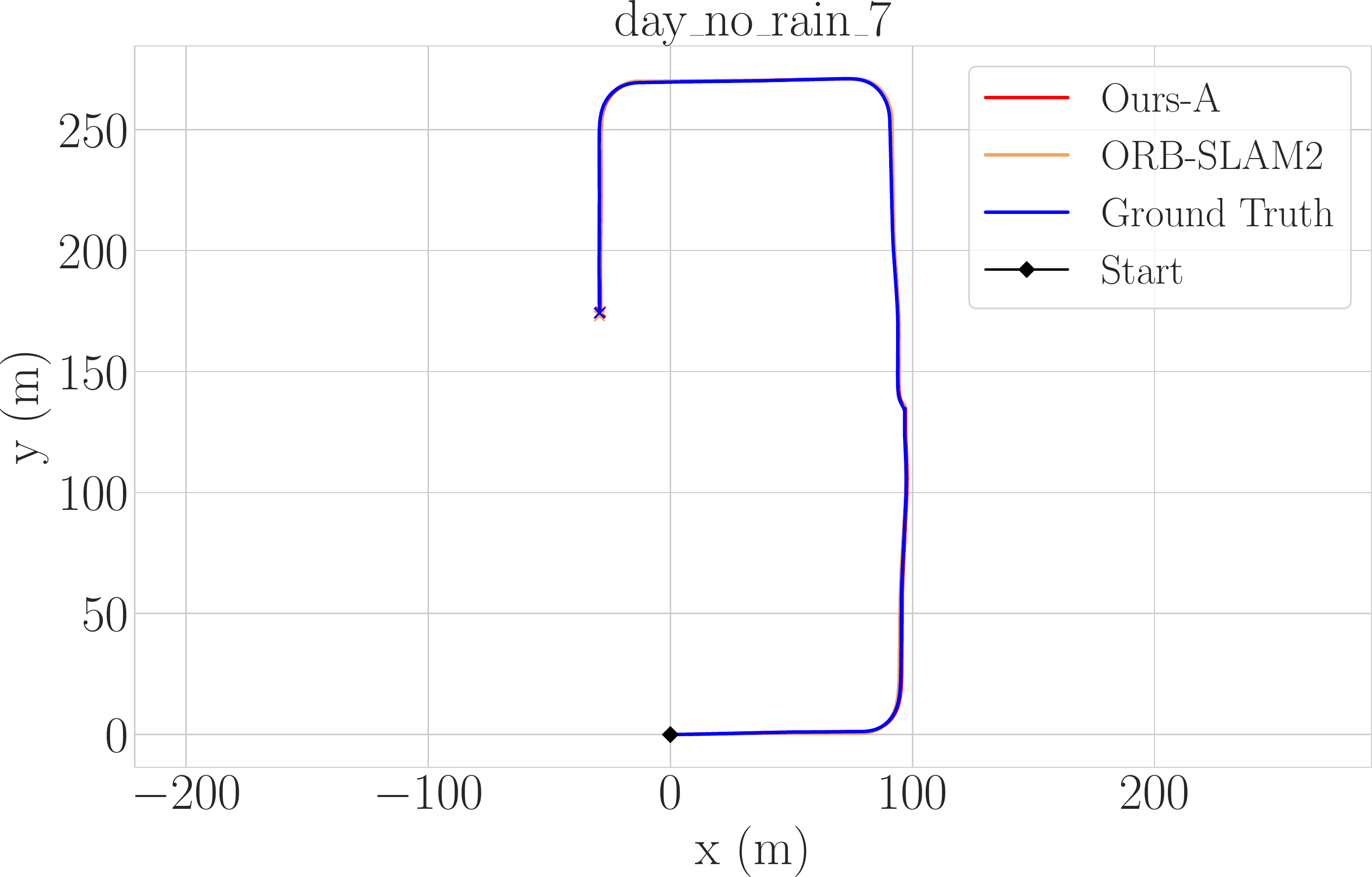}
    \includegraphics[width=0.24\linewidth]{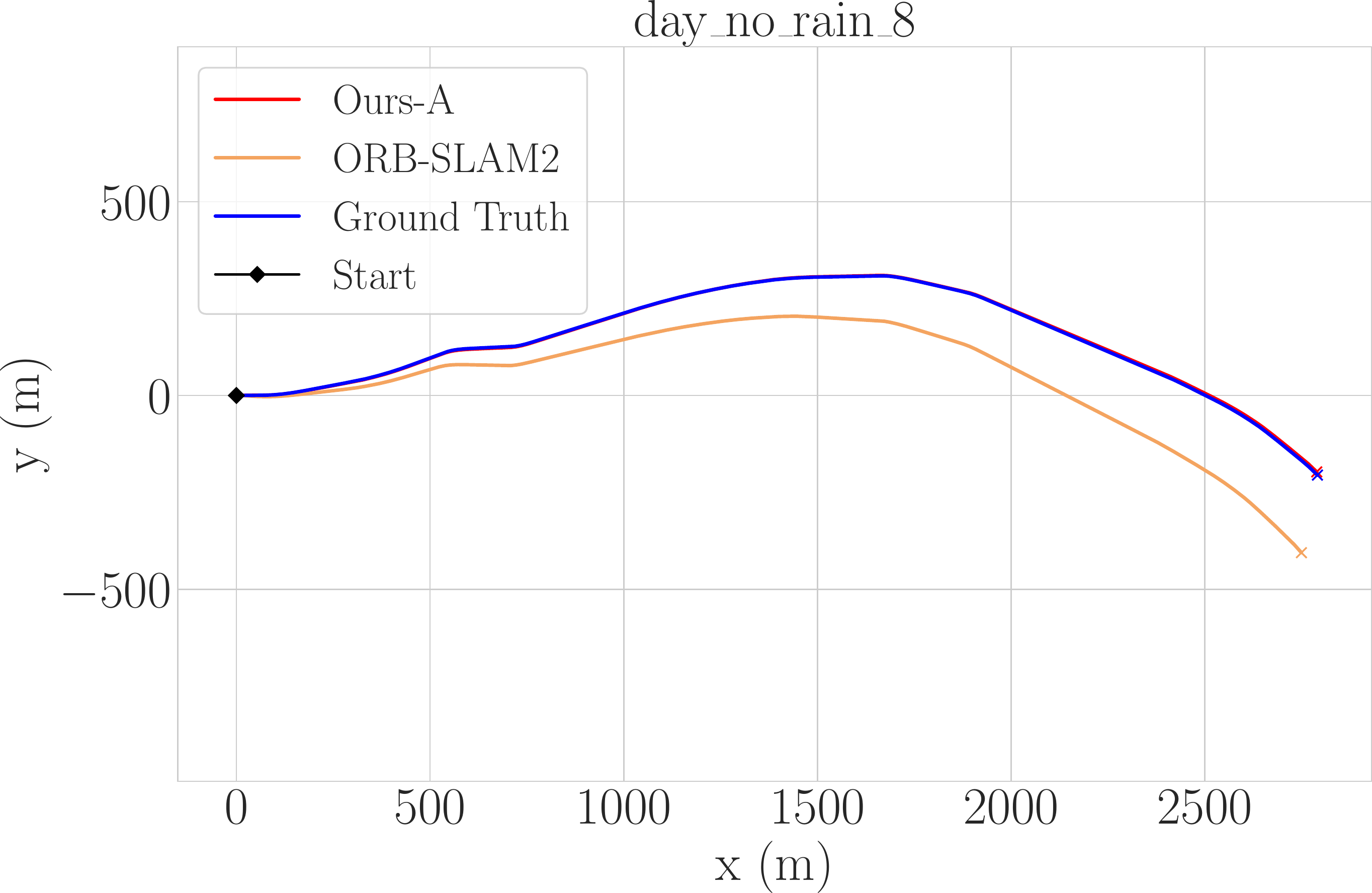}
    \includegraphics[width=0.24\linewidth]{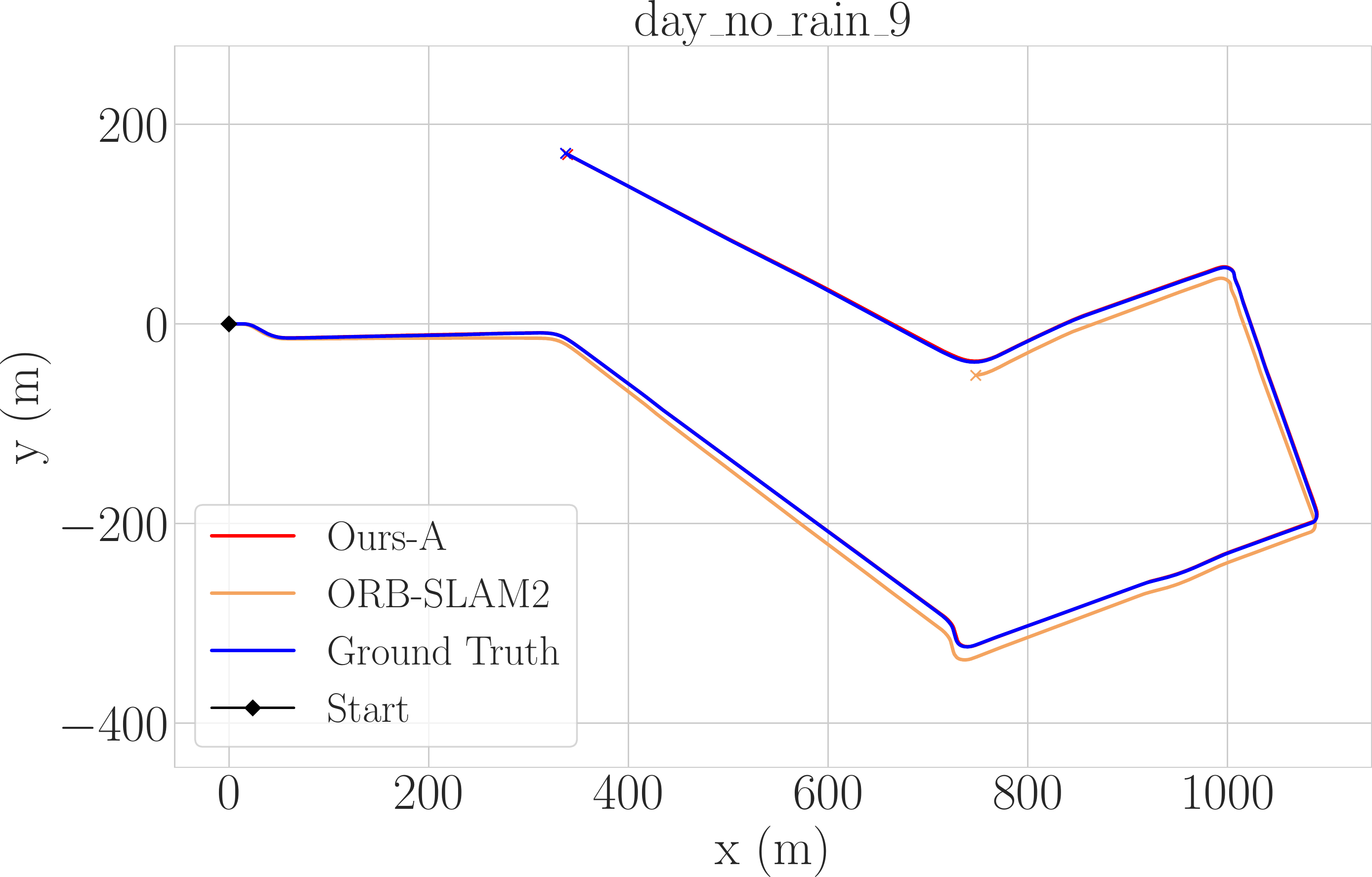}
    \includegraphics[width=0.24\linewidth]{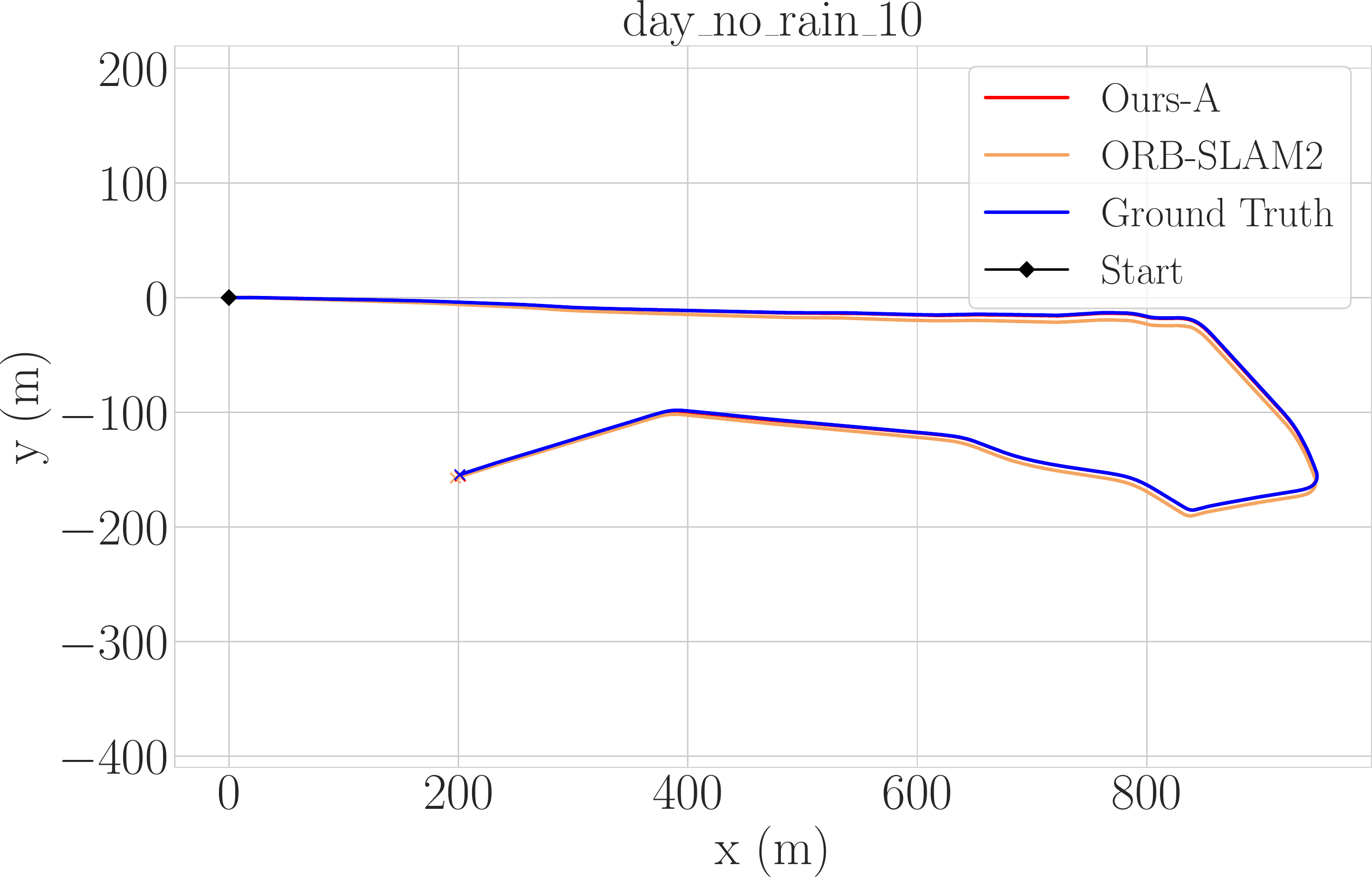}
    \includegraphics[width=0.24\linewidth]{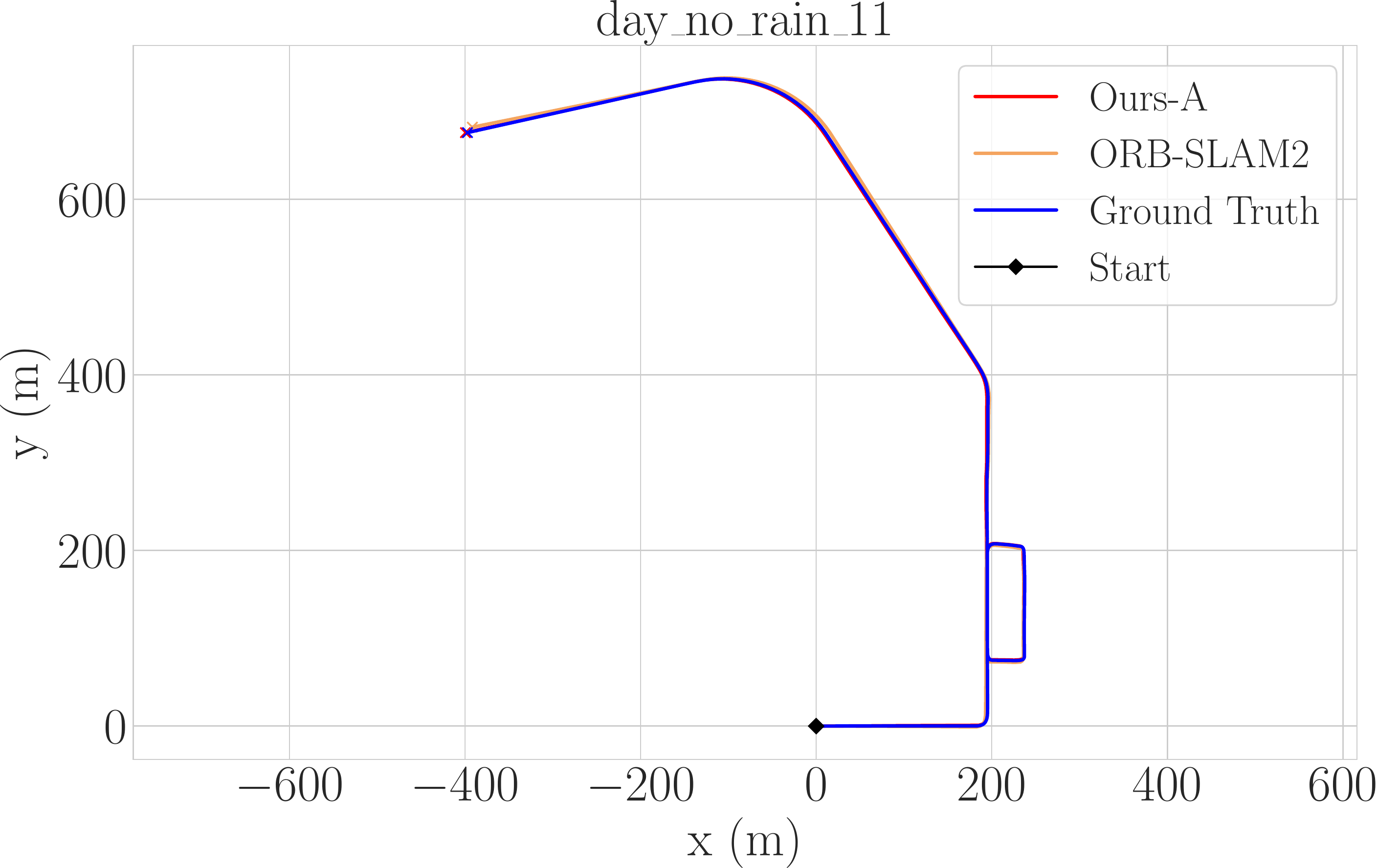}
    \includegraphics[width=0.24\linewidth]{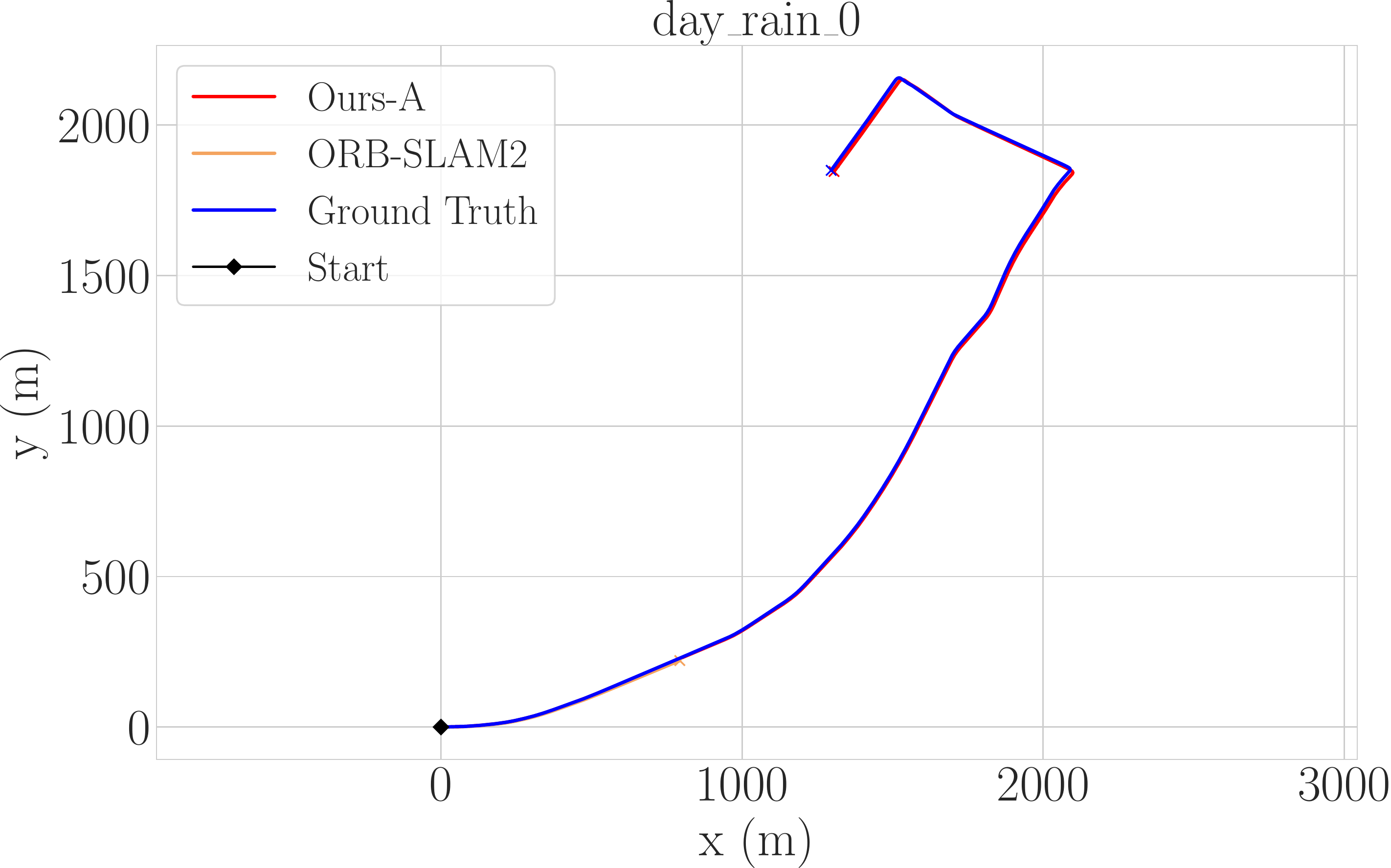}
    \includegraphics[width=0.24\linewidth]{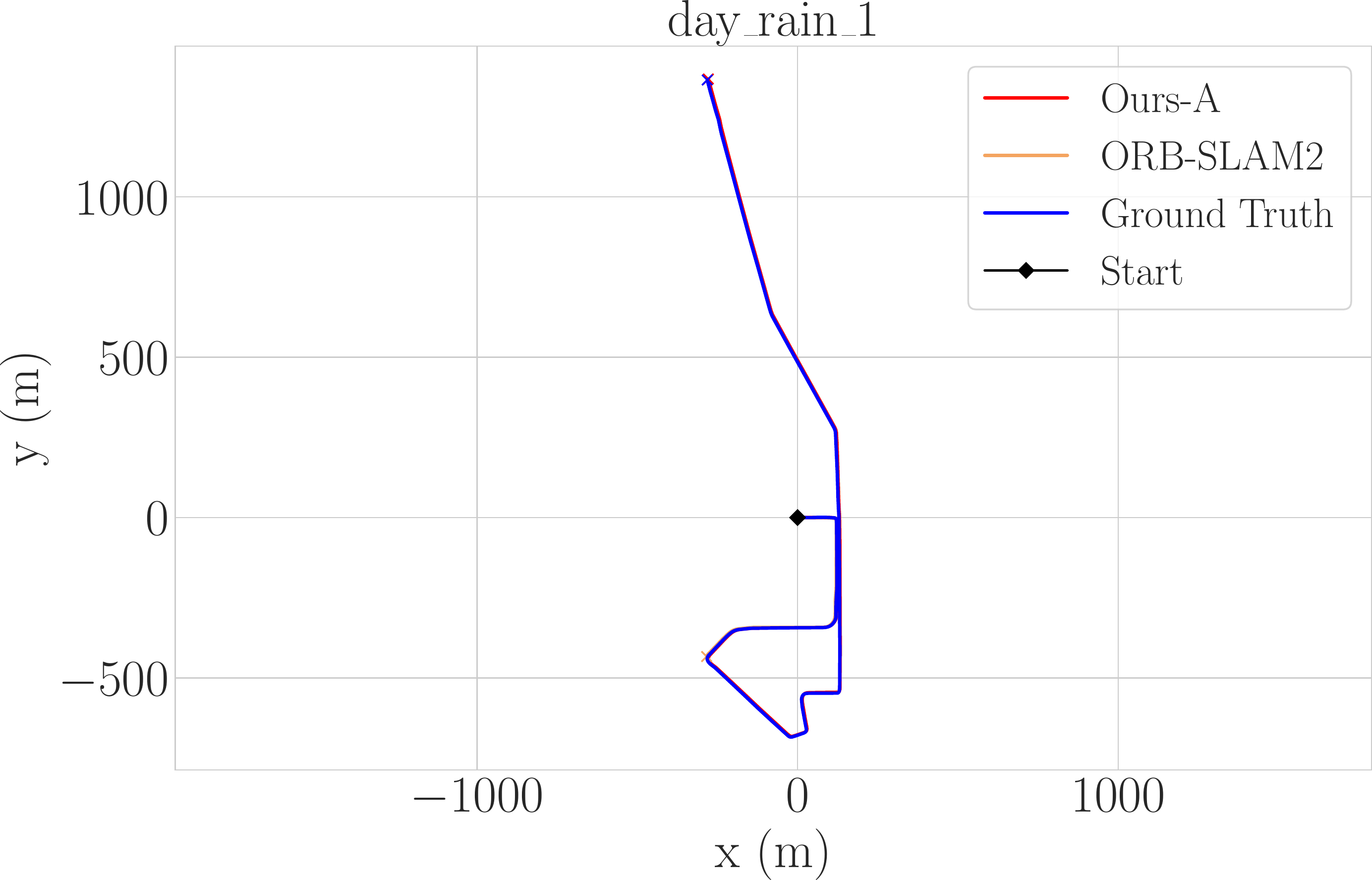}
    \includegraphics[width=0.24\linewidth]{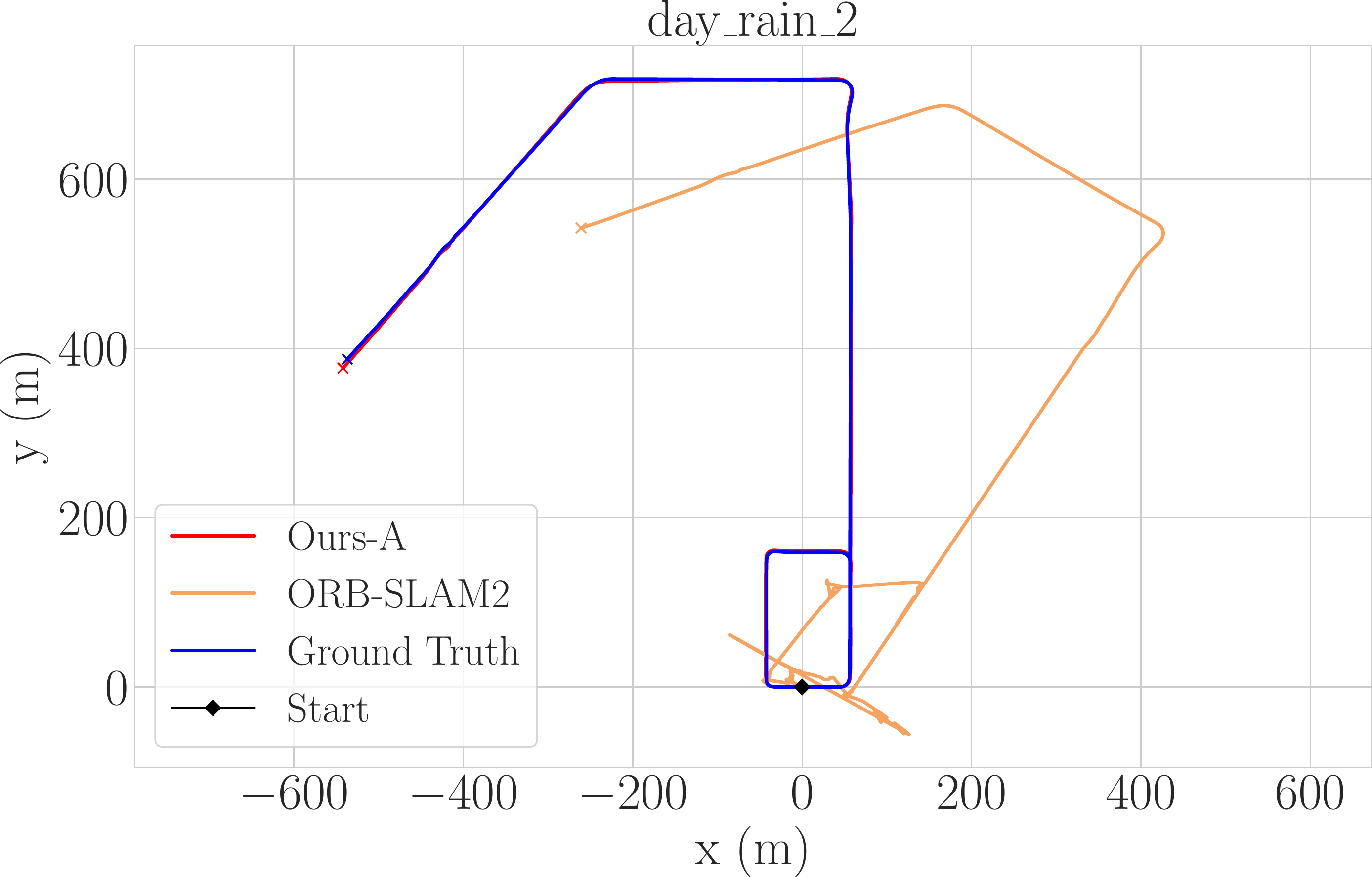}
    \includegraphics[width=0.24\linewidth]{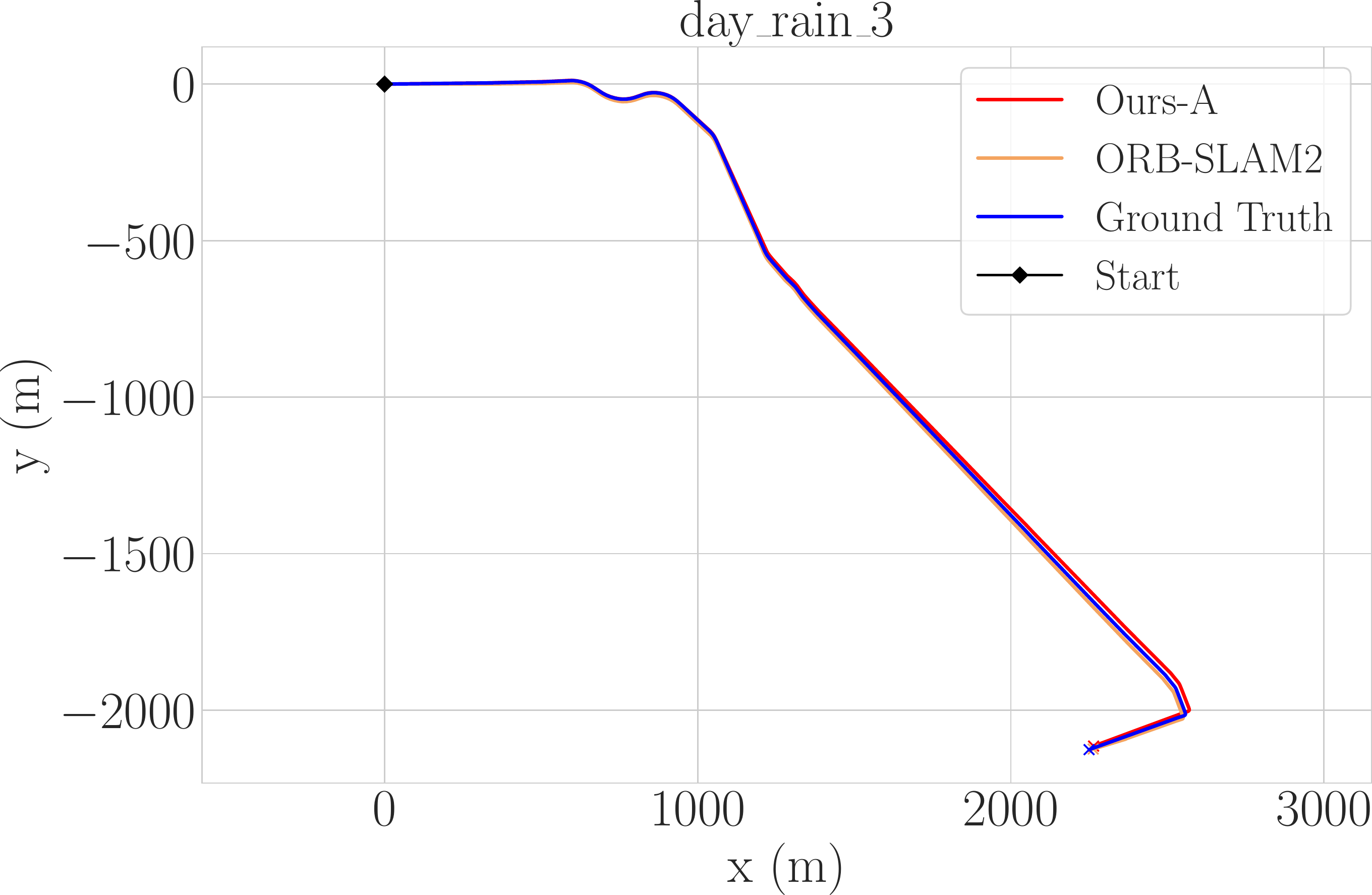}
    \includegraphics[width=0.24\linewidth]{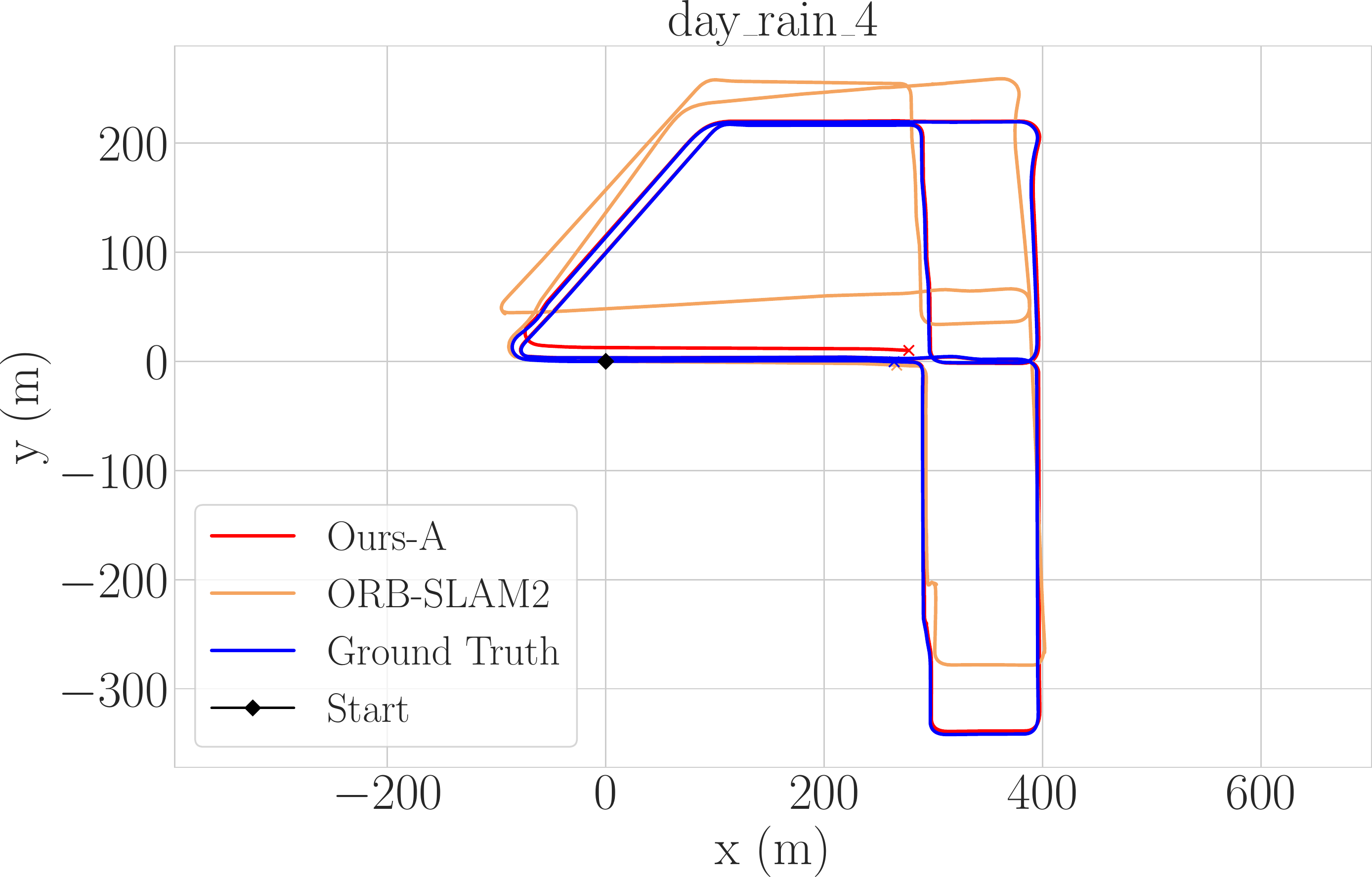}
    \includegraphics[width=0.24\linewidth]{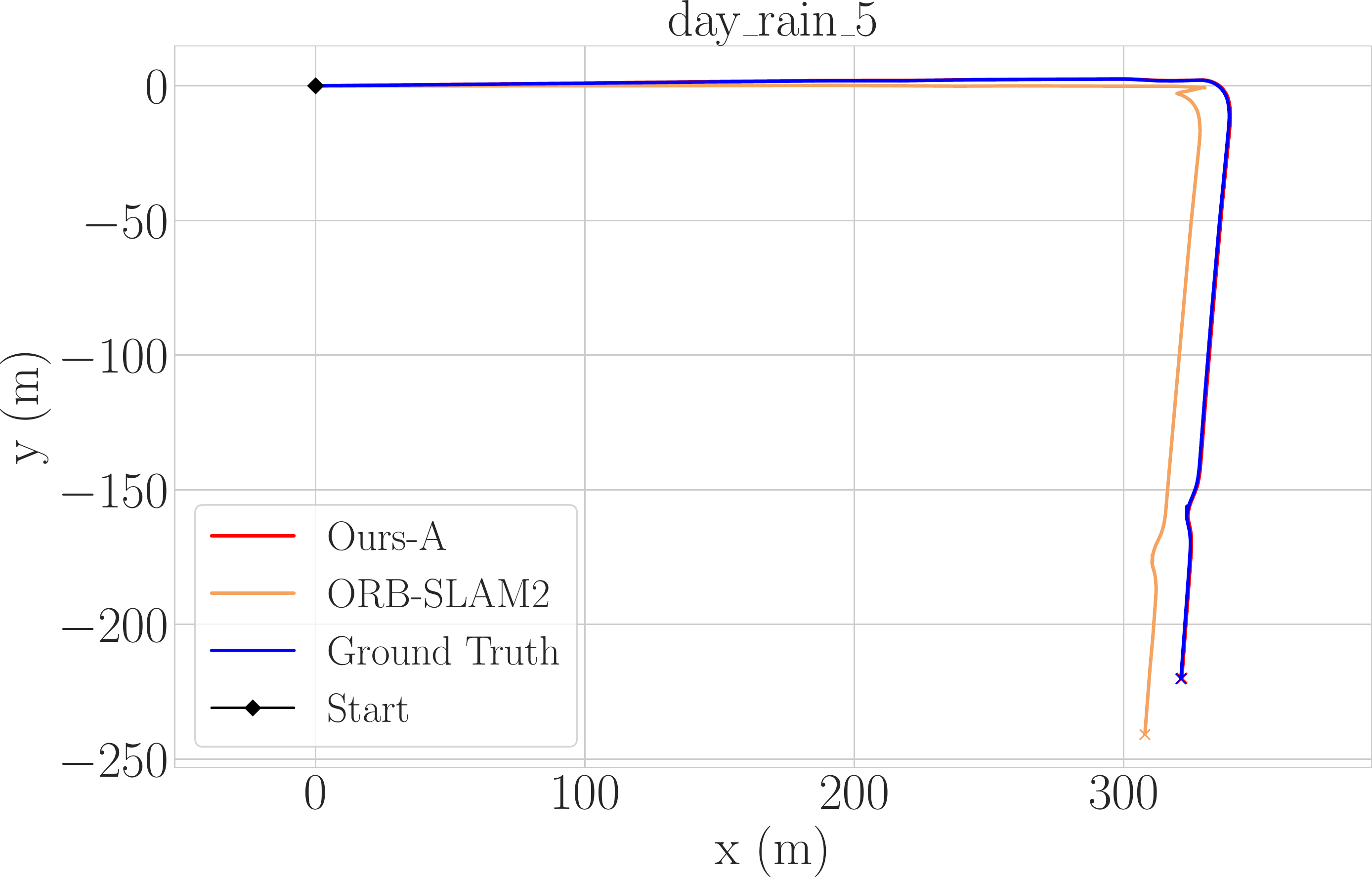}
    \includegraphics[width=0.24\linewidth]{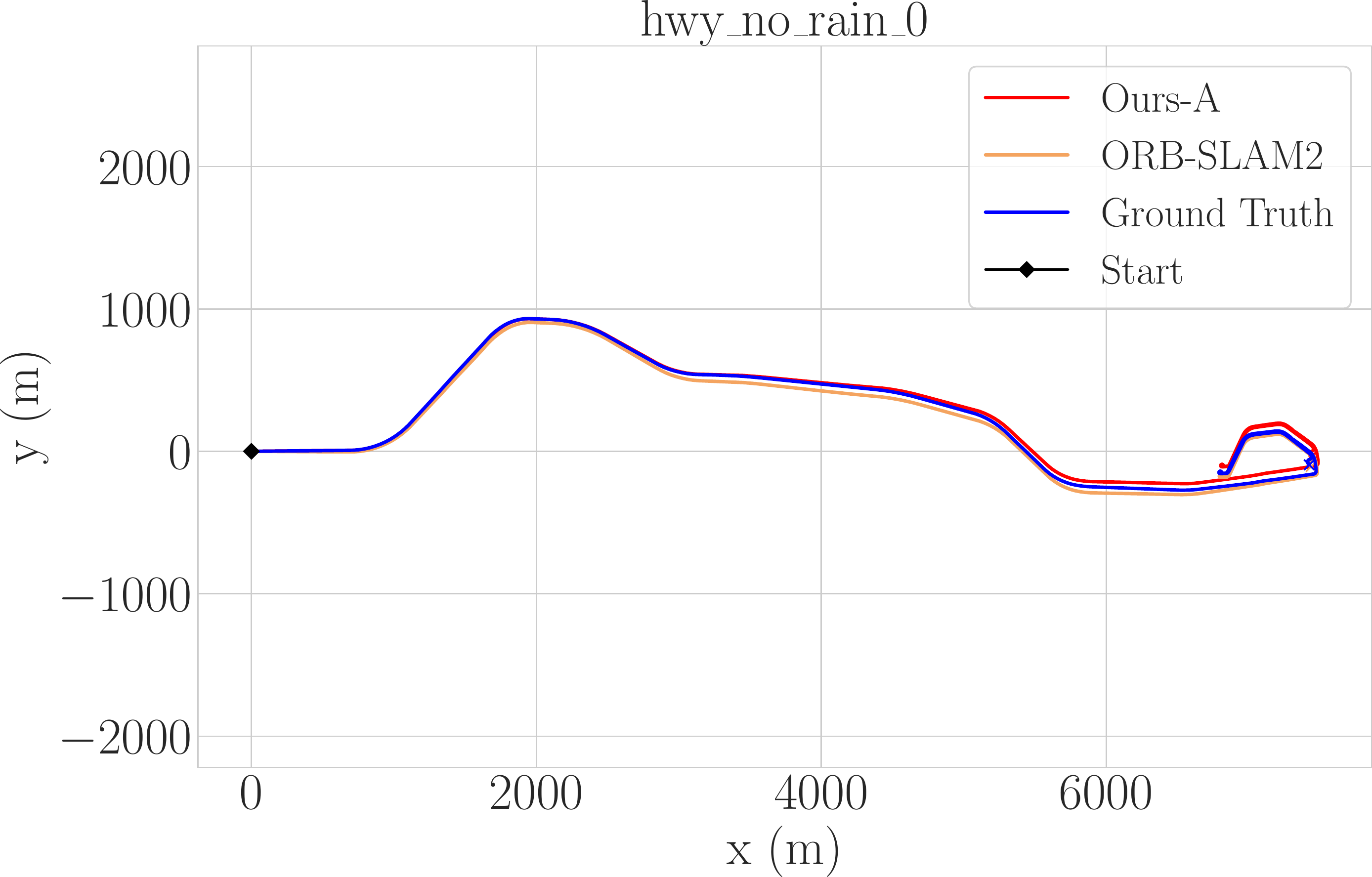}
    \includegraphics[width=0.24\linewidth]{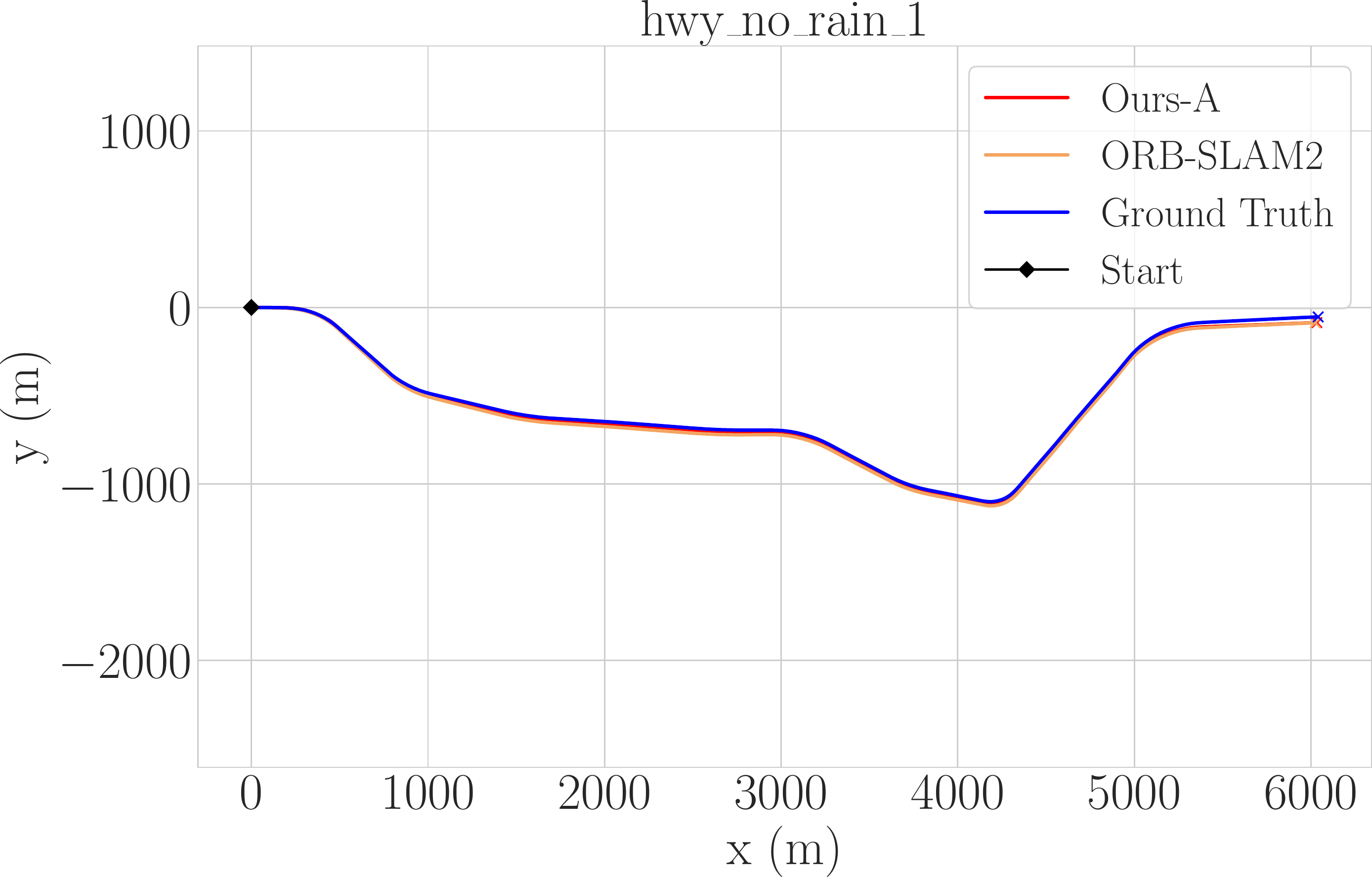}
    \includegraphics[width=0.24\linewidth]{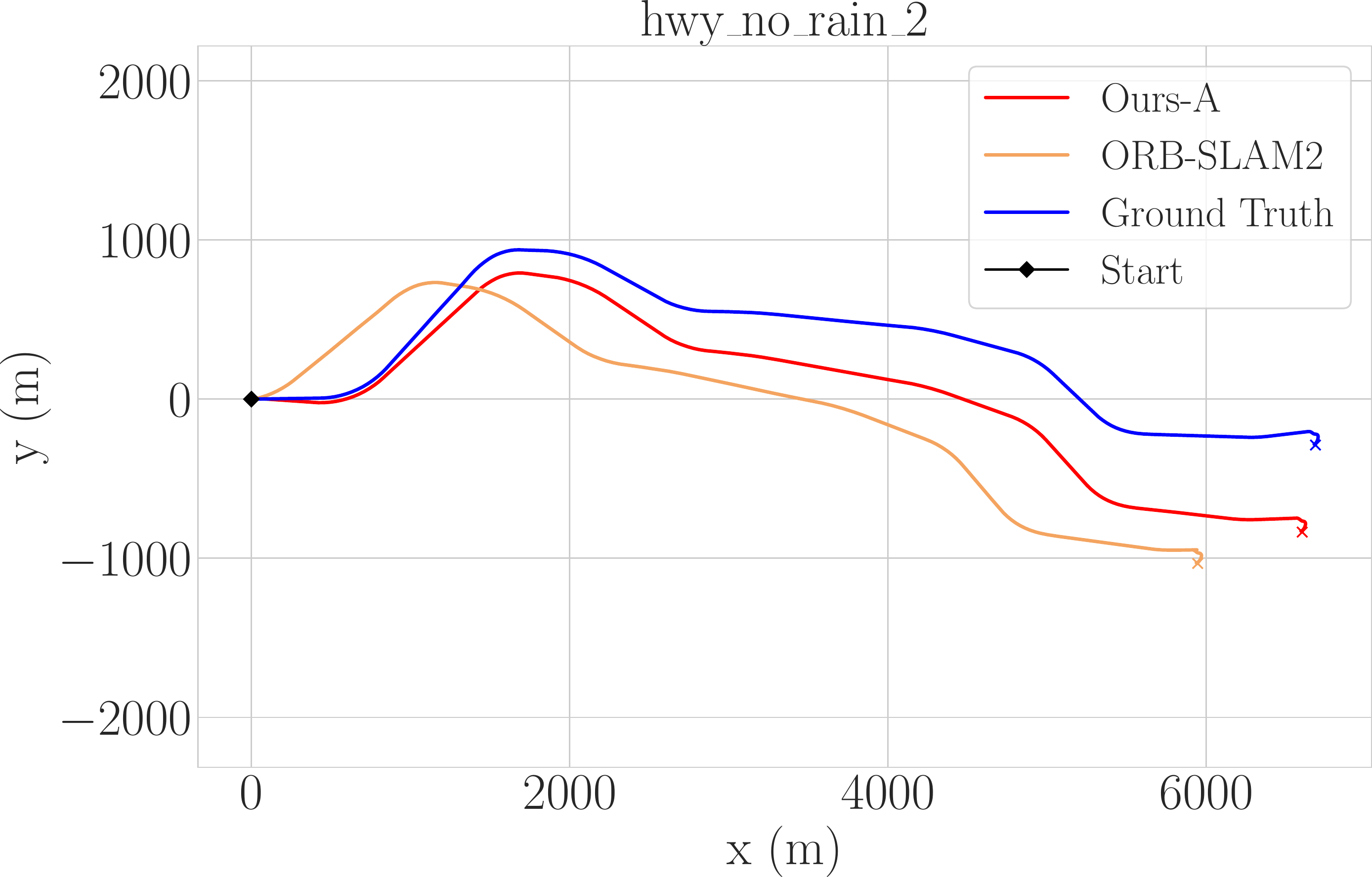}
    \includegraphics[width=0.24\linewidth]{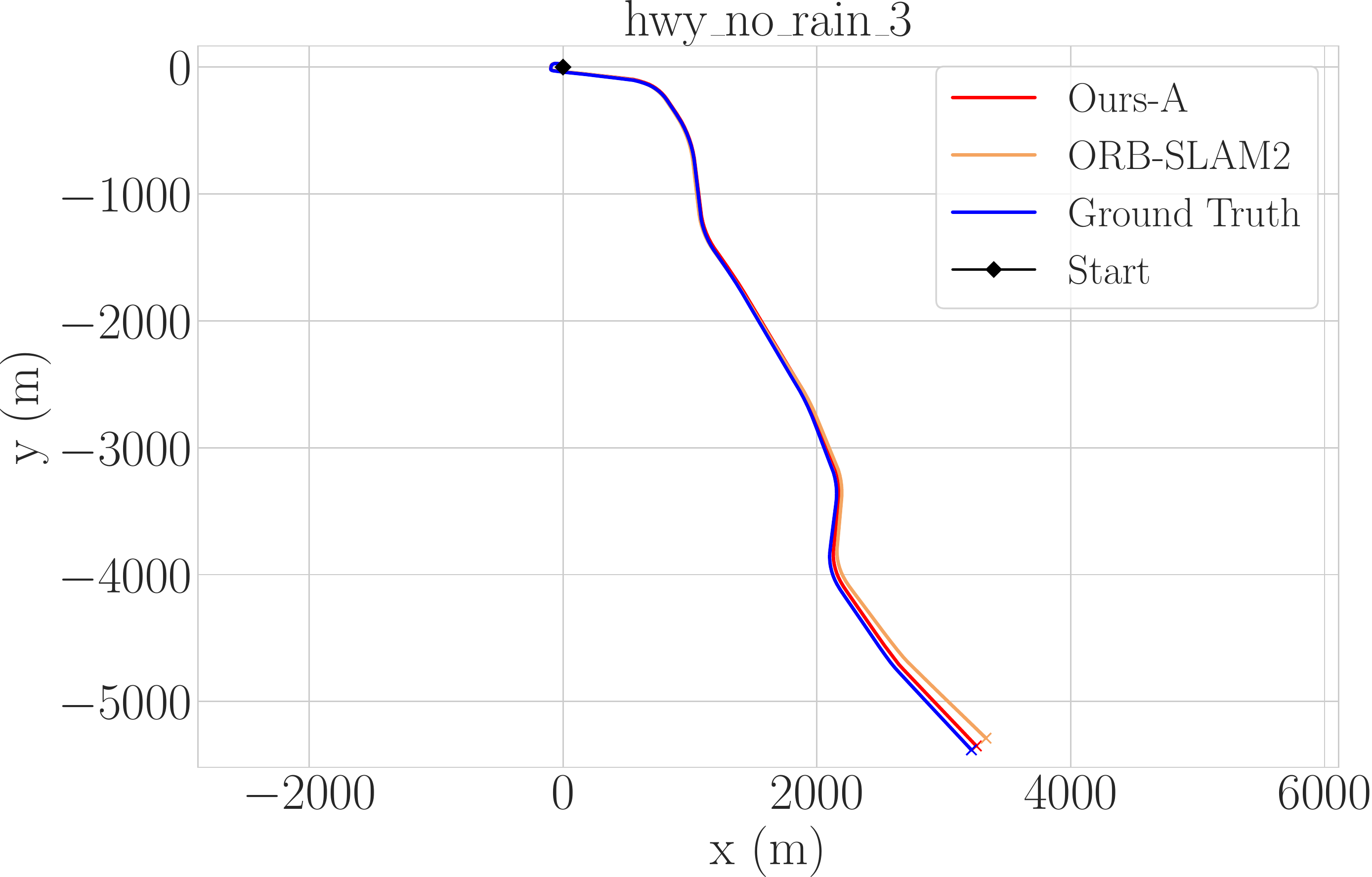}
    \includegraphics[width=0.24\linewidth]{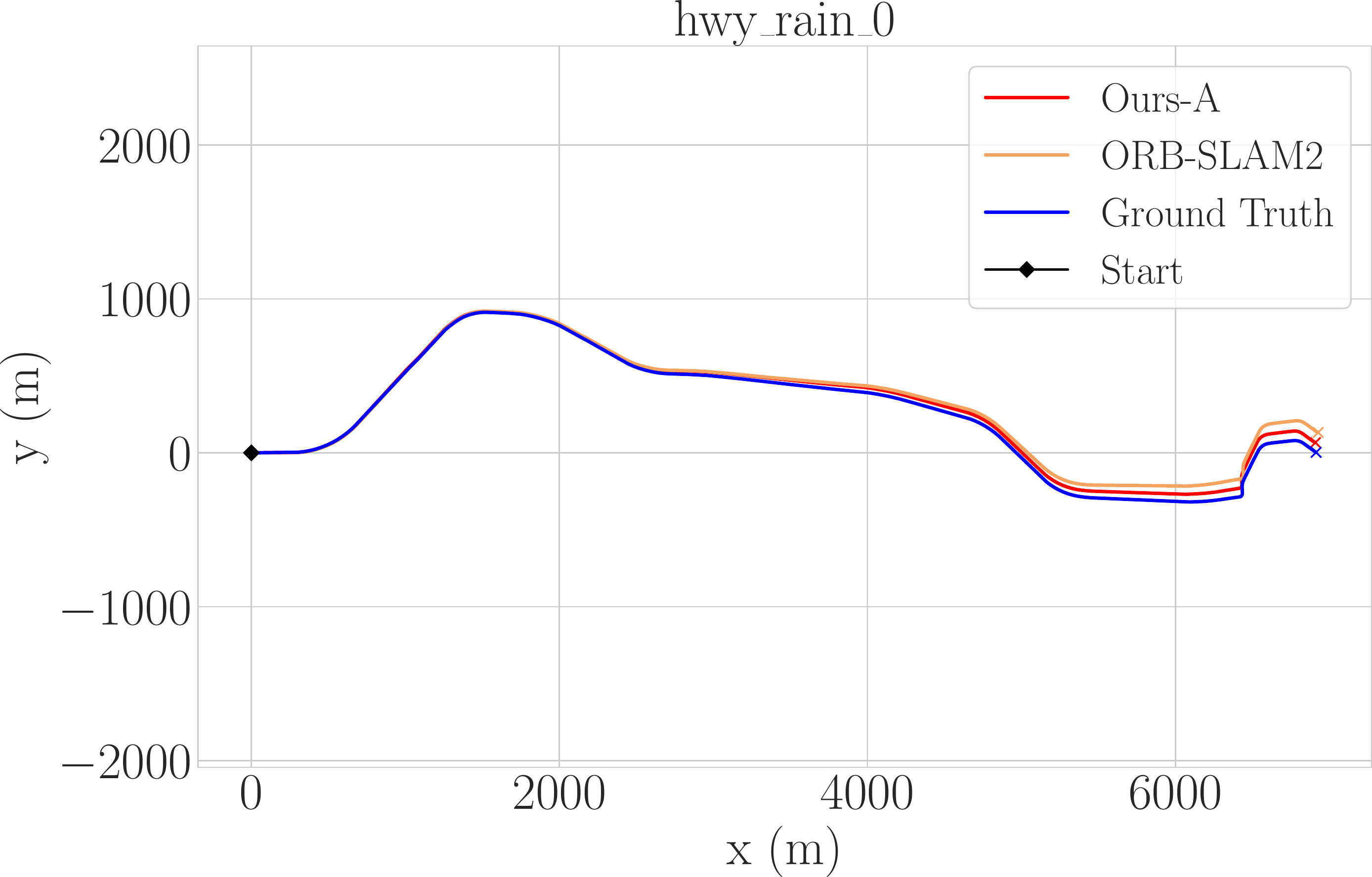}
    \includegraphics[width=0.24\linewidth]{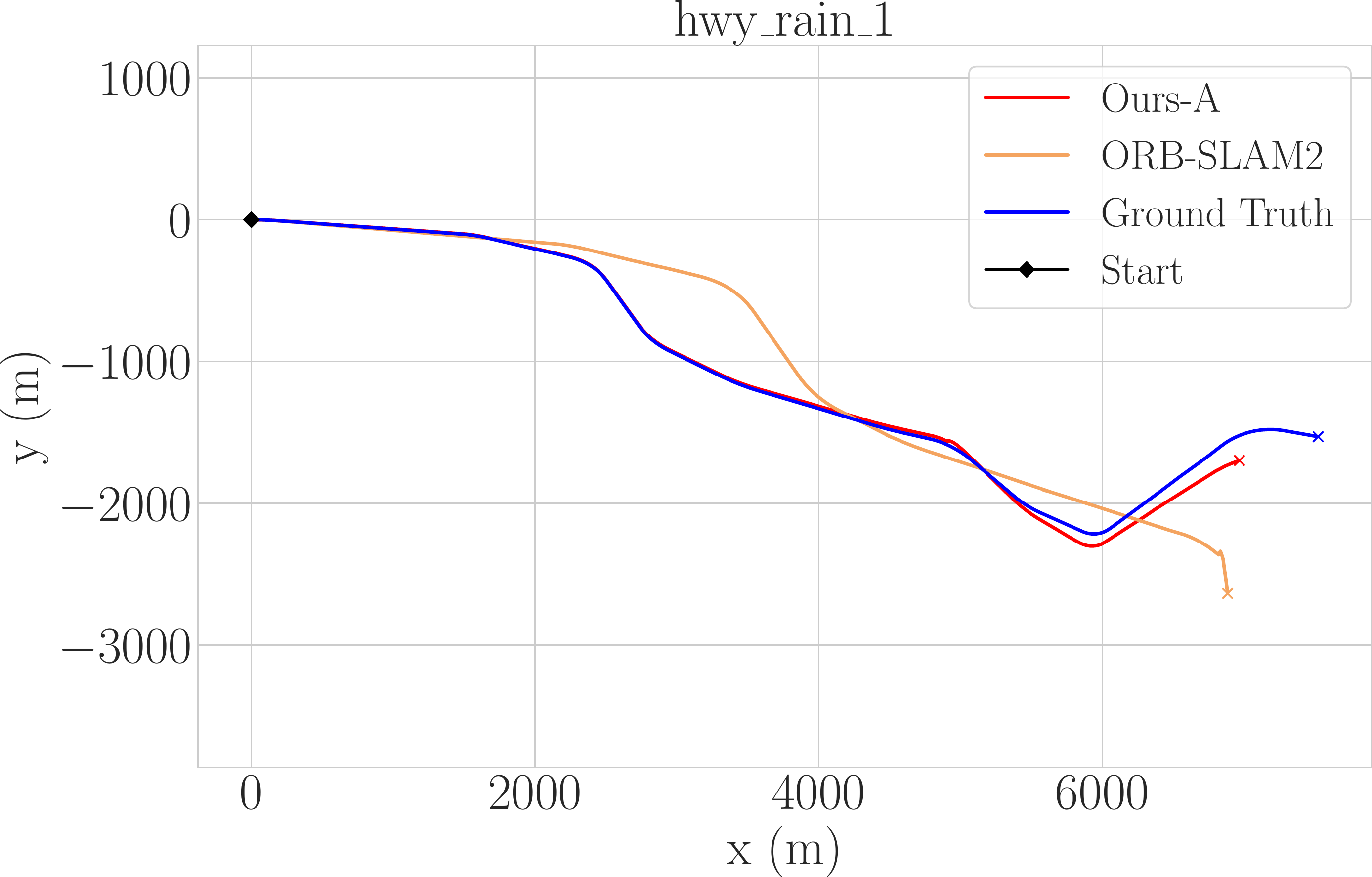}
    \includegraphics[width=0.24\linewidth]{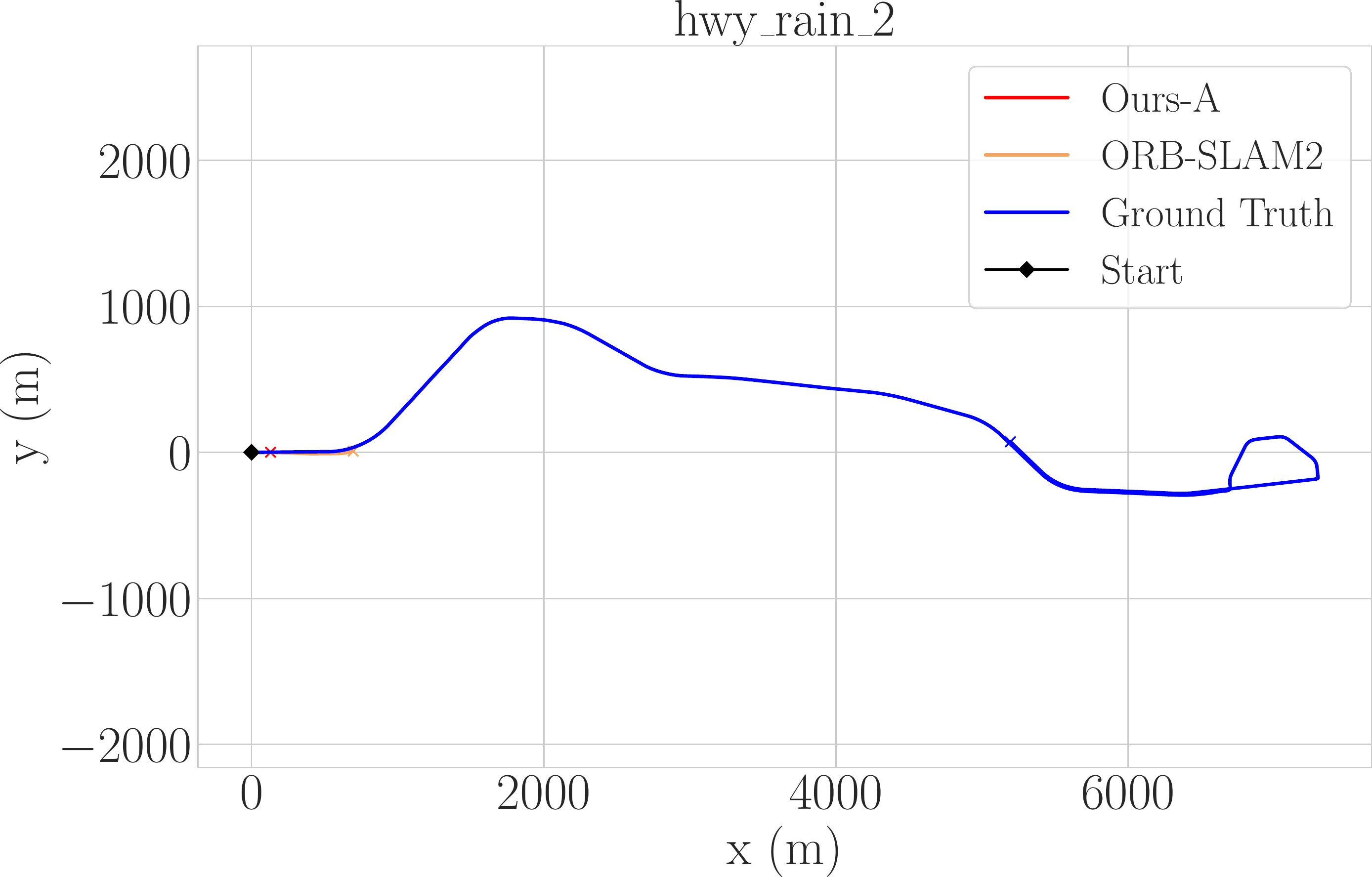}
    \caption{Estimated trajectories in all 25 validation sequences, comparing
    ORB-SLAM2 (stereo) with our full system with all 7 cameras.
    \label{fig:full_qualitative_traj}}
\end{figure*}

\begin{figure*}
    \centering
    \includegraphics[width=0.16\linewidth]{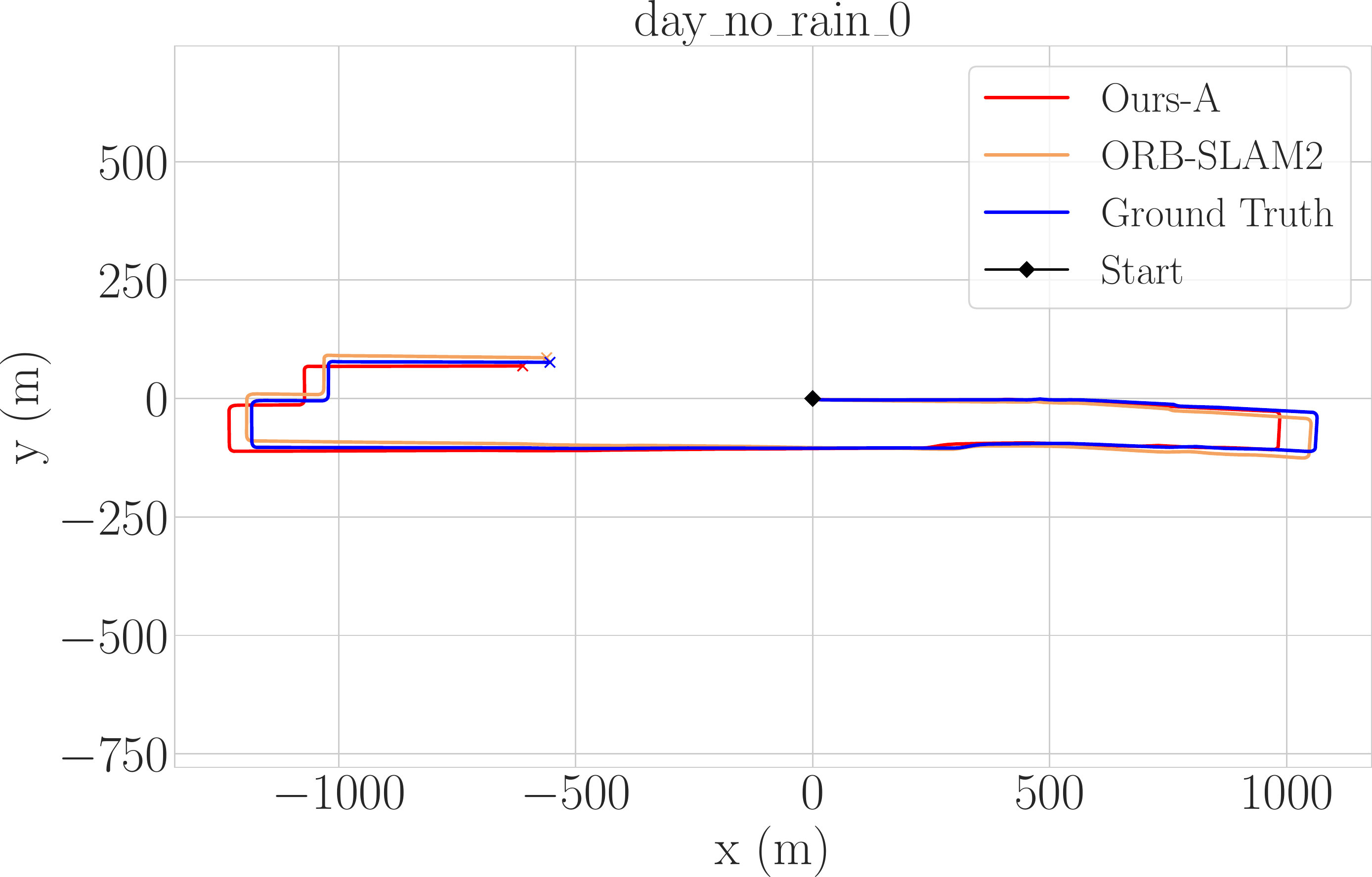}
    \includegraphics[width=0.16\linewidth]{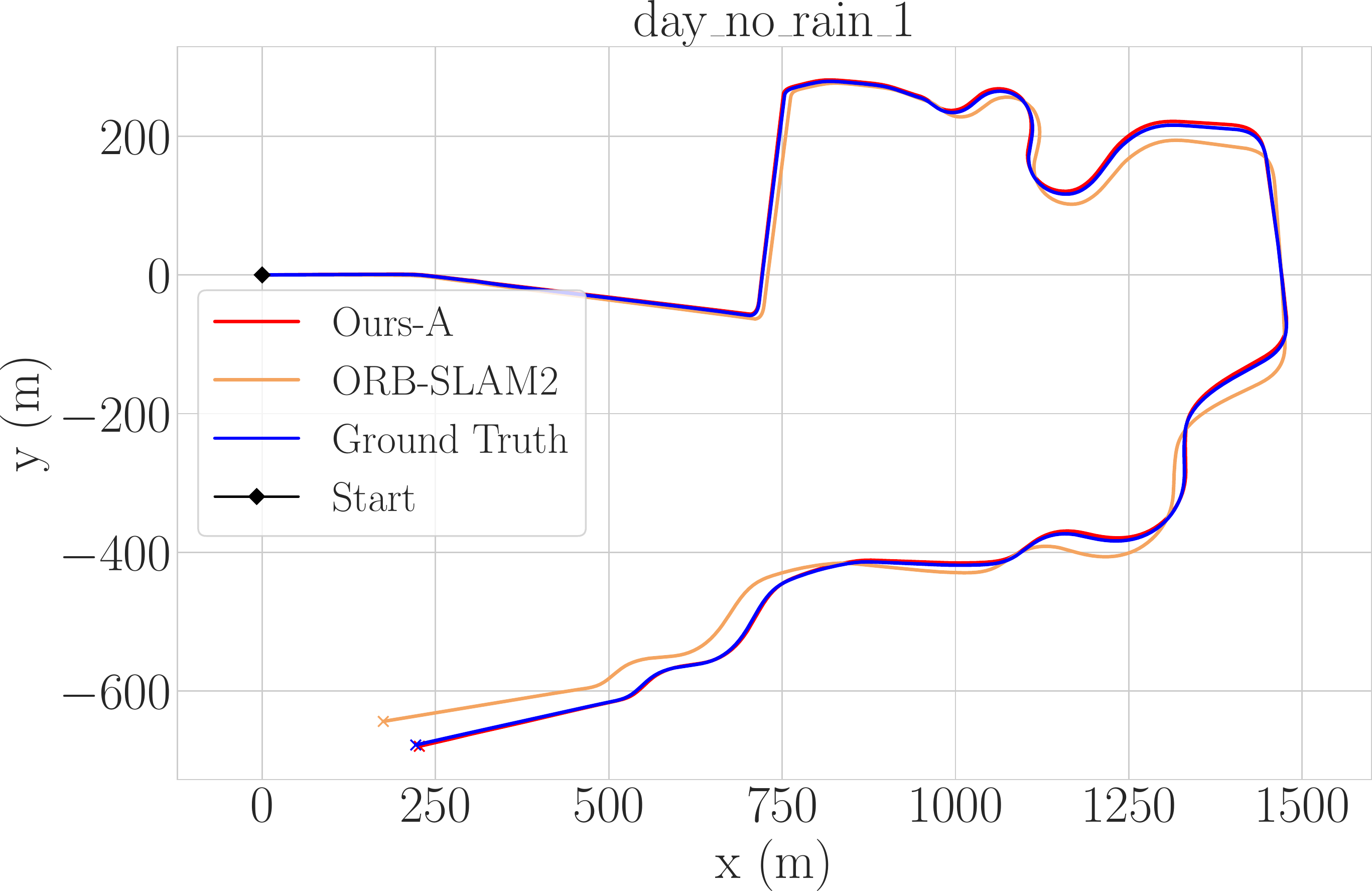}
    \includegraphics[width=0.16\linewidth]{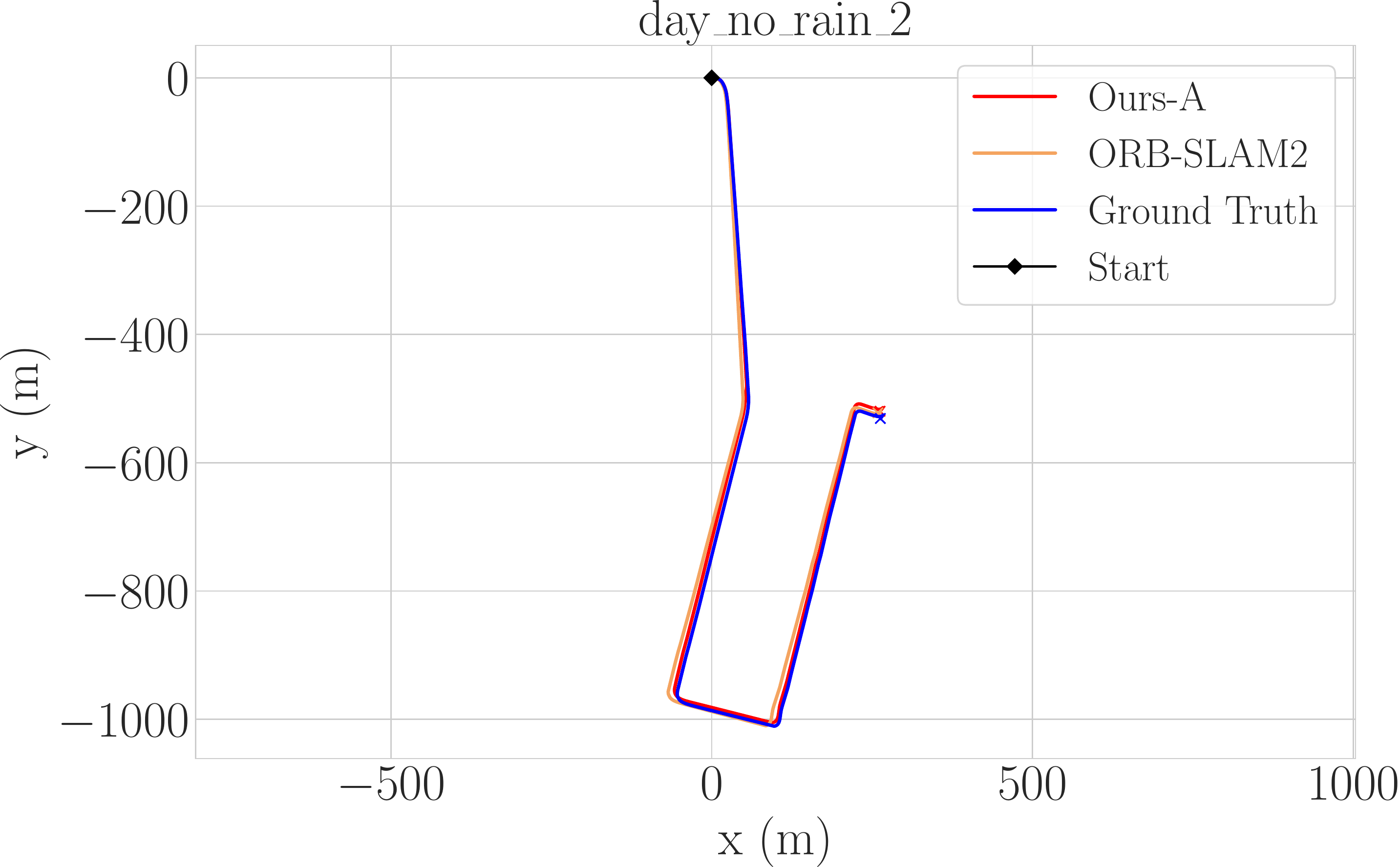}
    \includegraphics[width=0.16\linewidth]{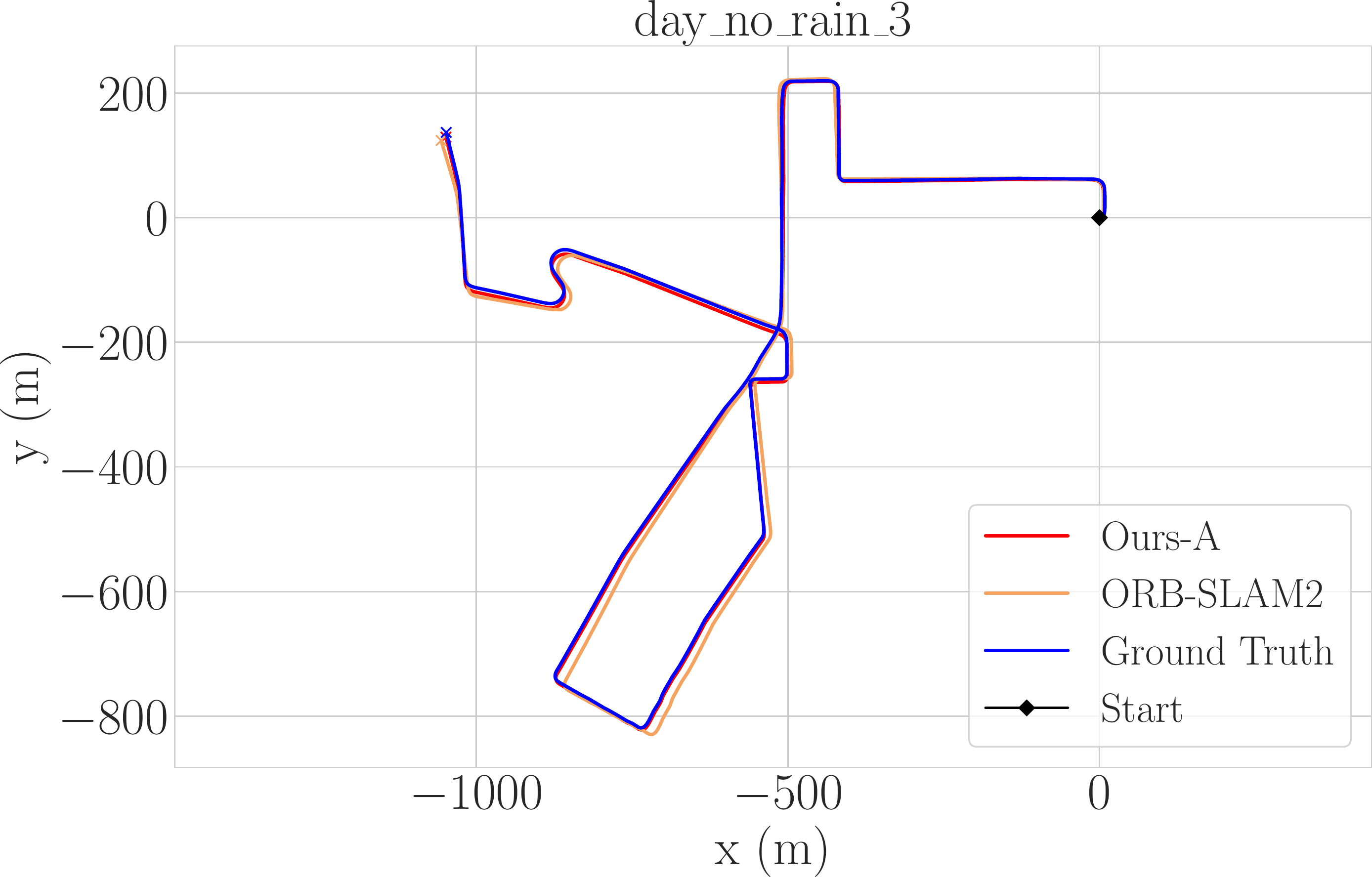}
    \includegraphics[width=0.16\linewidth]{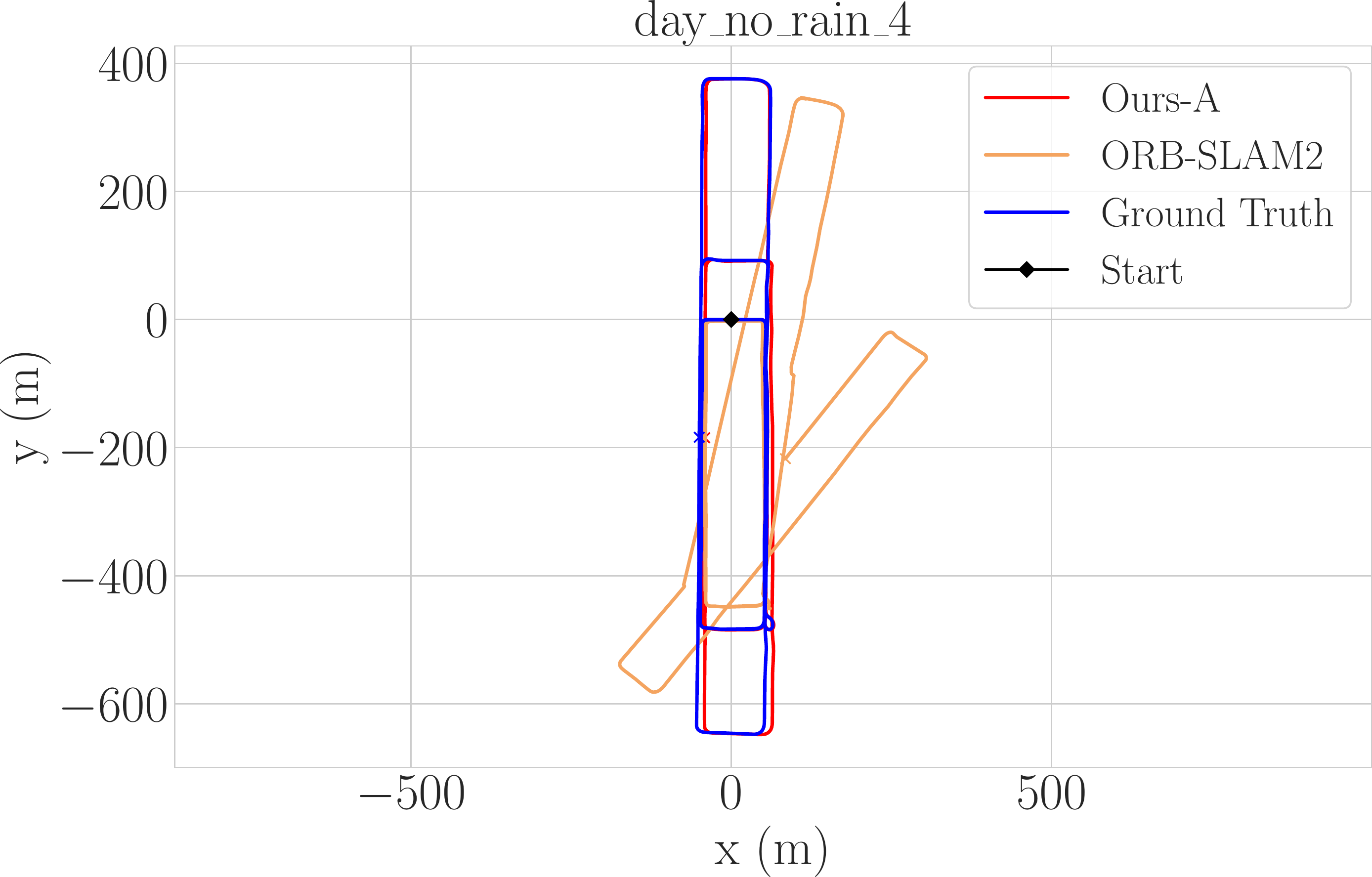}
    \includegraphics[width=0.16\linewidth]{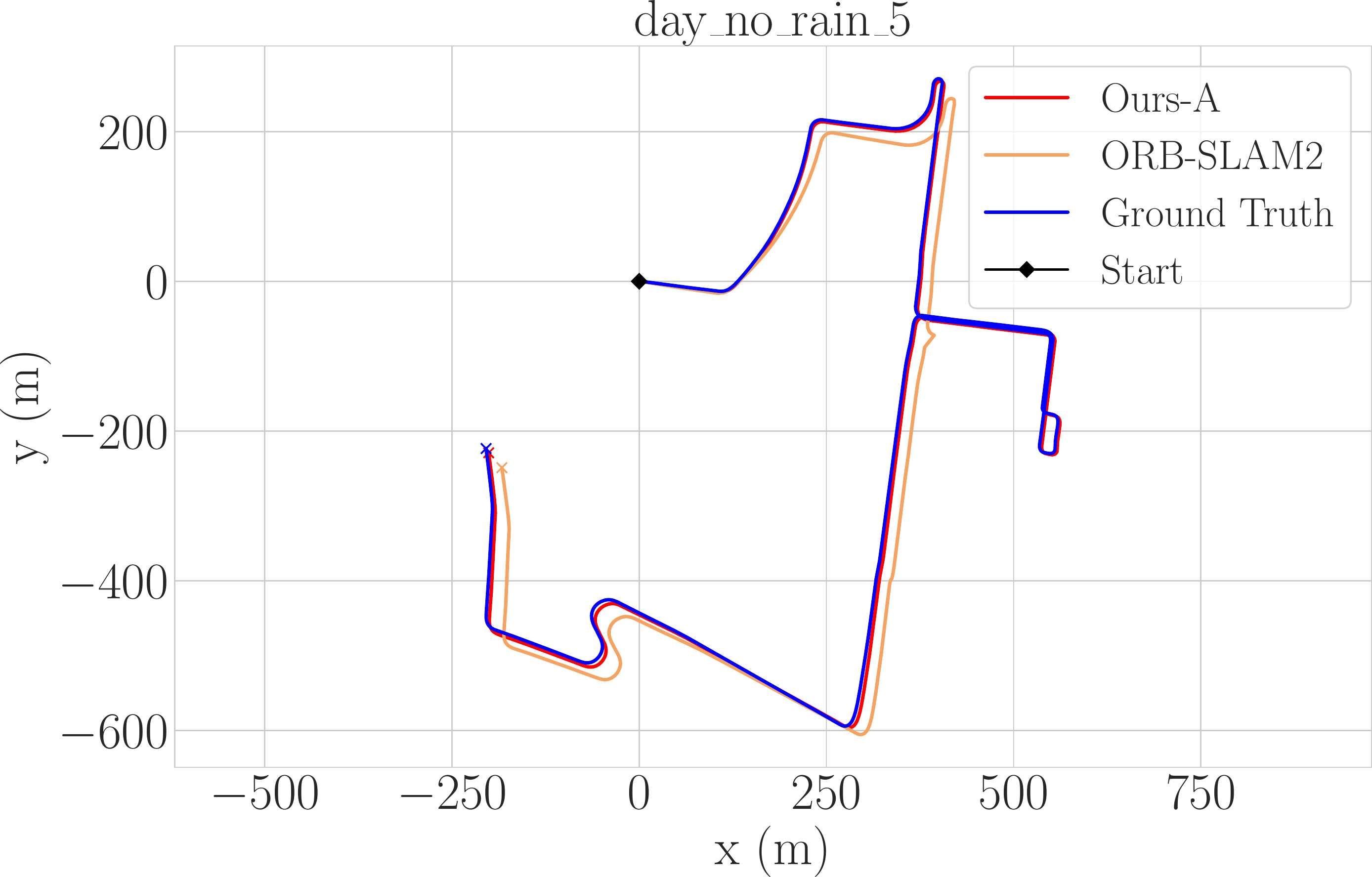}
    \includegraphics[width=0.16\linewidth]{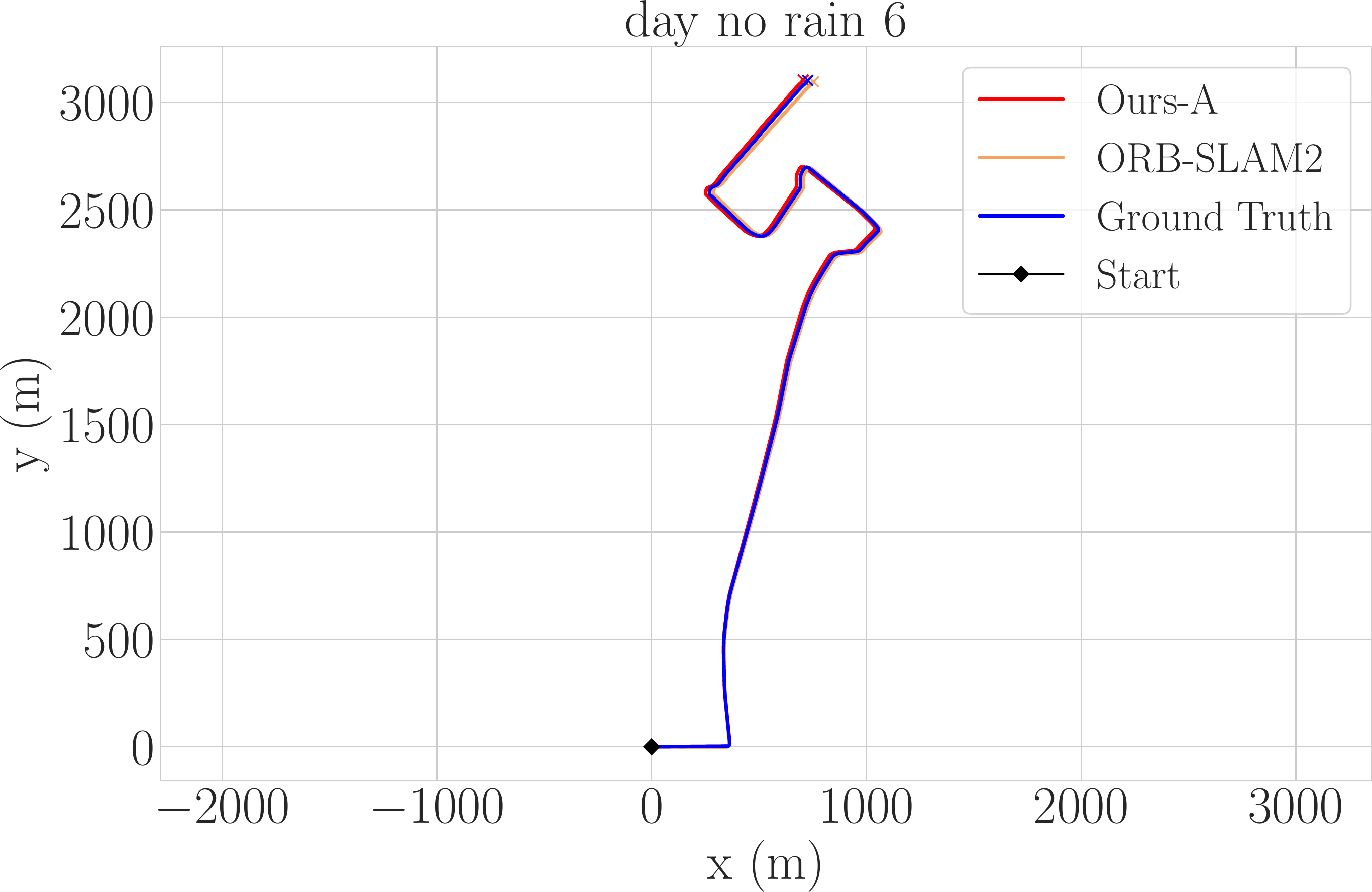}
    \includegraphics[width=0.16\linewidth]{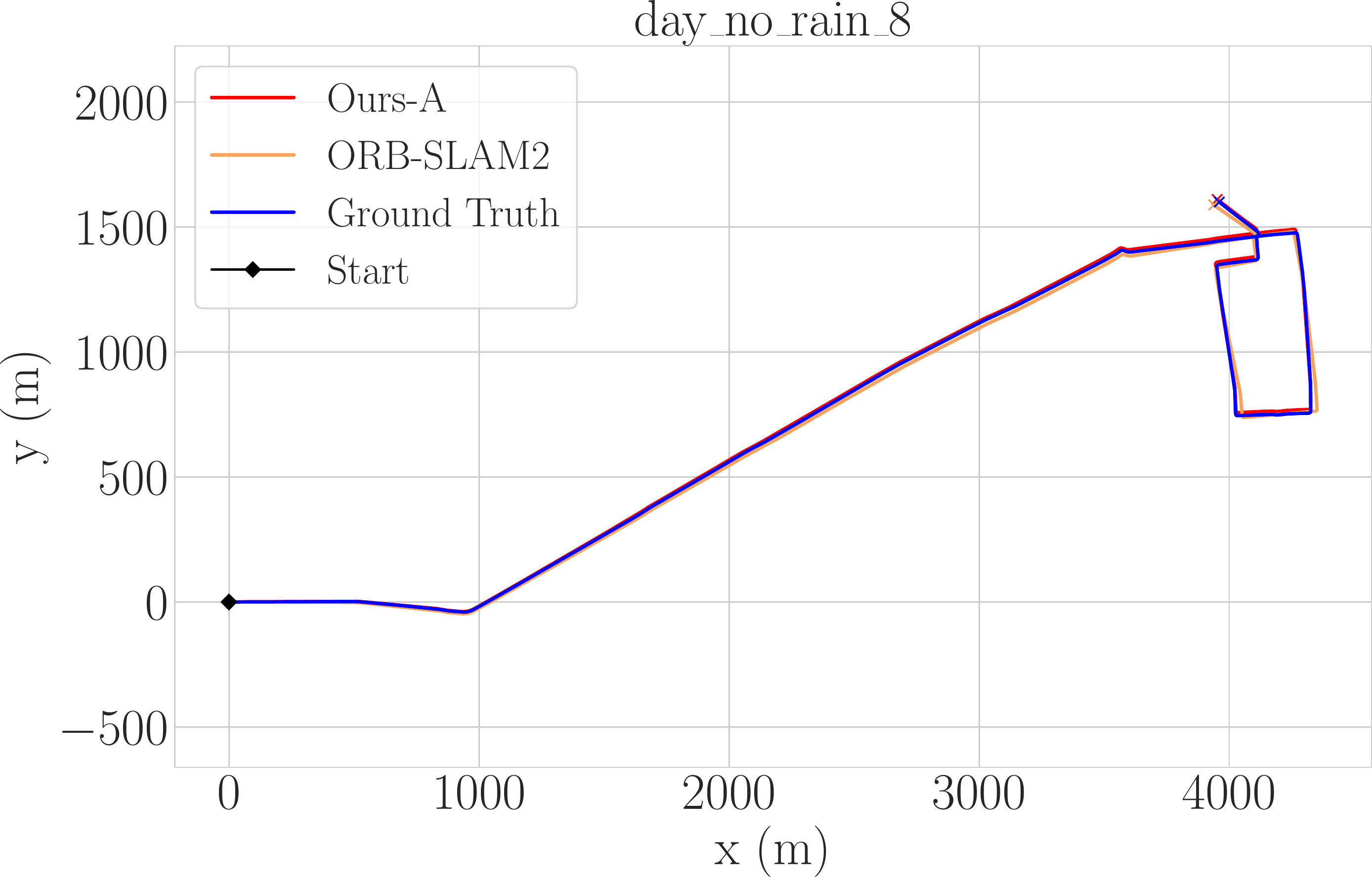}
    \includegraphics[width=0.16\linewidth]{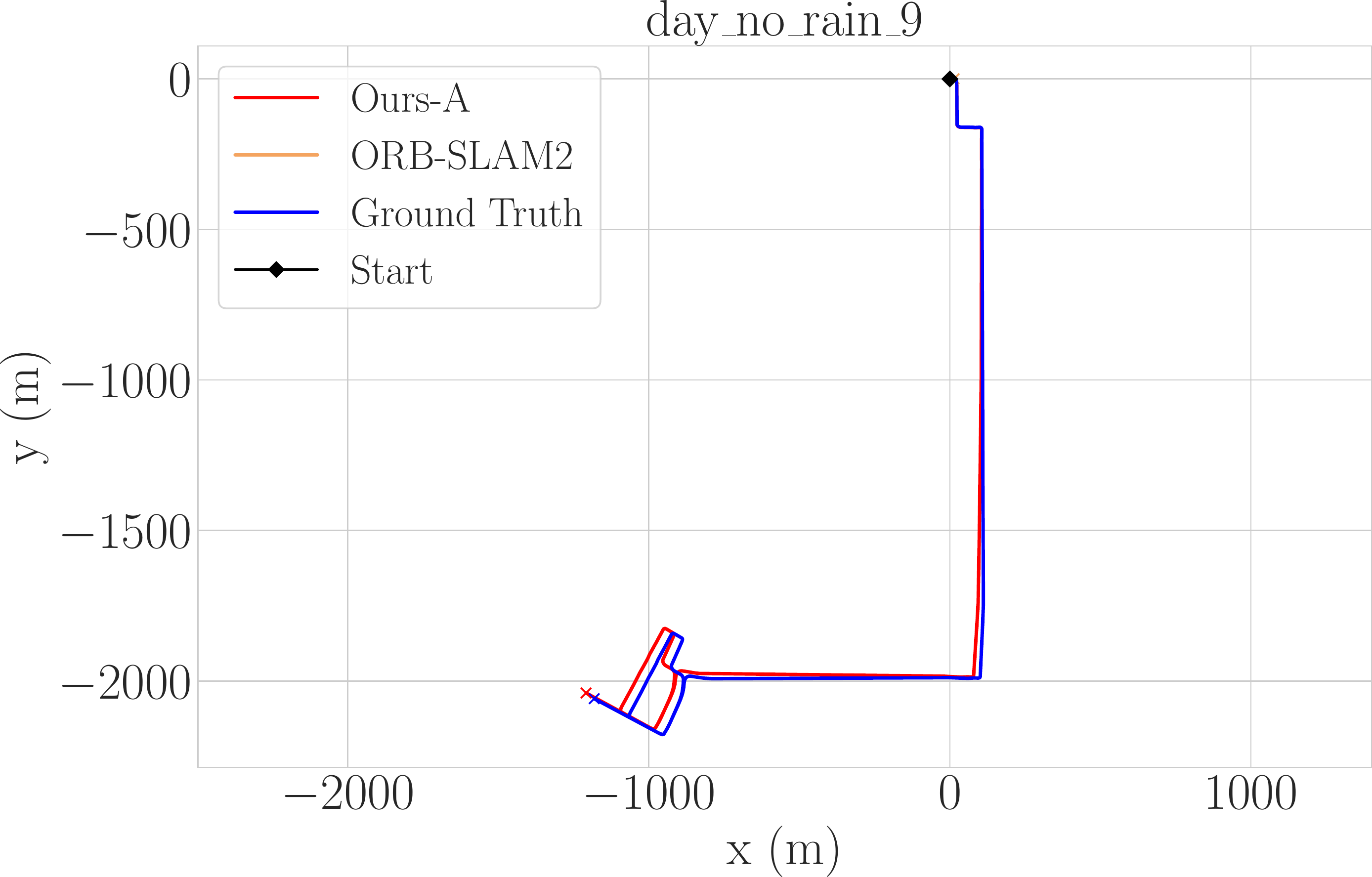}
    \includegraphics[width=0.16\linewidth]{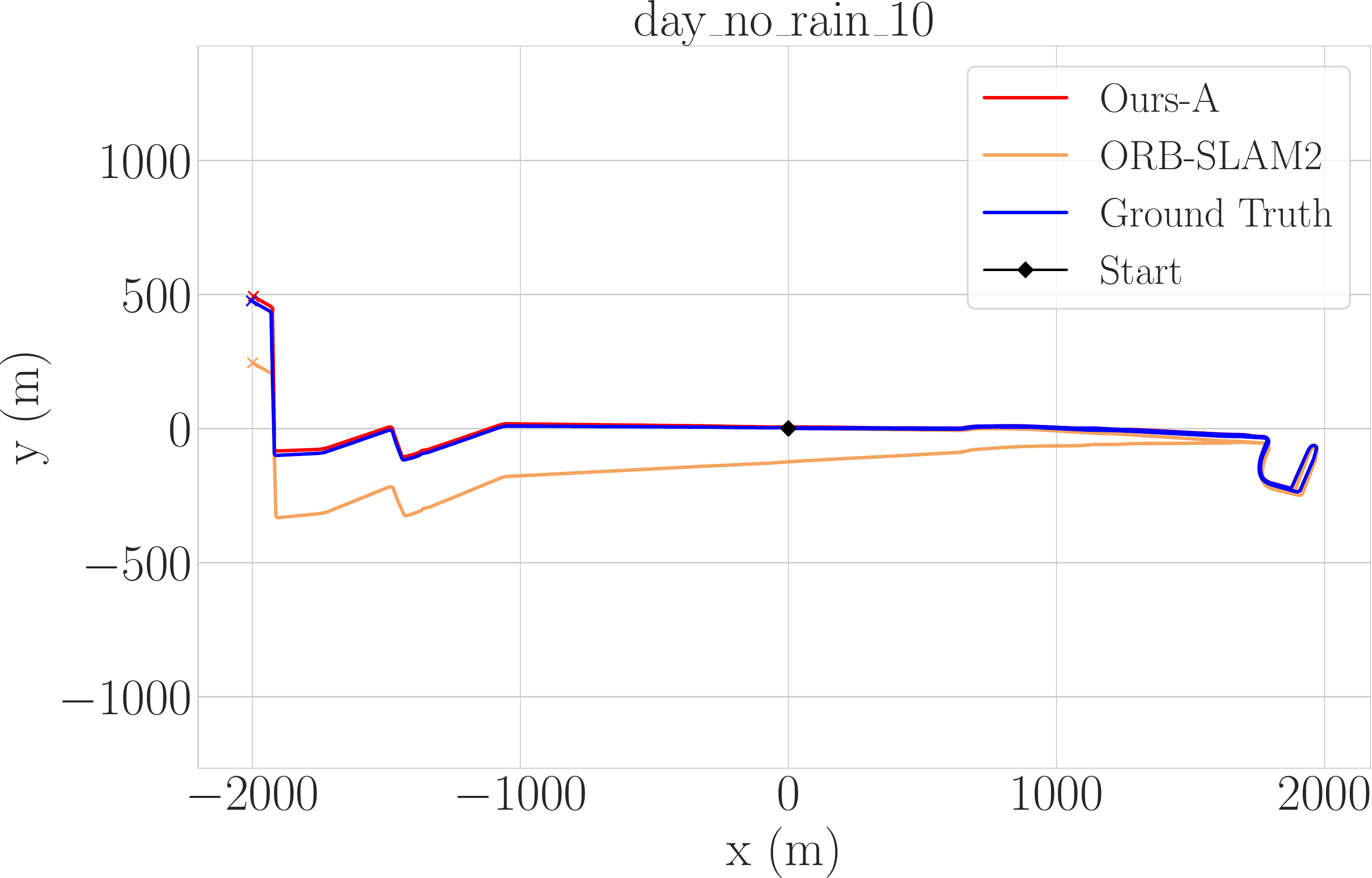}
    \includegraphics[width=0.16\linewidth]{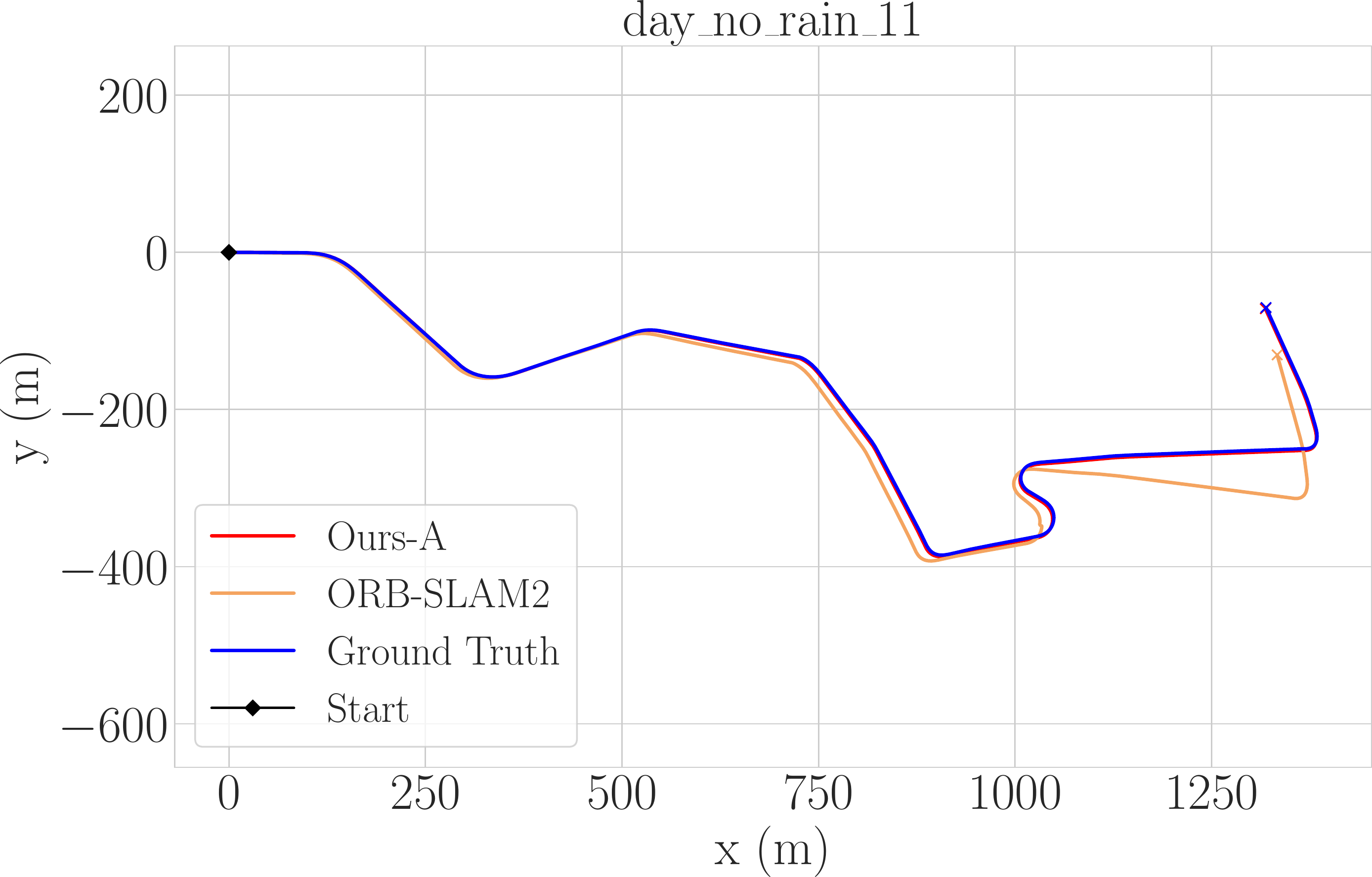}
    \includegraphics[width=0.16\linewidth]{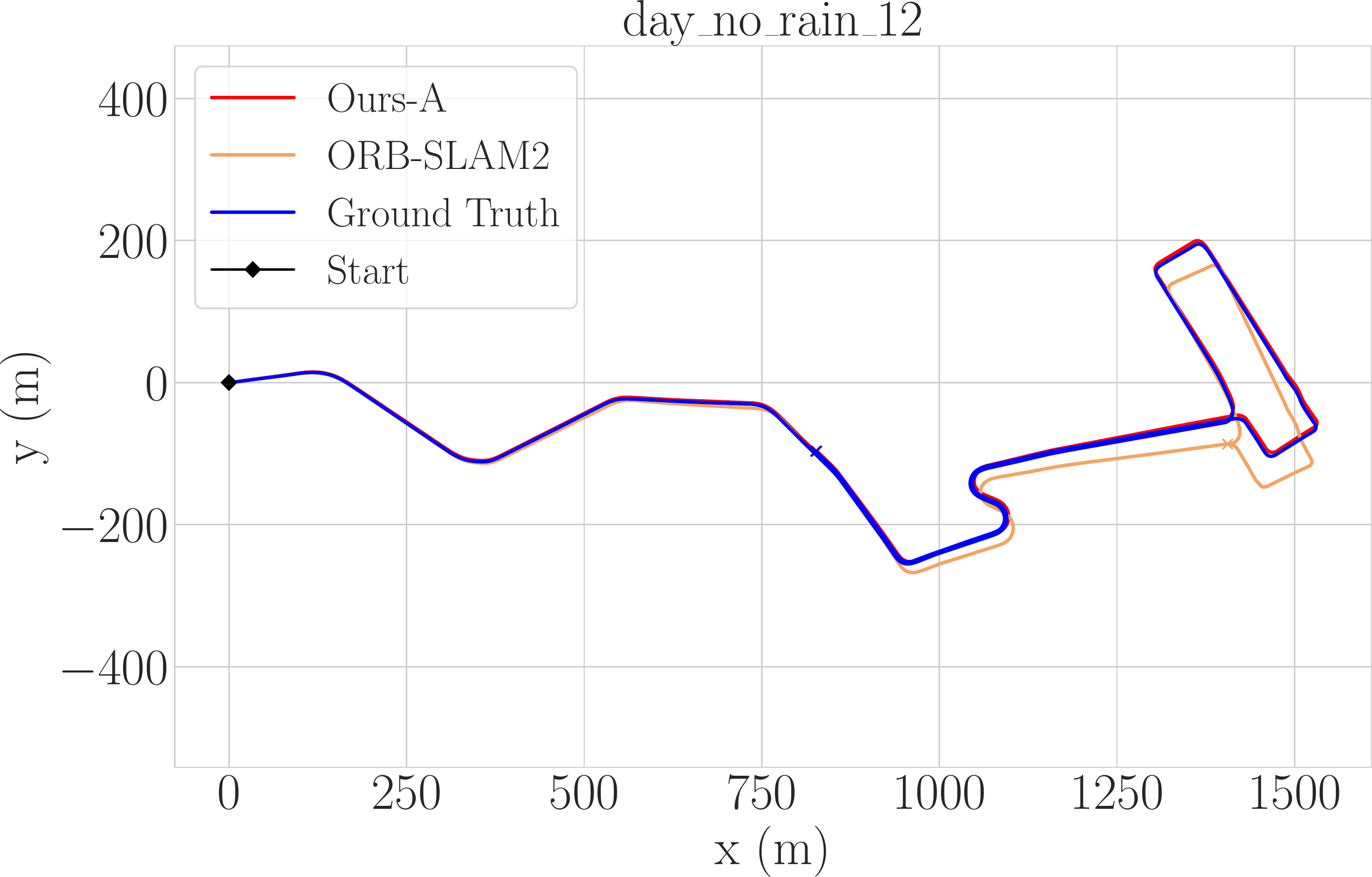}
    \includegraphics[width=0.16\linewidth]{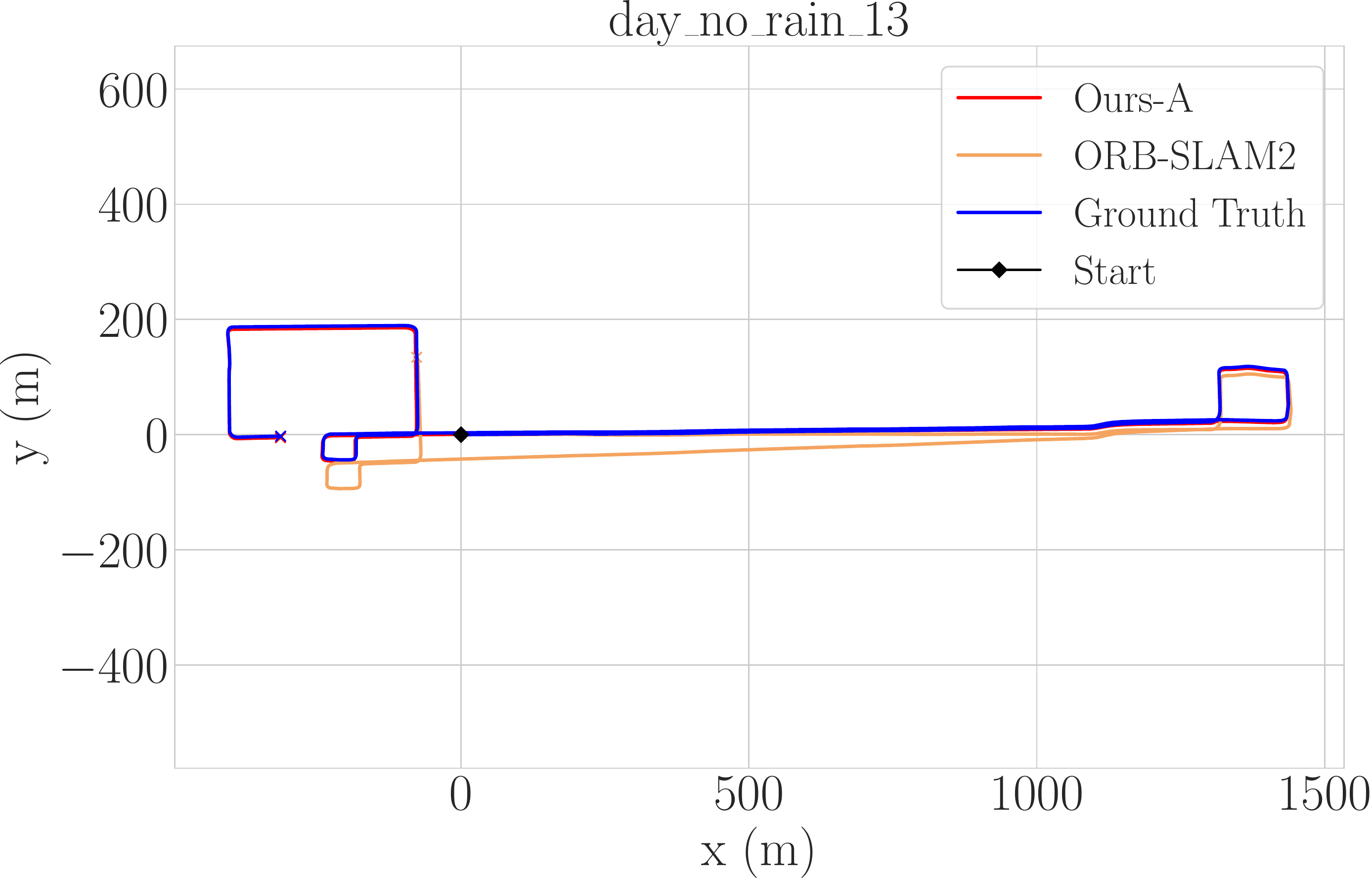}
    \includegraphics[width=0.16\linewidth]{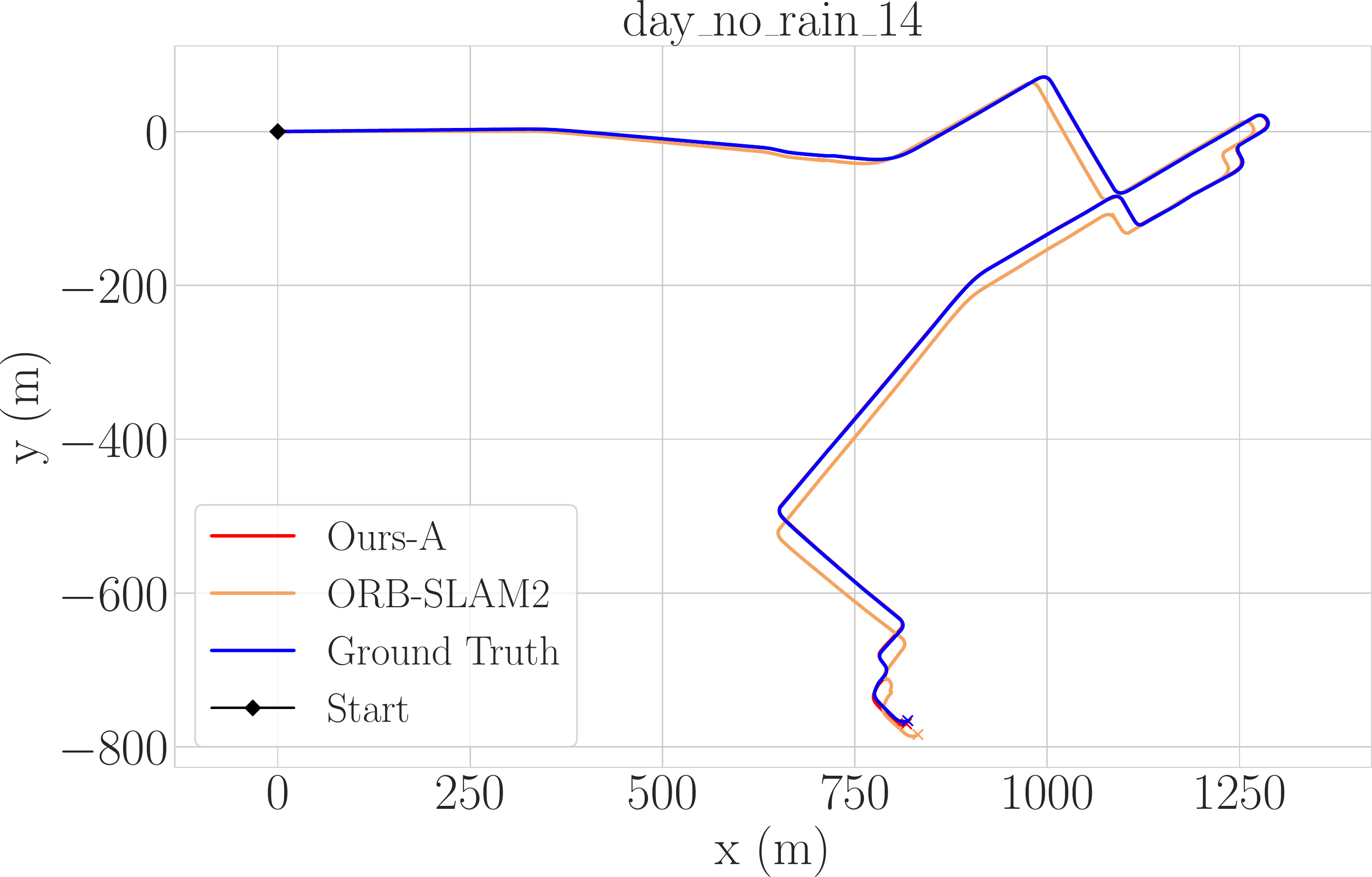}
    \includegraphics[width=0.16\linewidth]{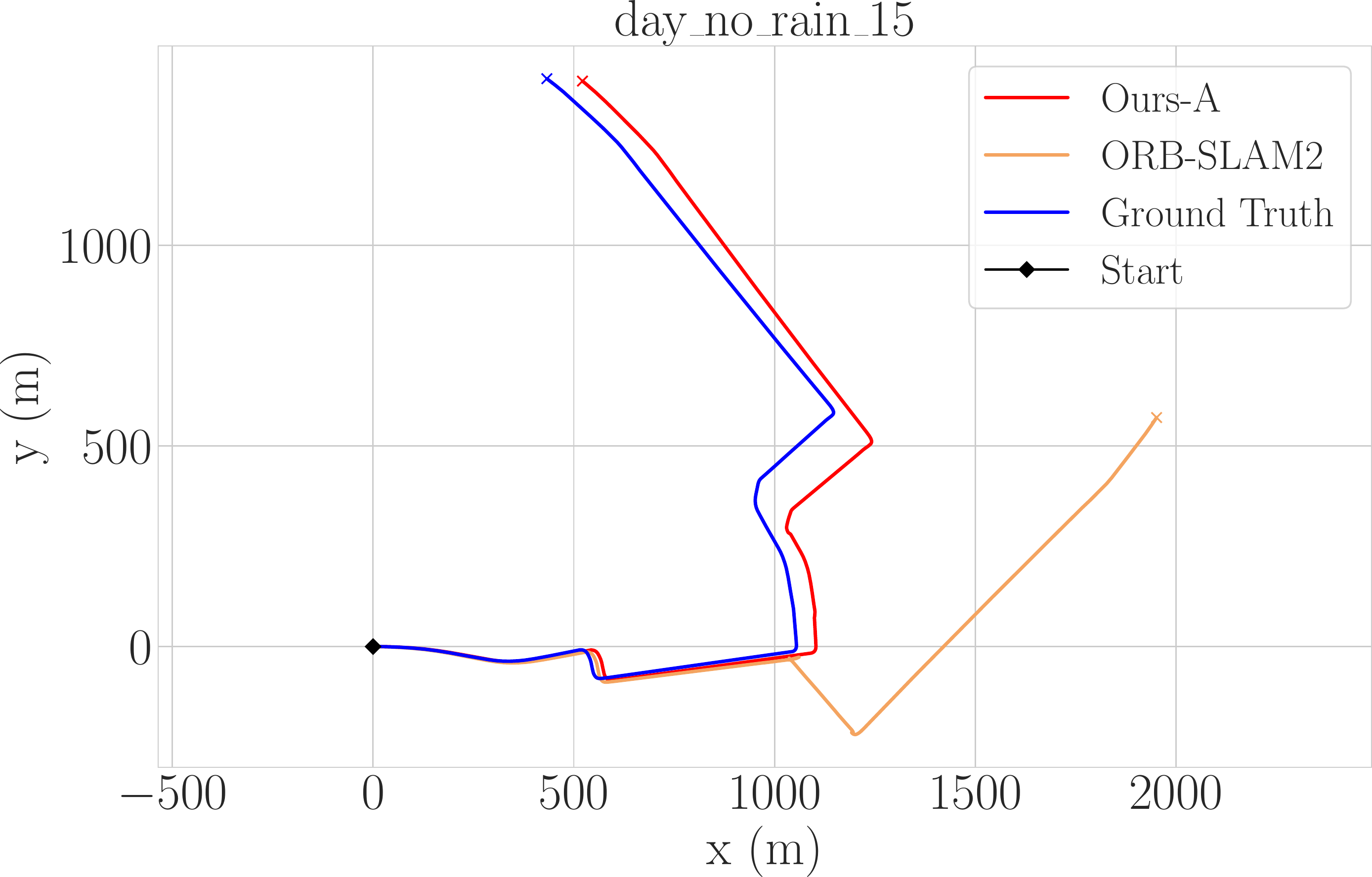}
    \includegraphics[width=0.16\linewidth]{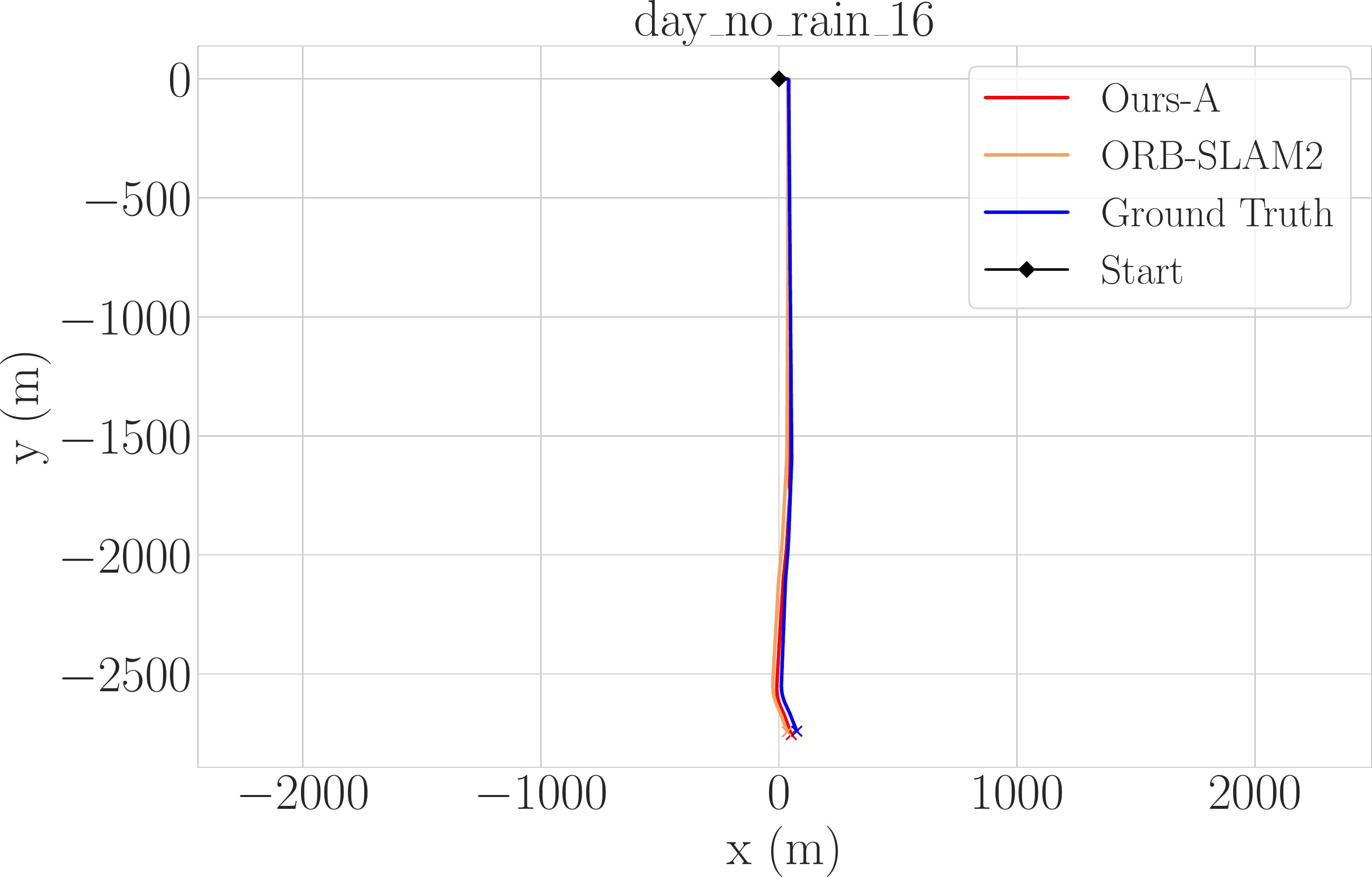}
    \includegraphics[width=0.16\linewidth]{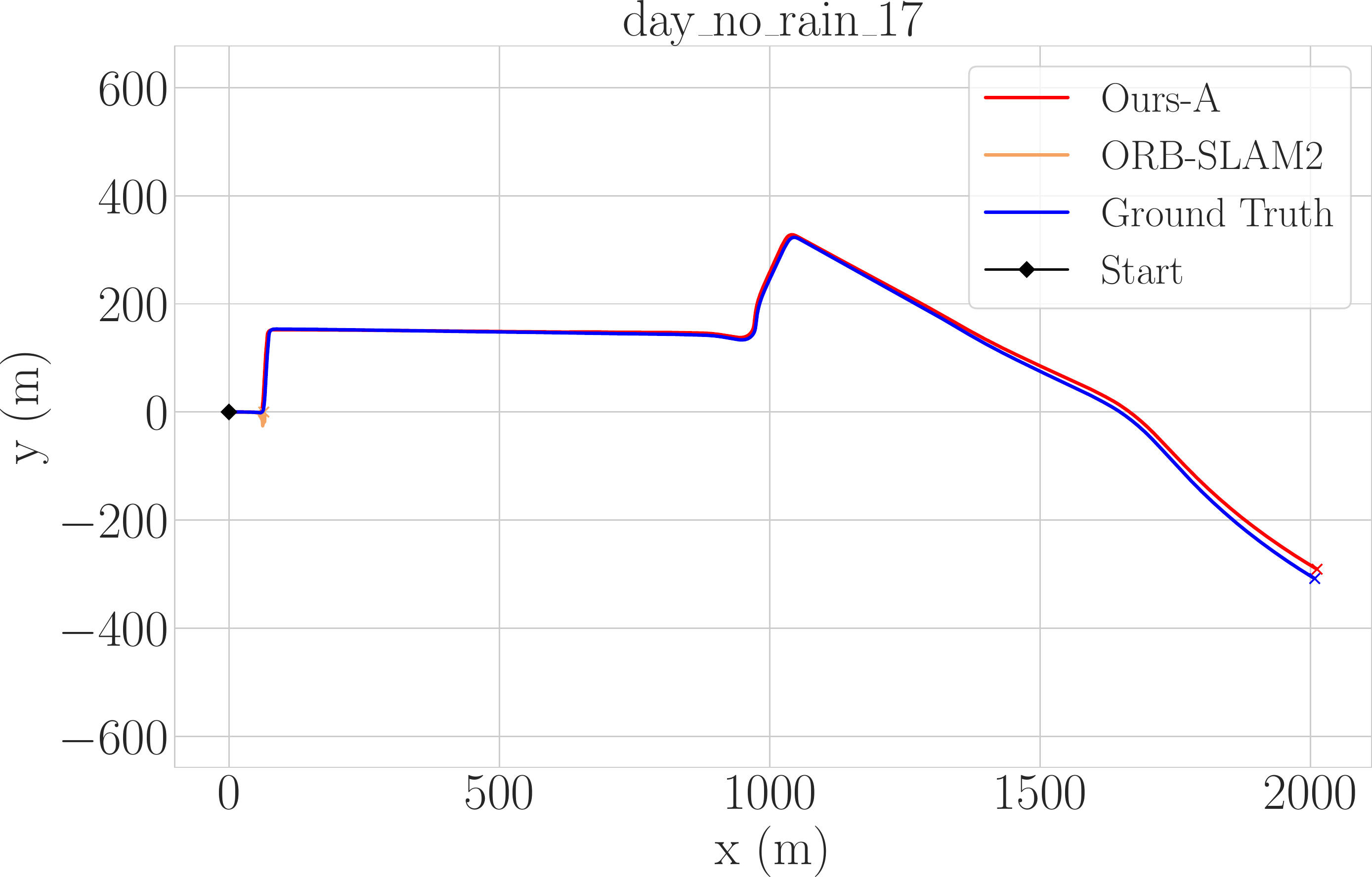}
    \includegraphics[width=0.16\linewidth]{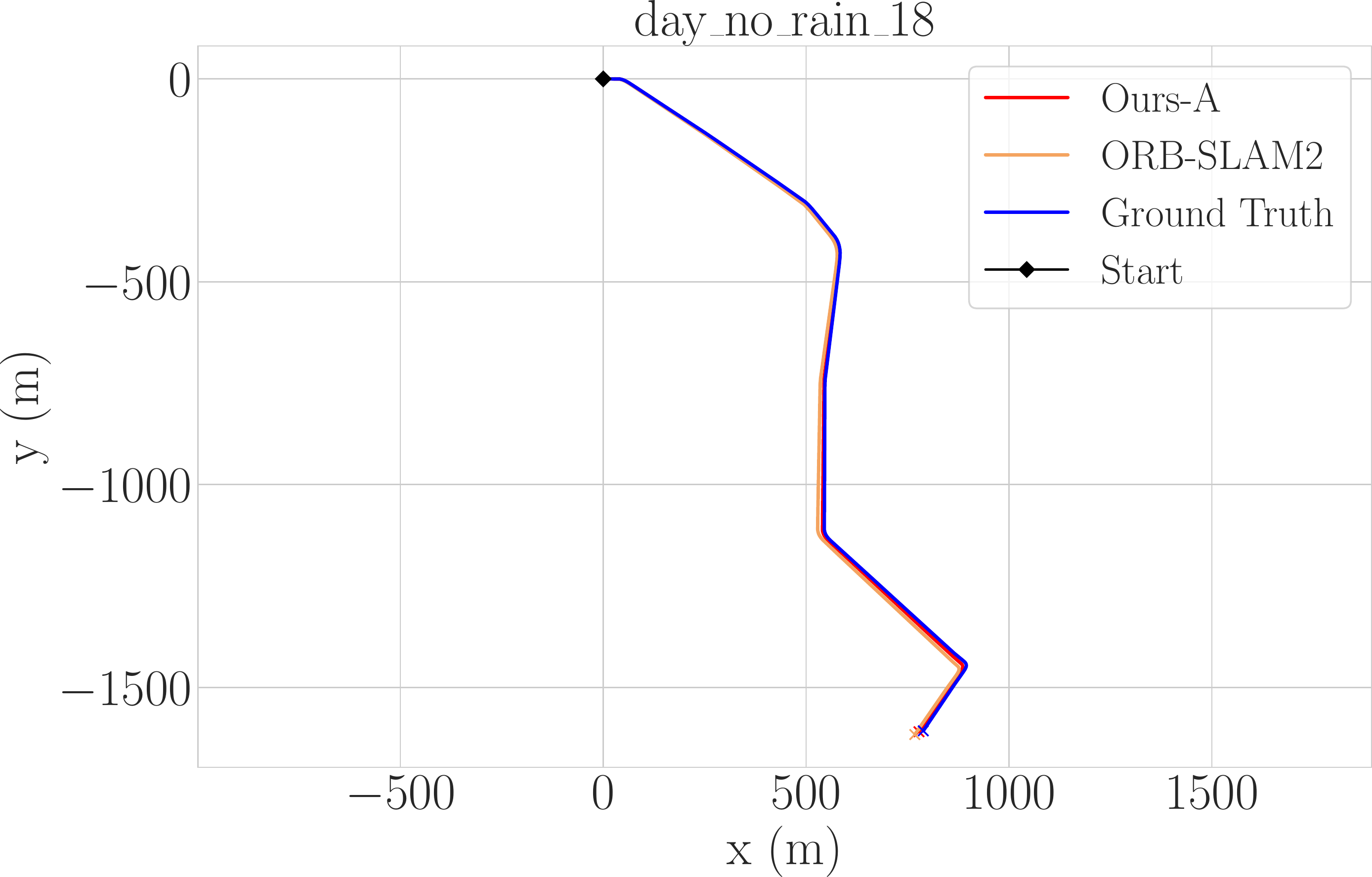}
    \includegraphics[width=0.16\linewidth]{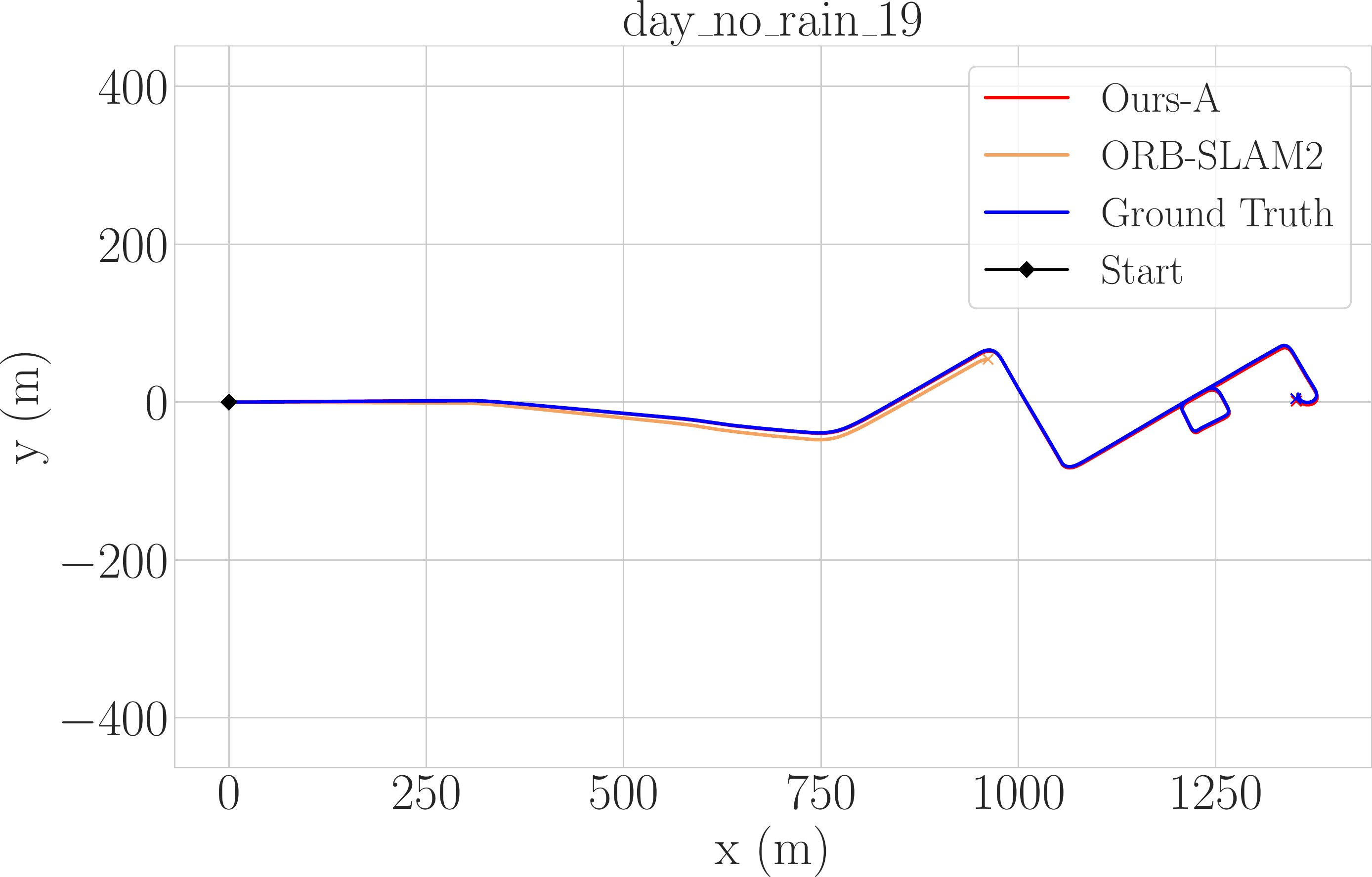}
    \includegraphics[width=0.16\linewidth]{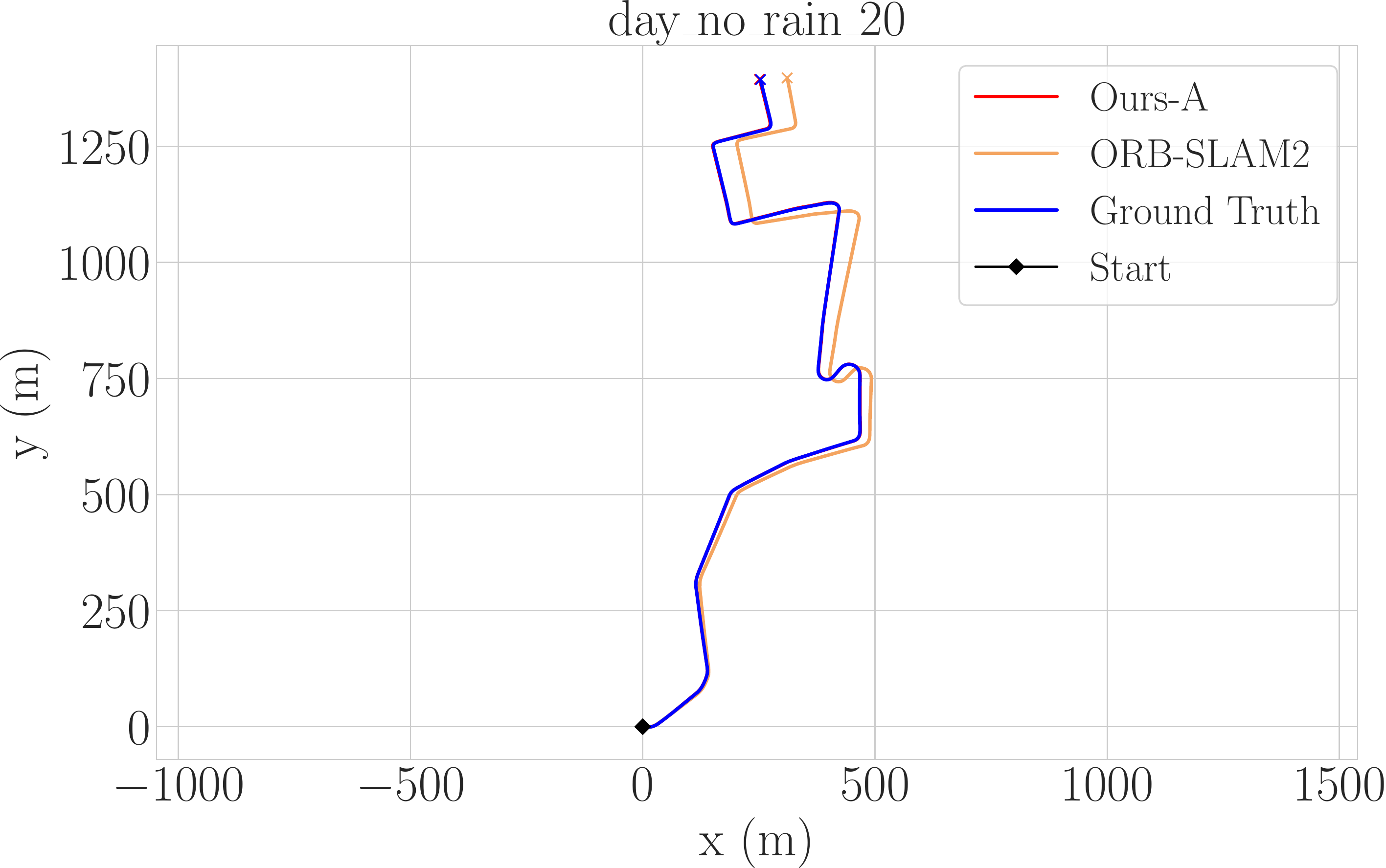}
    \includegraphics[width=0.16\linewidth]{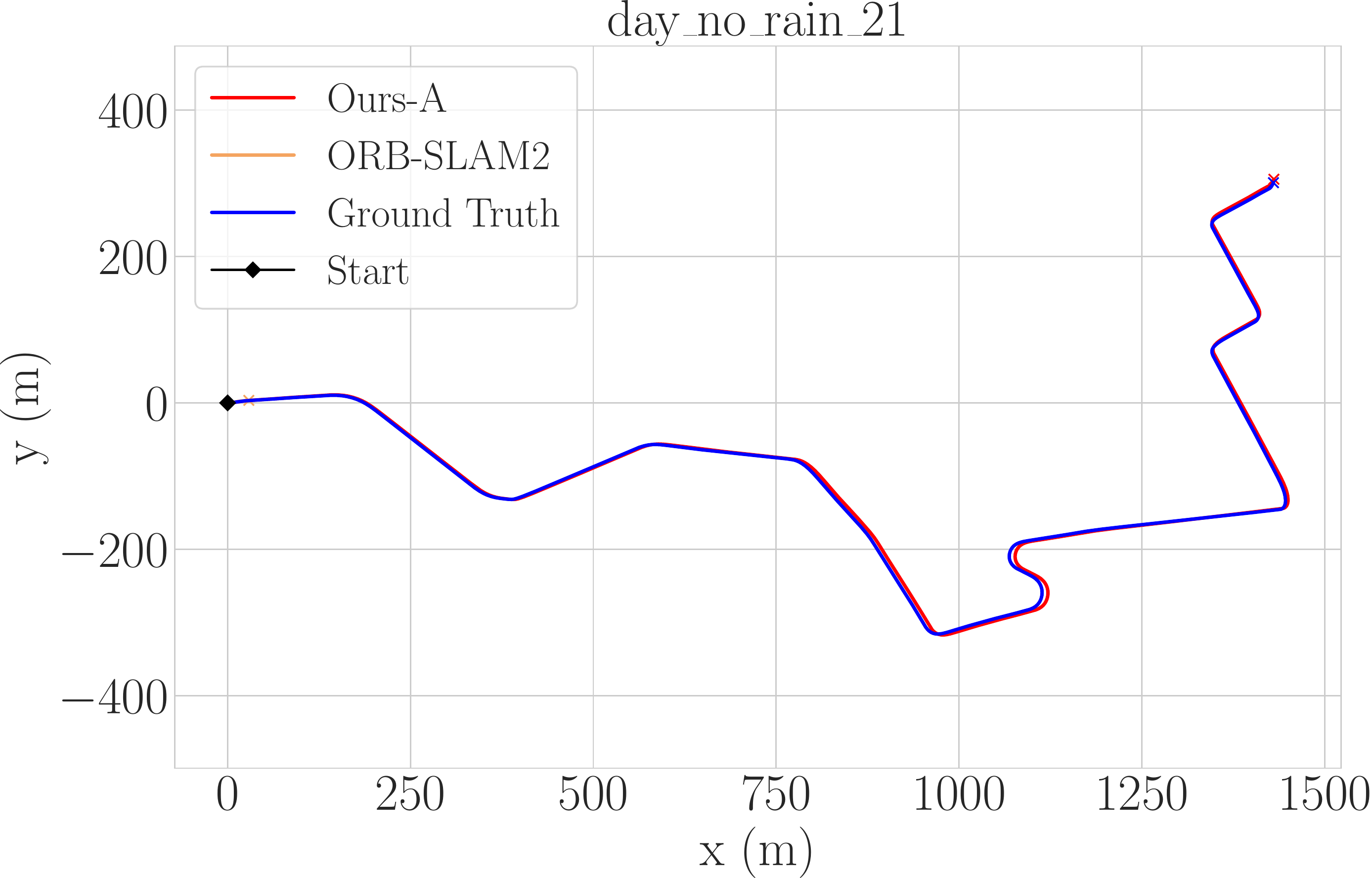}
    \includegraphics[width=0.16\linewidth]{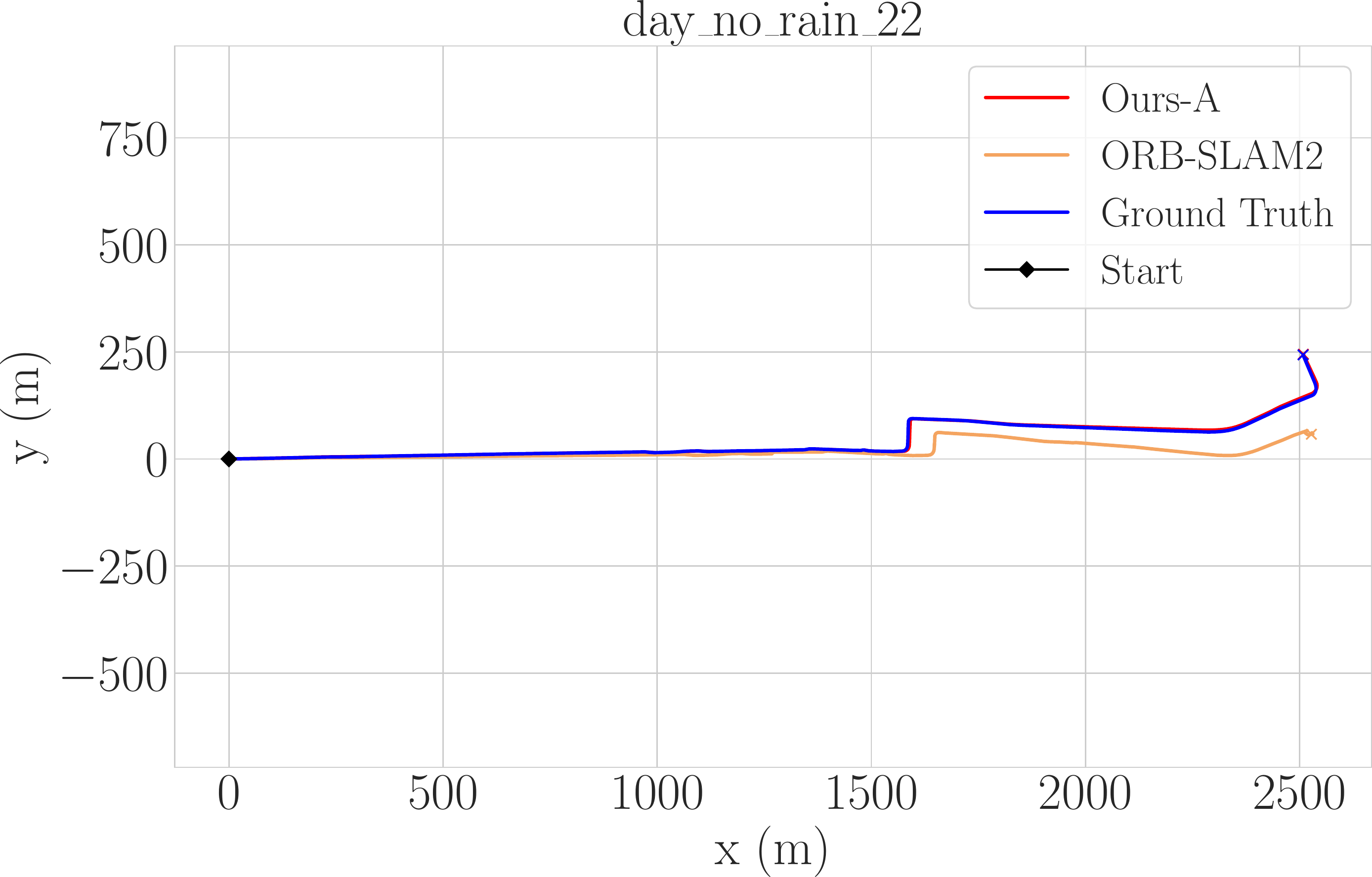}
    \includegraphics[width=0.16\linewidth]{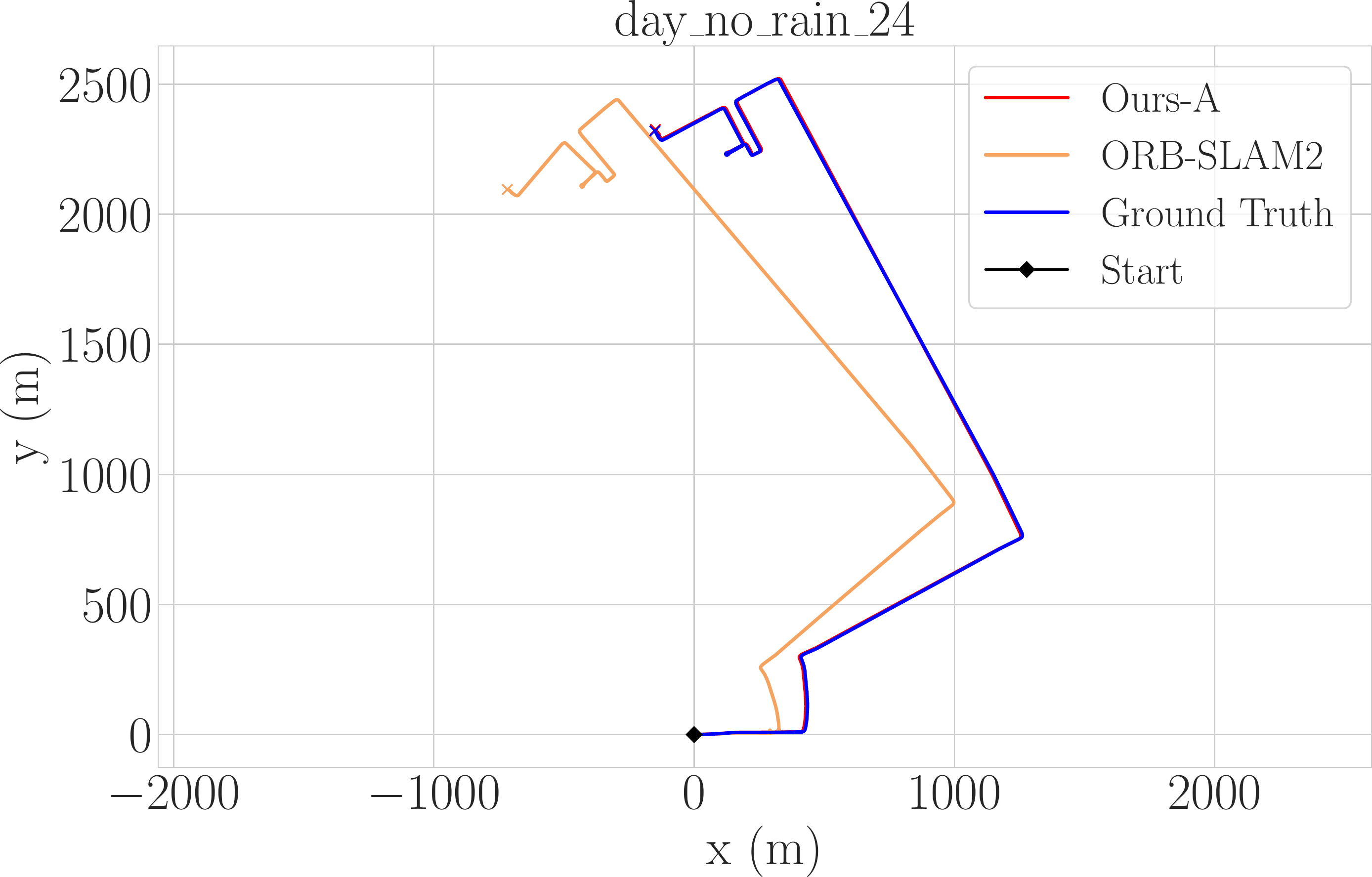}
    \includegraphics[width=0.16\linewidth]{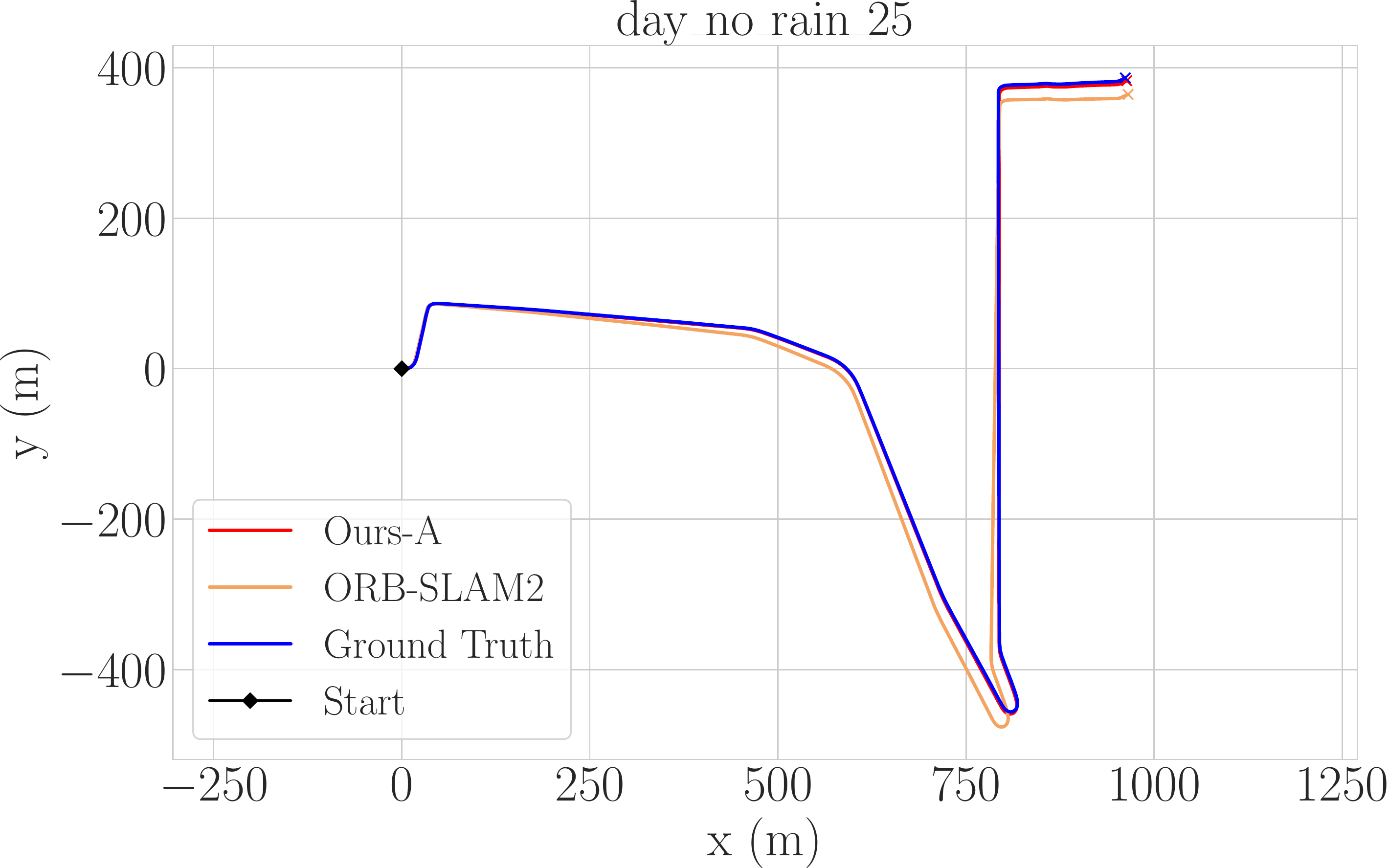}
    \includegraphics[width=0.16\linewidth]{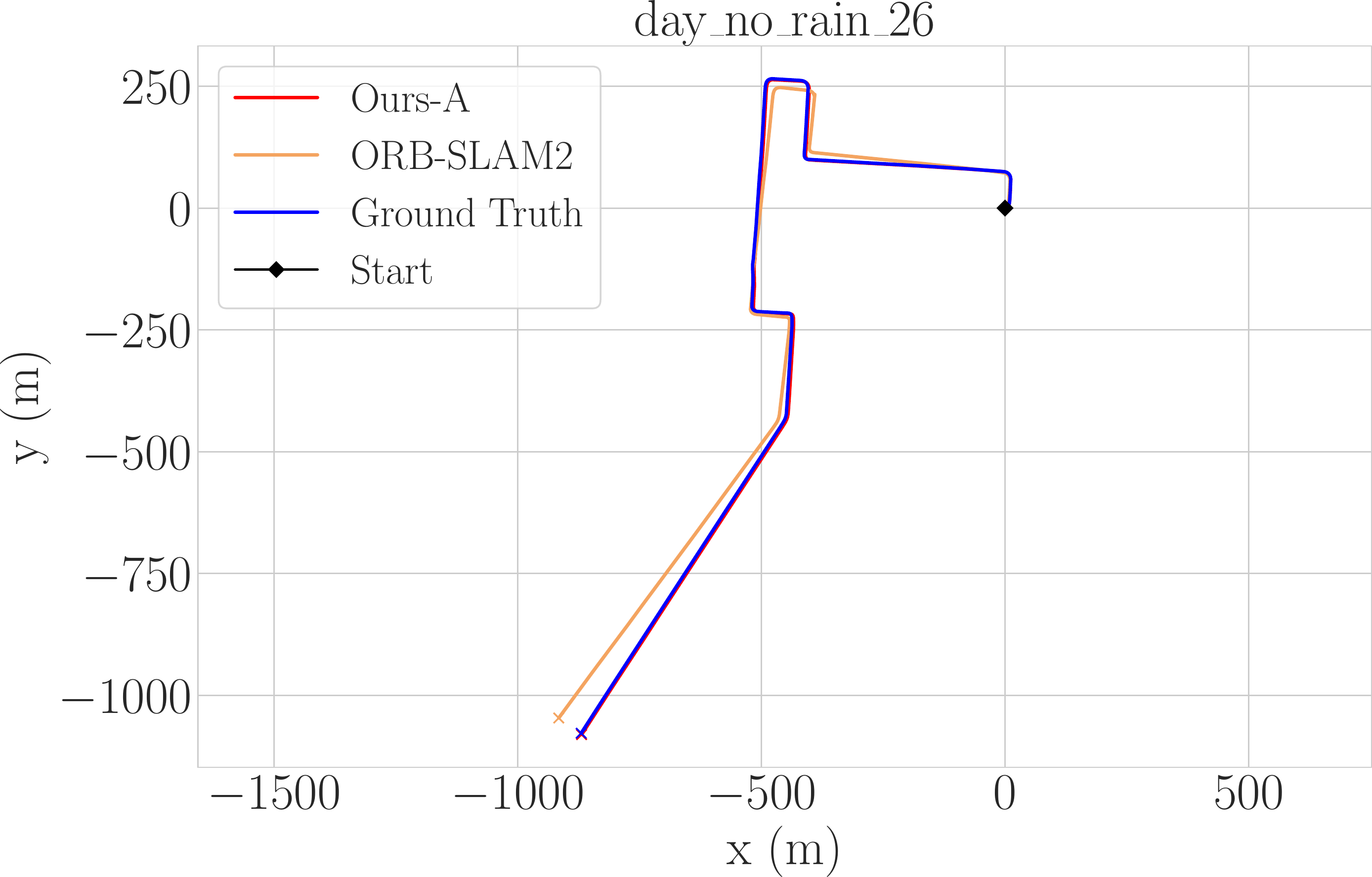}
    \includegraphics[width=0.16\linewidth]{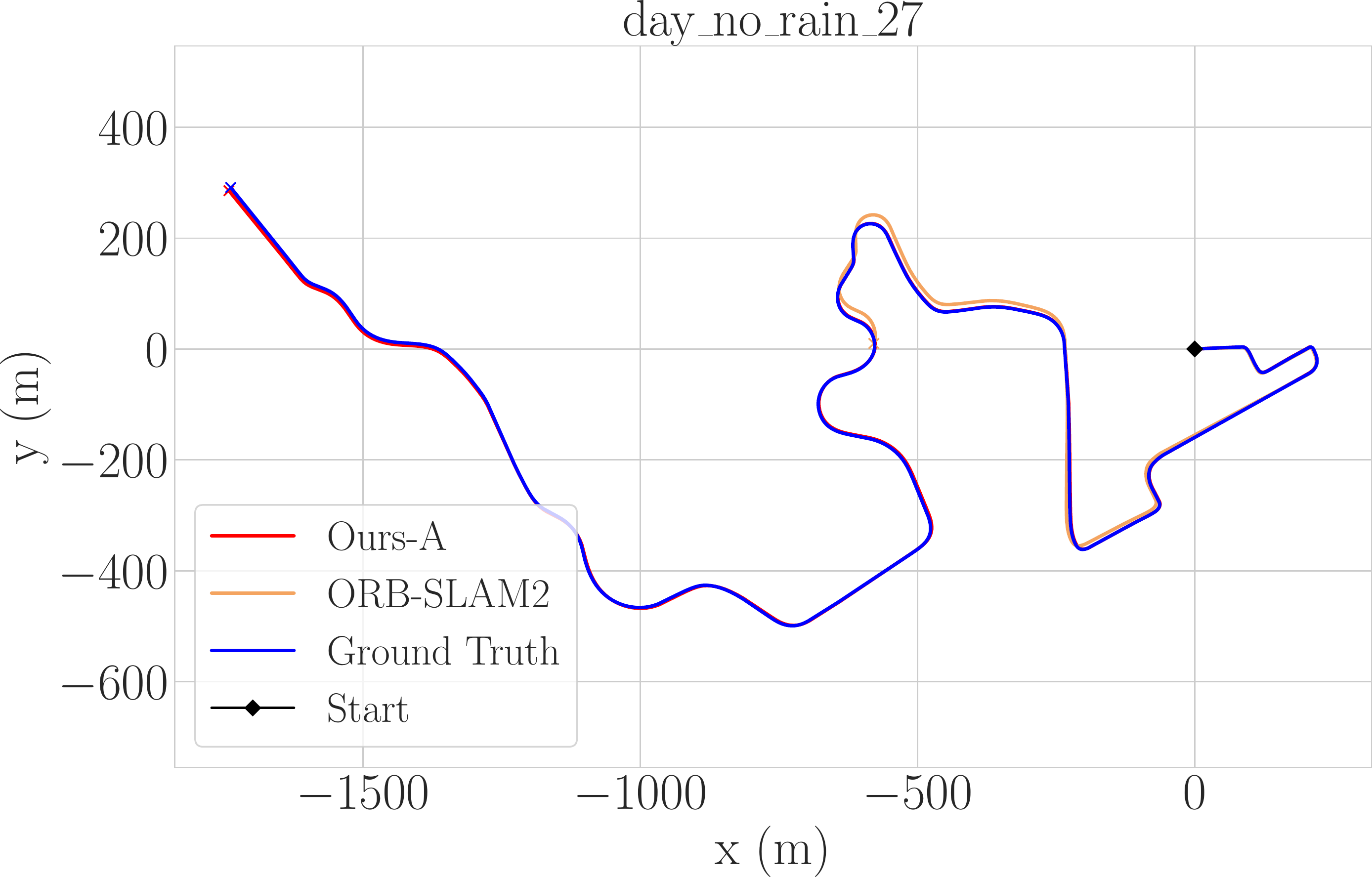}
    \includegraphics[width=0.16\linewidth]{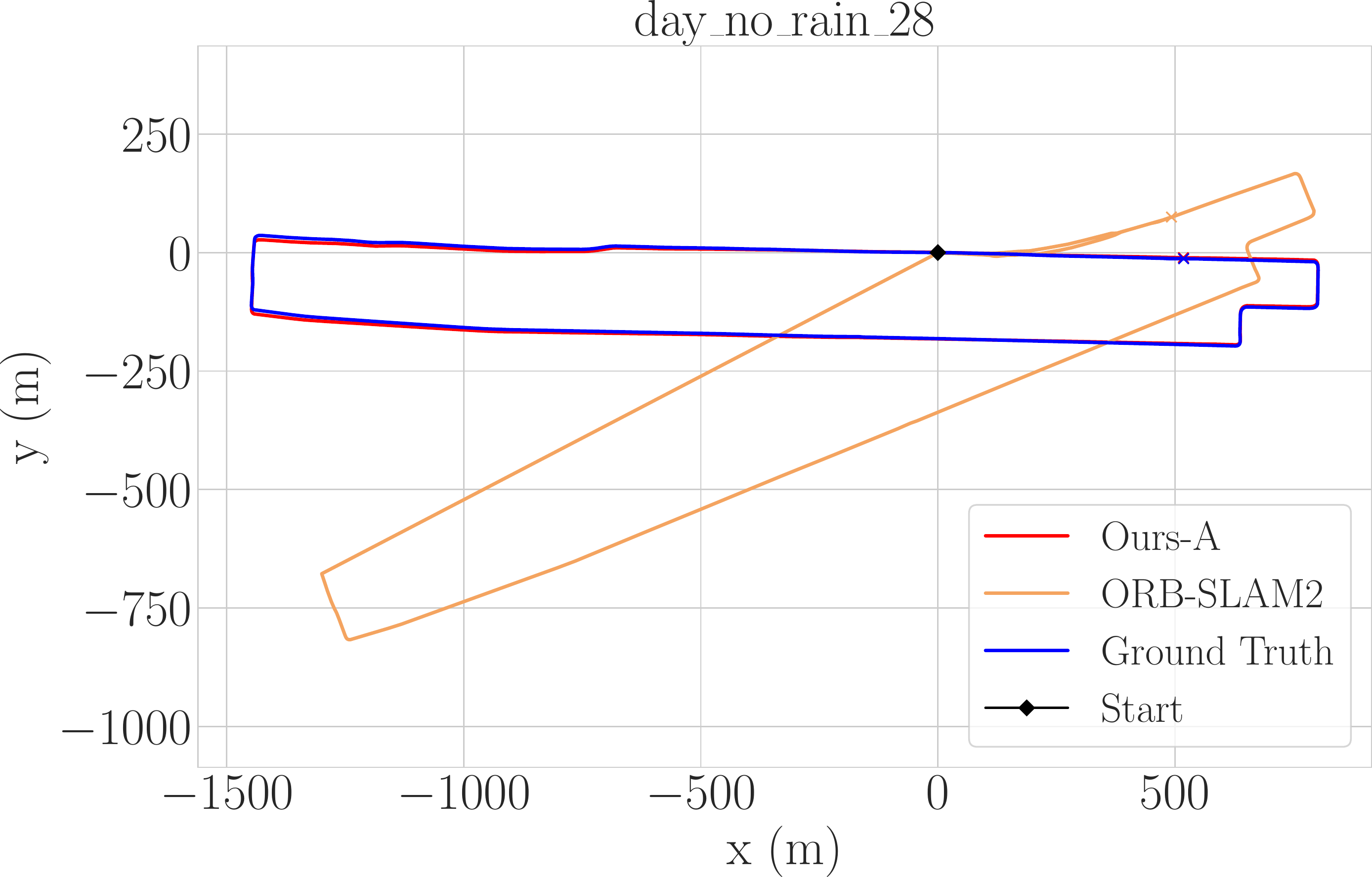}
    \includegraphics[width=0.16\linewidth]{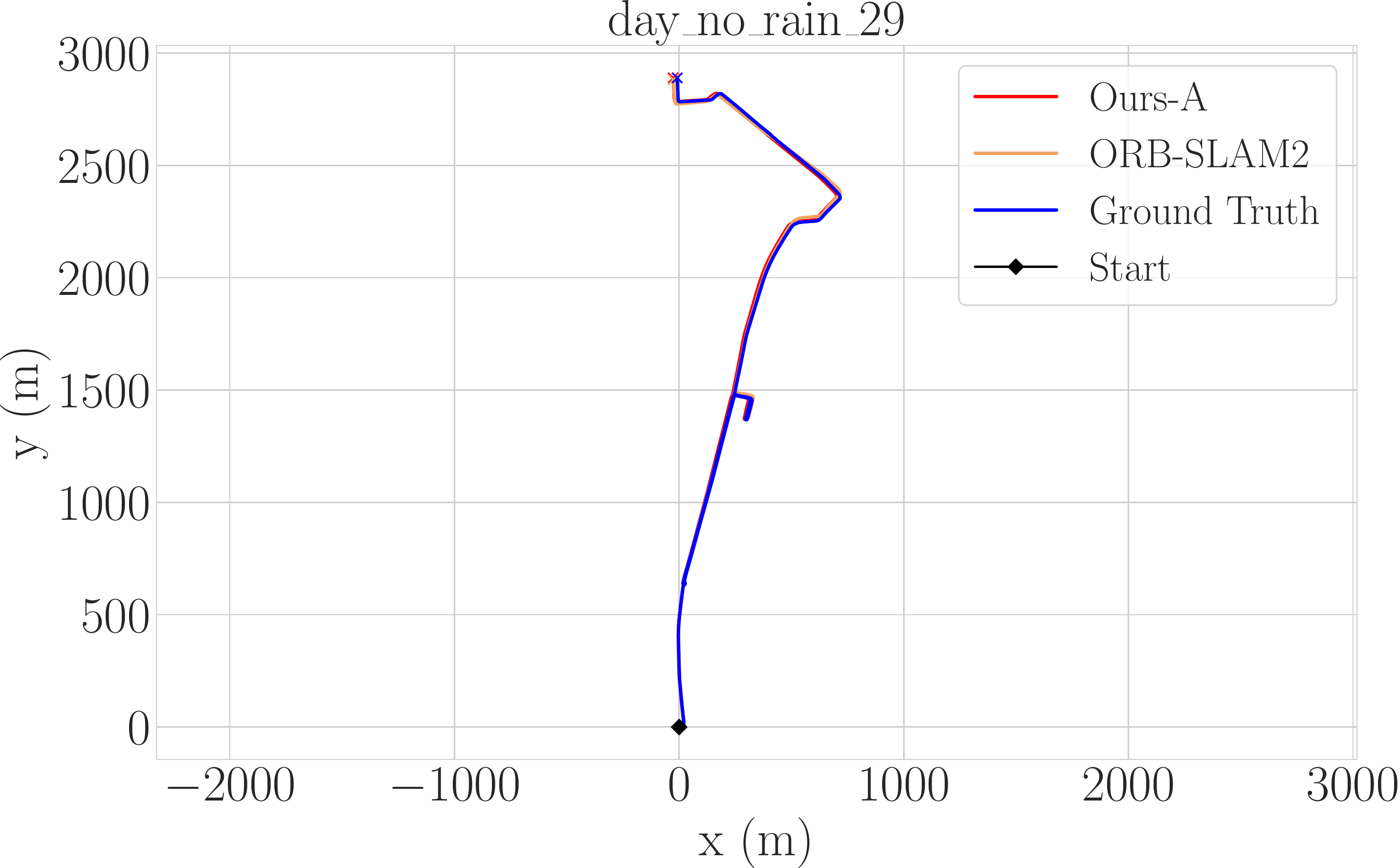}
    \includegraphics[width=0.16\linewidth]{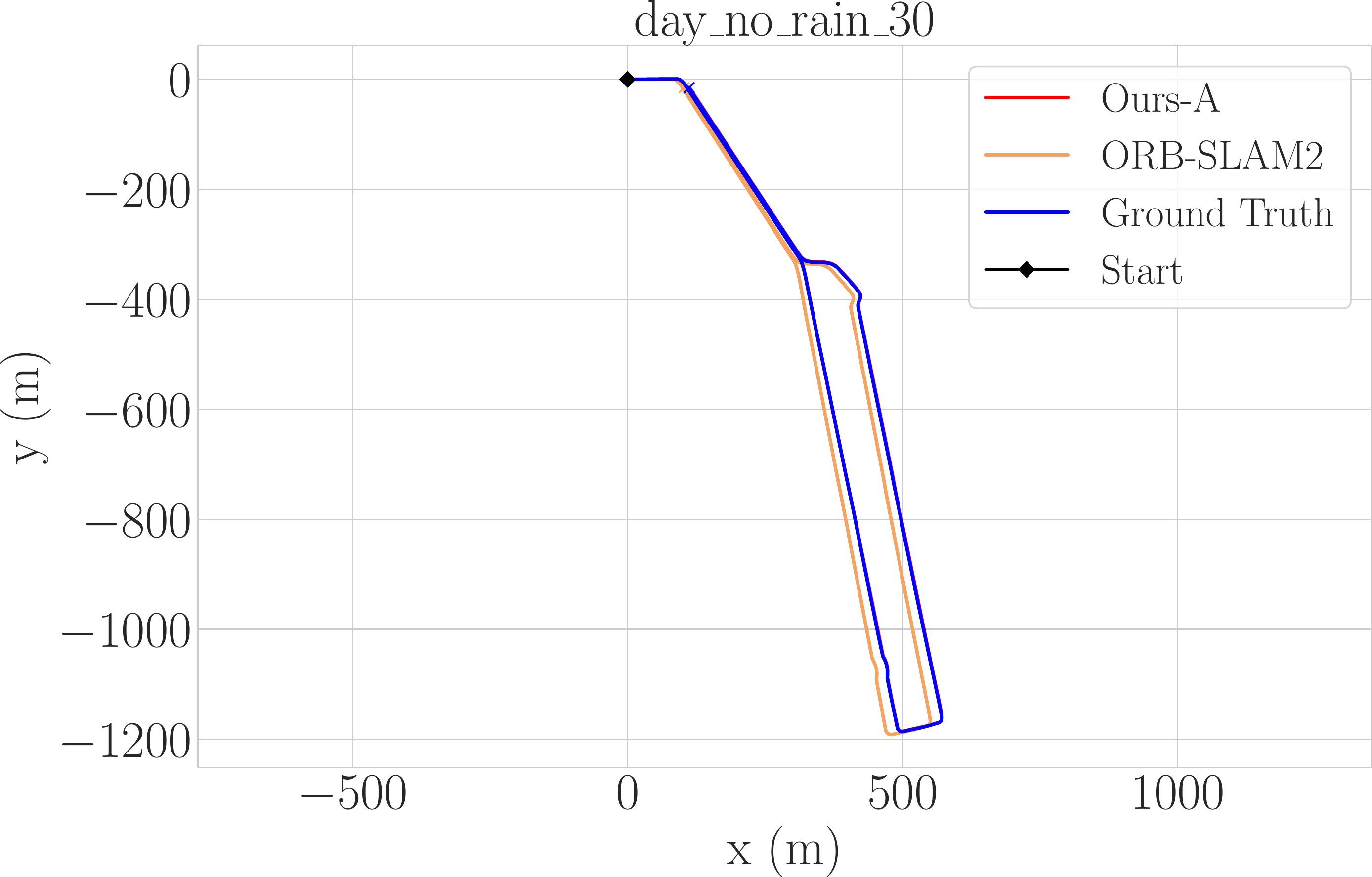}
    \includegraphics[width=0.16\linewidth]{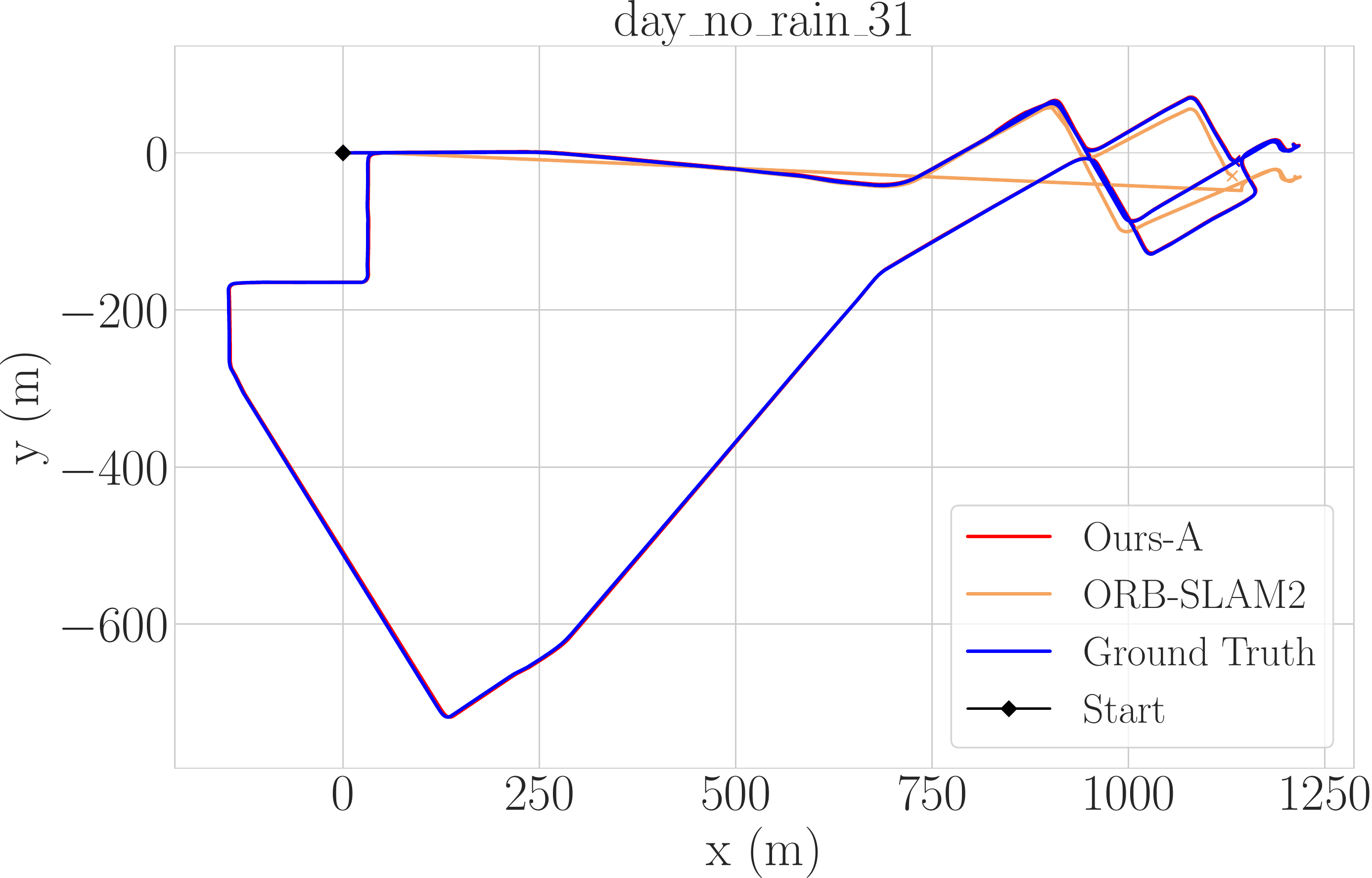}
    \includegraphics[width=0.16\linewidth]{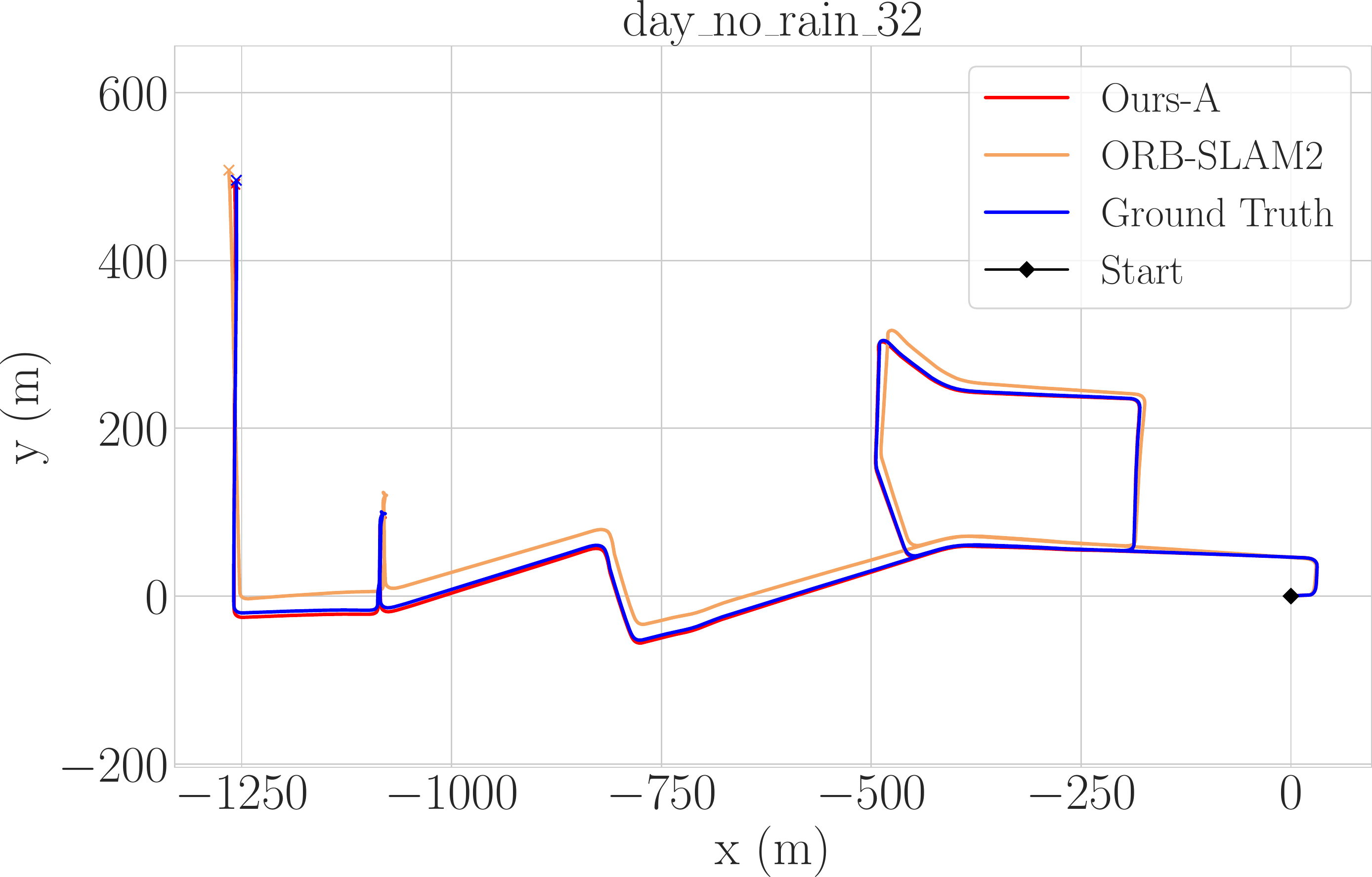}
    \includegraphics[width=0.16\linewidth]{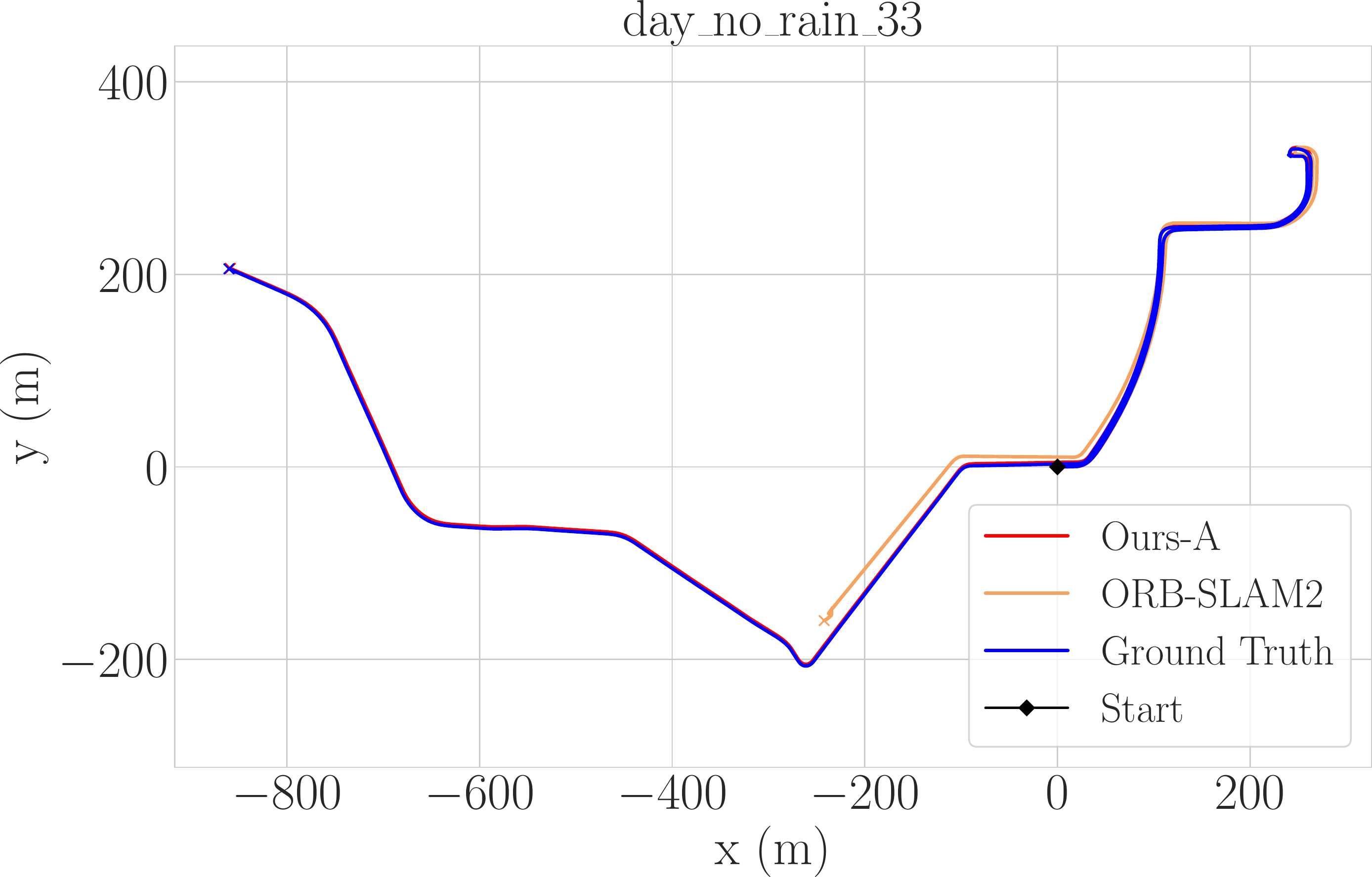}
    \includegraphics[width=0.16\linewidth]{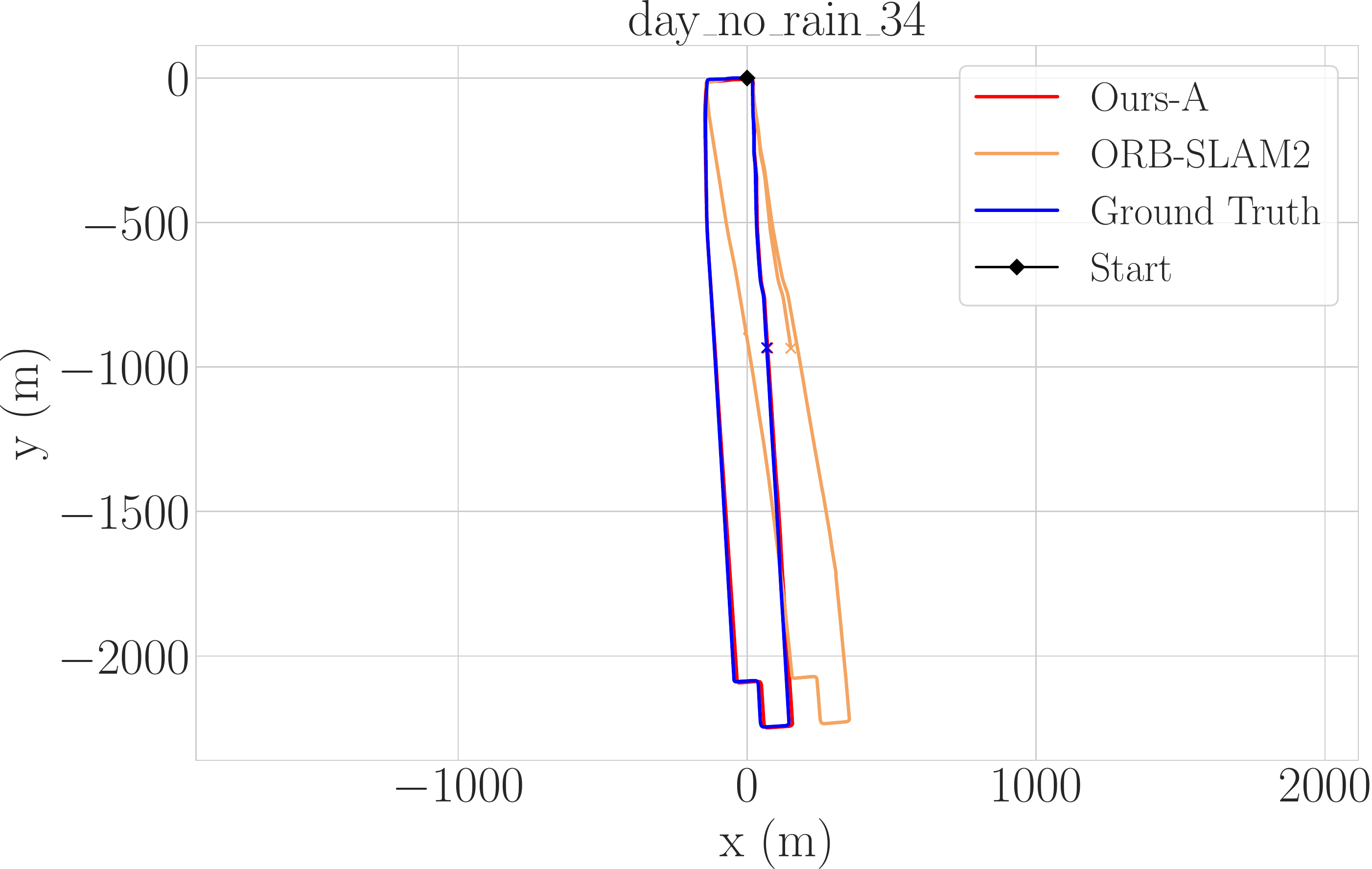}
    \includegraphics[width=0.16\linewidth]{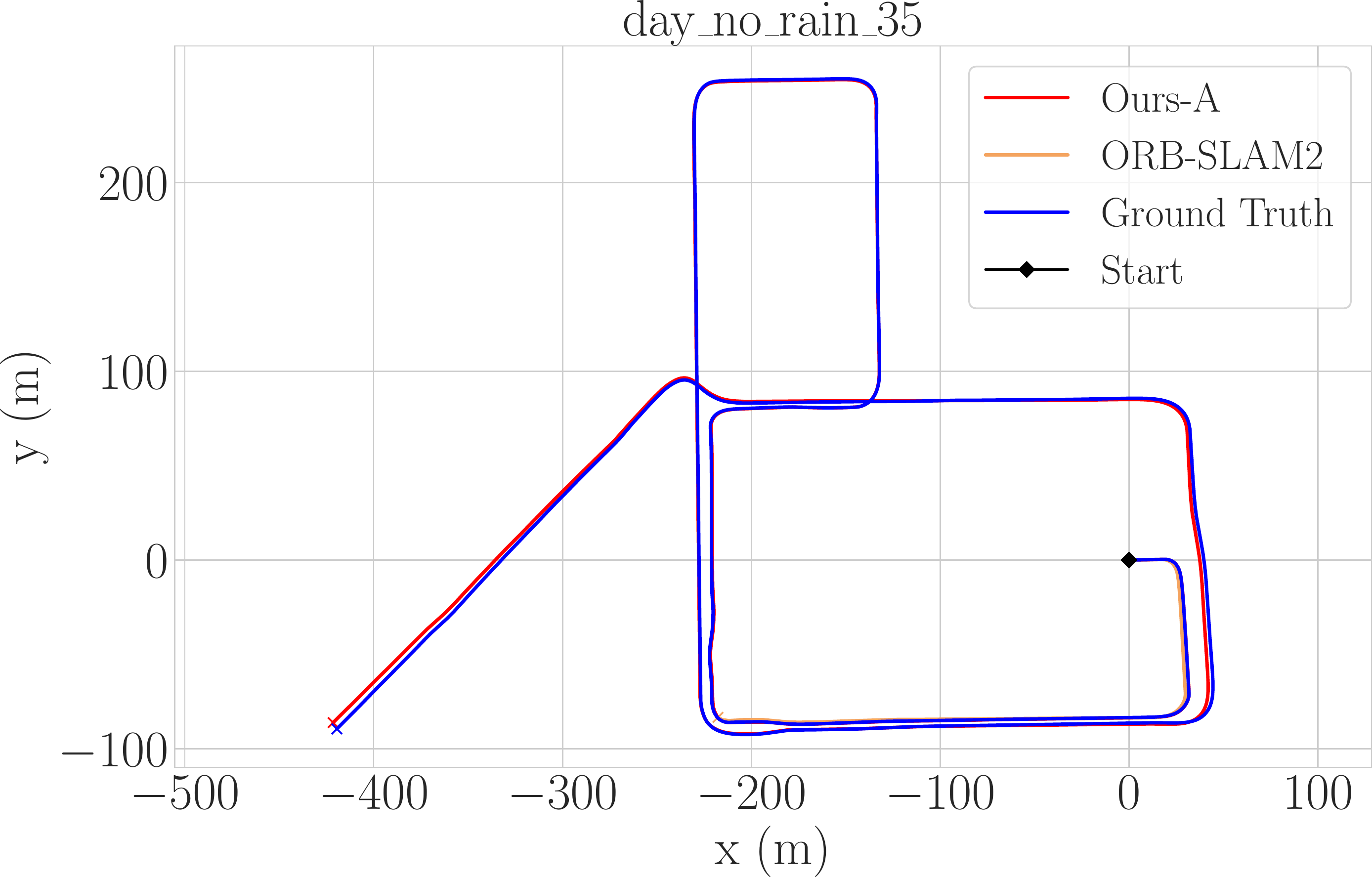}
    \includegraphics[width=0.16\linewidth]{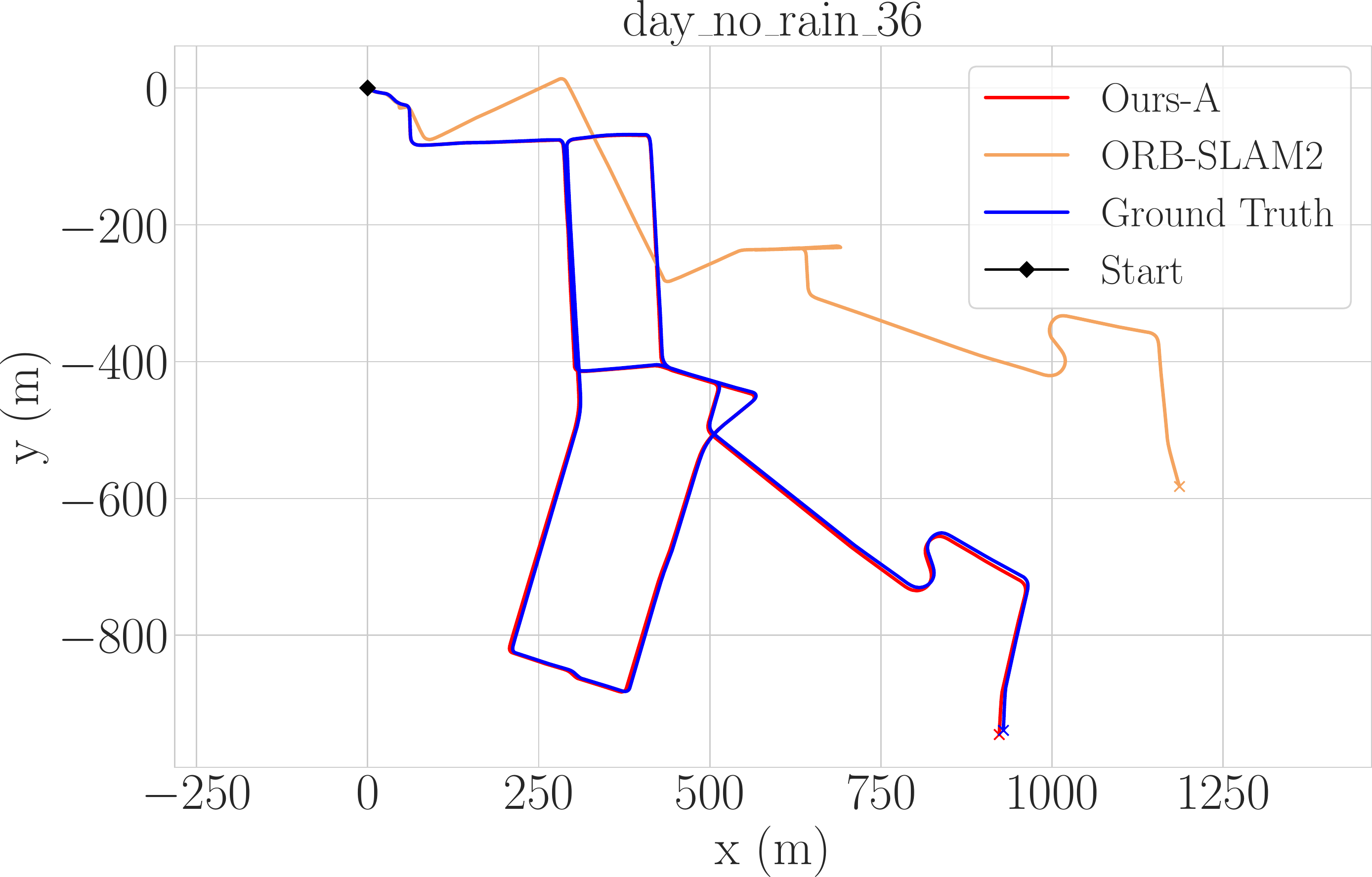}
    \includegraphics[width=0.16\linewidth]{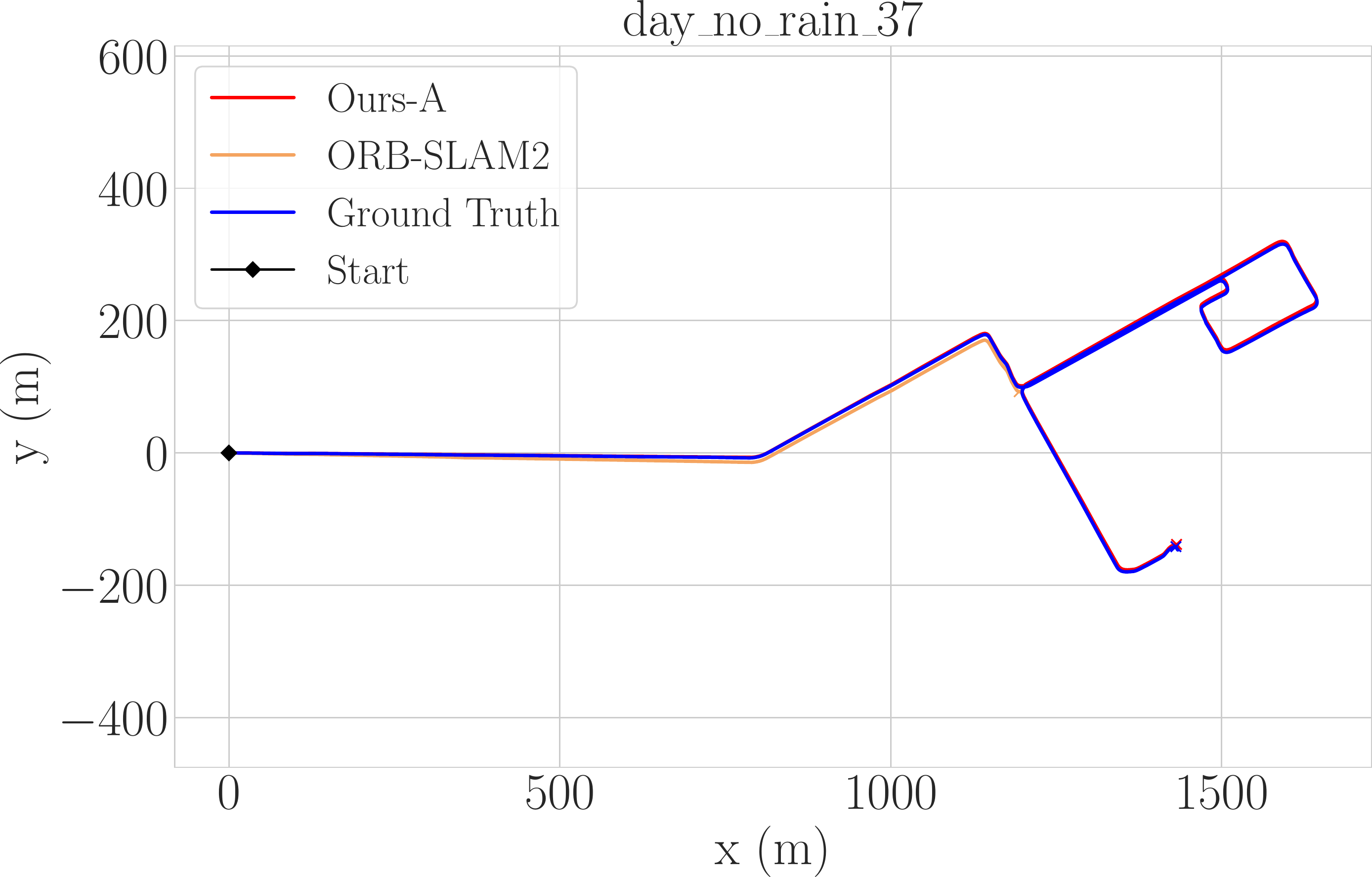}
    \includegraphics[width=0.16\linewidth]{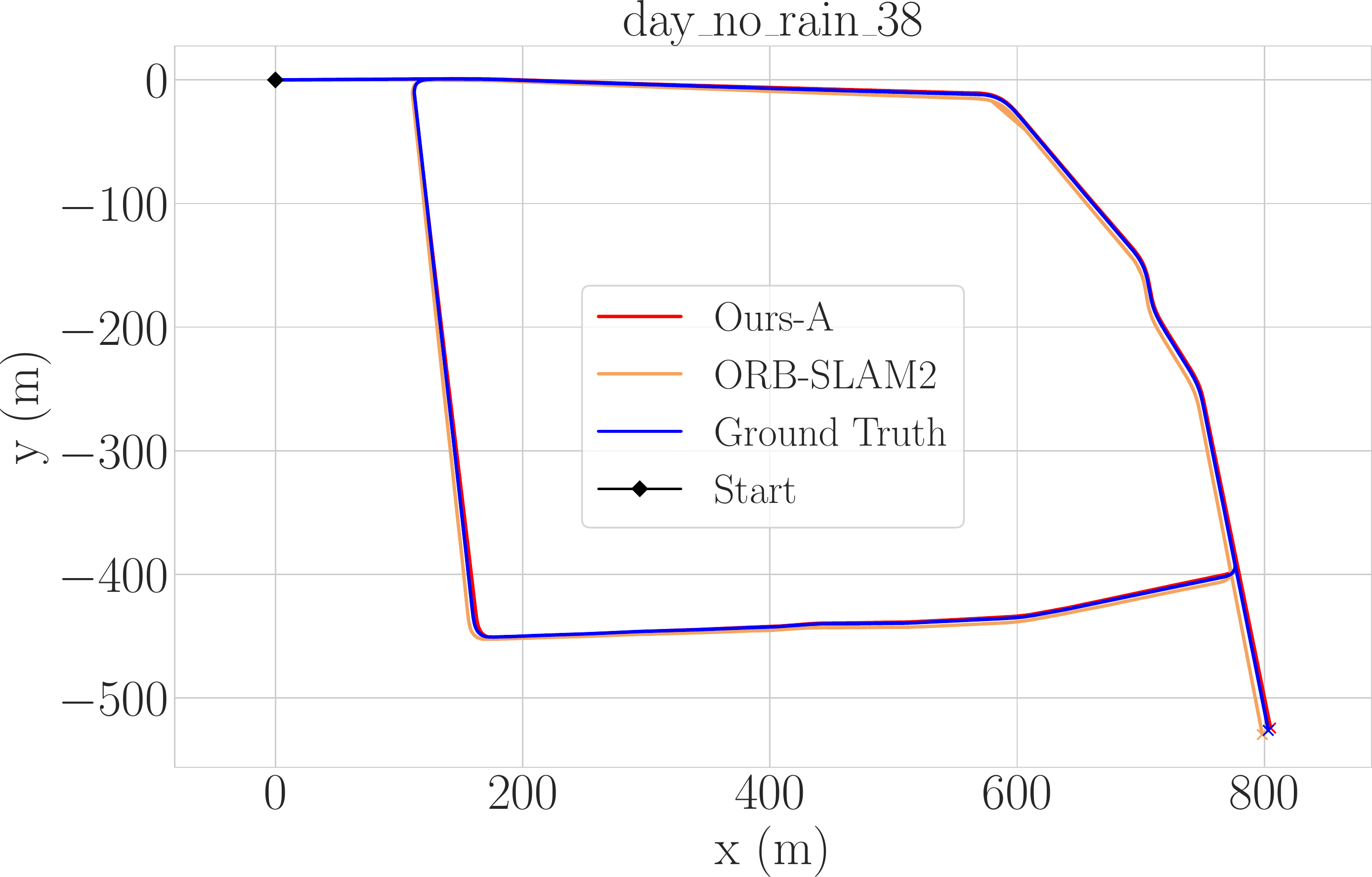}
    \includegraphics[width=0.16\linewidth]{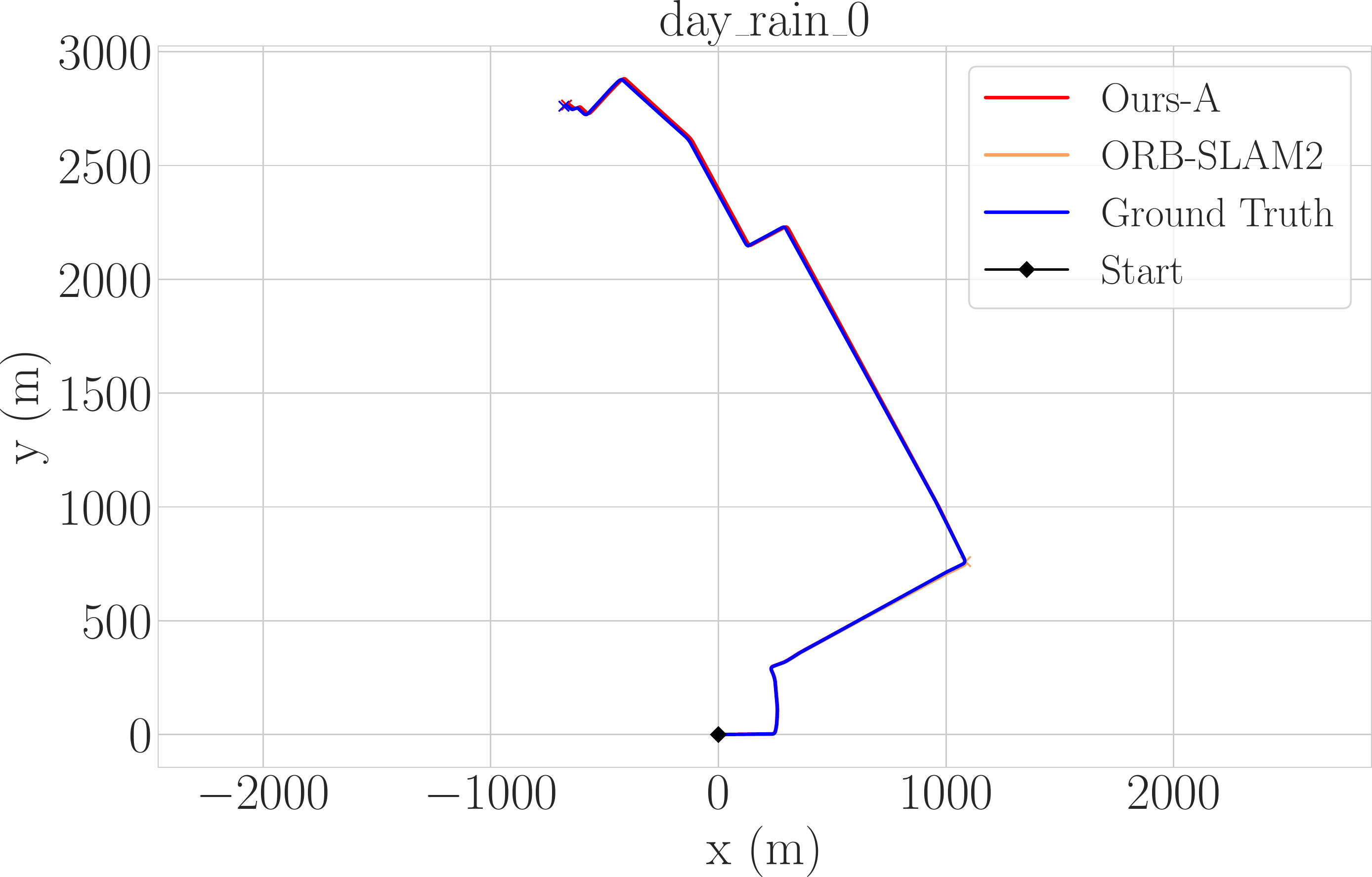}
    \includegraphics[width=0.16\linewidth]{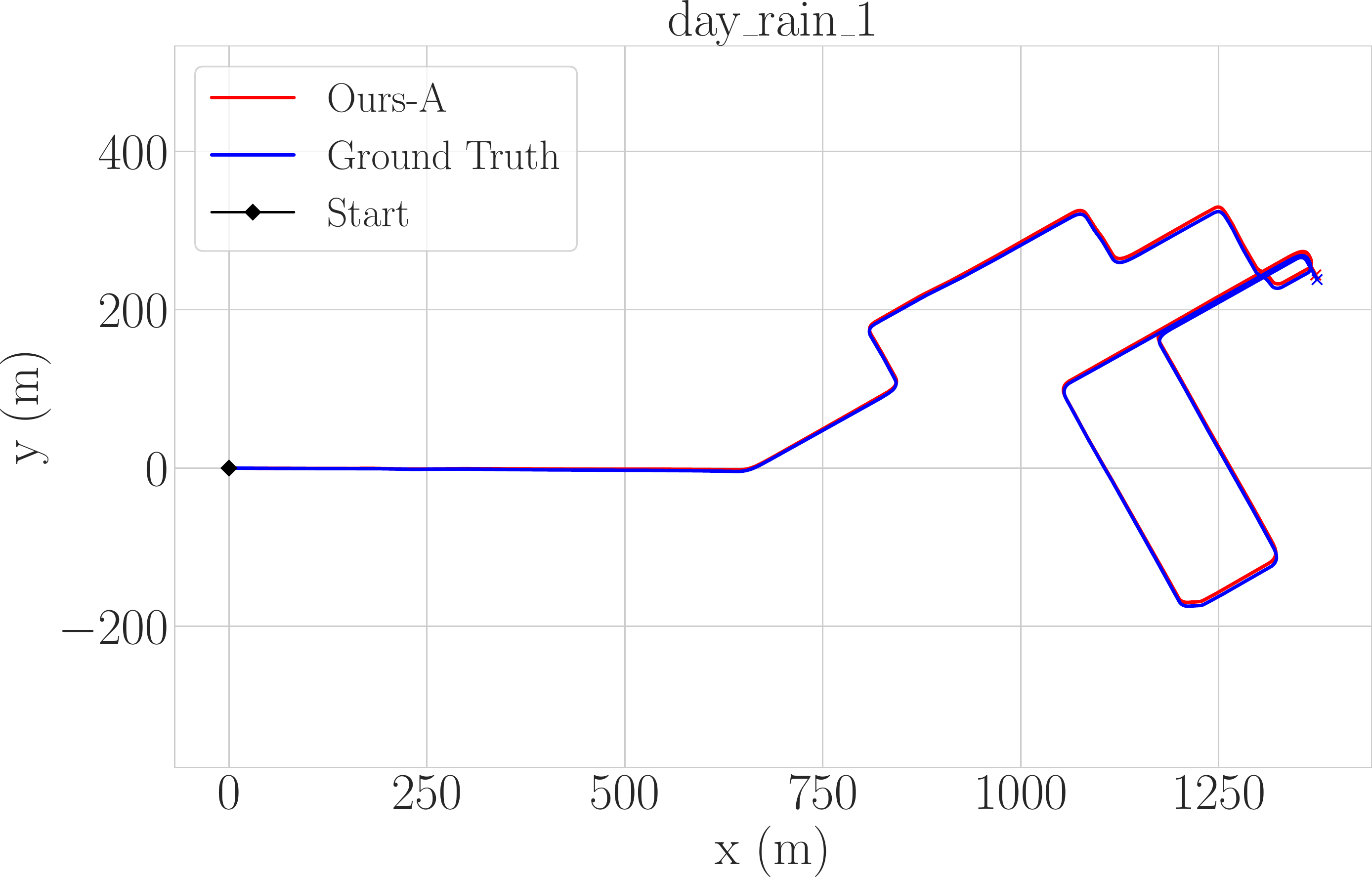}
    \includegraphics[width=0.16\linewidth]{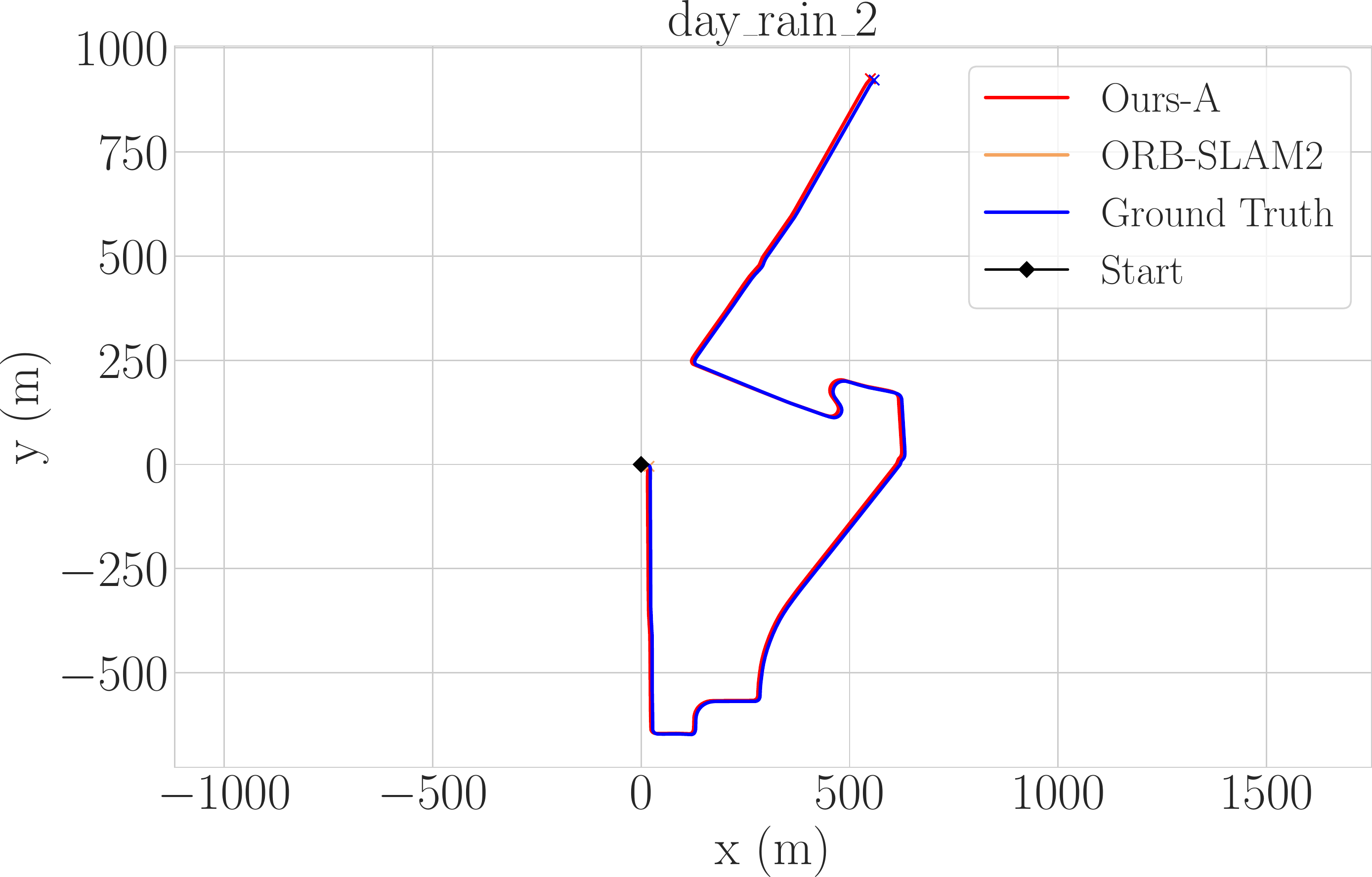}
    \includegraphics[width=0.16\linewidth]{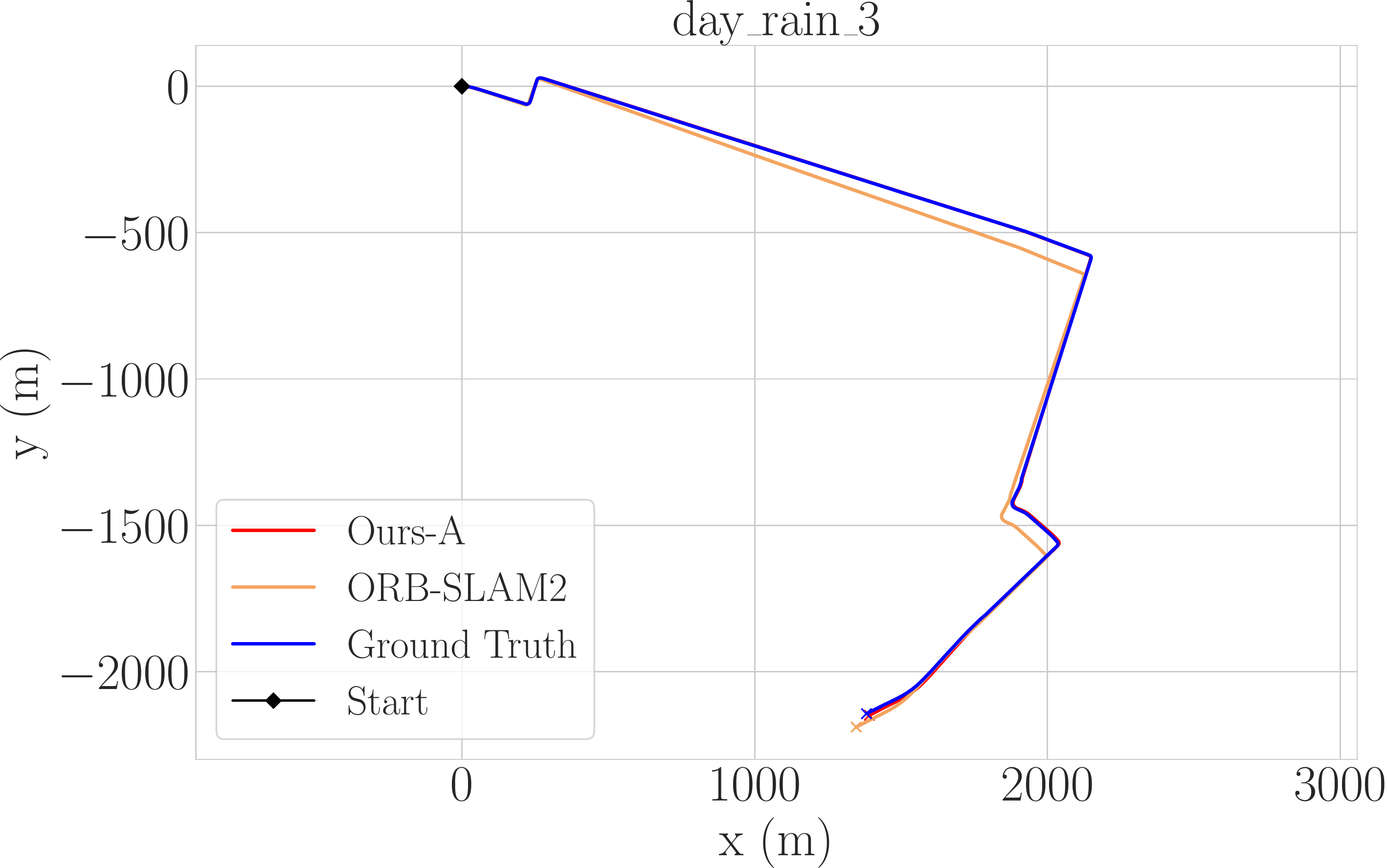}
    \includegraphics[width=0.16\linewidth]{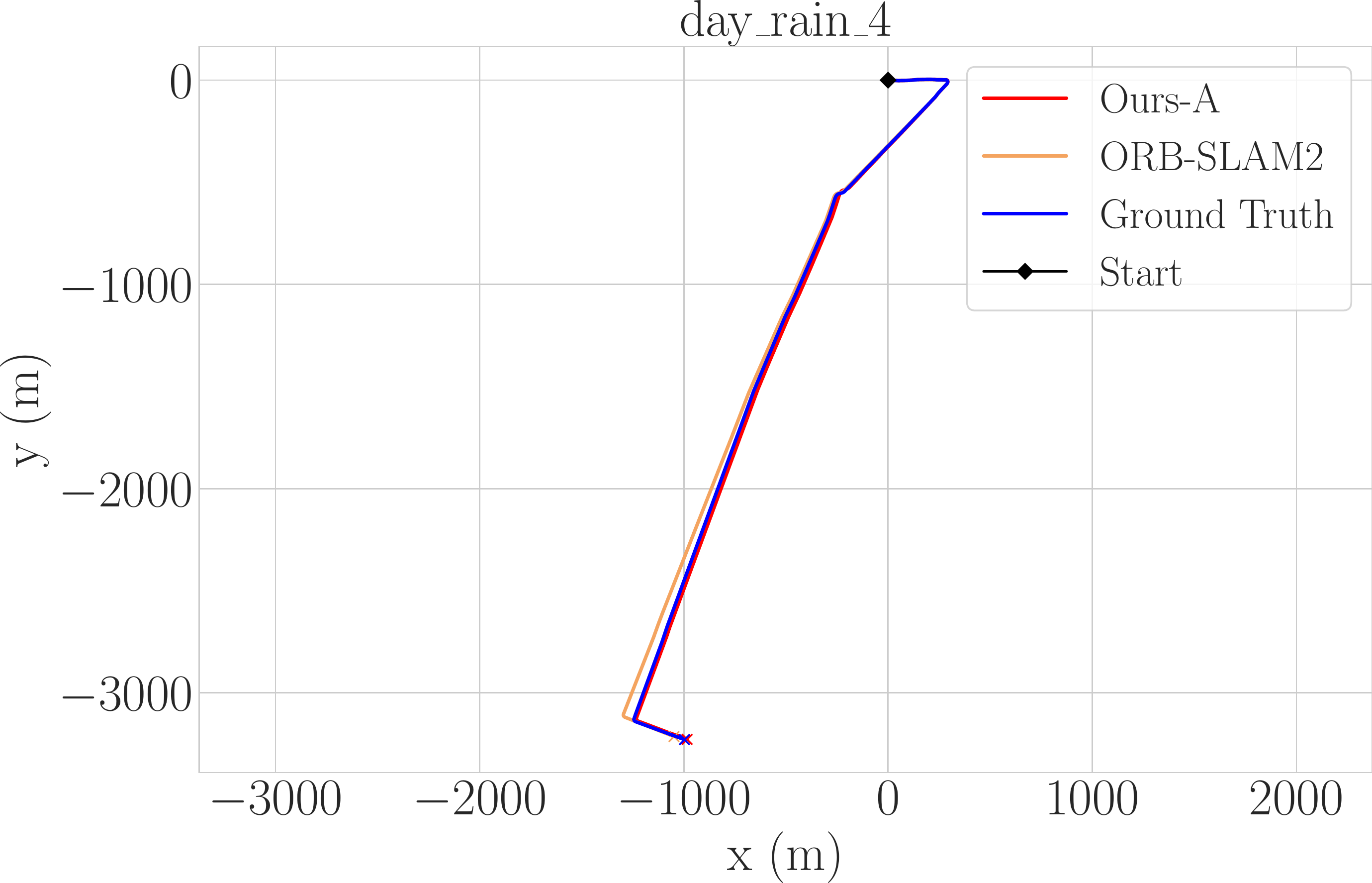}
    \includegraphics[width=0.16\linewidth]{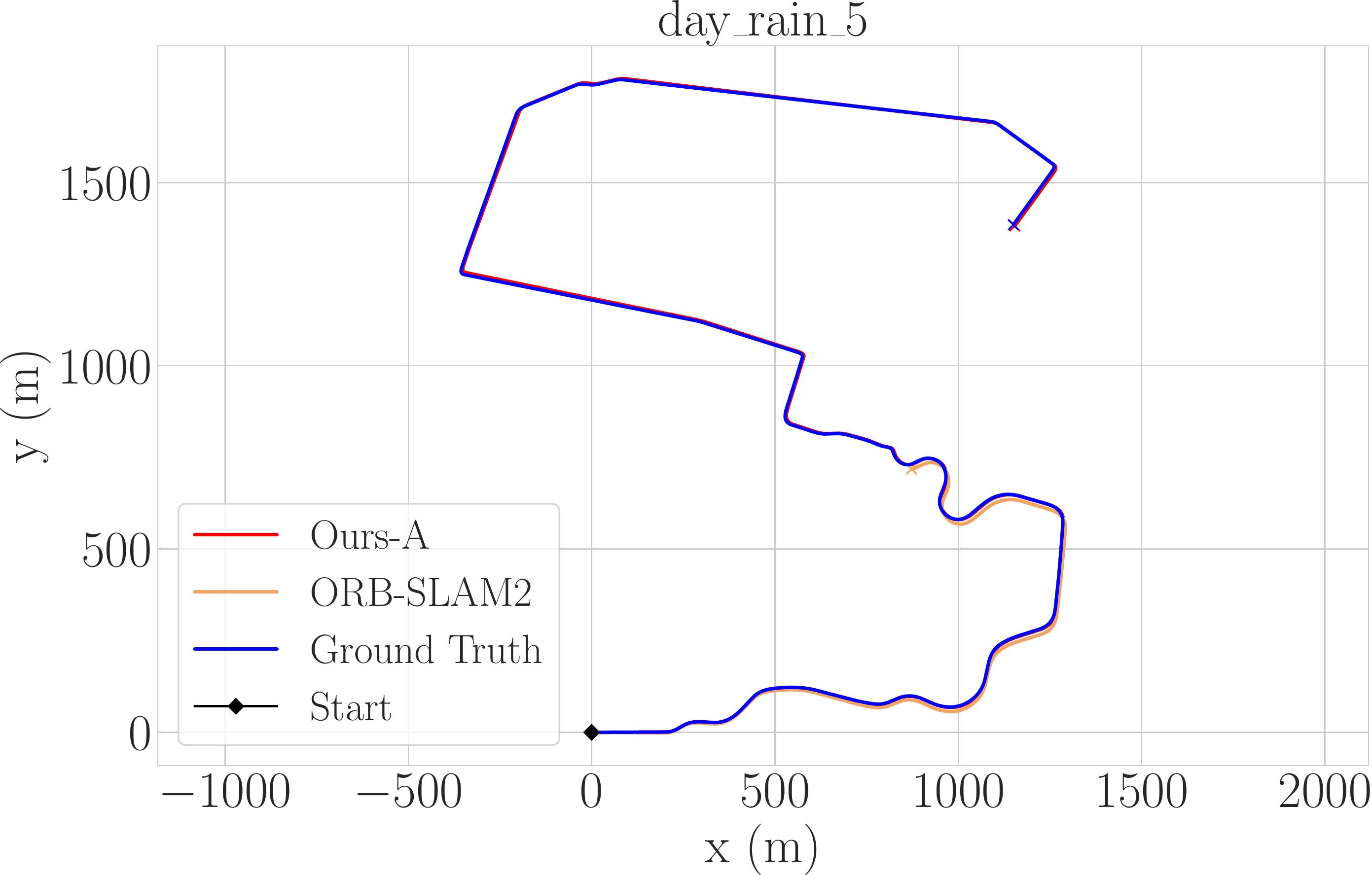}
    \includegraphics[width=0.16\linewidth]{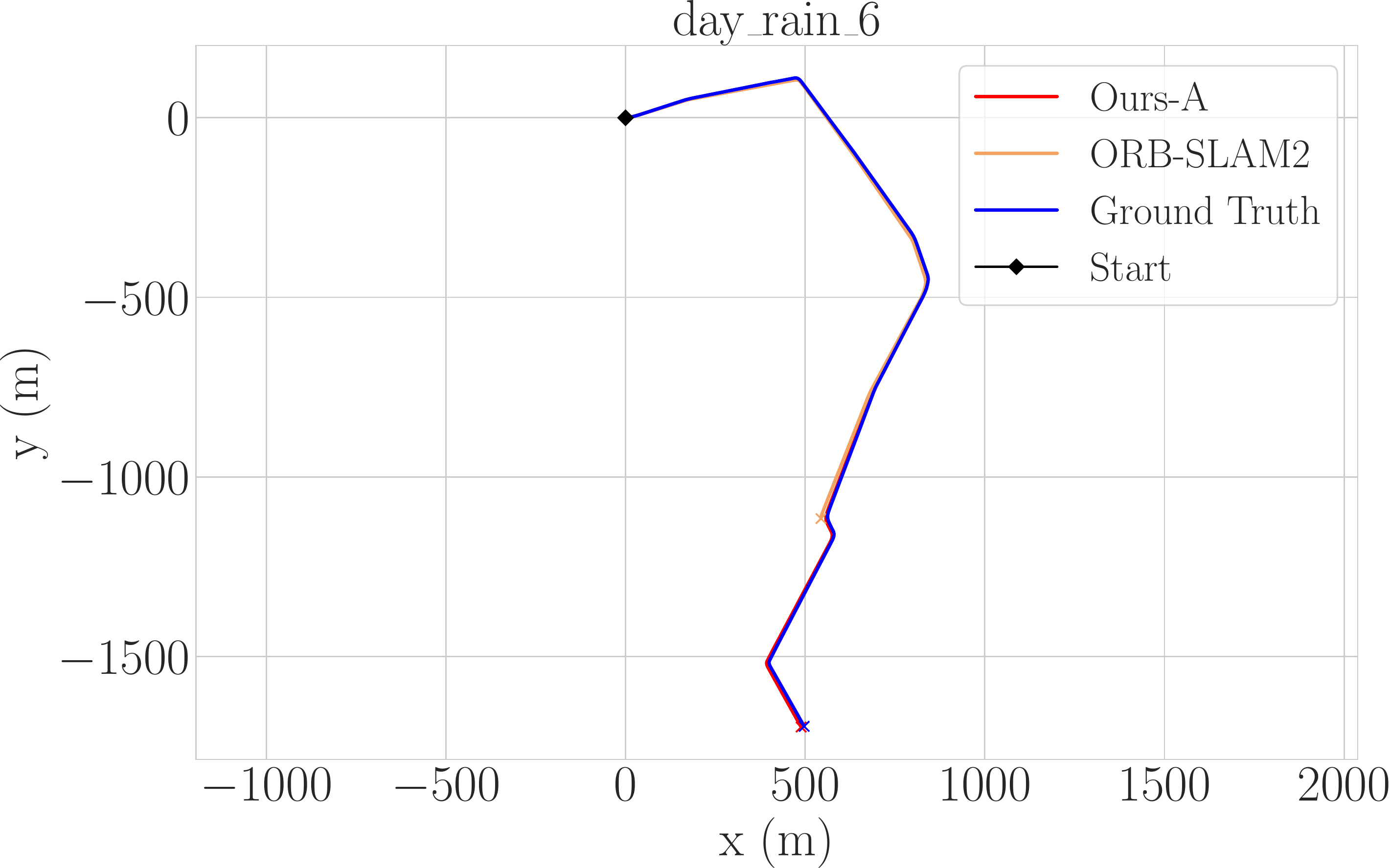}
    \includegraphics[width=0.16\linewidth]{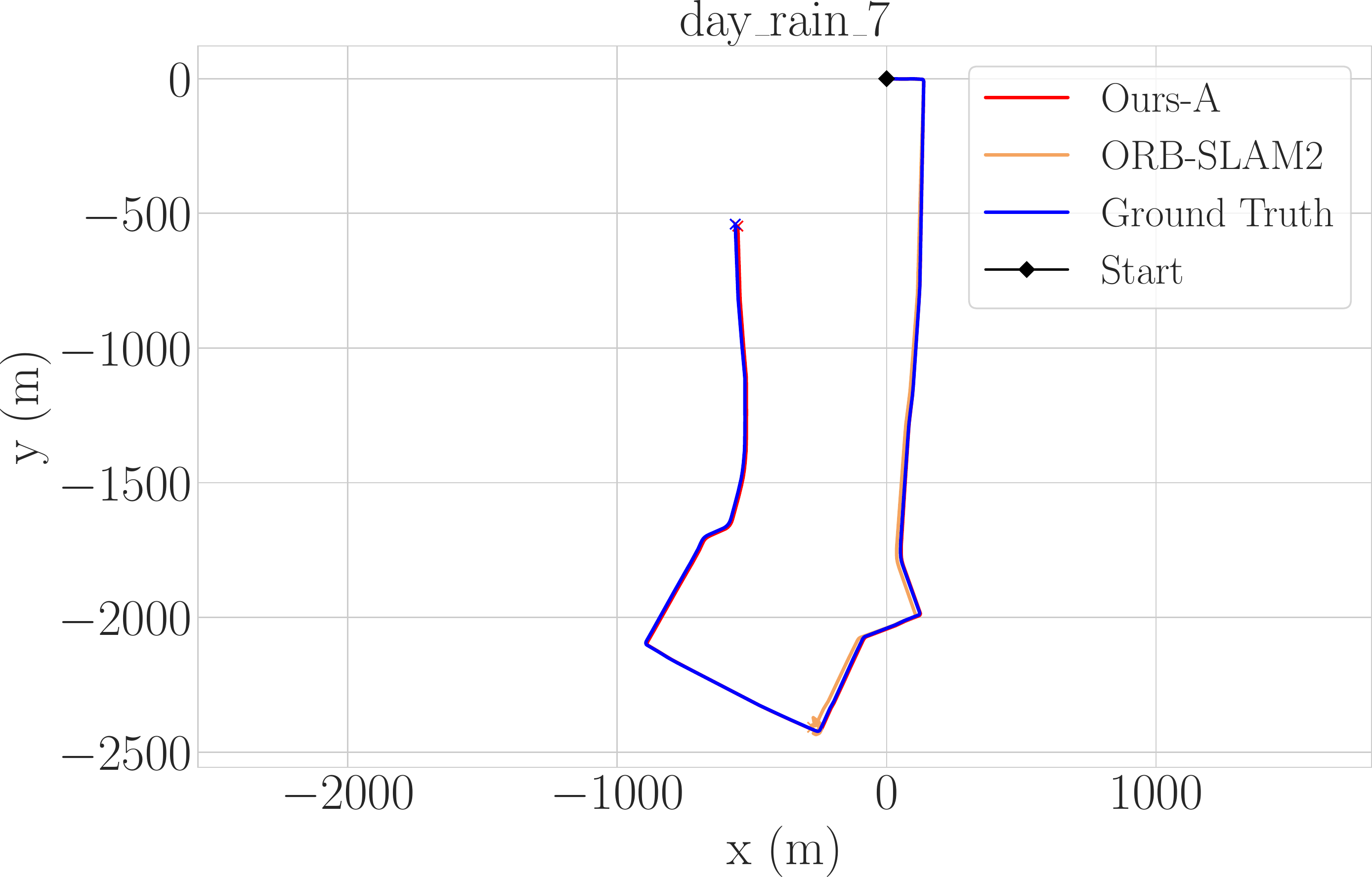}
    \includegraphics[width=0.16\linewidth]{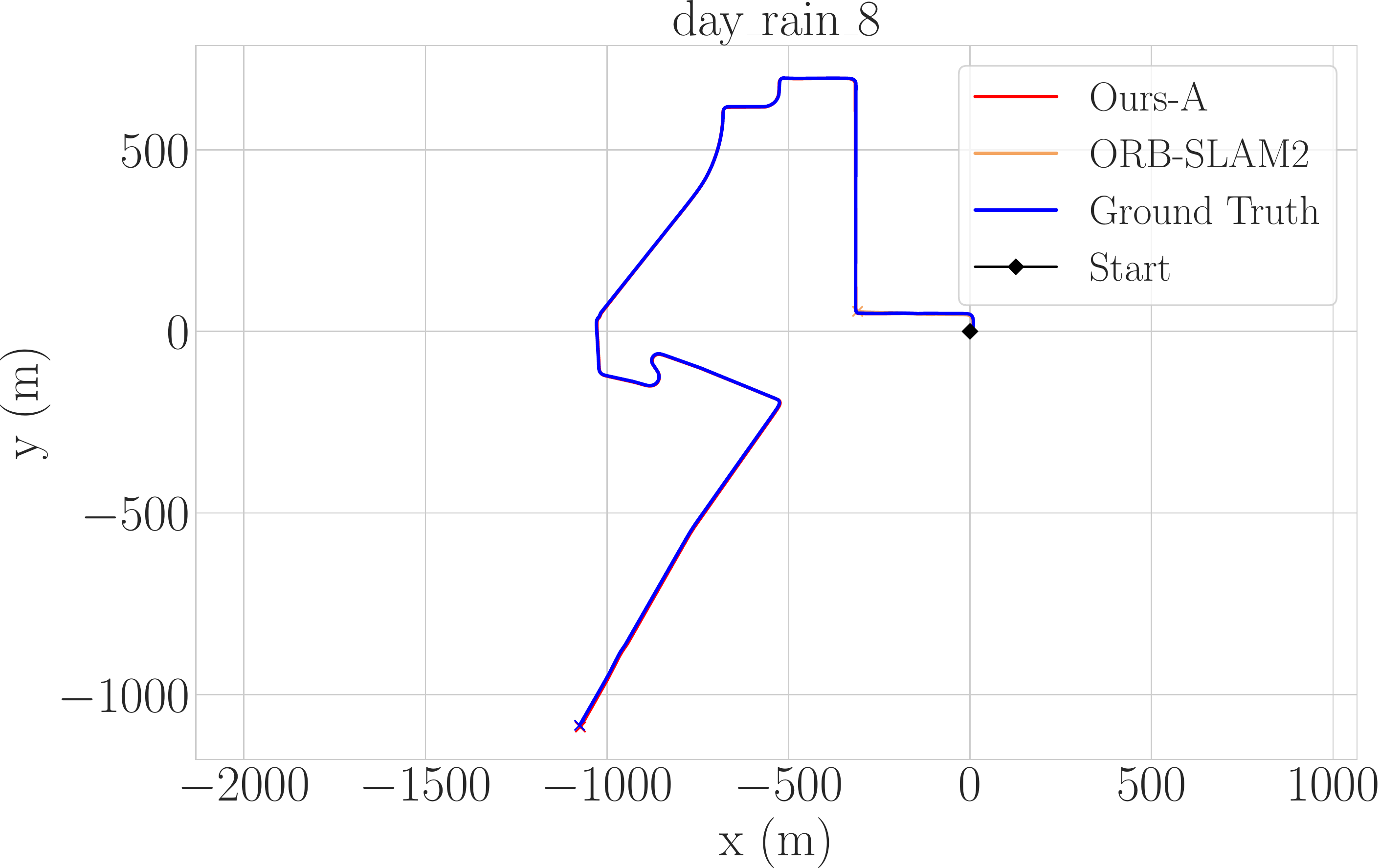}
    \includegraphics[width=0.16\linewidth]{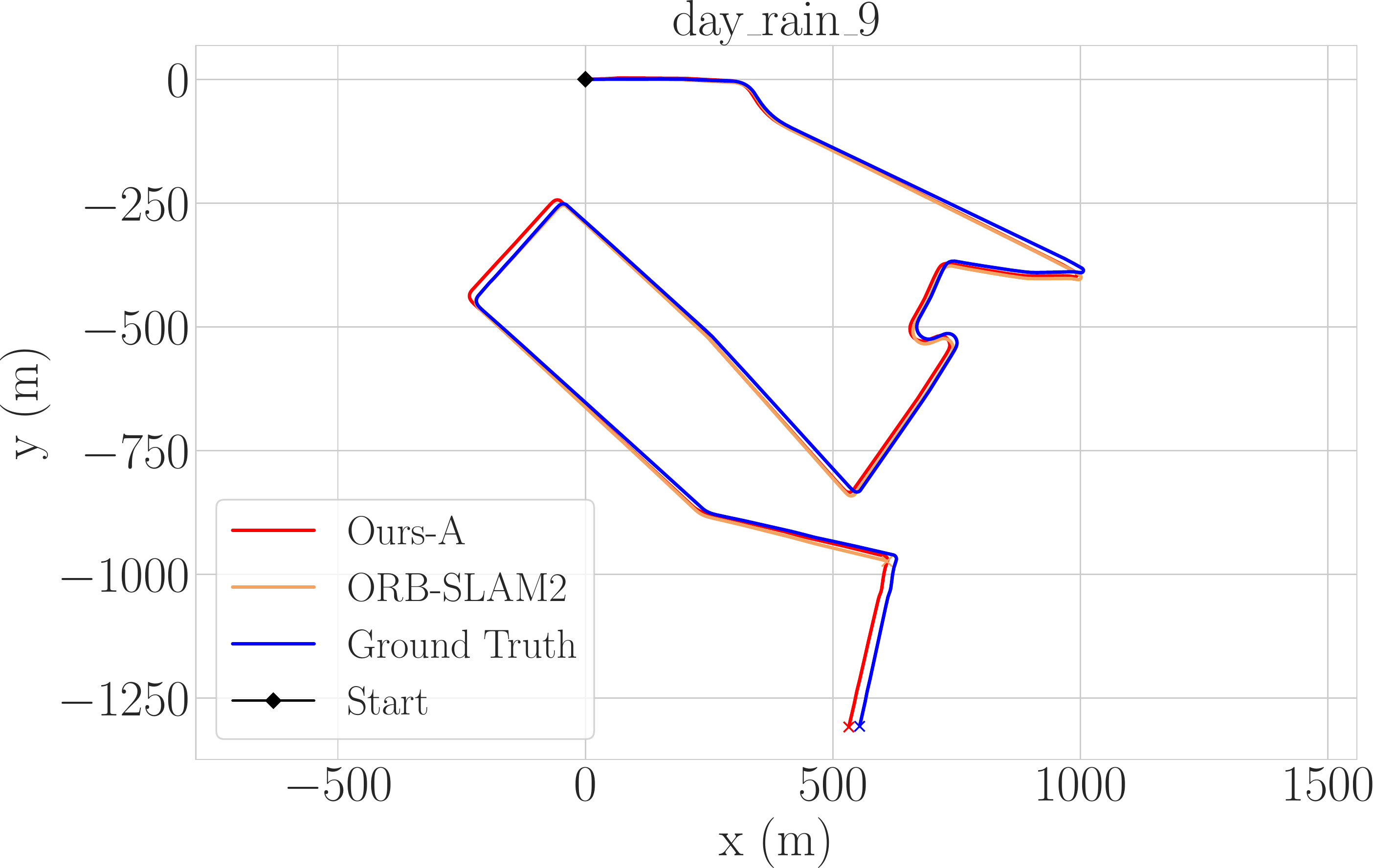}
    \includegraphics[width=0.16\linewidth]{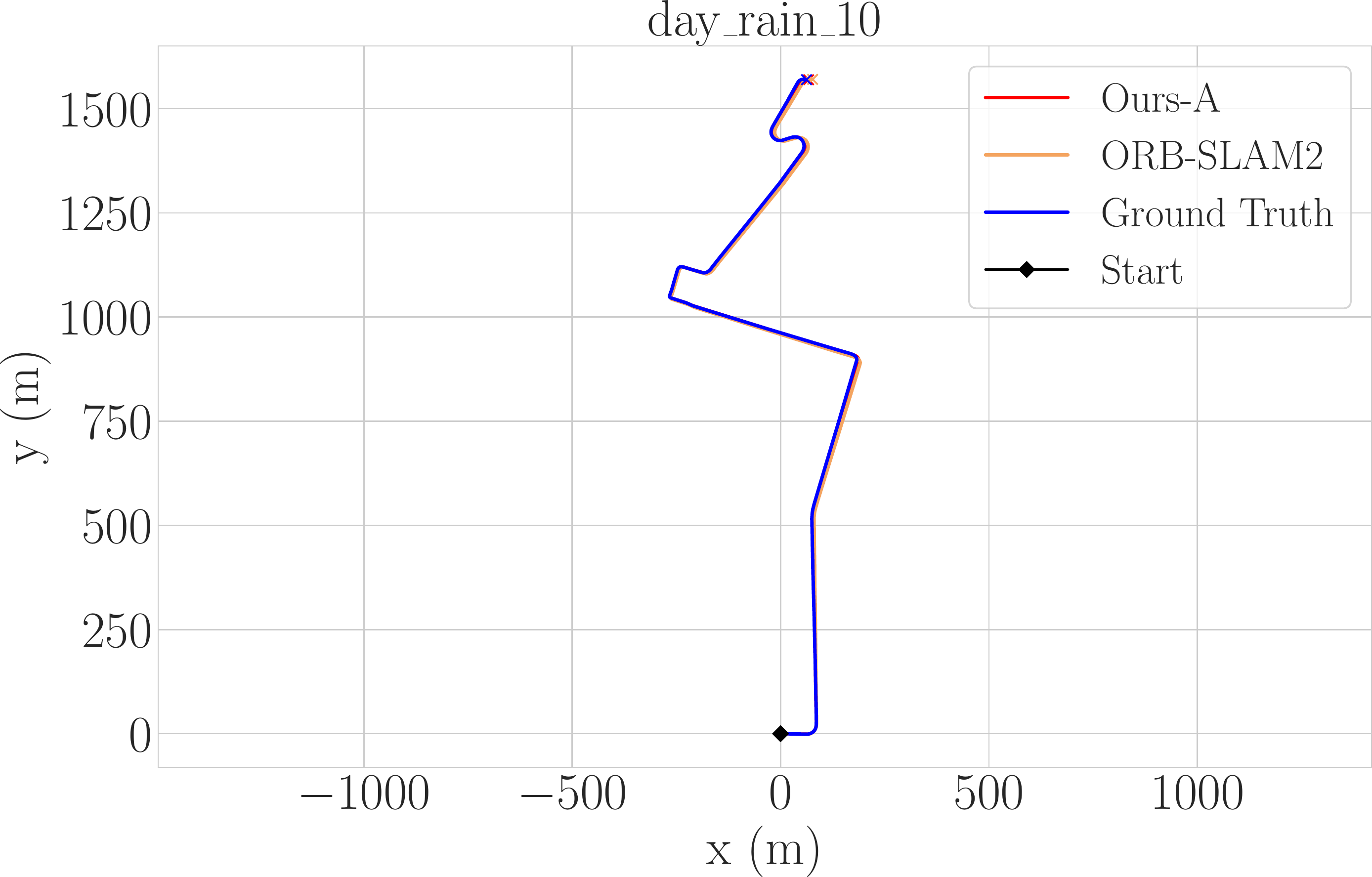}
    \includegraphics[width=0.16\linewidth]{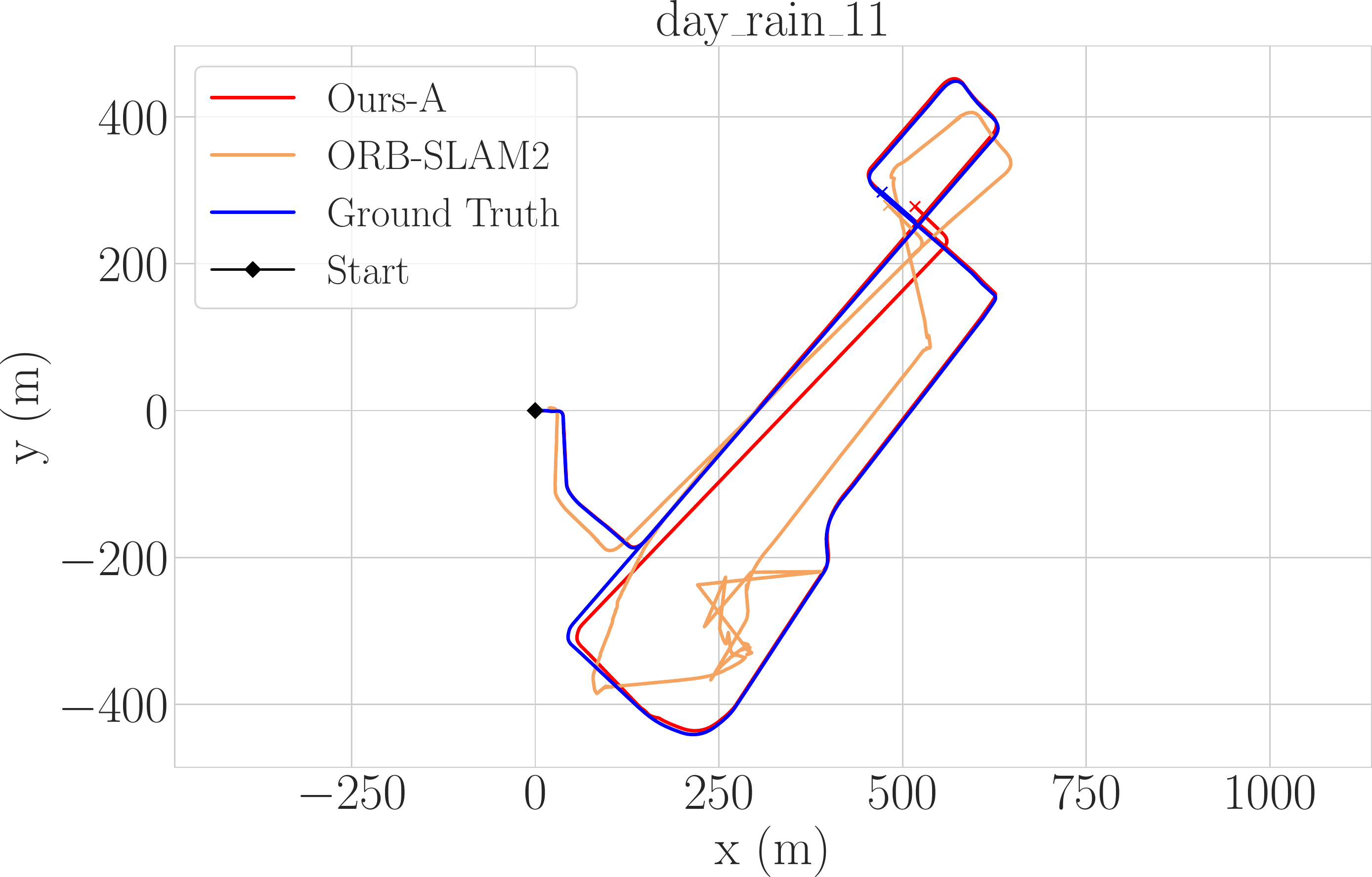}
    \includegraphics[width=0.16\linewidth]{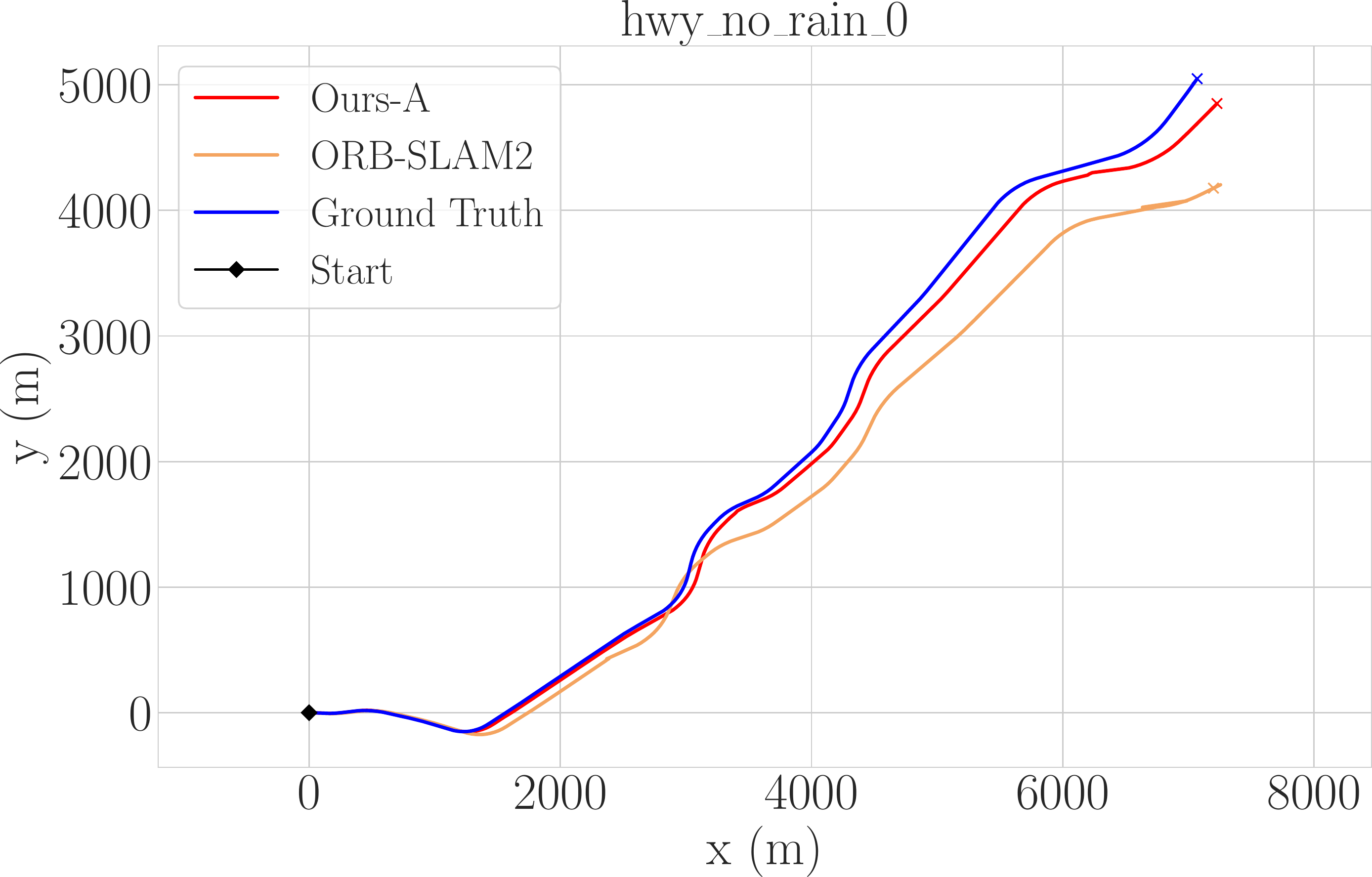}
    \includegraphics[width=0.16\linewidth]{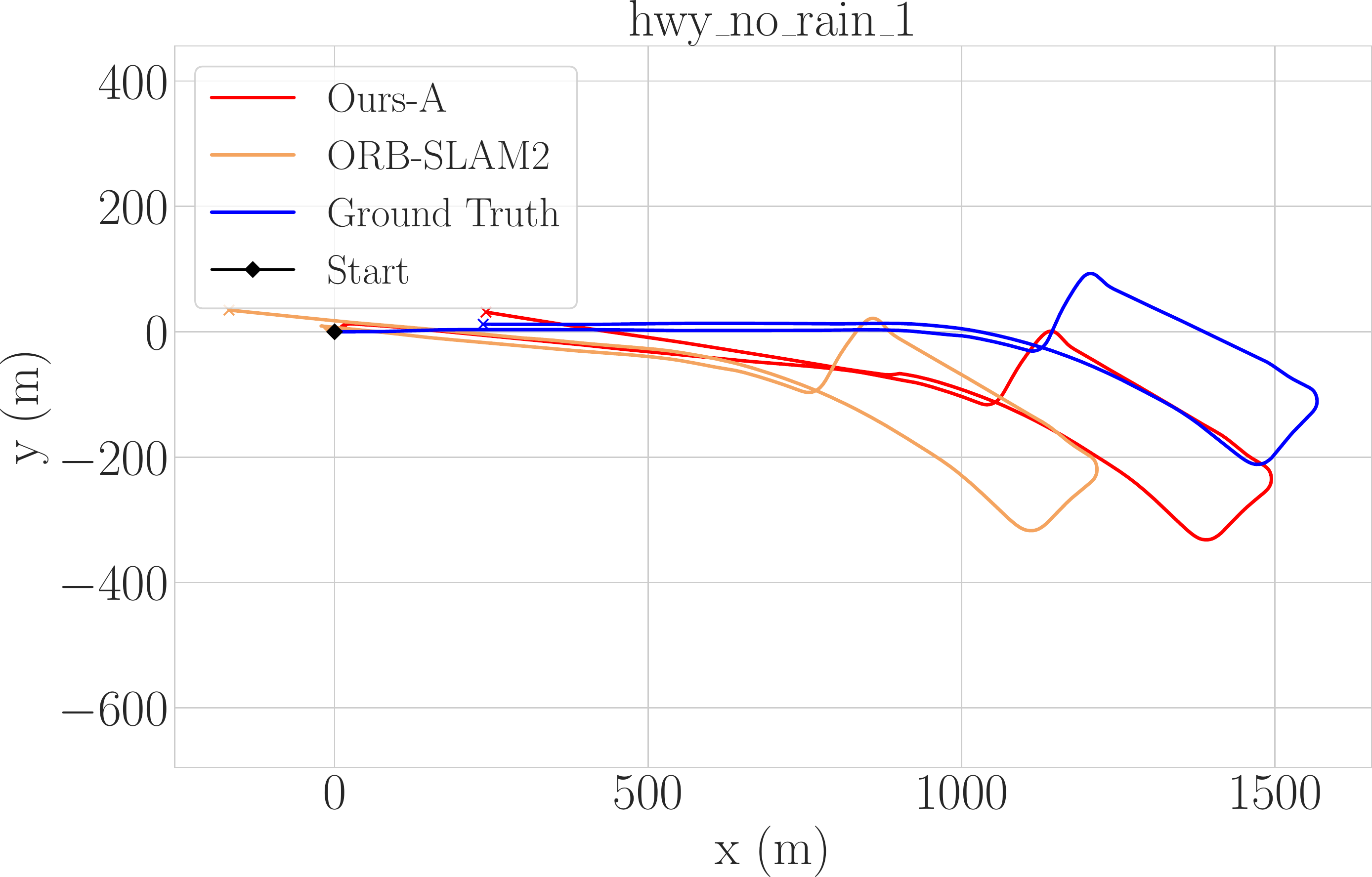}
    \includegraphics[width=0.16\linewidth]{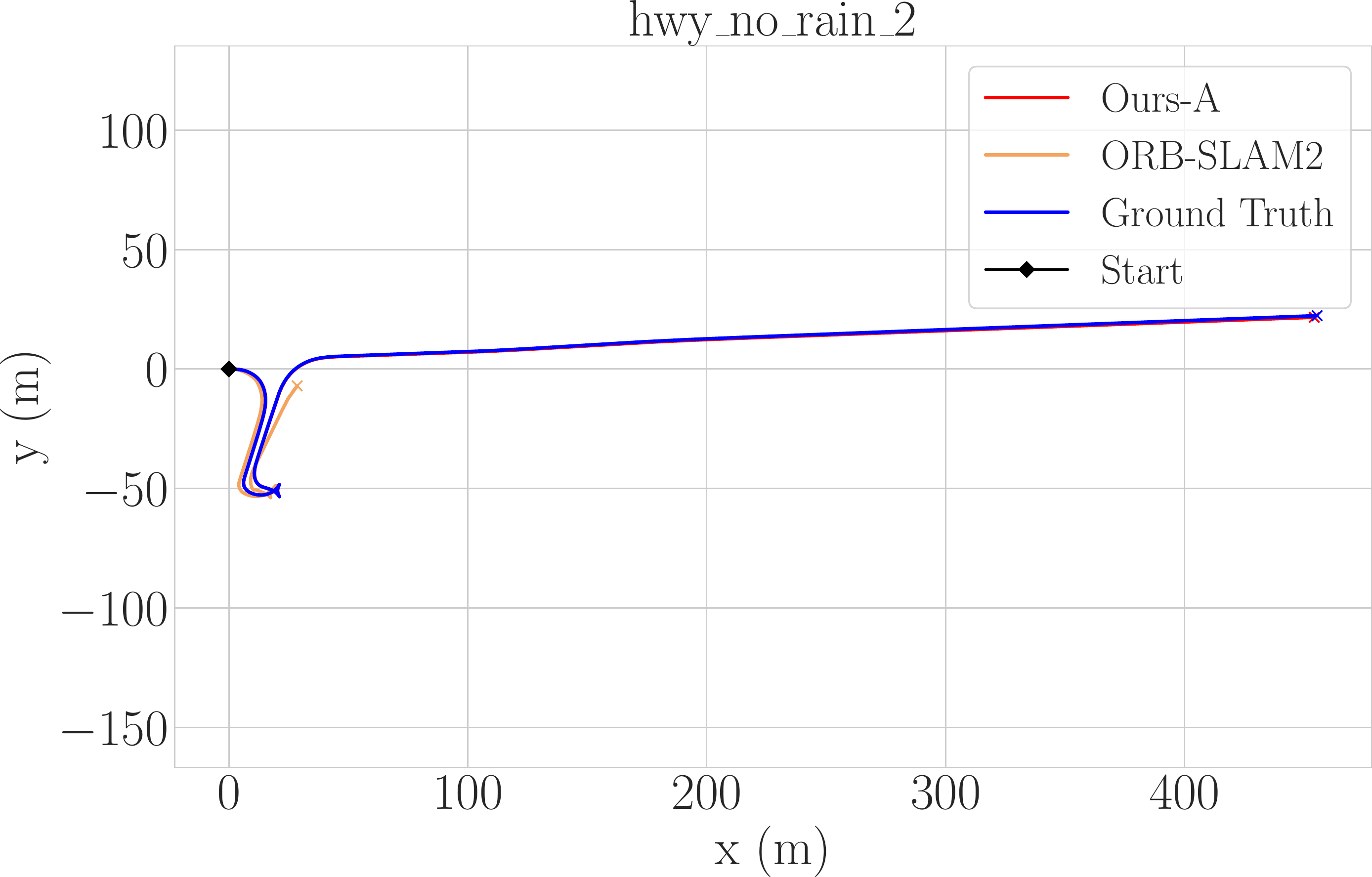}
    \includegraphics[width=0.16\linewidth]{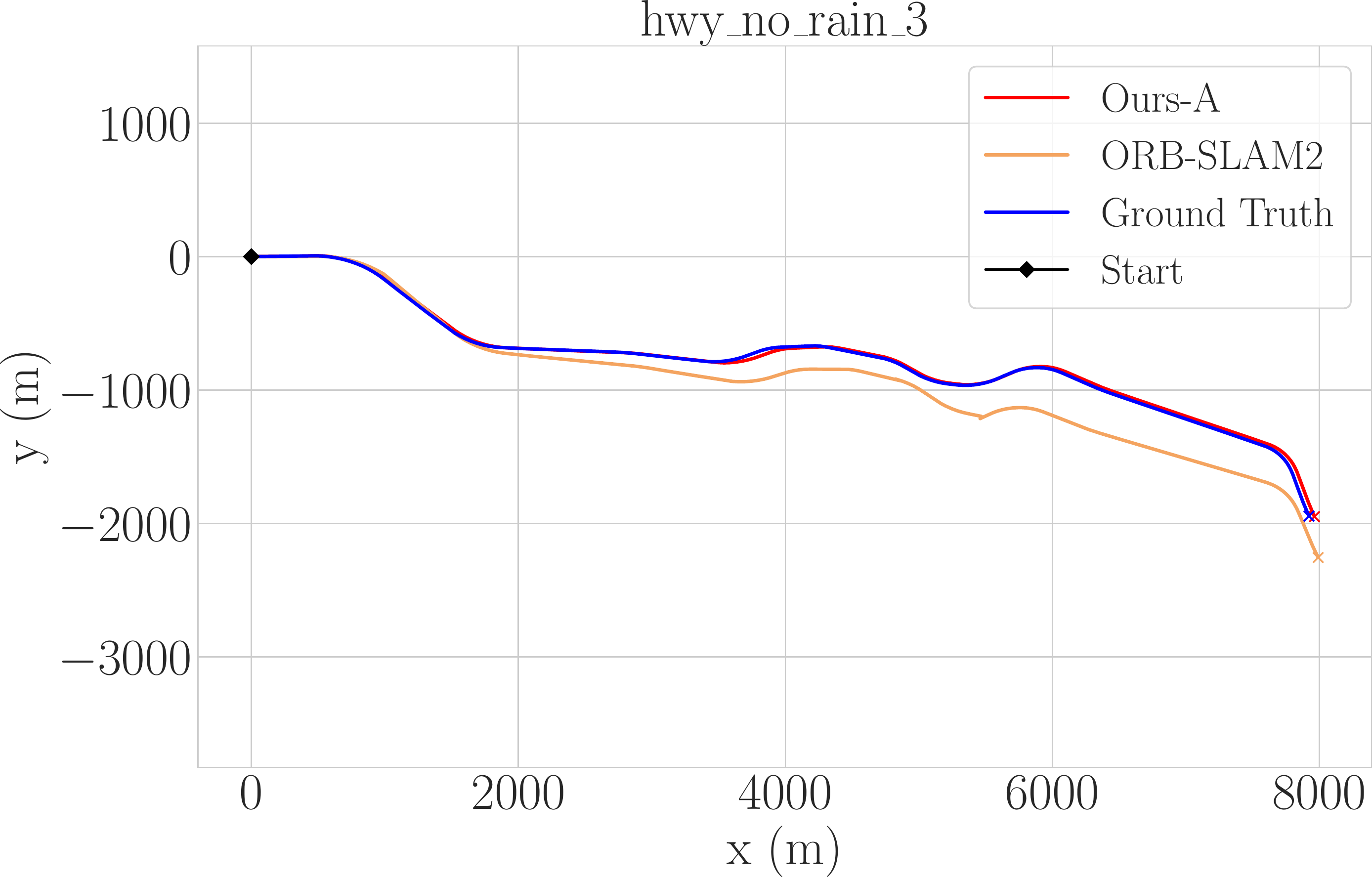}
    \includegraphics[width=0.16\linewidth]{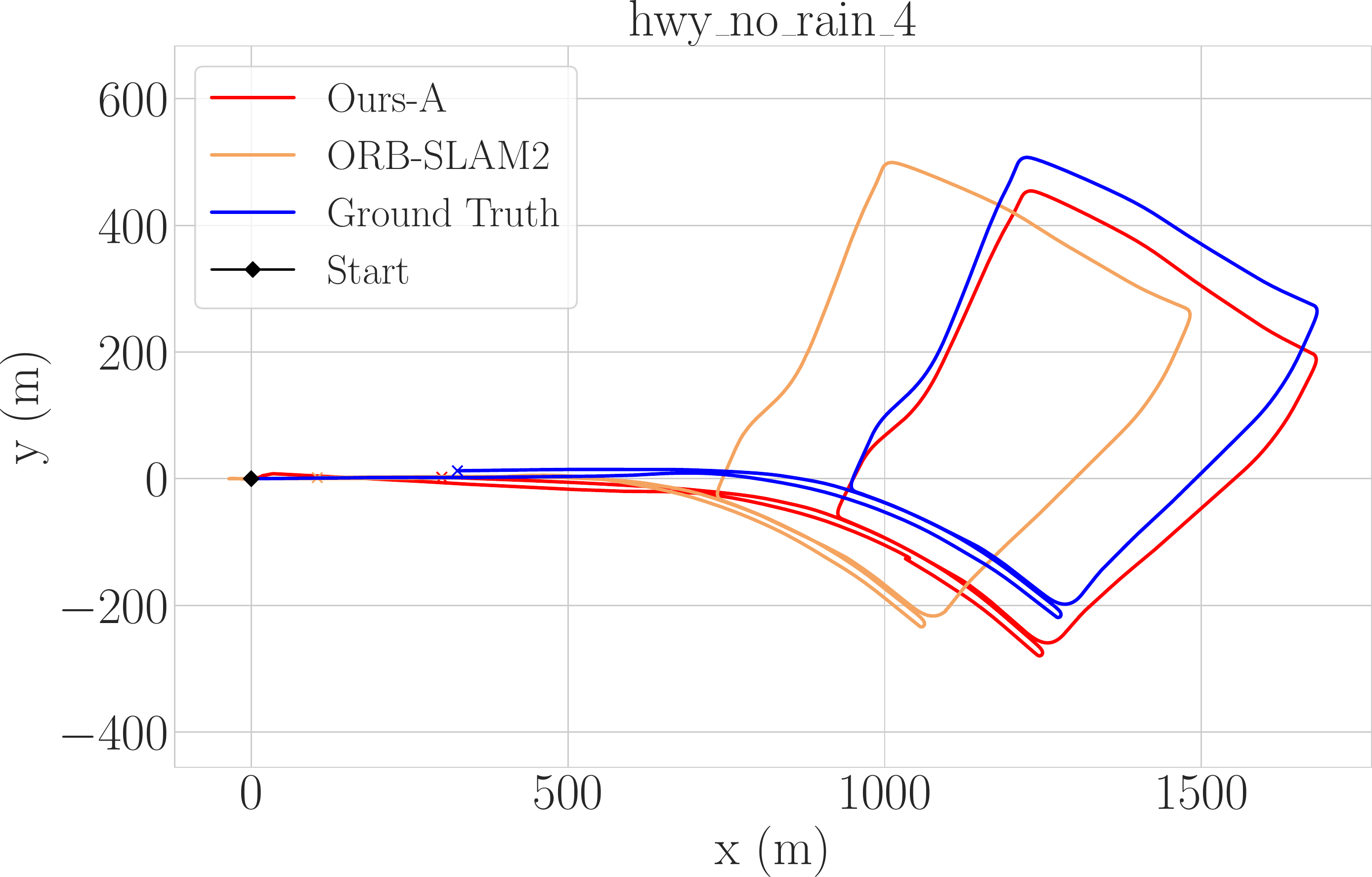}
    \includegraphics[width=0.16\linewidth]{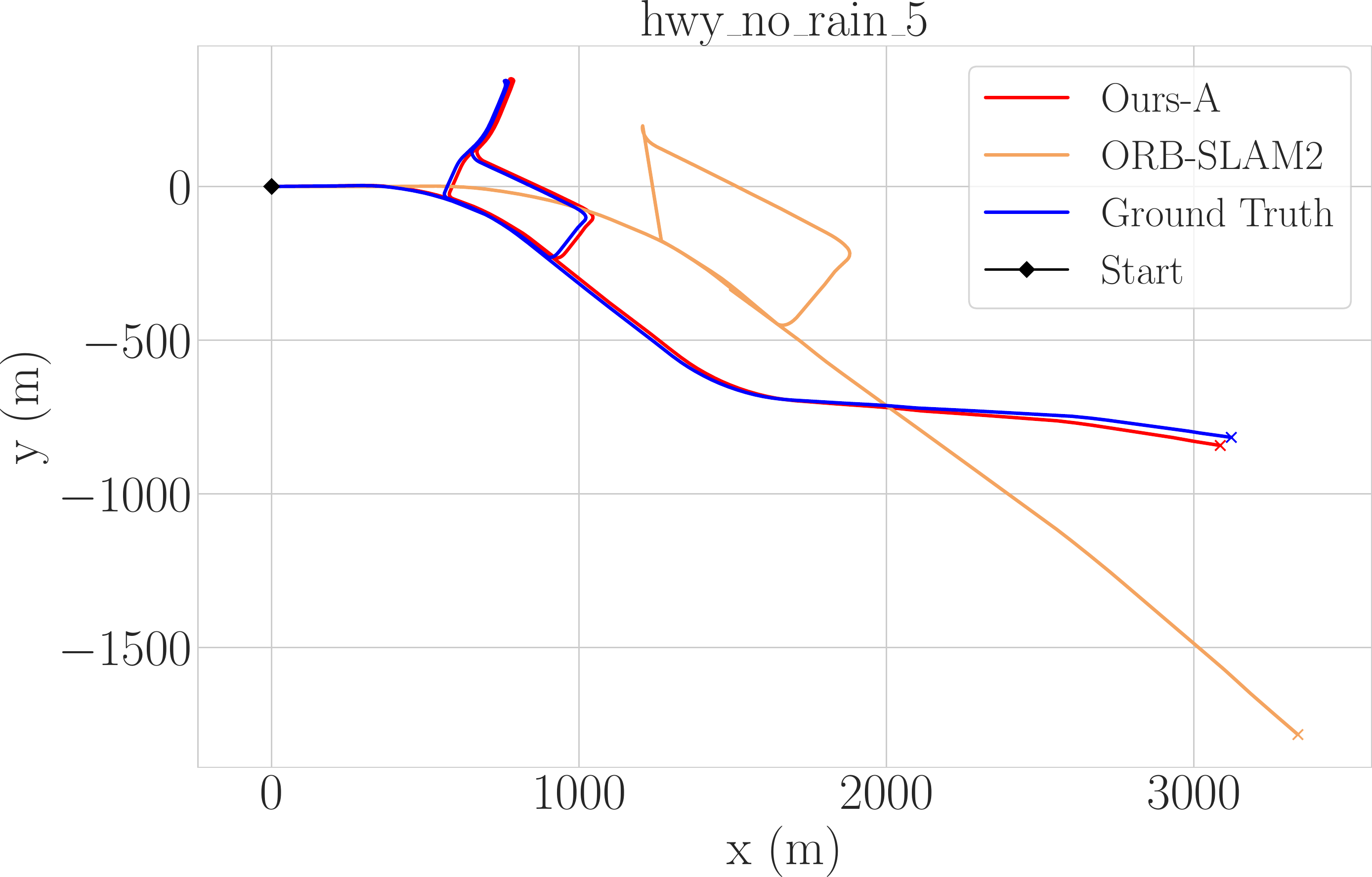}
    \includegraphics[width=0.16\linewidth]{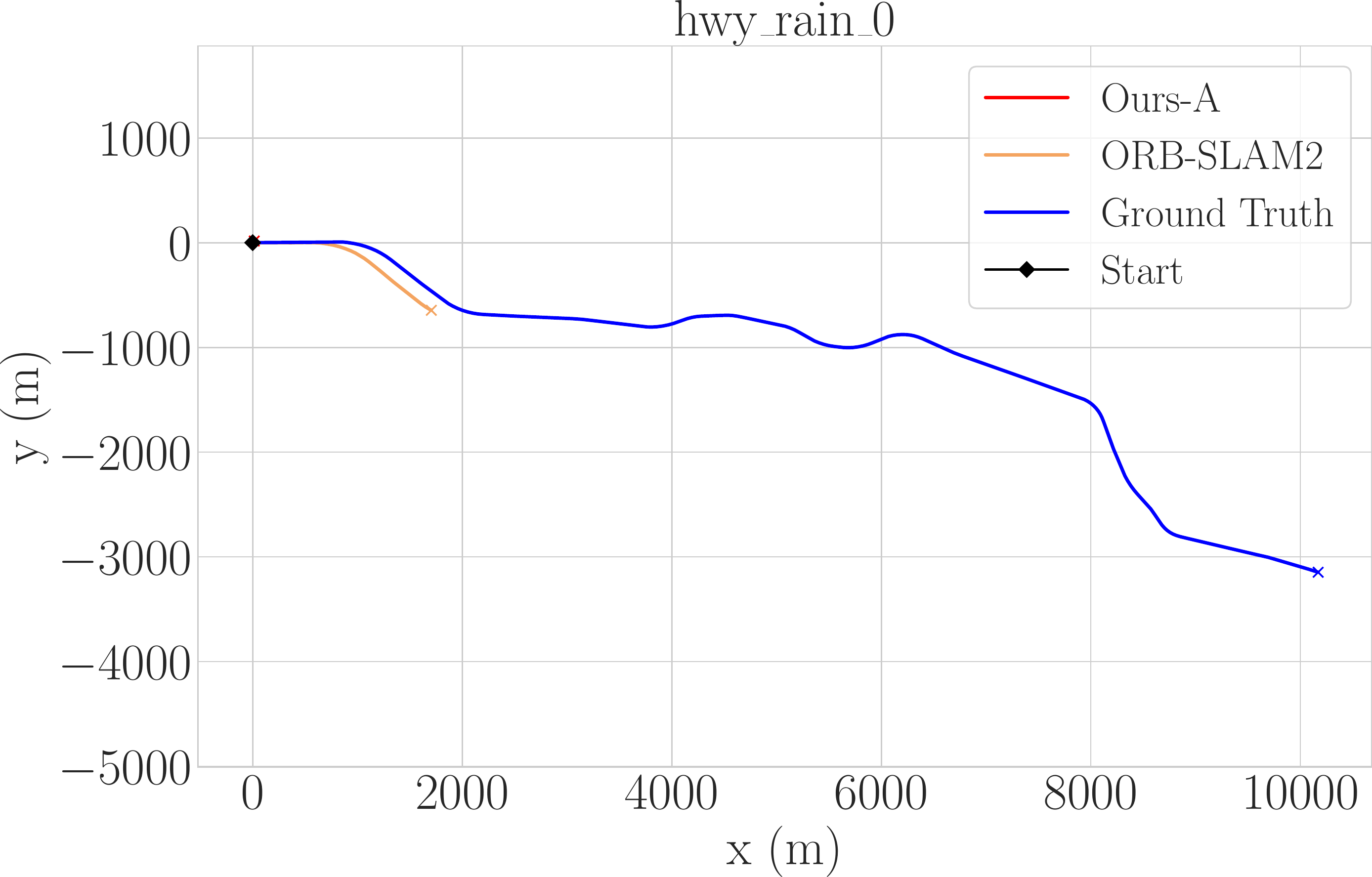}
    \includegraphics[width=0.16\linewidth]{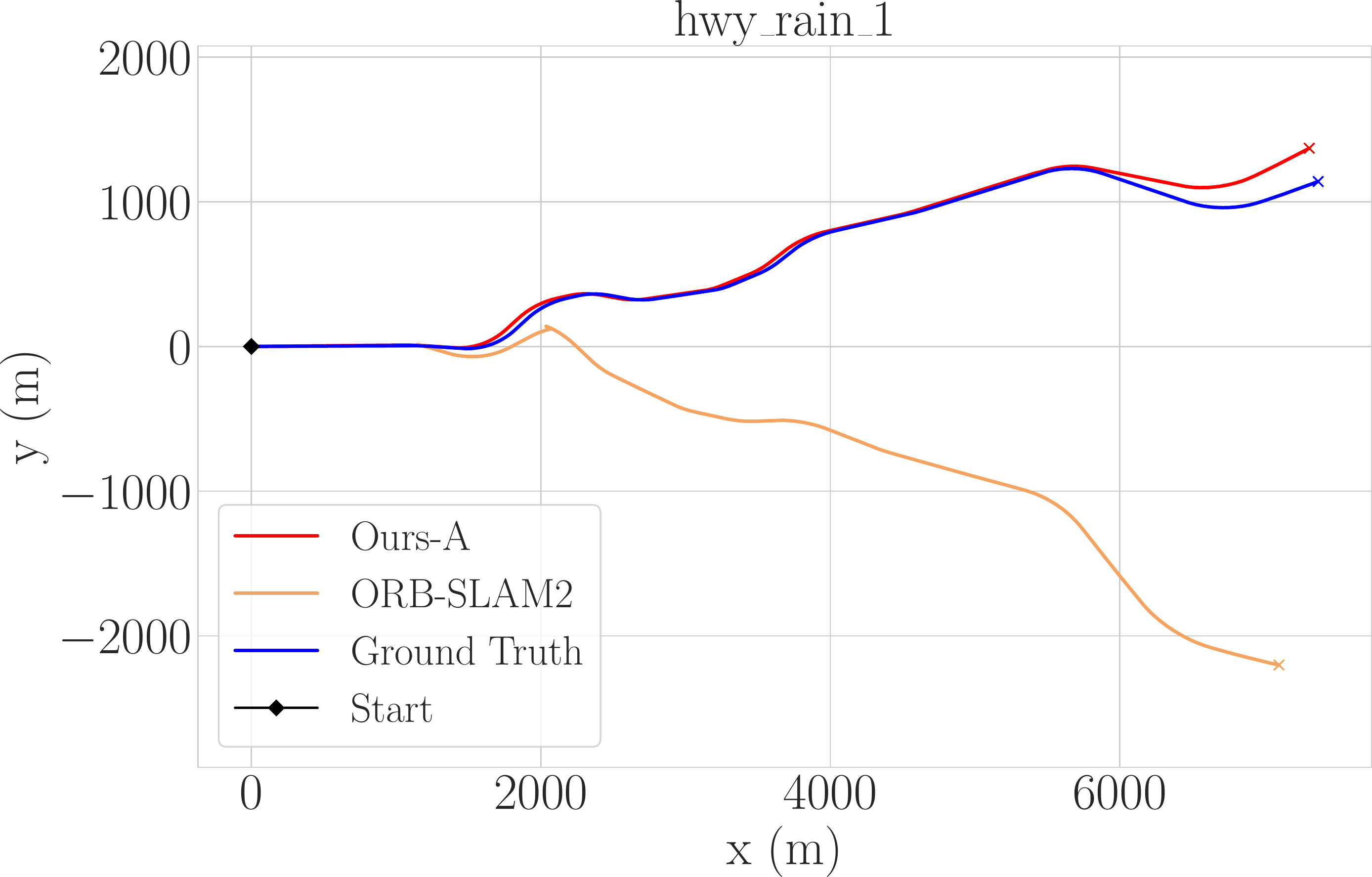}
    \includegraphics[width=0.16\linewidth]{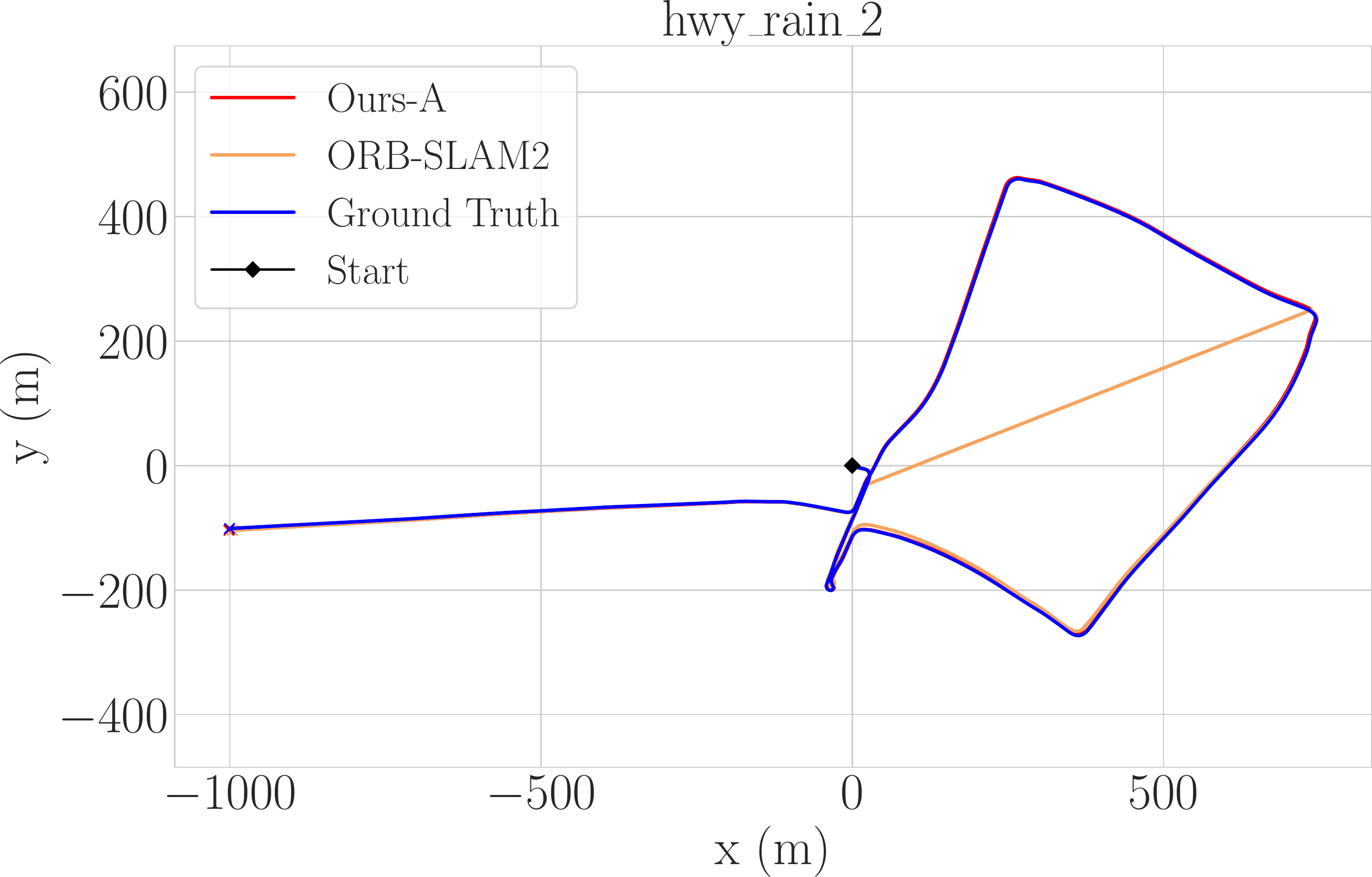}
    \includegraphics[width=0.16\linewidth]{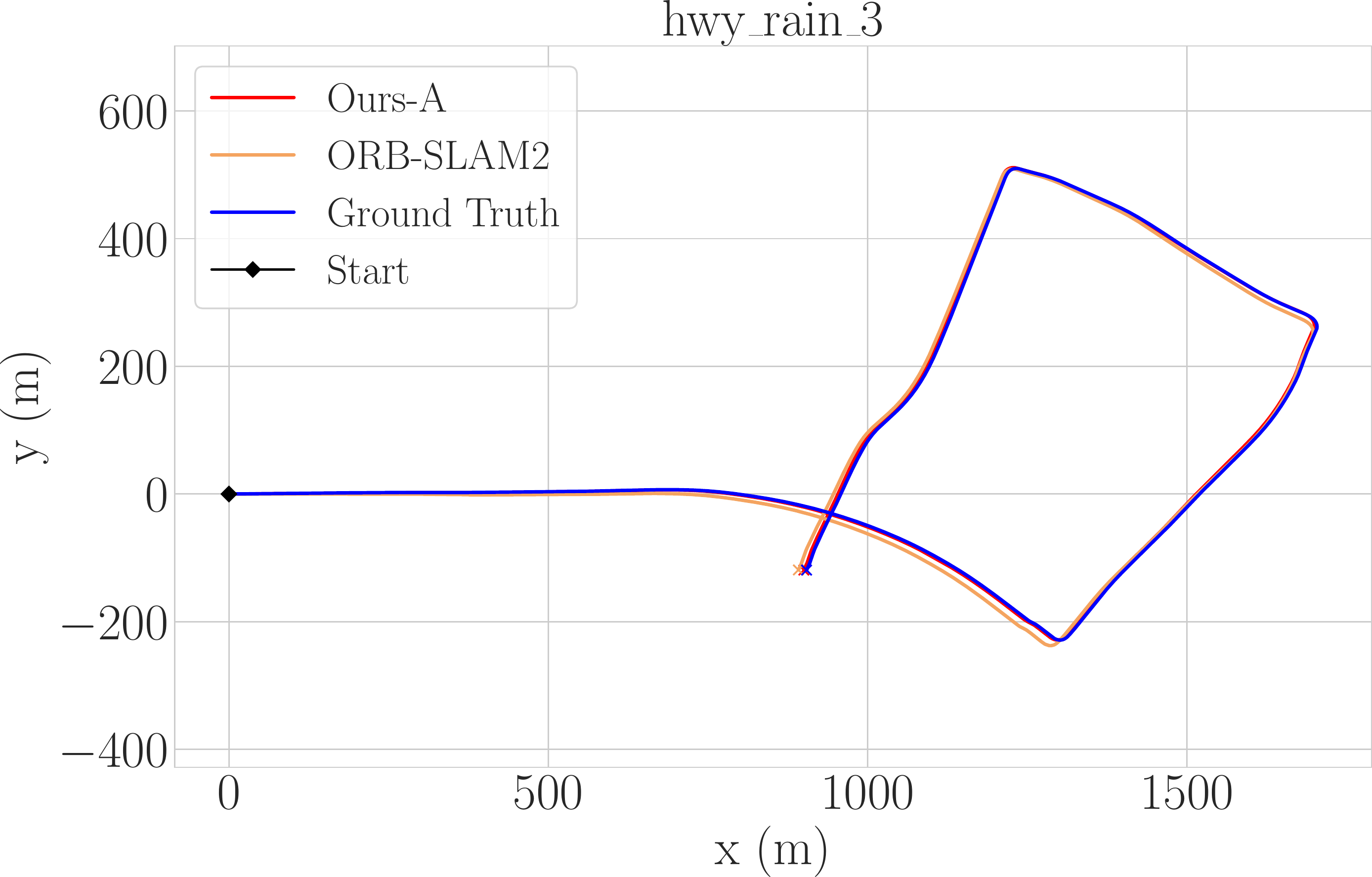}
    \includegraphics[width=0.16\linewidth]{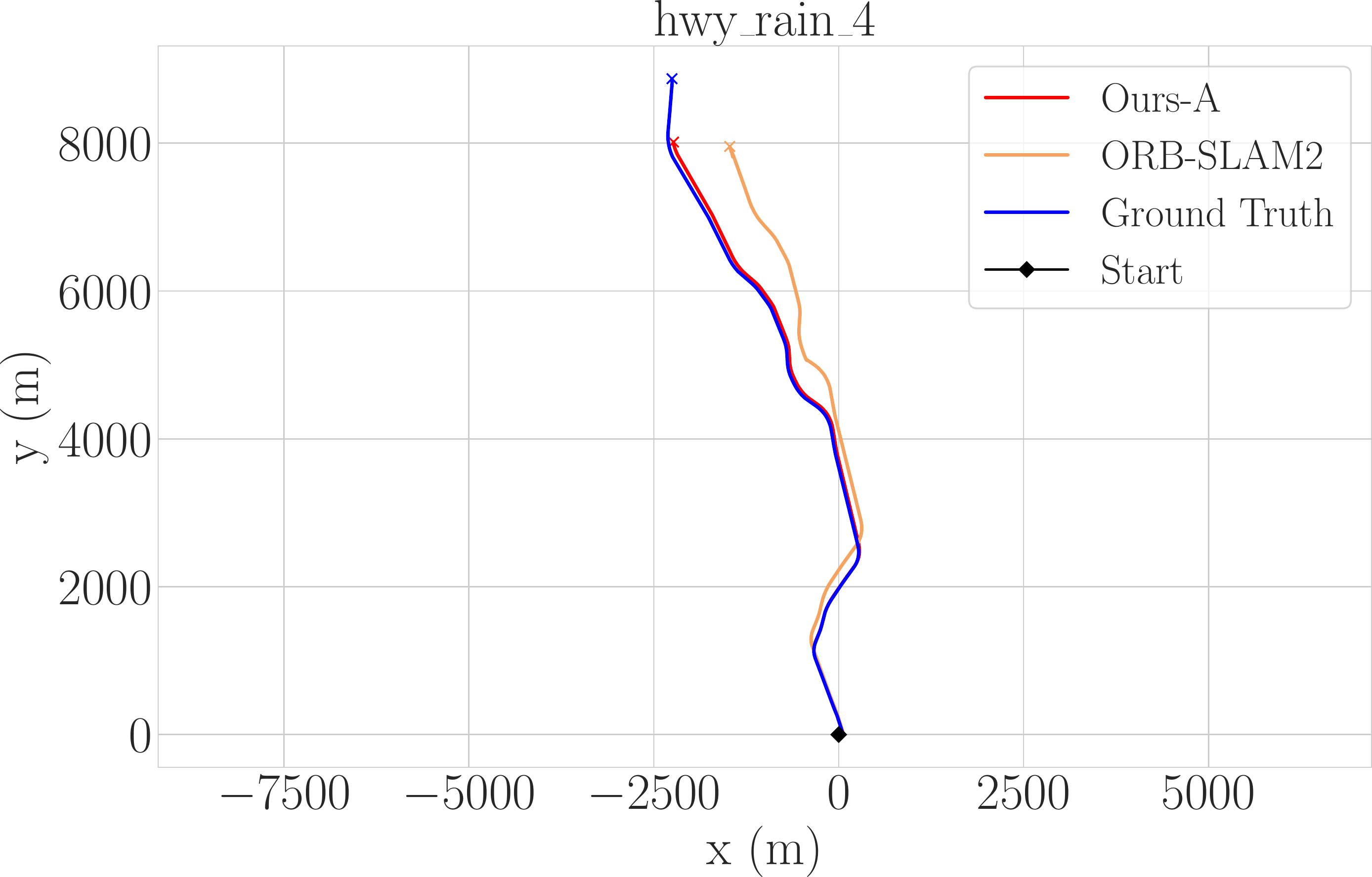}
    \includegraphics[width=0.16\linewidth]{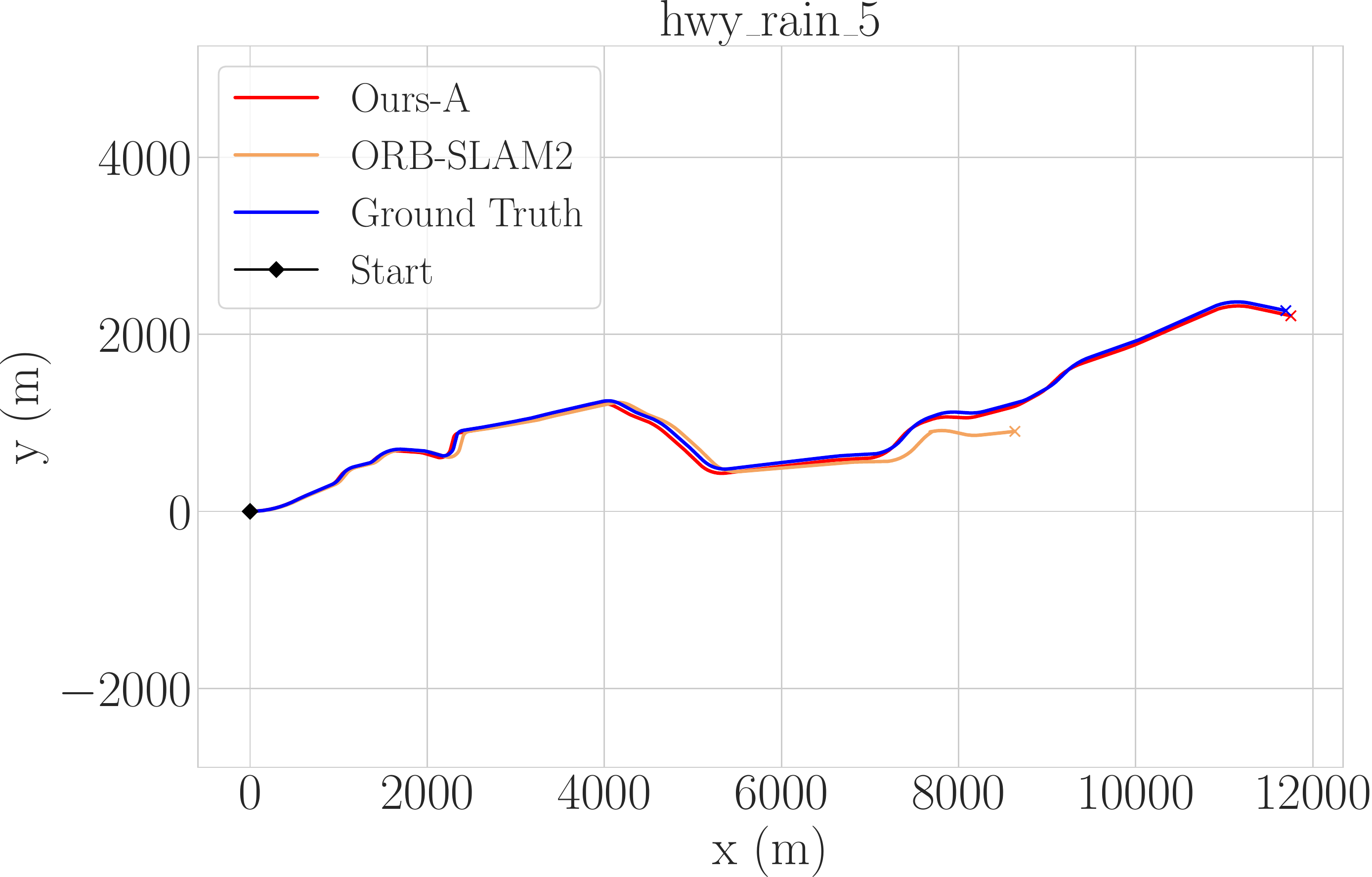}
    \includegraphics[width=0.16\linewidth]{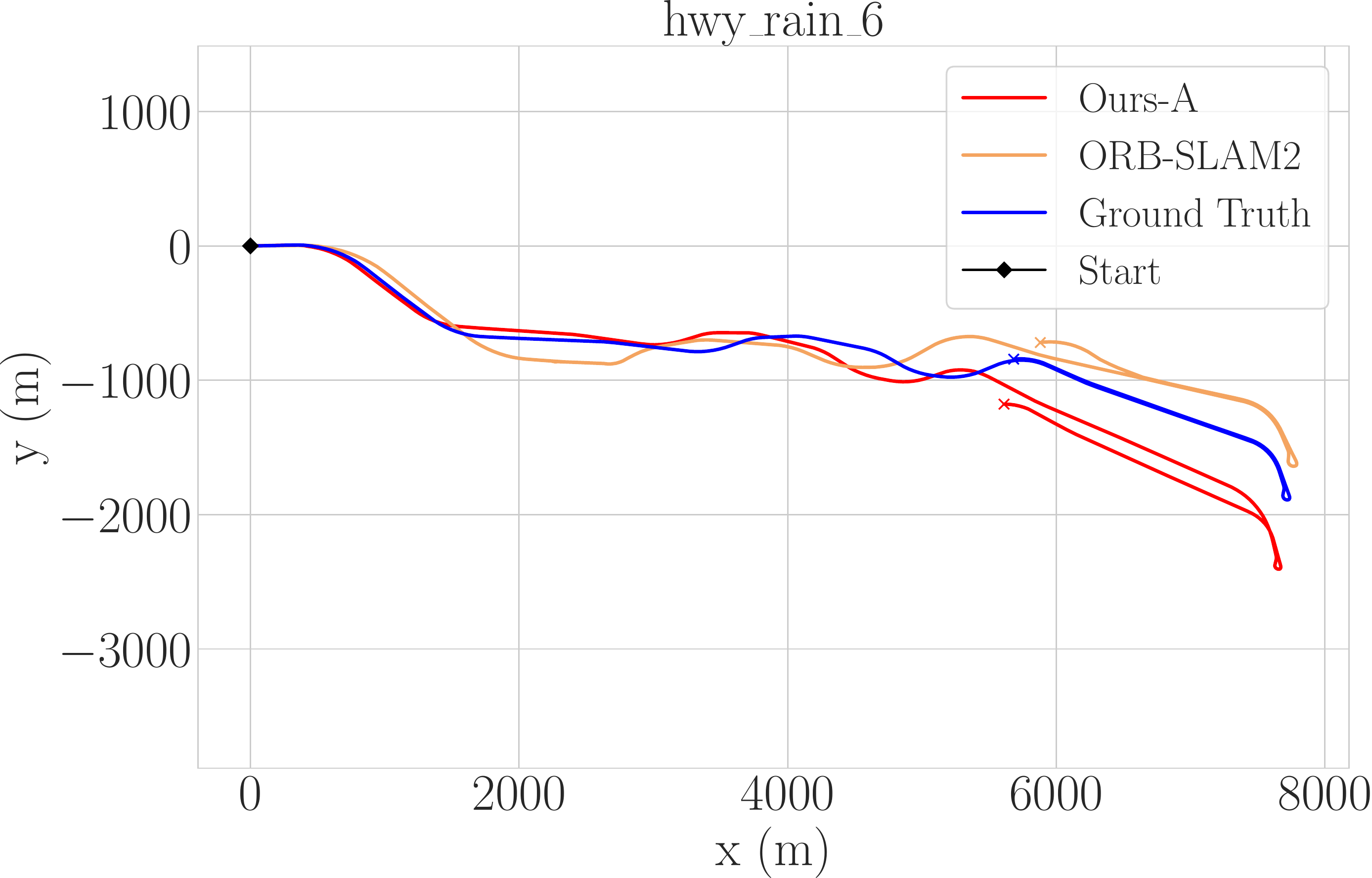}
    \includegraphics[width=0.16\linewidth]{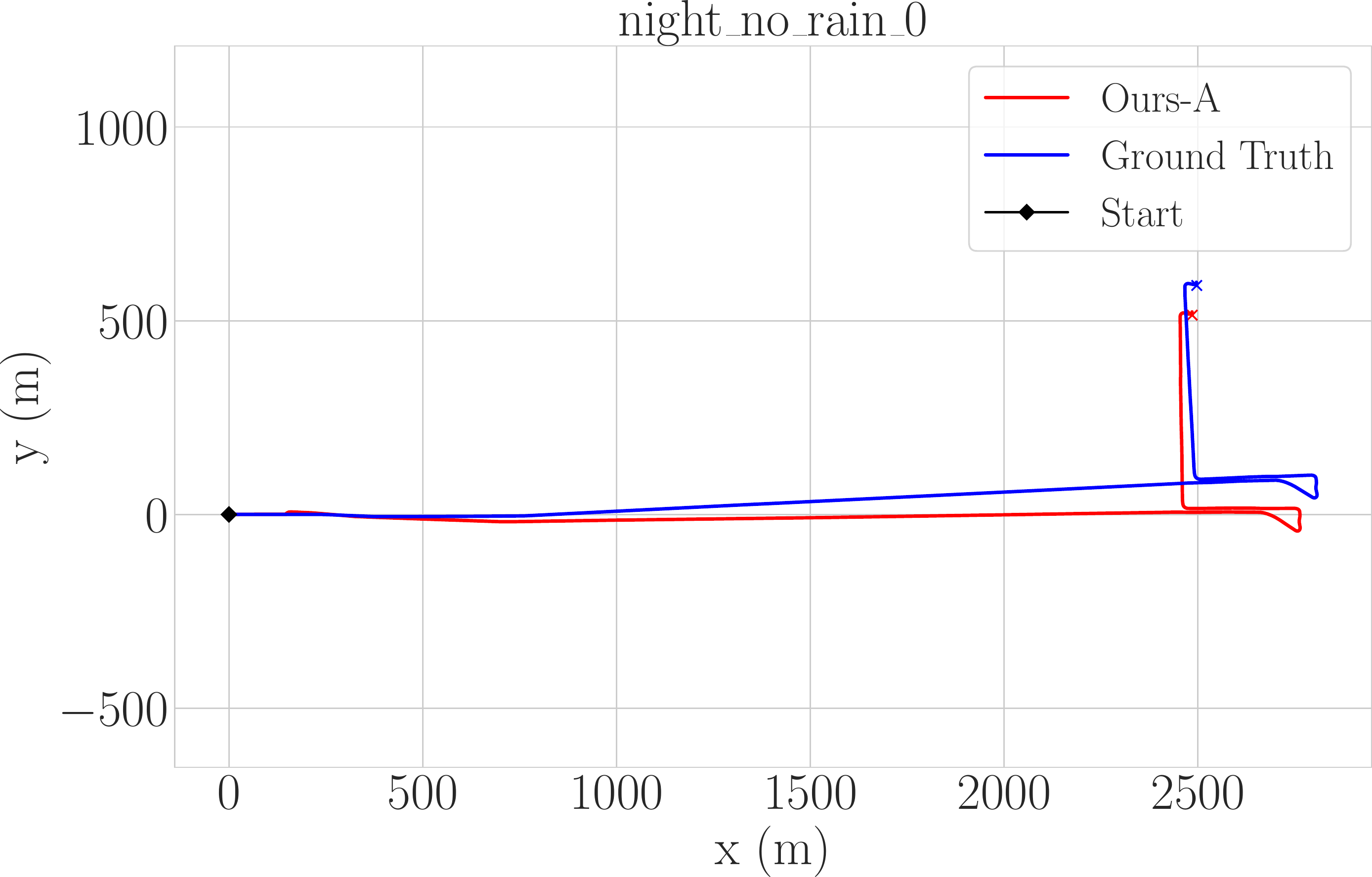}
    \includegraphics[width=0.16\linewidth]{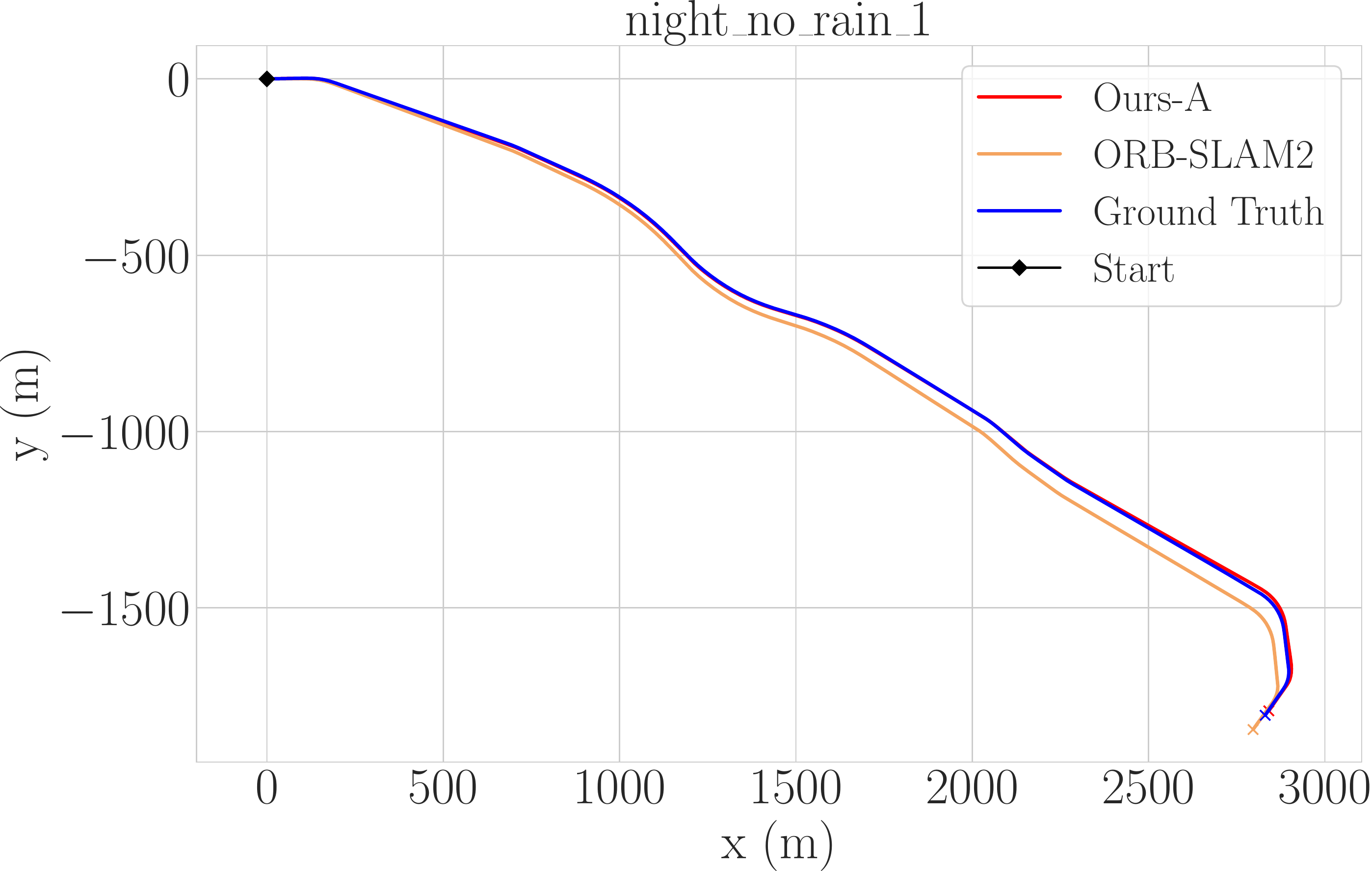}
    \includegraphics[width=0.16\linewidth]{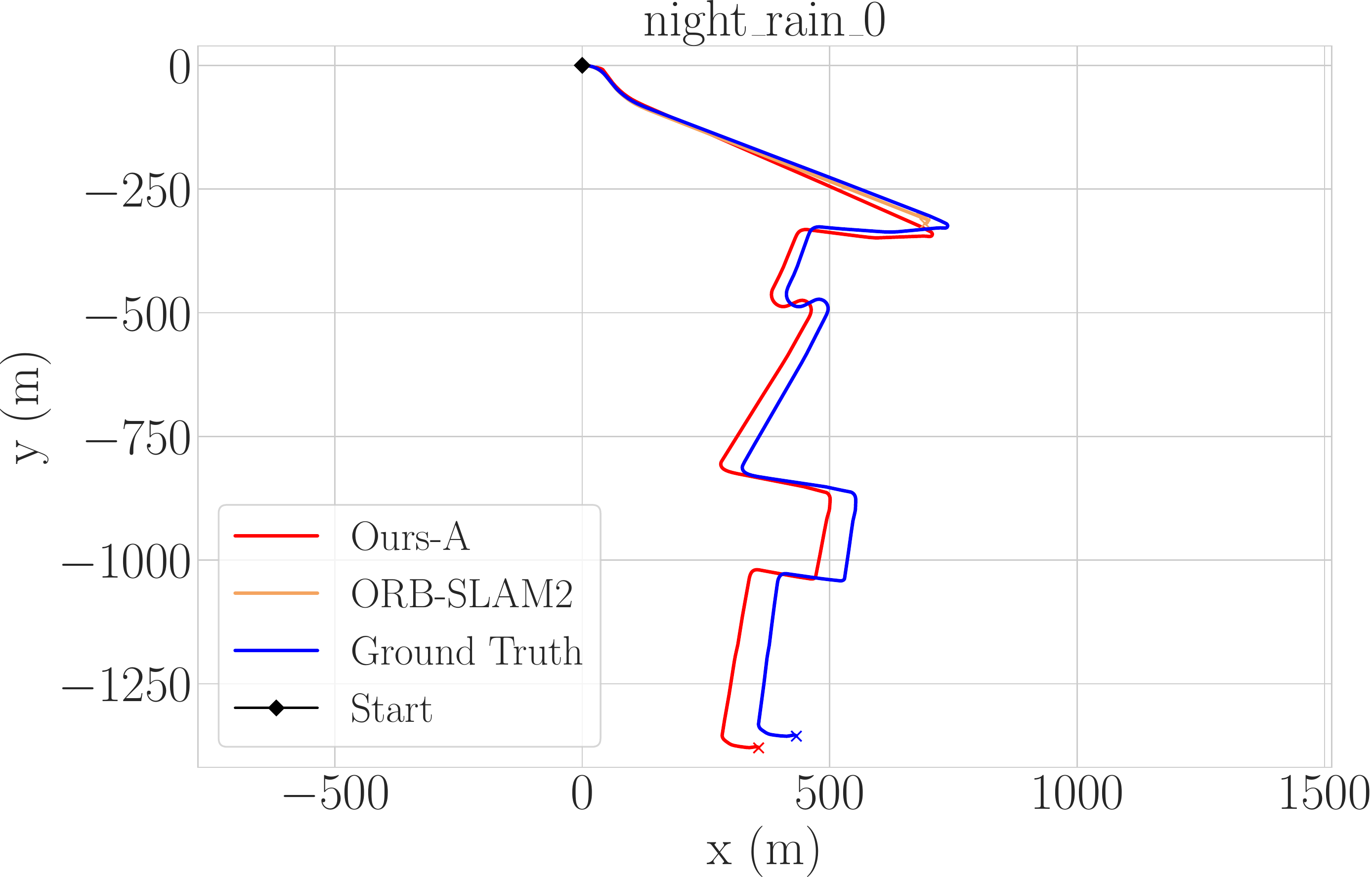}
    \caption{Estimated trajectories in all 65 train sequences, comparing
      ORB-SLAM2 (stereo) with our full system (all seven cameras).
    \label{fig:train_full_qualitative_traj}}
\end{figure*}

\begin{figure*}
    \centering
    \includegraphics[width=0.32\linewidth]{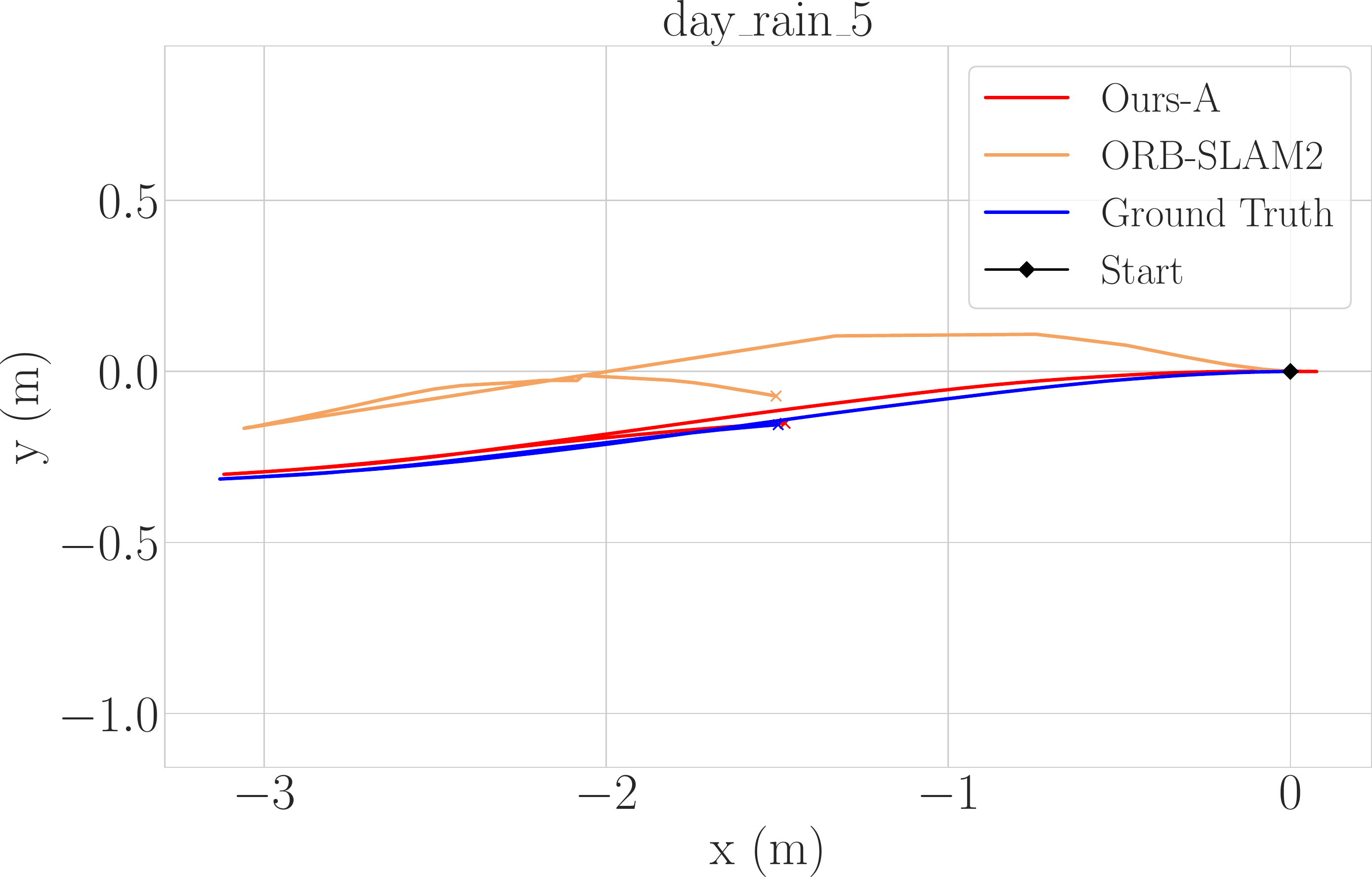}
    \includegraphics[width=0.32\linewidth]{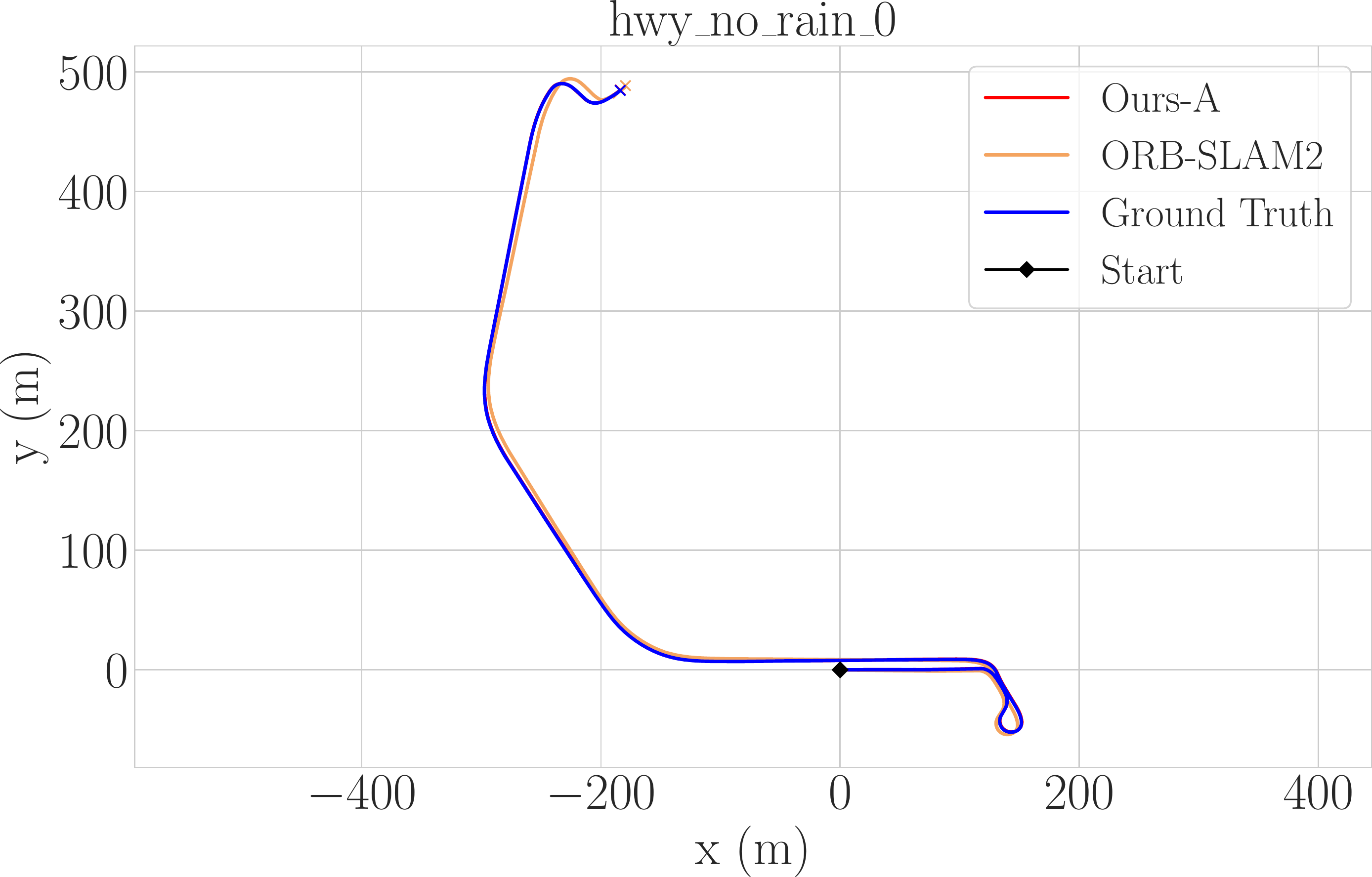}
    \includegraphics[width=0.32\linewidth]{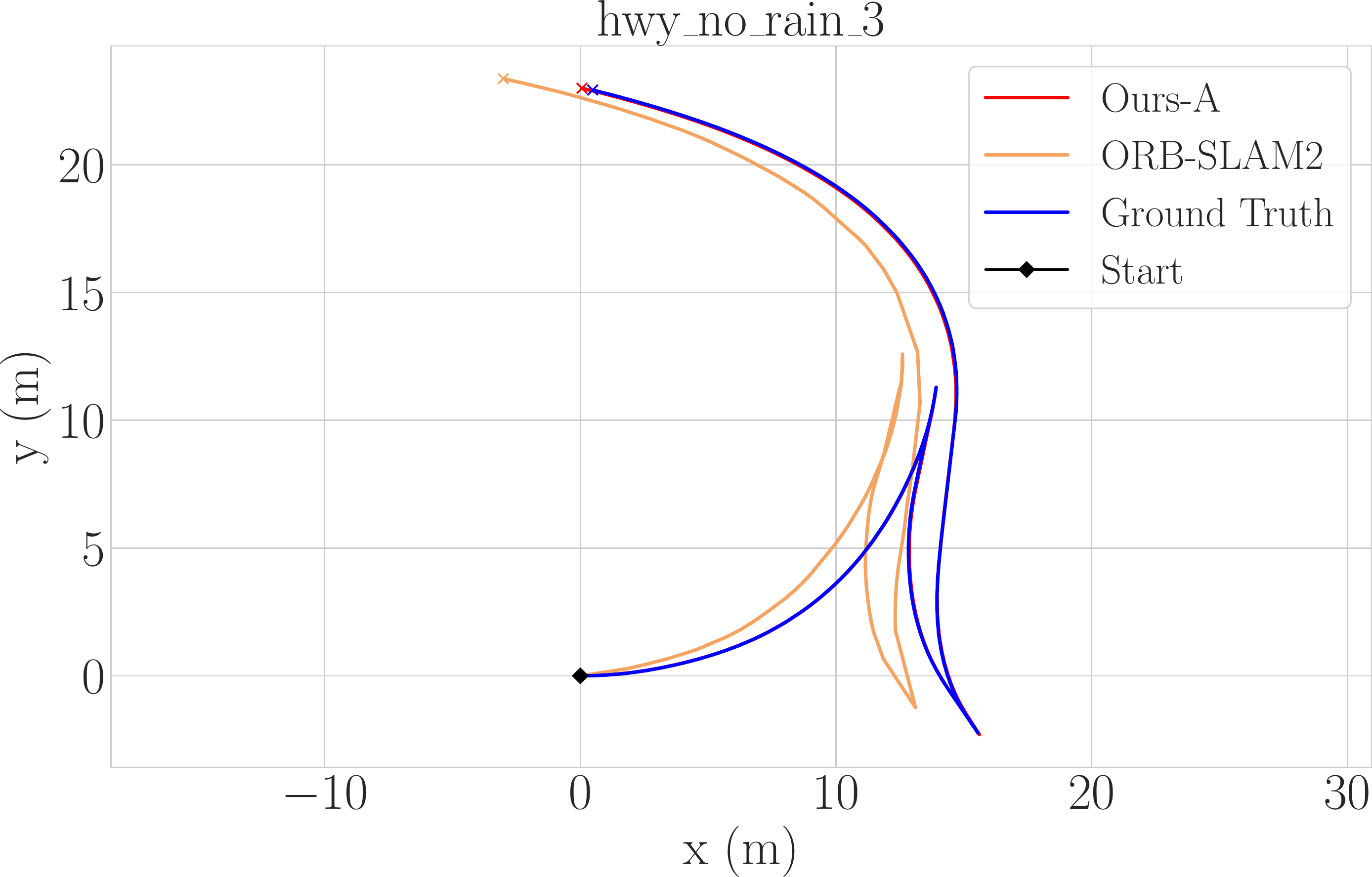}
    \caption{Example maneuvers in the validation set, comparing ORB-SLAM2 and our full system with all
    7 cameras. (Left) Reversing into a parallel parking spot.
    (Middle \& Right) Maneuvers in a parking lot.
    \label{fig:full_qualitative_maneuver}}
    \vspace{-3mm}
\end{figure*}

\begin{figure*}
    \centering
    \includegraphics[width=0.24\linewidth]{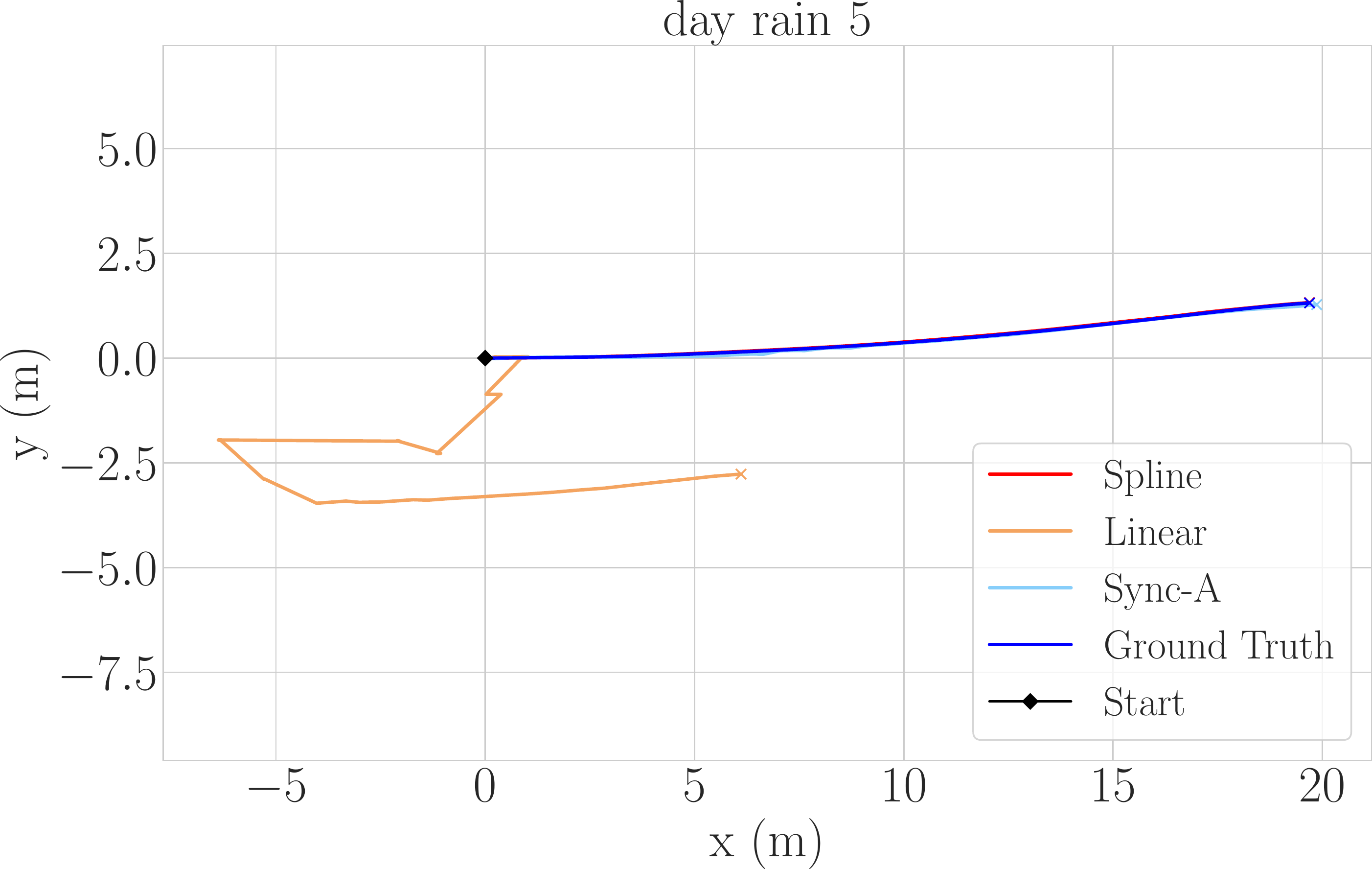}
    \includegraphics[width=0.24\linewidth]{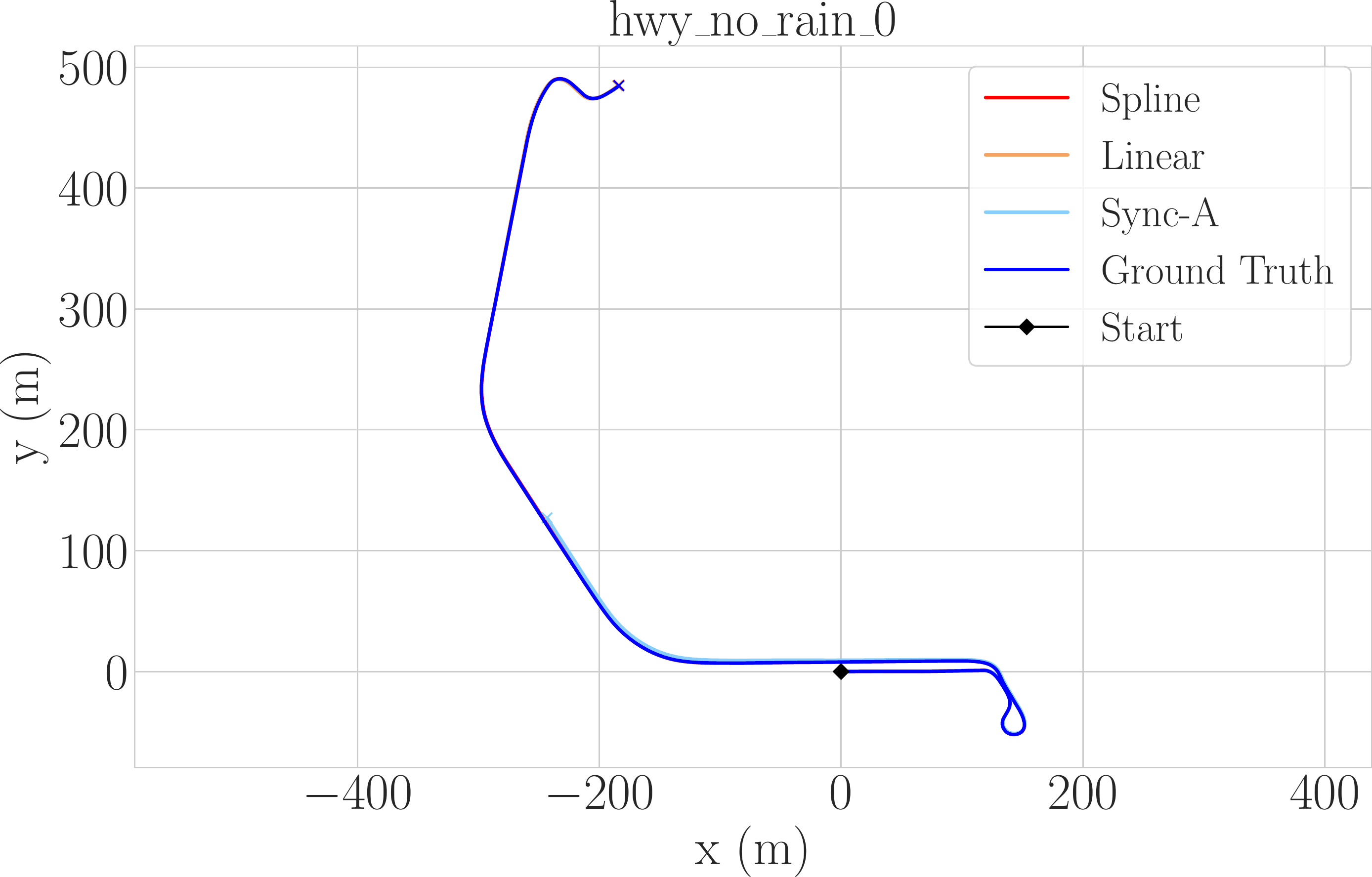}
    \includegraphics[width=0.24\linewidth]{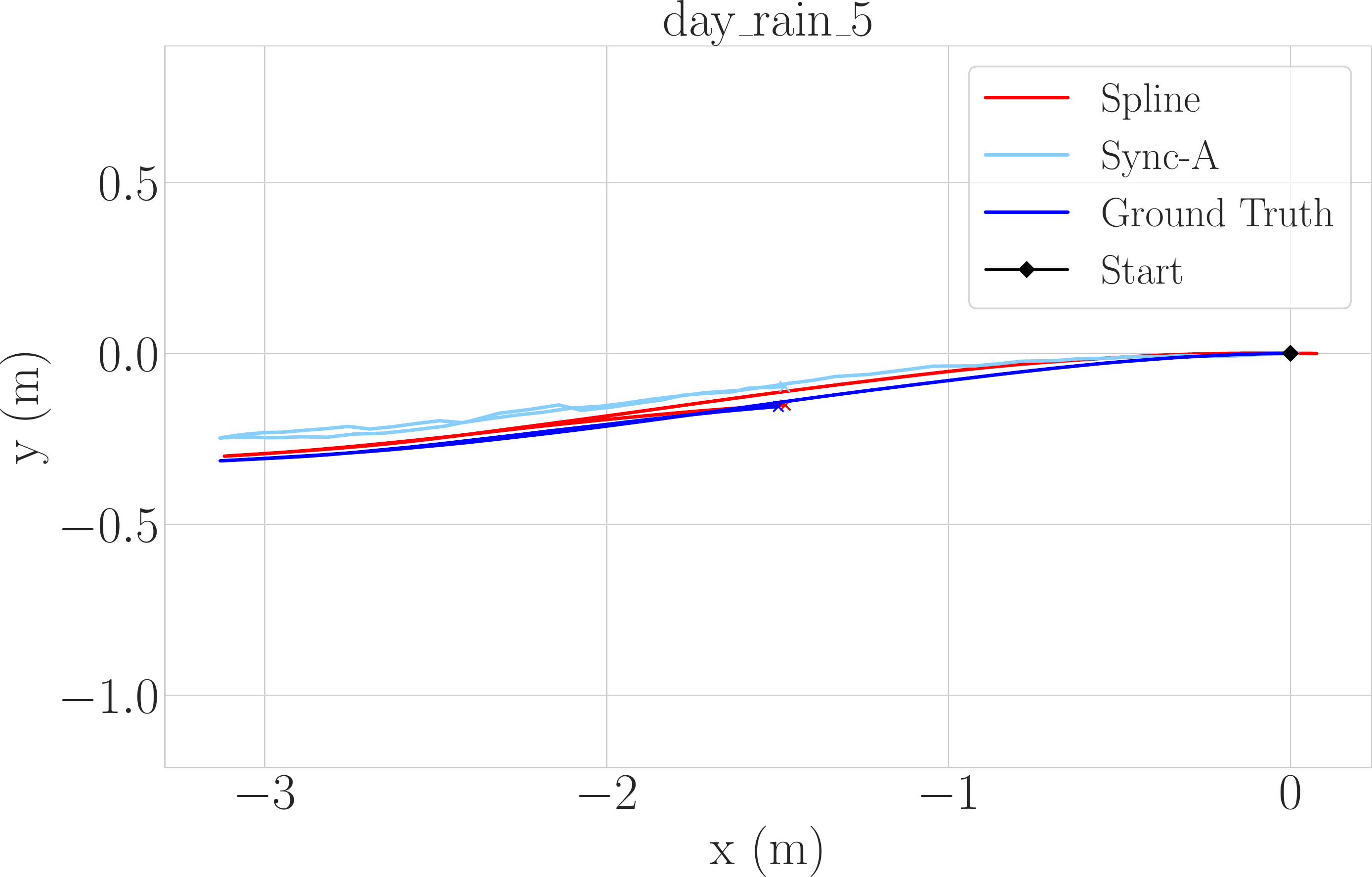}
    \includegraphics[width=0.24\linewidth]{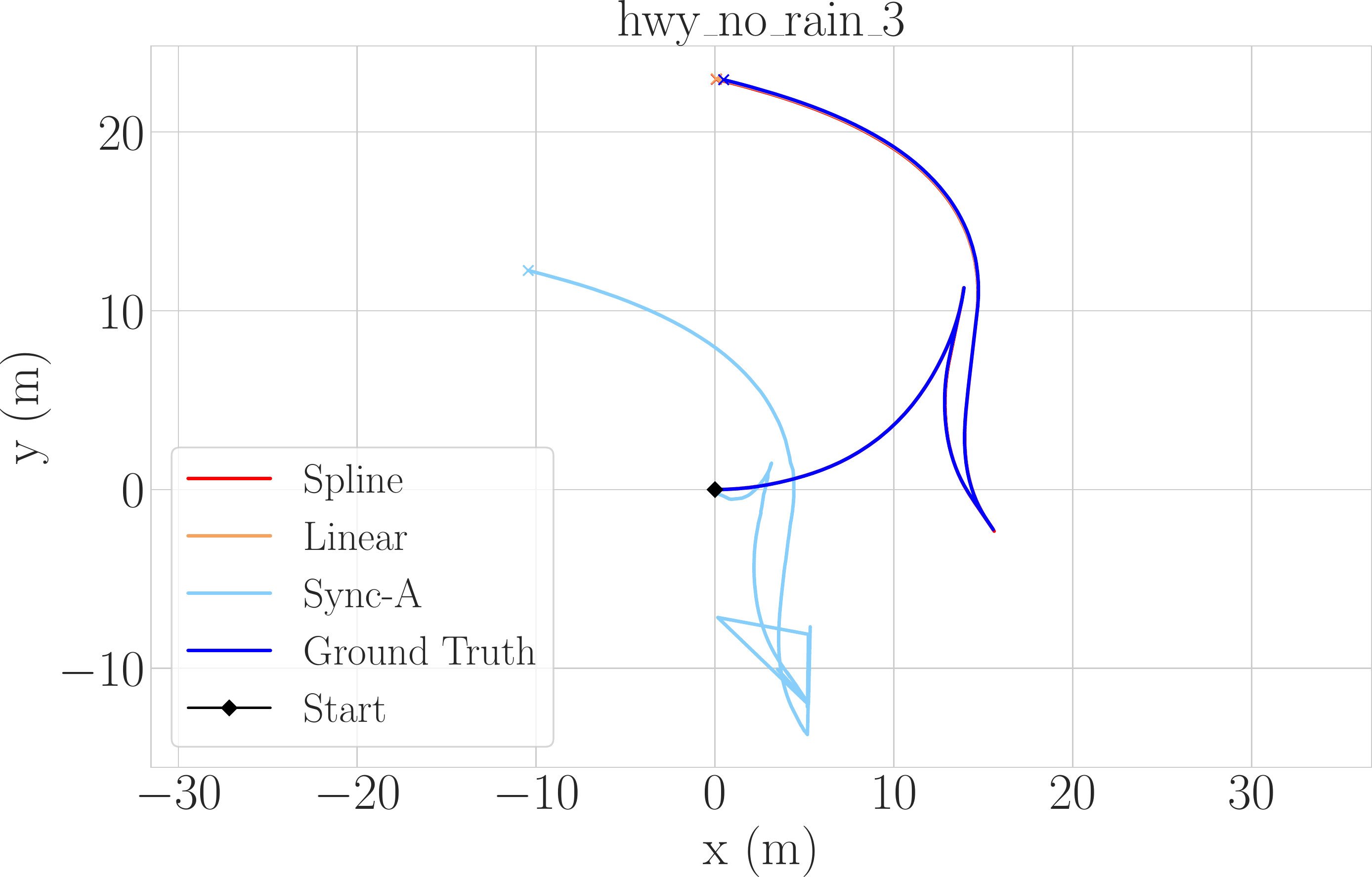}
    \caption{Motion model ablation trajectories comparing our asynchronous cubic B-spline model,
    an asynchronous linear motion model, and a discrete-time motion model
    falsely assuming
    all cameras are synchronous. (Leftmost) Zoomed-in view on a segment where the linear motion model
    failed. The vehicle was at an intersection with many dynamic objects.
    (Right) Maneuvers in the validation set. The linear motion model is missing in the
    middle sequence because it failed before reaching the maneuver.
    \label{fig:motion_model_qualitative}}
    \vspace{-3mm}
\end{figure*}

\begin{figure*}
    \centering
    \includegraphics[width=0.24\linewidth]{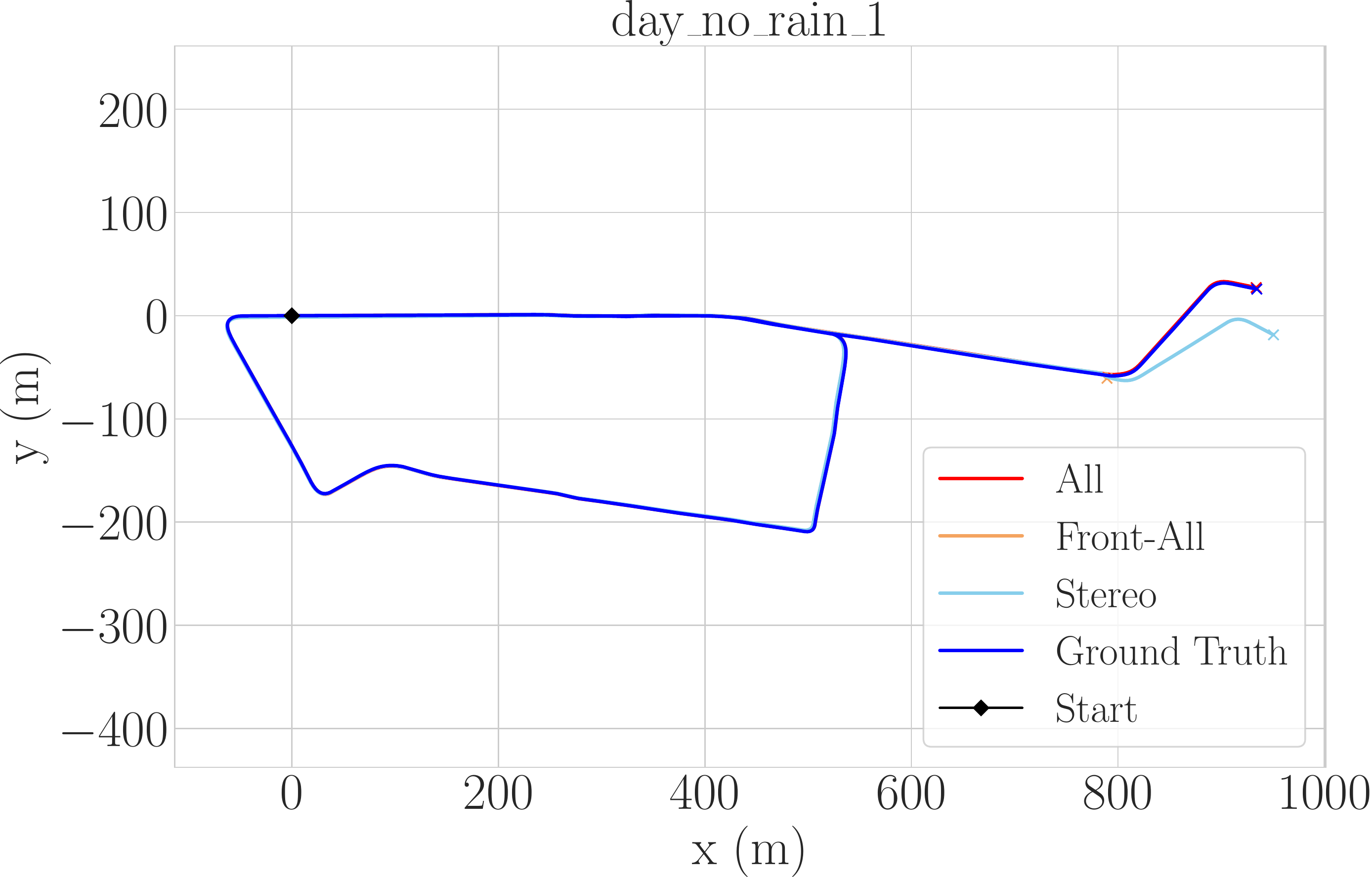}
    \includegraphics[width=0.24\linewidth]{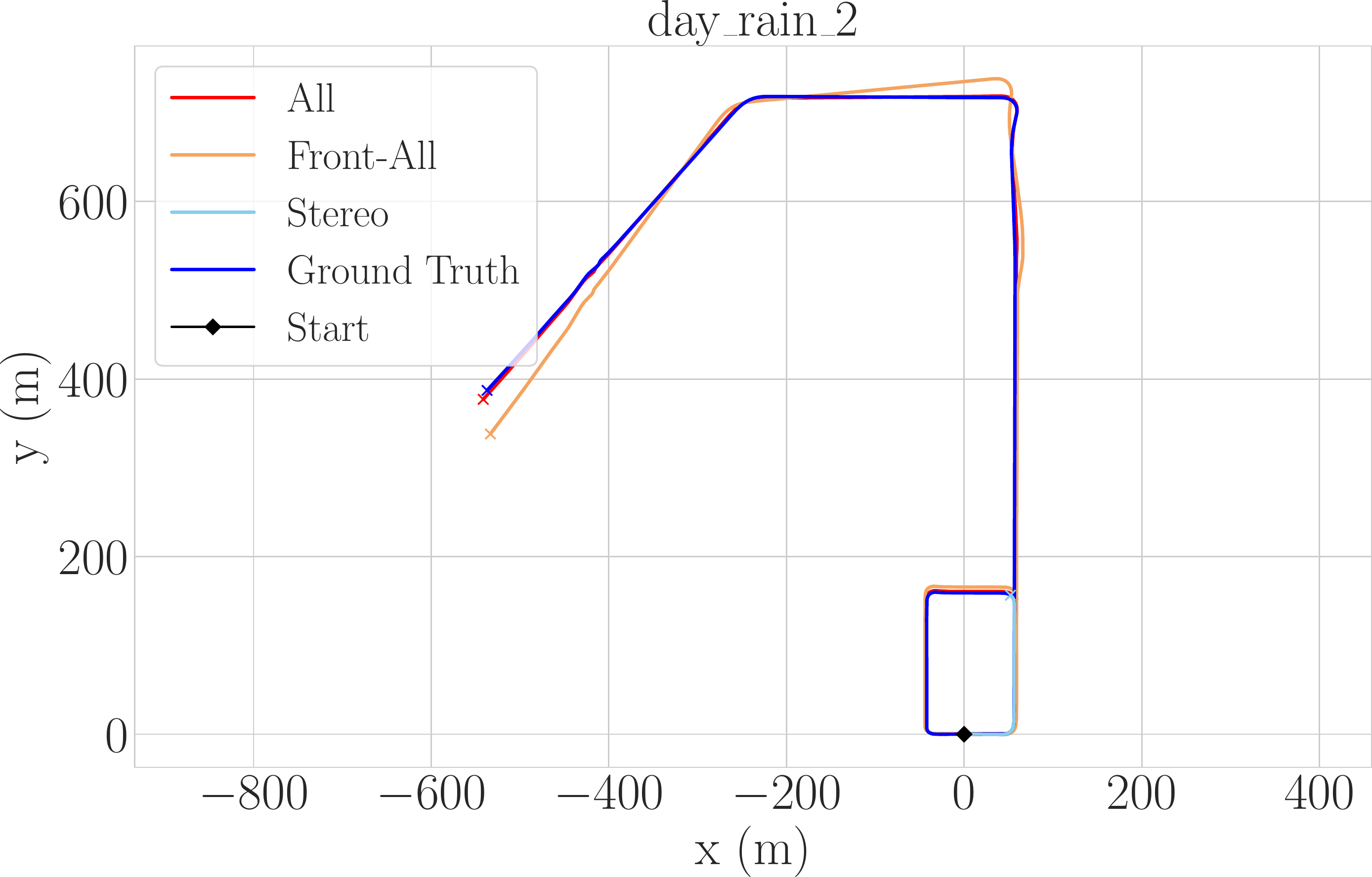}
    \includegraphics[width=0.24\linewidth]{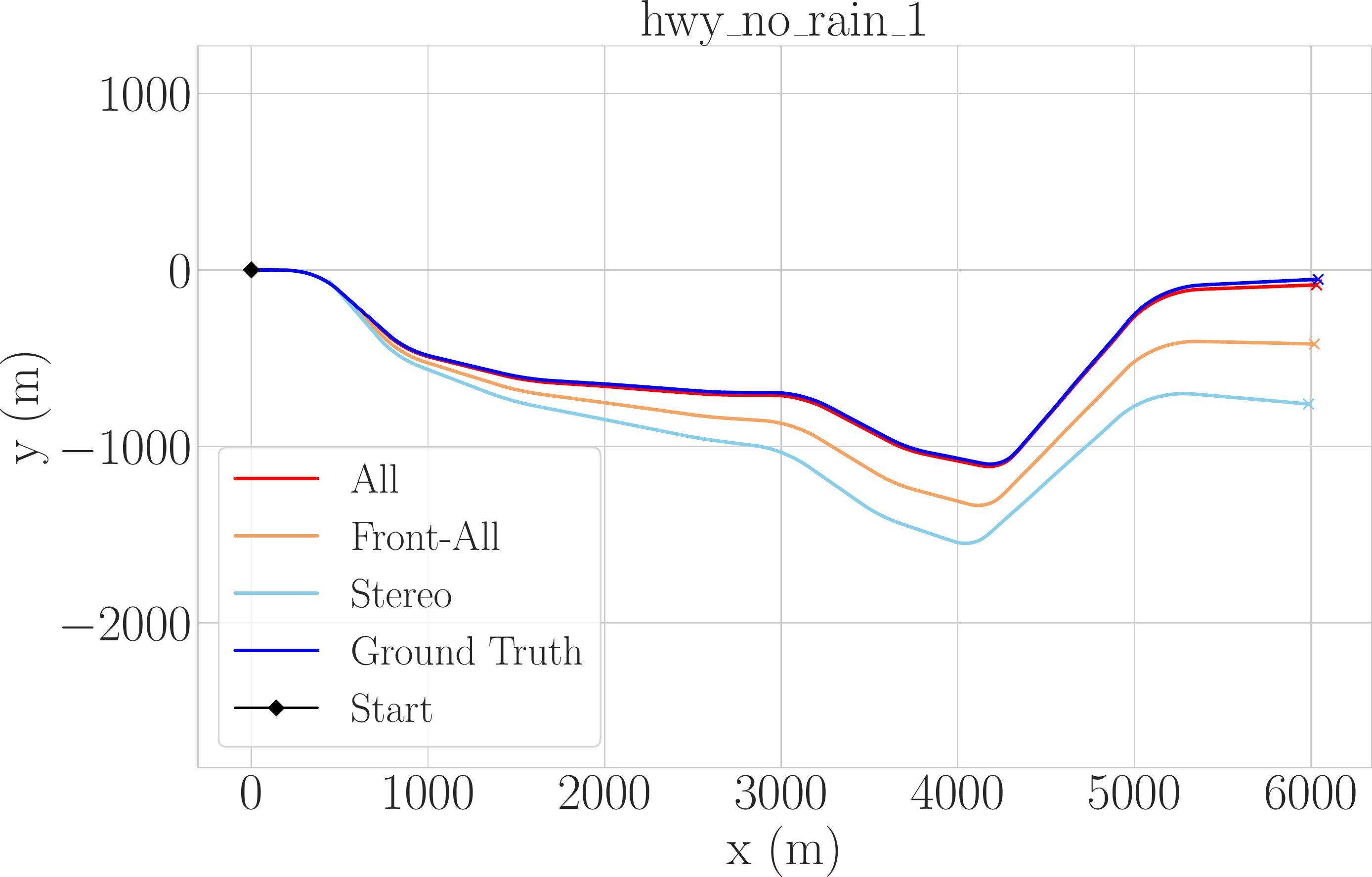}
    \includegraphics[width=0.24\linewidth]{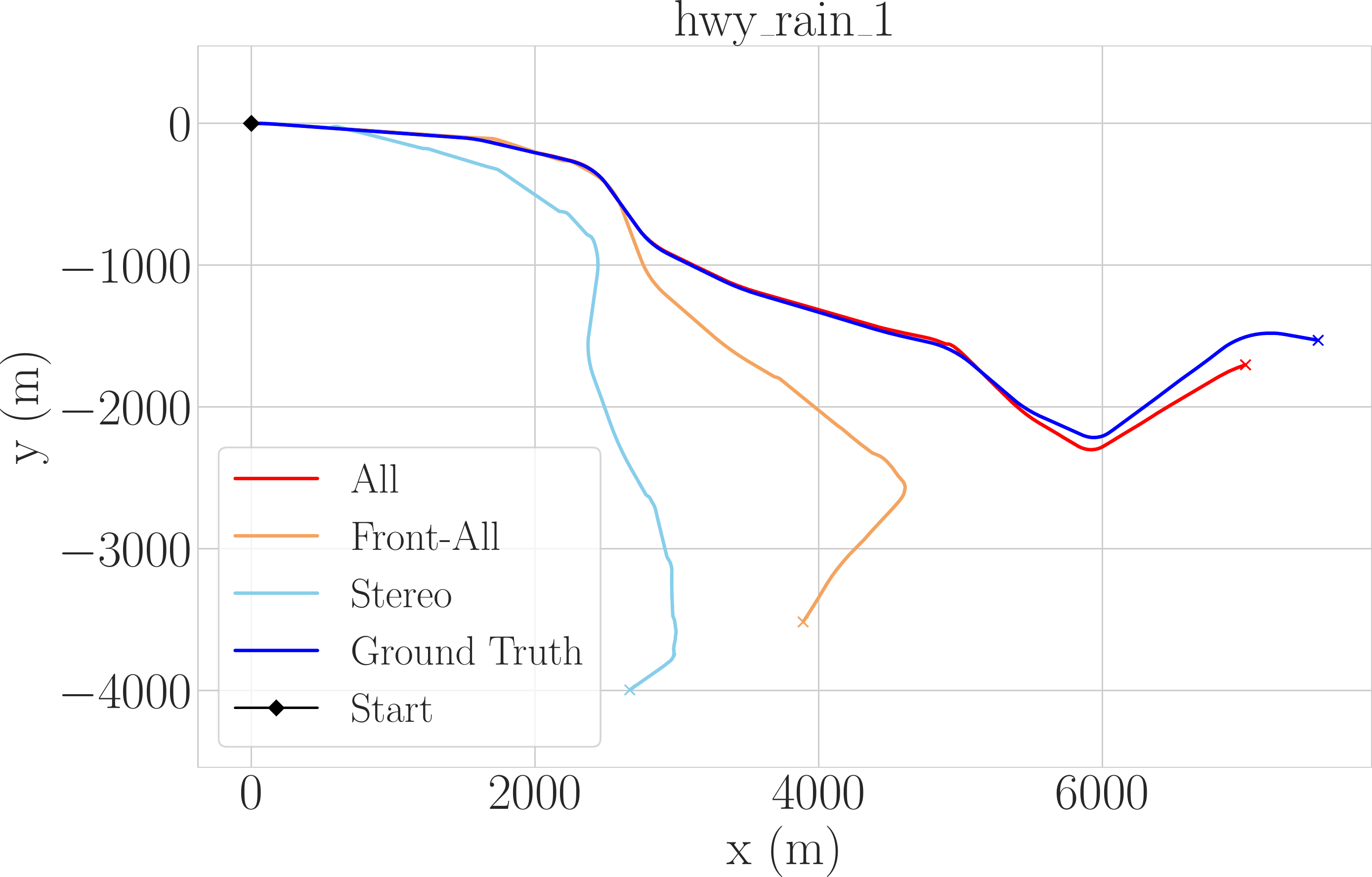}
    \caption{Camera ablation trajectories estimated with (1) all 7 cameras,
    (2) all 3 wide front cameras + the stereo pair, and (3) the stereo pair only.
    The leftmost scenario happened at an intersection where the front view was obstructed
    by a huge truck that was making a turn. Front-all failed due to repeatedly inconsistent bundle
    adjustment results, while stereo persisted with a visible rotation error.
    The second-left scenario is a rainy dusk environment with high volume of traffic.
    The two scenarios on the right correspond to fast highway driving.
    \label{fig:cam_qualitative}}
    \vspace{-3mm}
\end{figure*}

\begin{figure*}
    \centering
    \includegraphics[width=0.24\linewidth]{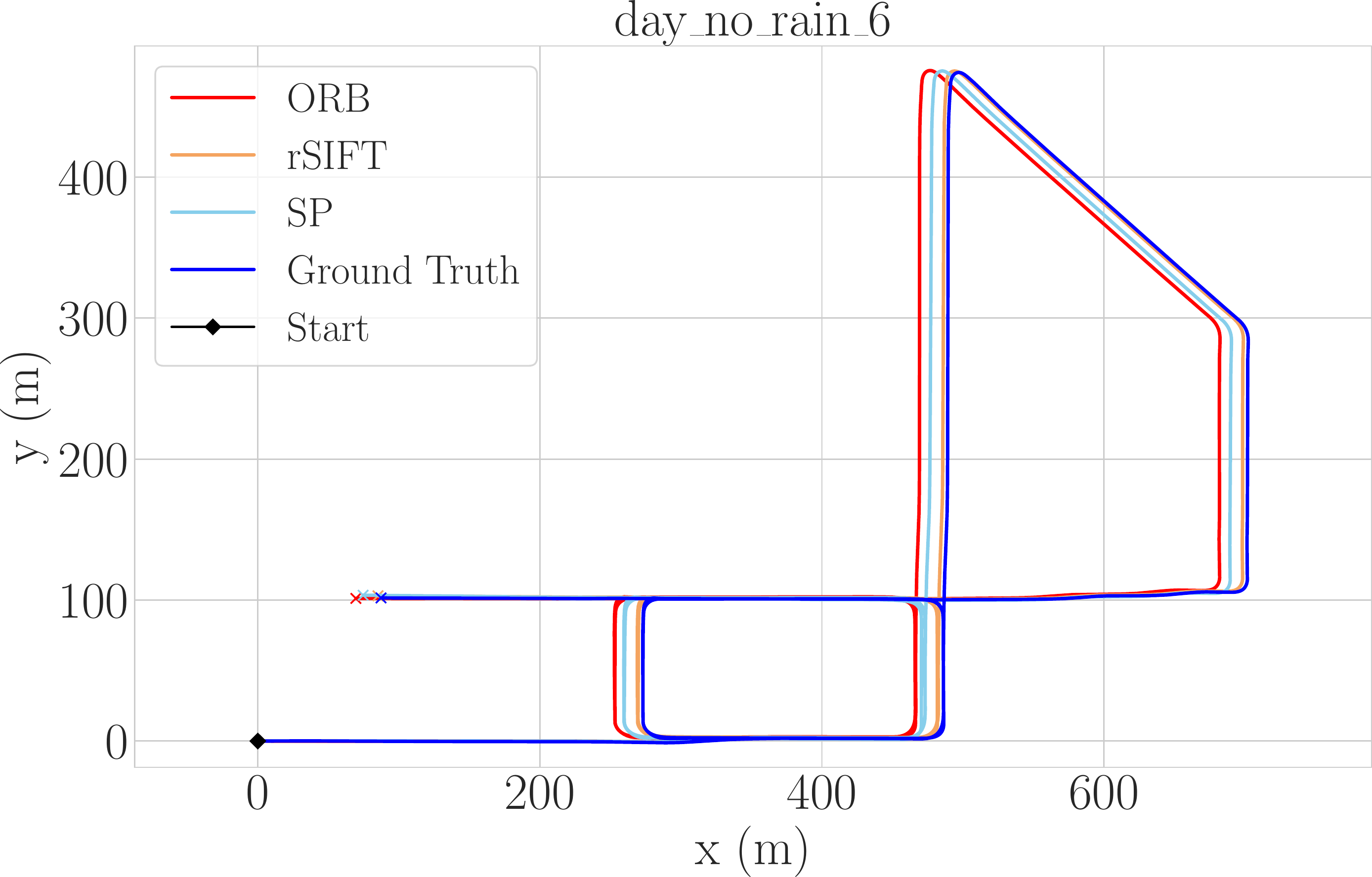}
    \includegraphics[width=0.24\linewidth]{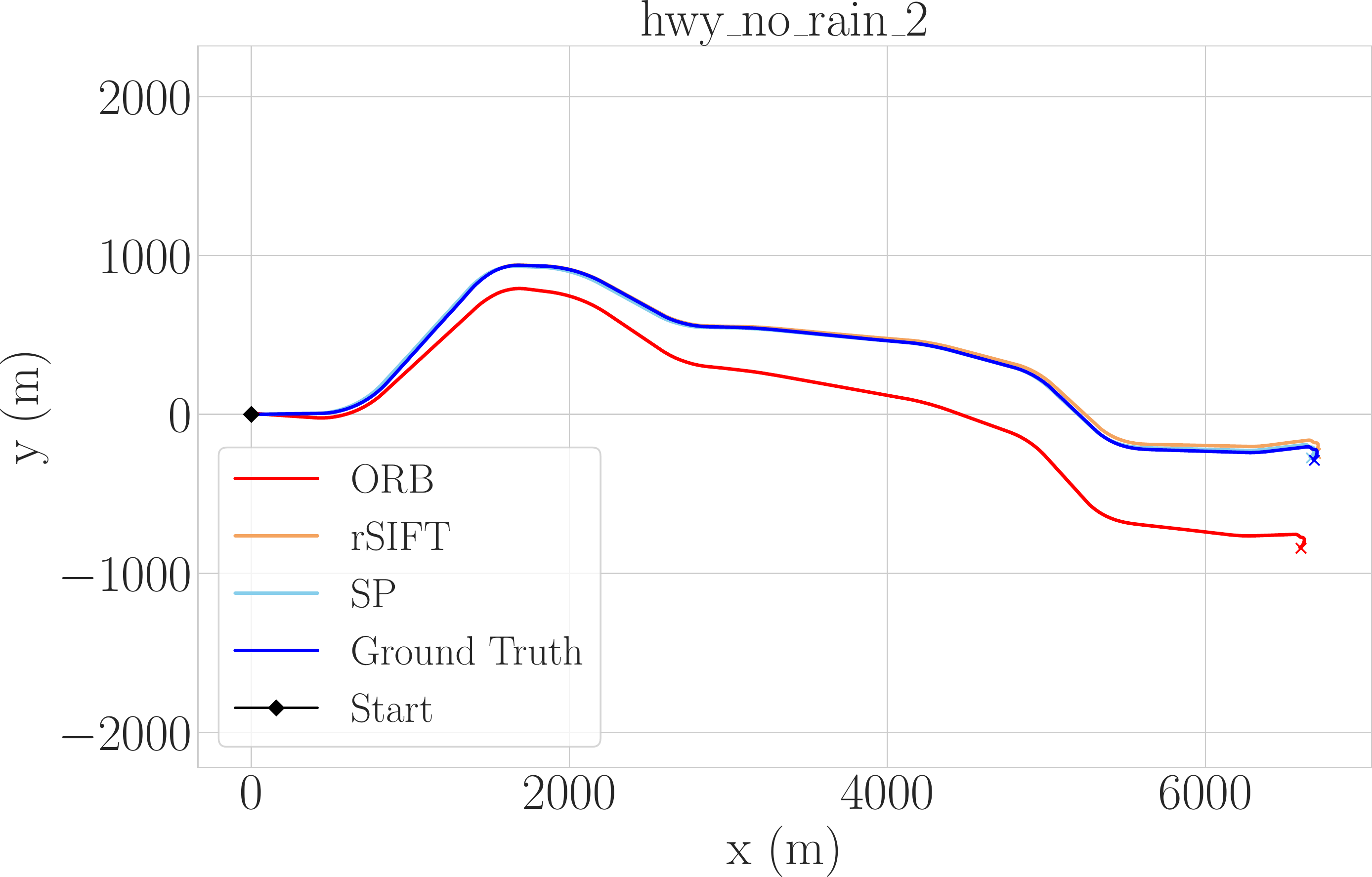}
    \includegraphics[width=0.24\linewidth]{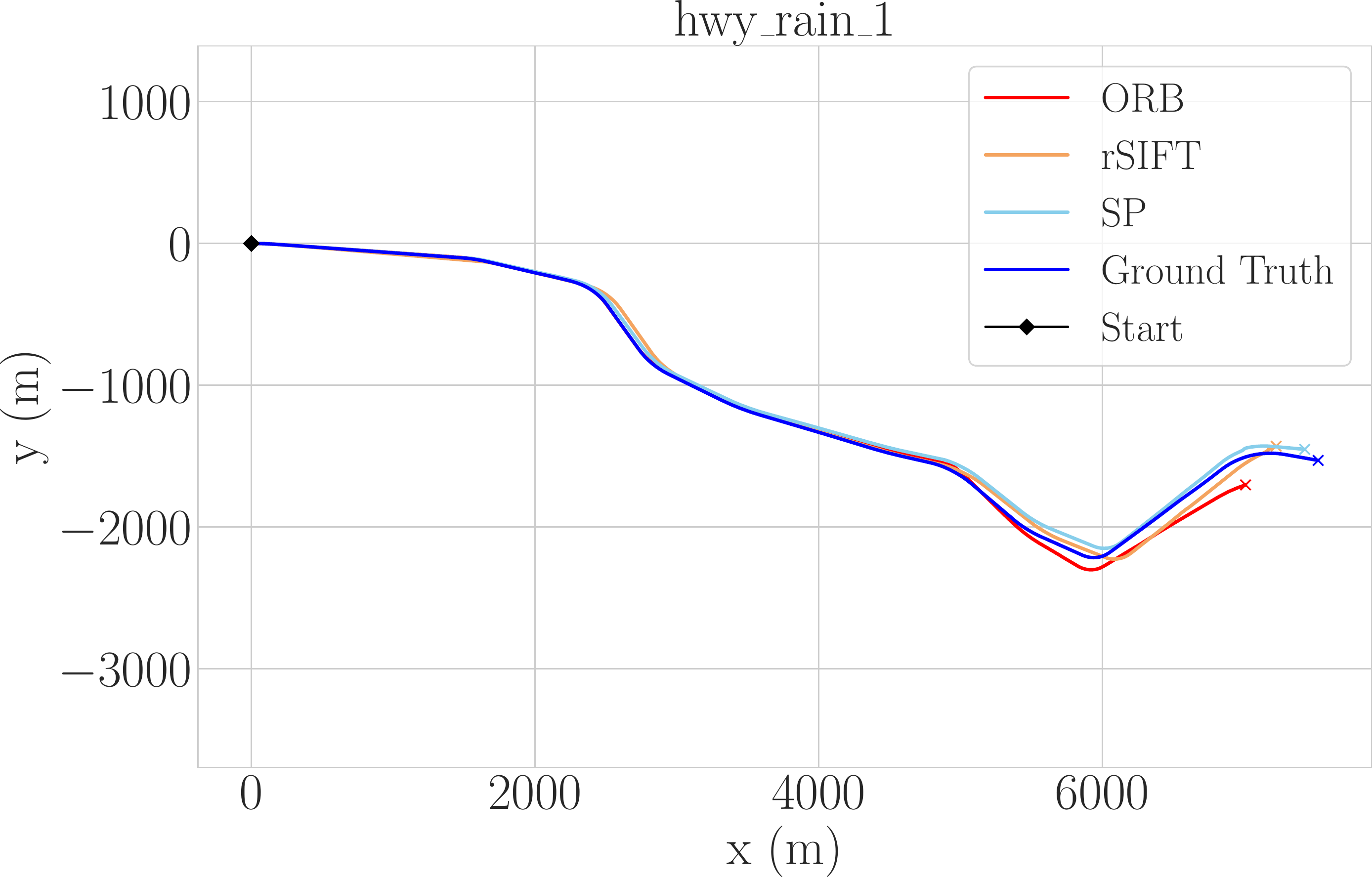}
    \includegraphics[width=0.24\linewidth]{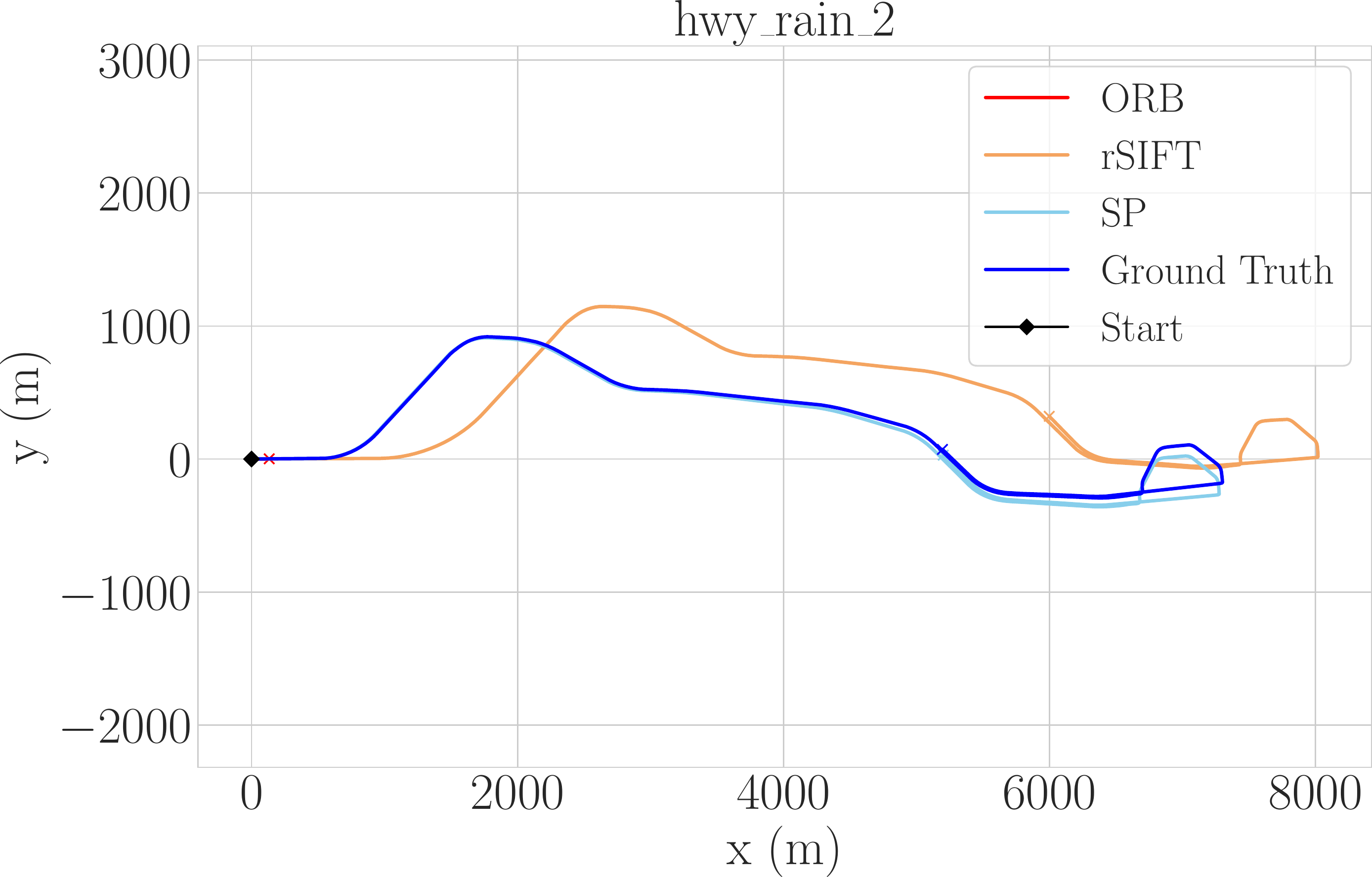}
    \caption{Keypoint extractor ablation trajectories estimated with ORB, RootSIFT and SuperPoint.
    RootSIFT and SuperPoint have smaller absolute errors overall and finish
    a higher percentage of the challenging rainy highway sequences.
    \label{fig:features_qualitative}}
    \vspace{-4mm}
\end{figure*}

\paragraph{Full Results for Ours-A vs. ORB-SLAM2}
Fig.~\ref{fig:full_qualitative_traj} and Fig.~\ref{fig:full_qualitative_maneuver}
showcase the trajectories in all 25 validation sequences,
comparing ORB-SLAM2 using only the stereo cameras to our full system using
all 7 asynchronous cameras.
Fig.~\ref{fig:train_full_qualitative_traj} depicts the trajectories
ins all 65 training sequences.

In the following paragraphs, we qualitatively
showcase the failure cases of our
main paper ablation study baselines.
\paragraph{Motion Model Ablation}
Fig.~\ref{fig:motion_model_qualitative}
plots failure cases of the linear motion model and the discrete-time motion model
with a wrong synchronous assumption. The linear motion model trial failed early due to
repeated mapping failures in a challenging case with dynamic objects, and the synchronous
model had huge estimation errors during complex maneuvers such as reversing and parking.
\paragraph{Camera Ablation}
Fig.~\ref{fig:cam_qualitative} plots trajectories estimated with different camera
configurations, highlighting failure cases resulted from camera configurations with a narrower
field of view in challenging conditions
like view obstruction, low light, rainy environments and low-textured
highway driving.
\paragraph{Keypoint Extractor Ablation}
Fig.~\ref{fig:features_qualitative} plots trajectories estimated by
SLAM systems that use ORB, RootSIFT~\cite{rootsift} and SuperPoint~\cite{superpoint2018}
respectively as the keypoint extractor.
RootSIFT and SuperPoint trajectories visually align better with the ground truth and are able to
complete more challenging rainy highway sequences.

\subsubsection{Loop Closure}

\begin{figure*}[]
  \vspace{3mm}
  \centering
  \includegraphics[width=0.40\linewidth]{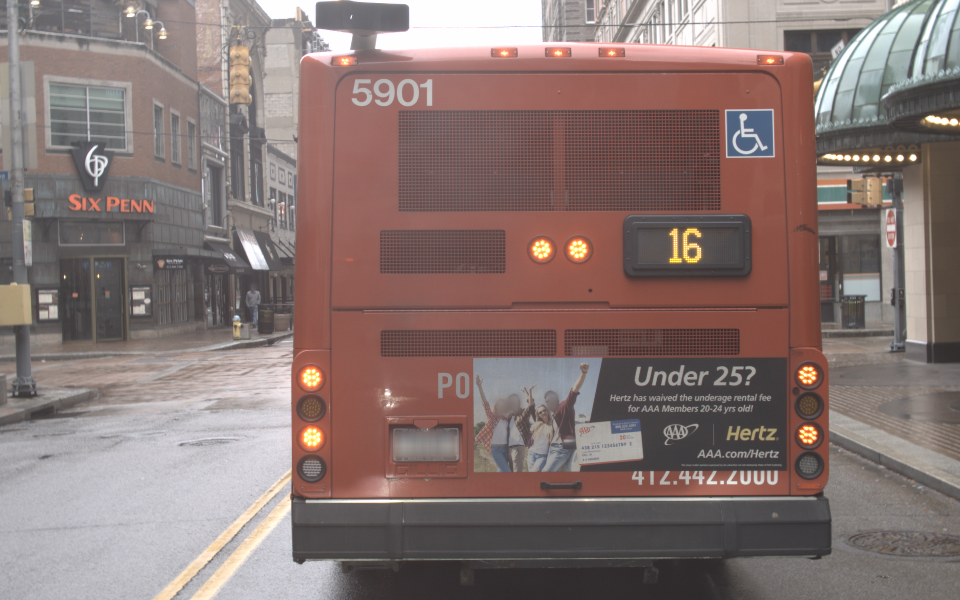}~
  \includegraphics[width=0.40\linewidth]{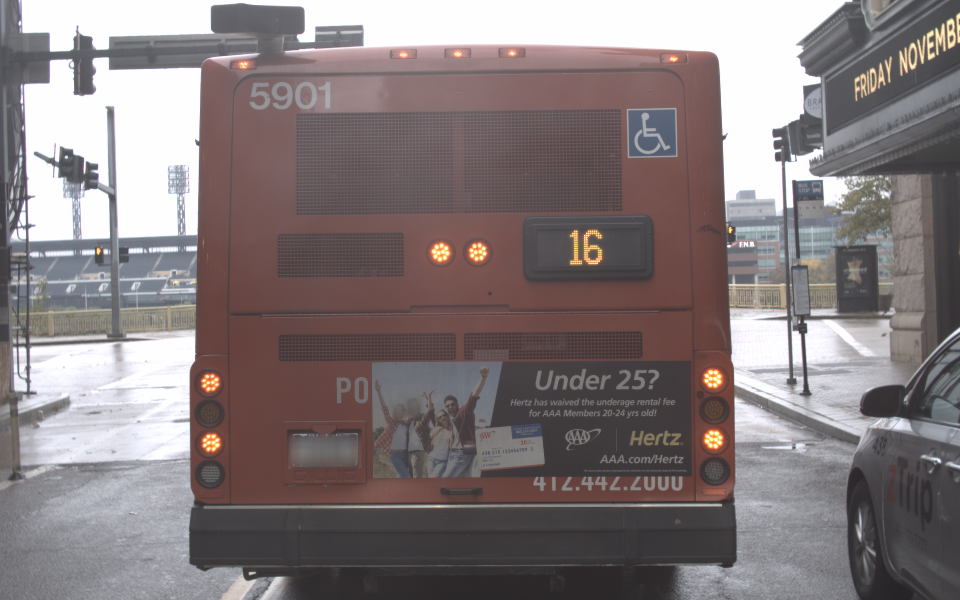}
  \caption{Example where stereo-only loop detection fails due to the presence of
  the same large bus in two geographically distant frames. This sample is from
the training set sequence titled \texttt{day\_rain\_7}.}%
  \label{fig:stereo_loop_failure_0}
\end{figure*}

Figure~\ref{fig:stereo_loop_failure_0} shows a failure case consisting in
a false positive loop detection in the stereo setting. The large bus dominates
the field of view of both cameras while also having rich texture due to the
lights, ad, \etc, causing a loop to be incorrectly closed. Multi-view loop
closure correctly rejects this case and many similar others. This highlights the
importance of multiple cameras for robust SLAM in the real world.
For more qualitative results on the loop closure in the main system,
please refer to the supplementary video.

\subsubsection{Qualitative Map}
Figures~\ref{fig:q_detailed_road},
\ref{fig:q_day_rain_5}, \ref{fig:q_day_no_rain_0},
and \ref{fig:q_day_rain_7}
showcase visualizations of some of the maps produced by AMV-SLAM. Please refer
to our supplementary video for additional qualitative results.

\begin{figure*}[]
  \centering
  \includegraphics[width=0.8\linewidth,trim={6cm 0 15cm 3cm},clip]{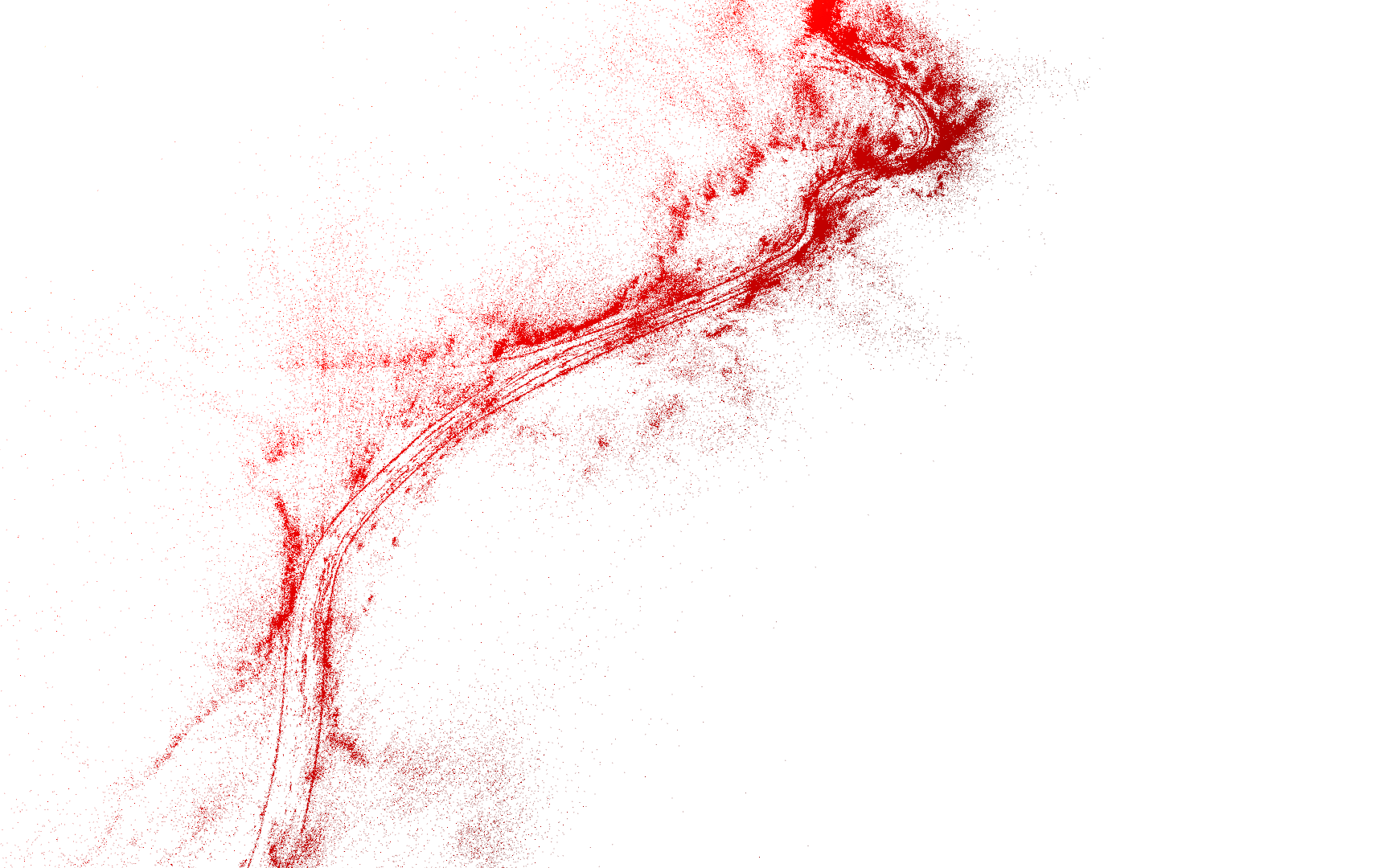}
  \caption{Qualitative example of the 3D structure produced by the AMV-SLAM system. Note
  the system's ability to sharply reconstruct the road boundaries, in addition
  to the surrounding vegetation. This example is from the training set sequence
  titled \texttt{day\_rain\_5}.}%
  \label{fig:q_detailed_road}
\end{figure*}

\begin{figure*}[]
  \centering
  \includegraphics[width=0.8\linewidth,trim={4cm 0 8cm 0},clip]{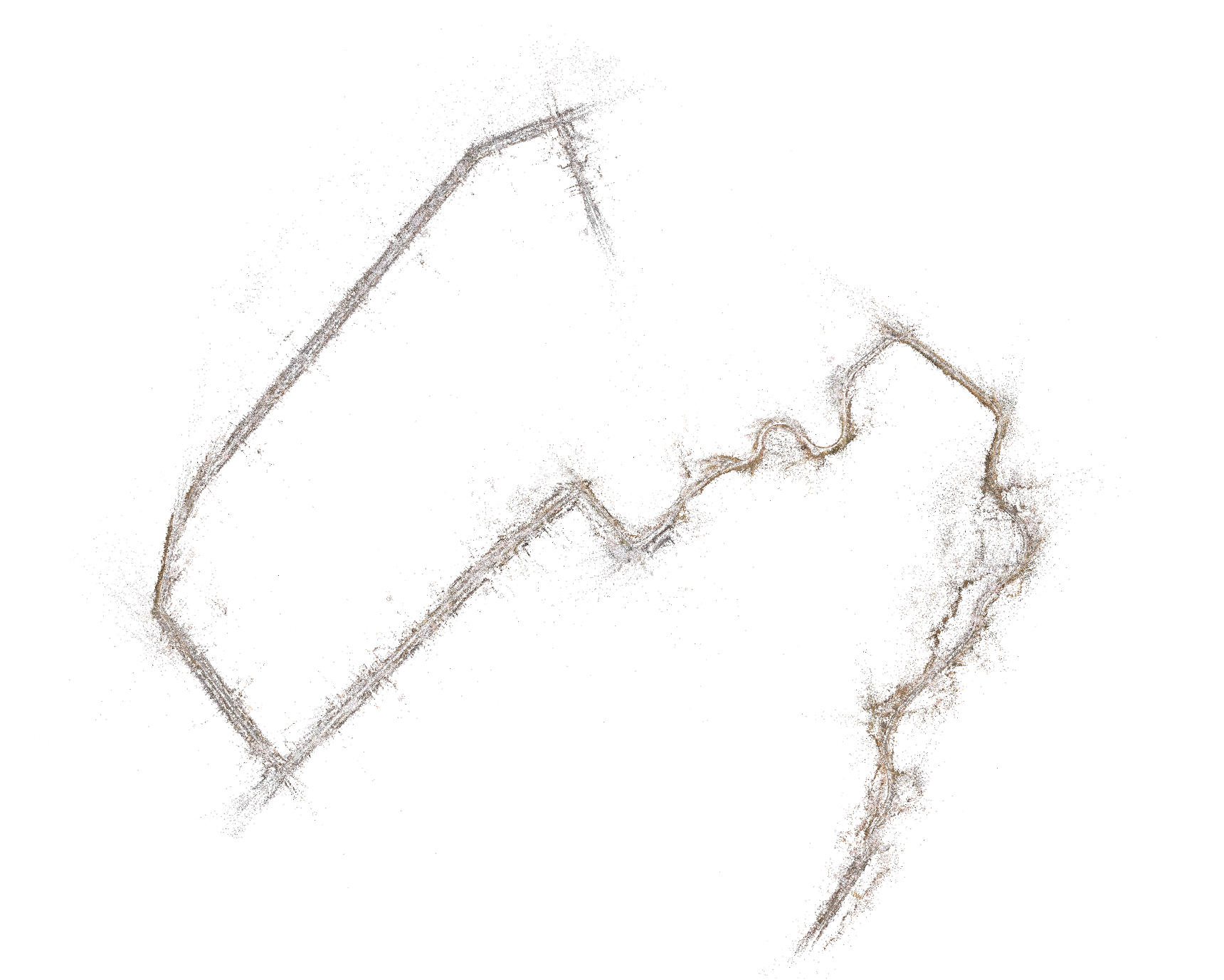}
  \caption{
    Reconstructed point cloud from the training sequence \texttt{day\_rain\_5}.
    Post-processed to include colors from the original camera images. Best
    viewed in electronic format.
  }%
  \label{fig:q_day_rain_5}
\end{figure*}

\begin{figure*}[]
  \centering
  \includegraphics[width=0.8\linewidth,trim={5cm 0 10cm 0},clip]{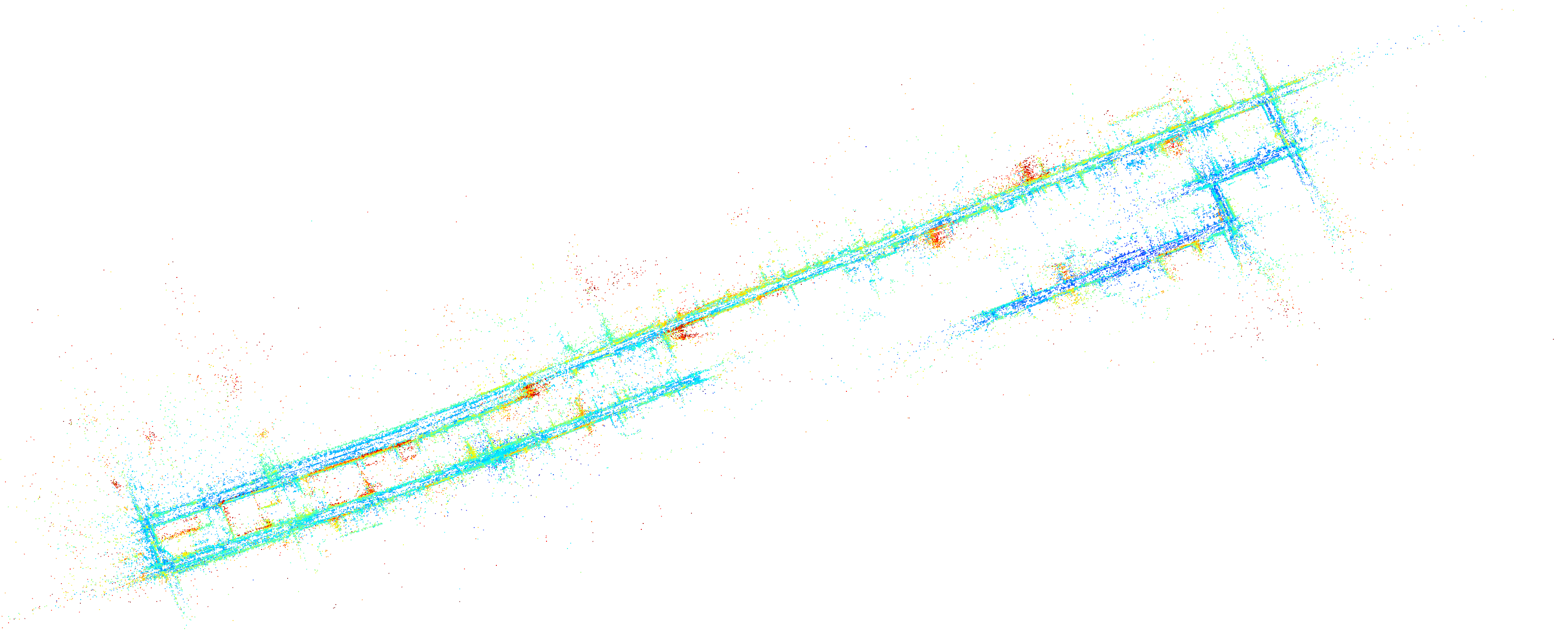}
  \caption{Reconstructed point cloud from the training sequence
  \texttt{day\_no\_rain\_0}.}%
  \label{fig:q_day_no_rain_0}
\end{figure*}

\begin{figure*}[]
  \centering
  \includegraphics[width=0.8\linewidth,trim={6cm 0 8cm 0},clip]{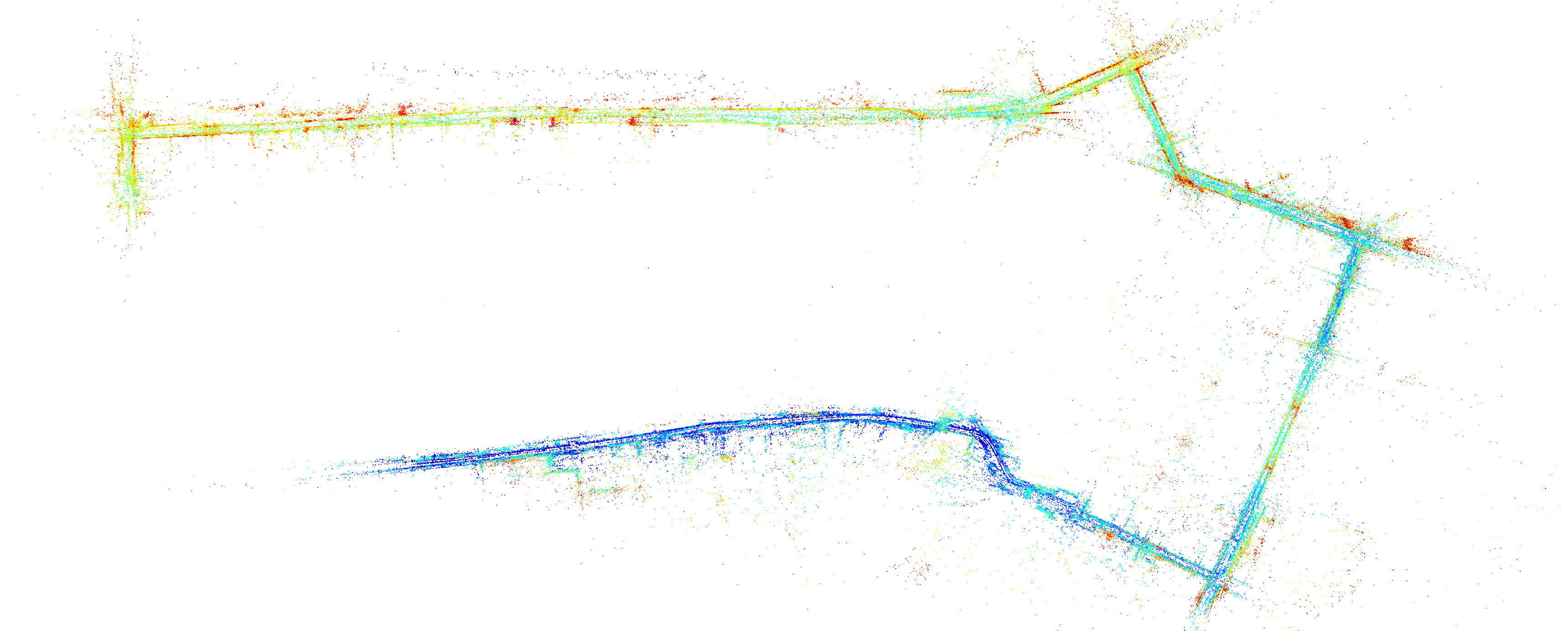}
  \caption{Overview of a reconstruction produced from the training set sequence
  \texttt{day\_rain\_7} by our system. The point cloud is colored by the height
(Z) of each point, in the map reference frame.}%
\label{fig:q_day_rain_7}
\end{figure*}

  \section*{ACKNOWLEDGMENTS}
  The authors would like to thank Julieta Martinez, Davi Frossard, and Wei-Chiu
Ma for their valuable input on the writing of the paper, Jack Fan for his
contributions to the metadata code used to select the segments in the
benchmark, and Rui Hu for his help on improving the learned feature
inference speed. We would also like to thank Prof. Tim Barfoot for the detailed
and thoughtful feedback on the project and paper, as well as our anonymous 
ICRA 2021 reviewers for their thorough comments and suggestions.

  \FloatBarrier
  \clearpage
\else
\fi

\bibliographystyle{IEEEtran}
\bibliography{IEEEabrv,CVConferencesAbrv,refs}

\end{document}